\documentclass[lettersize,journal, openany]{IEEEtran}

\usepackage{booktabs}
\usepackage{amssymb}
\usepackage{algorithmic}
\usepackage{algorithm}
\usepackage{array}
\usepackage{textcomp}
\usepackage{url}

\usepackage{verbatim}
\usepackage{graphicx}
\usepackage{cite}
\usepackage{relsize}
\usepackage{etoolbox}
\usepackage{flushend}
\usepackage{balance} 
\usepackage{cuted}     
\usepackage{amsthm}
\newtheoremstyle{case}{}{}{}{}{}{:}{ }{}
\usepackage{mathtools}
\usepackage[margin = 0.9in]{geometry}
\usepackage{diagbox}
\usepackage{multirow}
\usepackage{blindtext}
\usepackage{comment}
\usepackage{caption}
\usepackage{hyperref}

\makeatletter
\patchcmd{\@makecaption}
  {\scshape}
  {}
  {}
  {}
\makeatother
\usepackage[section]{placeins}
\usepackage{placeins}
\ifCLASSOPTIONcompsoc
    \usepackage[caption=false, font=normalsize, labelfont=sf, textfont=sf]{subfig}
\else
\usepackage[caption=false, font=footnotesize]{subfig}
\fi
\begin{document}

\title{Optimum Output Long Short-Term Memory Cell for High-Frequency Trading Forecasting}

\author{Adamantios Ntakaris, Moncef Gabbouj, and Juho Kanniainen~\IEEEmembership{~IEEE,}
        
\thanks{This paper was produced by the IEEE Publication Technology Group. They are in Piscataway, NJ.}
}

\author{\IEEEauthorblockN{Adamantios Ntakaris\IEEEauthorrefmark{1}, Moncef Gabbouj\IEEEauthorrefmark{2},~\IEEEmembership{Fellow,~IEEE}, and Juho Kanniainen\IEEEauthorrefmark{2}  
}\\
\IEEEauthorblockA{\IEEEauthorrefmark{1}Business School,
University of Edinburgh, Edinburgh, EH8 9JS, UK}\\
\IEEEauthorblockA{\IEEEauthorrefmark{2} Department of Computing Sciences, Tampere University, Tampere, 33720, Finland}\\
}



\maketitle
\begin{abstract}
High-frequency trading requires fast data processing without information lags for precise stock price forecasting. This high-paced stock price forecasting is usually based on vectors that need to be treated as sequential and time-independent signals due to the time irregularities that are inherent in high-frequency trading. A well-documented and tested method that considers these time-irregularities is a type of recurrent neural network, named long short-term memory neural network. This type of neural network is formed based on cells that perform sequential and stale calculations via gates and states without knowing whether their order, within the cell, is optimal. In this paper, we propose a revised and real-time adjusted long short-term memory cell that selects the best gate or state as its final output. Our cell is running under a shallow topology, has a minimal look-back period, and is trained online. This revised cell achieves lower forecasting error compared to other recurrent neural networks for online high-frequency trading forecasting tasks such as the limit order book mid-price prediction as it has been tested on two high-liquid US and two less-liquid Nordic stocks. 
\end{abstract}


\section{Introduction}
\IEEEPARstart{H}{igh-Frequency} trading (HFT), which represents more than 50\% of the trading activity in the US stock market, is the process where signals and trades are analyzed and executed in a fraction of a second (i.e., double-digit nanoseconds). This speed race creates several opportunities for the participants who can capitalize on their infrastructure, like fiber cables and microwave signals, and also on their scientific and technical expertise. One way to organize this trading activity is the so-called limit order book (LOB) where participants in the form of liquidity providers and liquidity takers, with the former ones being more active in the trading scene(s), form its order flow dynamics. This symbiotic relationship between these two parties creates competition based on the order flow information analysis and trading strategy execution. In this paper, we focus on the first part which is the fast order flow information analysis for forecasting tasks such as the prediction of the next LOB's mid-price (i.e., the average of the best ask and bid LOB's price level) and is the main objective of our experimental protocol. The forecasting of the next mid-price, as part of the fast order flow information analysis, is equivalent to online or else tick-by-tick (i.e., forecasting based on every possible trading event disconnected from the time) prediction.  

Every forecasting task in the HFT LOB universe is a challenge in terms of data size digestion (i.e., several million trading events per trading day) and time irregularities inspection. These two factors create a non-linear environment that can be handled properly by specific types of neural networks (NN). A NN type that has exhibited good predictive power is a variant that belongs to the recurrent neural network (RNN) family, named long short-term memory neural network (LSTM) \cite{lstmoriginal}. LSTM filters current and look-back/past information of the input data. The effectiveness of LSTMs is quite extensive in several fields such as acoustics \cite{lstmacc, lstmacc2, lstmacc3}, natural language processing \cite{lstmnlp, nlp2, nlp3}, biology and medicine \cite{lstmmed, bio2, bio3}, and computer vision \cite{lstmvis, lstmvis2, vis3} among others. One particular area of research, where LSTMs have been applied extensively, is finance and algorithmic trading \cite{fin1, fin2, fin3, fin4, fin5, siri2}. 

The study of HFT LOB datasets requires the modeler to parse all the relevant information without any processing lags. There are several examples where the machine learning (ML) experimental protocol was organized around NNs that had to predict lagged or hidden lagged supervised classification labels \cite{deeplob, siri1}. The term hidden lagged supervised classification task refers to the process where the forecasting seems to be based on tick data but the ML HFT trader has to wait until the next change in the bid (or ask) price is going to take place without any parallel countdown mechanism in place like in \cite{ntak1}. 

An additional line of challenge in building real online machine learning experimental protocols is not just the engineering of the objective (i.e., LOB's mid-price forecasting) but also the development of dynamically-adjusted NNs that are suitable for short training based on fewer training epochs. To address the challenge of creating real online forecasting models following the RNN architecture we have to better understand the internal computational processes of their cells, such as the LSTM cell. The LSTM cell is a wrapper system that contains several gates and states that filter the input vector(s). These gates and states are combinations of activation functions that act as the source of information for the next LSTM cell. The order of these calculations is fixed, stale during training/learning, and their position within the cell is not justified by any specific metric. That means that the LSTM cell and the LSTM NN in general exhibit a static behavior against a dynamic and ultra-fast trading setting (i.e., HFT LOB universe). A critical question that the ML HFT trader has to answer is whether the LSTM NN should be utilized as it is (i.e., forecasting will be based on NN's stale behavior and online prediction will be based on processing data using a sliding window) or an autonomous and dynamic cell intervention will allow a more systematic and computationally cheaper, in terms of a shallow NN topology, forecasting outcome. 

In this paper, we address these issues (i.e., truly dynamic and shallow NN topology) by re-arranging and selecting online the optimal LSTM's cell final output. We do that by evaluating online the importance of the internal gates and states. This internal (i.e., within the LSTM cell) feature importance mechanism has been developed around two critical components: a simple optimization method that follows the gradient descent (GD) learning algorithm and a non-forecasting supervised regression problem that has similar but not the same targets/labels as the main objective of our experimental protocol. The GD method and the non-forecasting supervised regression problem are combined internally (i.e., within the LSTM cell) and the optimized outcome is passed to the next LSTM cell. Since we capitalize on the existing LSTM cell architecture without adding, removing, or changing the core LSTM cell calculations we name our cell architecture as Optimum Output LSTM (OPTM-LSTM) cell.\footnote{The coding part is based on TensorFlow \cite{tensorflow} and Keras \cite{keras} libraries. Code available at \url{https://github.com/DeQmE/OPTMLSTM}.}

The rest of this paper is organized as follows. In \hyperref[section:Related]{Section \ref{section:Related}}, we expand on related work about different RNN formations. In \hyperref[section:Method]{Section \ref{section:Method}}, we provide the technical details of the OPTM-LSTM cell and in \hyperref[section:Expe]{Section \ref{section:Expe}} we provide the experimental details and the comparative performance results. \hyperref[section:Con]{Section \ref{section:Con}} , summarizes our findings and discusses limitations and future lines of research.       
\section{Related Work}\label{section:Related}
\noindent So far, RNNs such as LSTM NN (e.g., shallow or deep topology) have been applied extensively in the LOB literature for several different forecasting tasks. For instance, authors in \cite{pass1, pass2} utilized, among several ML and deep learning (DL) methods, LSTMs for mid-price movements forecasting with LSTMs being among the best performers in terms of predictive power. LSTMs also exhibit good forecasting performance for ask and bid future movements as highlighted by the authors in \cite{deeplob}. Authors in \cite{Ch1} enhanced the LSTM's predictive power by attaching an attention mechanism. In an earlier study authors in \cite{Maki1} combined LSTM NNs with an attention mechanism for stock jump price prediction. LSTMs have been recently implemented for market-making strategies as this can be seen in \cite{markmak}. 

The LSTM cell architecture is based on a pre-defined sequence of operations that were set based on ad-hoc assumptions about the ordering of the internal calculations, the number of gates/states, and the cell's output information. This is a common approach for other RNN variants such as standard recurrent cells, gated recurrent units (GRU), and several other LSTM variants. There are only a few studies, outside the HFT LOB universe, where the internal gates were adjusted according to the information flow but with a few shortcomings. For instance, the authors in \cite{Goo} evaluated over ten thousand RNN architectures with their findings suggesting that a combination of some of the calculations, similar to a GRU development, performed better but close to the performance of the standard LSTM NN. Although the authors offered an impressive number of different architectures, for the problems of arithmetic predictions, XML, and language modelling, their results are not suitable for HFT LOB forecasting tasks. The extensive topological grid search is time-consuming until a suitable sequence and type of linear algebra calculations have to be used. Another problem is that the re-arrangement or the elimination of the RNN's gate(s) is not directly connected to the final objective(s). That means that the RNN's weights will be updated first on the full batch (or mini-batch) sequence and then a different architecture will be tested. On a different note, several other authors suggested variants of RNN cells that exhibited good predicting performance but their architecture remained unchanged when the training process started. For instance, authors in \cite{simple} suggested three simplified LSTM cells with similar or better performance to the original LSTM cell structure. In the same fashion authors in \cite{chrono} and \cite{improve} proposed modifications to the most important LSTM's cell gates. Lighter RNN cells are also suggested by authors in \cite{zhang, Unitary, ororbia, lee, greff}. 

More complicated structures were also proposed. For instance, authors in \cite{gers} attached an additional term to several internal LSTM gates, known as peephole LSTM NN. Other variants of such complicated cells can be found in \cite{he, zoneout, phased, stoch, yao, rot}. Few additional RNN variants exist, such as bidirectional LSTM, which were utilized effectively in \cite{bidi} for predicting the open, high, low, and close stock prices. Similarly, authors in \cite{bidi_2}, and \cite{bidi_3} also employed bidirectional LSTMs architectures for daily stock price forecasting. An additional line of research of LSTMs is their use as hybrid models since there are several examples, such as \cite{cnn_1}, \cite{cnn_2}, \cite{cnn_3}, and \cite{cnn_4}, that they combined with convolutional neural networks (CNN) for the task of stock price forecasting. These models, despite their forecasting efficacy, are not agile in terms of high-paced information flow analysis since the cells remain stale during training and also during learning and are disconnected from their predicting task.

\section{Proposed Method}\label{section:Method}
\noindent The behavior of an HFT LOB is characterized by the rapid changes in its stock inventory which affects the stock price. As a result, the LSTM NN needs to be ready to identify, in a fraction of a second, these dynamics and provide the most optimized suggestion/information. This can happen effectively if we improve, for instance, the original mechanics of the LSTM cell. The cell, as part of the LSTM NN, so far was over-engineered, either by adding or removing complexity in terms of the number and order of its internal states and gates. For this reason, we propose a real adaptive LSTM cell architecture, named OPTM-LSTM cell, that is different from the existing ones (i.e., RNN cell mechanics) in two key areas: 
\medskip
\begin{itemize}
    \item LSTM's cell gates and states are treated as features and we measure their ability to handle the online information flow via an internal feature importance mechanism. More specifically, the LSTM cell is equipped with this internal (i.e., within the cell) feature importance mechanism which acts as a non-forecasting supervised regression problem. That means that this non-forecasting supervised regression relies on labels that represent the current (i.e., already known) LOB's mid-price and act as a guarantor for the importance of the internal LSTM's cell gates and states. The guarantor will attach trained weights, via the GD learning algorithm, to every LSTM's cell internal gates and states. 
    \medskip
    \item The guarantor mechanism, which is an internal block within the LSTM cell, operates based on labels that represent the current (i.e., already known) LOB's mid-price. The fact that these labels represent the current mid-price means that the OPTM-LSTM cell is equipped with a robust internal instructing method that is connected, conceptually (i.e., current and already known mid-price), to the overall forecasting objective of the present paper which is the prediction of the next mid-price. In other words, this instructing method is using a lagged version of the labels (i.e., current mid-price labels as part of the internal feature importance mechanism) that are utilized in the prediction of the next mid-price. It can also be stated that we have two supervised regression problems, the first one is part of the guarantor mechanism (i.e., use of the current mid-price as a label) within the OPTM-LSTM cell and the other one is part of the overall forecasting objective in our ML experimental protocol which is the prediction of the next LOB mid-price (i.e., use of the next mid-price as a label).   
 \end{itemize}
\medskip
This way, we do not add or remove any arbitrary calculations within the LSTM cell but instead, we capitalize on its existing formation. Thus, we can train fast without any extensive grid search calculations. The cell will be updated constantly based on the tick-by-tick/current mid-price (i.e., every incoming trading event that forms the current mid-price) inflow. That means that the internal gates and states will be evaluated according to the GD method with respect to the mean squared error (MSE) and this mechanism will measure the features' (i.e., gates and states) importance. Before we delve into the details of the method let us define the LSTM cell first together with some background information in terms of learning and training. 

\begin{figure*}[!htpb]
    \centering
    \includegraphics[width=11cm]{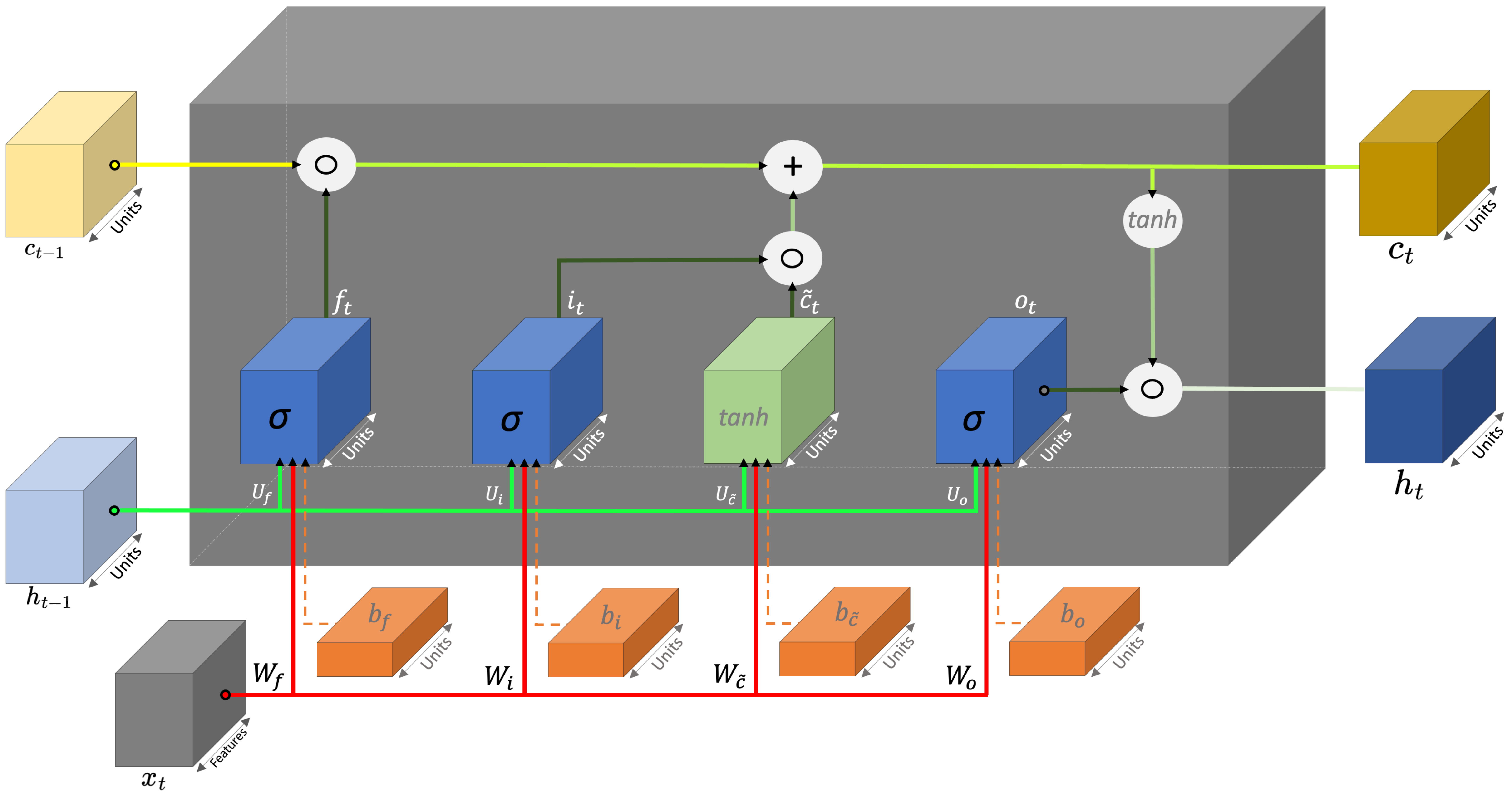}
    \caption{LSTM Cell: This is a three-dimensional interpretation of the core LSTM cell operations. There are four internal gates (i.e., $\textit{f}_t$, $\textit{i}_t$, $\tilde{\textit{c}}_t$, and $\textit{o}_t$) and four states divided into two groups named previous (i.e., $\textit{c}_{t-1}$ and $\textit{h}_{t-1}$) and next states (i.e., $\textit{c}_t$ and $\textit{h}_t$), respectively. There are three input tensors and these are: the feature vector $\textit{x}_t \in \mathbb{R}^{1 \times d}$  which represents just one time-step of the look-back period and the full range of the feature space, the previous LSTM cell output $\textit{h}_{t-1} \in \mathbb{R}^{1 \times U}$ where its dimensions stem from the previous time-step and the number of LSTM units (i.e., the units are based on Tensorflow's notation where the ML user usually selects 8, 16, 32, 64, 128, 256 or 512 units), and the third input is the previous cell state $\textit{c}_{t-1} \in \mathbb{R}^{1 \times U}$ with a similar dimension profile as the previous hidden state $\textit{h}_{t-1}$. Note: The LSTM cell is designed in a three-dimensional space. More specifically, every gate and state are presented as three-dimensional tensors but this is only for convenience to demonstrate the different building blocks. In reality, these are two-dimensional tensors with a batch size of 1.}
    \label{fig:lstmcell}
\end{figure*}

In \hyperref[fig:lstmcell]{Fig. \ref{fig:lstmcell}} we can see that every LSTM cell considers only one time-step input. An example of this input can be just the present full LOB level (i.e., the continual operation of a minimal time-step input length) or the current and previous full LOB levels (i.e., a sliding window based on larger historical input LOB data). These time-step inputs are sequential trading events of a specific length (i.e., usually of arbitrary length but authors in \cite{ntak1} specified, based on a hypothesis test, a logical time-steps vector length) and we can call it a look-back period. The look-back period will determine whether the number of sequential LSTM cells can retain any useful information in their memory. The LSTM's cell memory is based on nine components (i.e., four internal gates, four states, and one input feature vector), which are:

\begin{itemize}
    \item $\textit{f}_t \in \mathbb{R}^{1 \times U}$, forget gate
    \item $\textit{i}_t \in \mathbb{R}^{1 \times U}$, input gate
    \item $\tilde{\textit{c}}_t \in \mathbb{R}^{1 \times U}$, cell input gate
    \item $\textit{o}_t \in \mathbb{R}^{1 \times U}$, output gate
    \item $\textit{c}_t \in \mathbb{R}^{1 \times U}$, cell state
    \item $\textit{h}_t \in \mathbb{R}^{1 \times U}$, hidden state
    \item $\textit{h}_{t-1} \in \mathbb{R}^{1 \times U}$, previous hidden state
    \item $\textit{c}_{t-1} \in \mathbb{R}^{1 \times U}$, previous cell state
    \item $\textit{x}_{t} \in \mathbb{R}^{1 \times D}$, input vector
\end{itemize}

\noindent where the first dimension of the tensors above represents the current or the previous time-step of the look-back period, $\textit{U}$ are the number of LSTM units (e.g., under the TensorFlow framework), and $\textit{D}$ the number of input features. We also include the bias terms $\textit{b}_{f}$, $\textit{b}_{i}$, $\textit{b}_{\tilde{\textit{c}}}$, and $\textit{b}_{0}$ $\in \mathbb{R}^{1 \times U}$ (which are activated by default in TensorFlow). Each of these gates and states represents the following transformations:

\begin{equation}    
    \textit{f}_t = \sigma(W_f \ x_t + U_f \ h_{t-1} + b_f)
\end{equation}

\begin{equation}   
    \textit{i}_t = \sigma(W_i \ x_t + U_i \ h_{t-1} + b_i)
\end{equation}
    
\begin{equation}   
    \tilde{\textit{c}}_t = tanh(W_{\tilde{\textit{c}}} \ x_t +  U_{\tilde{\textit{c}}} \ h_{t-1} + b_{\tilde{\textit{c}}})
\end{equation}    
    
\begin{equation}   
    \textit{o}_t = \sigma(W_o \ x_t + U_o \ h_{t-1} + b_o)
\end{equation}

\begin{equation}\label{eq:cellstate}   
    \textit{c}_t = f_t \odot c_{t-1} + i_t \odot \tilde{c}_{t}
\end{equation}

\begin{equation}   
    \textit{h}_t = \textit{o}_t \odot tanh(c_t)
\end{equation}

\noindent where $\sigma$ is the sigmoid function, $\textit{tanh}$ is the hyperbolic tangent function, the recurrent weights $W_f$, $W_i$, $W_{\tilde{c}}$, $W_o$, $U_f$, $U_i$, $U_{\tilde{c}}$, and $U_o$ are multiplied with their corresponding input $x_t$ and previous hidden state $h_{t-1}$ vectors based on the dot product algebraic operation, and finally, the $\odot$ algebraic operation is the Hadamard product. The first dimension of the tensors represents the input time-step which in our case will be one, otherwise will be equal to the selected batch size. We need to mention here the importance of the relationship between the batch size and the look-back period in an online learning experimental protocol. The batch or mini-batch setting updates the NN weights with a lag equal to the length of the look-back period and this creates several contradictory facts about the online training scenarios. These scenarios are the sliding window (i.e., a length-specific rolling window) or the continual learning (i.e., the rolling window has a minimum length of one and represents the current LOB state) approaches. More specifically, if the length of the look-back period is selected based on some experimental criteria and not randomly, then the batch or mini-batch size will be less important and only then the attention will be paid to the topology of the LSTM NN. This is the main motivation that enables us to scrutinize the LSTM cell topology under an online continual learning protocol.   

\begin{figure*}[!htpb]
    \centering
    \includegraphics[width=12cm]{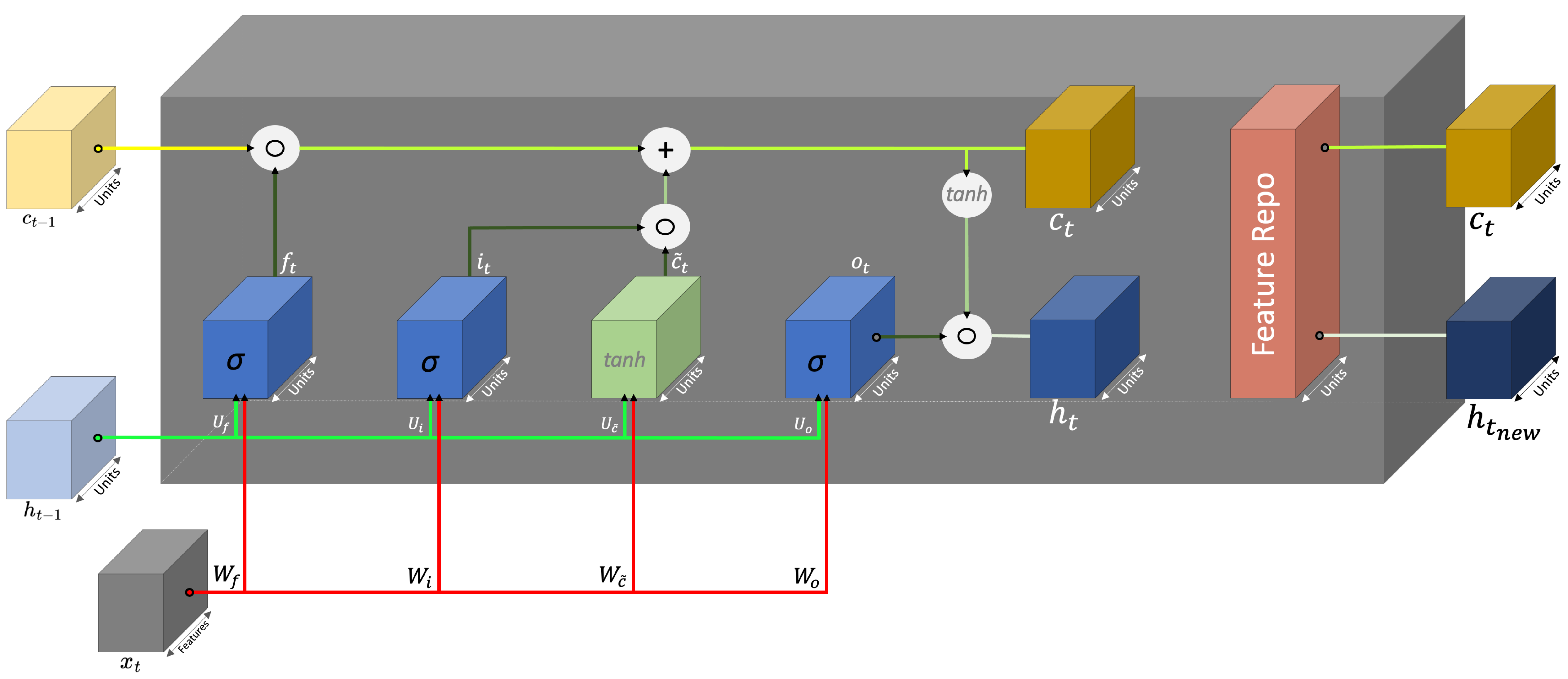}
    \caption{OPTM-LSTM Cell: The optimum output LSTM cell contains the same number of internal gates and states and the same number of outputs as the LSTM prototype. The only difference is that just before we release the two output tensors (i.e., hidden and cell states) we group (i.e., feature repository or Feature Repo) all the internal gates and states and perform a feature importance calculation based on a non-forecasting supervised regression task. We then select the most important feature according to the lowest MSE. The graph does not include any biases intentionally since we did not want to overload the visuals. Note: The OPTM-LSTM is designed in a three-dimensional space. More specifically, every gate and state are presented as three-dimensional tensors but this is only for convenience to demonstrate the different building blocks. In reality, these are two-dimensional tensors with a batch size of 1.}
    \label{fig:revlstmcell}
\end{figure*}

The main idea of our OPTM-LSTM cell is to enable the ML trader to focus only on the selection of the look-back period and discard the selection of the batch or mini-batch size per training iteration. This can happen by creating a dynamically adjusted LSTM cell that can capitalize on its existing internal gates and states and process the information flow without any lags. In \hyperref[fig:revlstmcell]{Fig. \ref{fig:revlstmcell}} we see an overview of the OPTM-LSTM cell. This cell contains the same number of gates and states as the original LSTM cell with the only difference that just before the generation of the two output tensors (i.e., hidden and cell states at time $t$) we employ a feature importance mechanism based on an internal non-forecasting supervised regression which acts as a feature importance guarantor. The guarantor considers only the current and already known mid-price and it will highlight the most important feature among $f_t$, $i_t$, $\tilde{c}_t$, $o_t$, $c_t$, and $h_t$. This mid-price will be the label/target for the internal non-forecasting supervised problem and it will define the importance/order of the LSTM's cell states and gates. This importance mechanism is utilizing the GD as the learning weights optimizer. 

The internal supervised problem is not a forecasting problem but a calibration problem. That means that the provided labels represent the current LOB mid-price. This approach offers a robust calibration of the feature repository (or else Feature Repo in \hyperref[fig:revlstmcell]{Fig. \ref{fig:revlstmcell}}) selection mechanism based on GD. The GD learning algorithm will converge after a few iterations (i.e., we tested that seven to ten iterations suffice) with the input data vector being the current full LOB data state. The feature repository then extracts the best gate or state and this will be the final/optimal output named hidden state $h_t$. The cell state $c_t$ stays intact. We observed that the replacement of the cell state $c_t$ by the second most important feature (according to the MSE score) did not exhibit significant improvements to the overall final objective, which is the prediction of the next mid-price (i.e., overall regression objective outside the OPTM-LSTM cell). 

\begin{figure*}[!htpb]
    \centering
    \includegraphics[width=12cm]{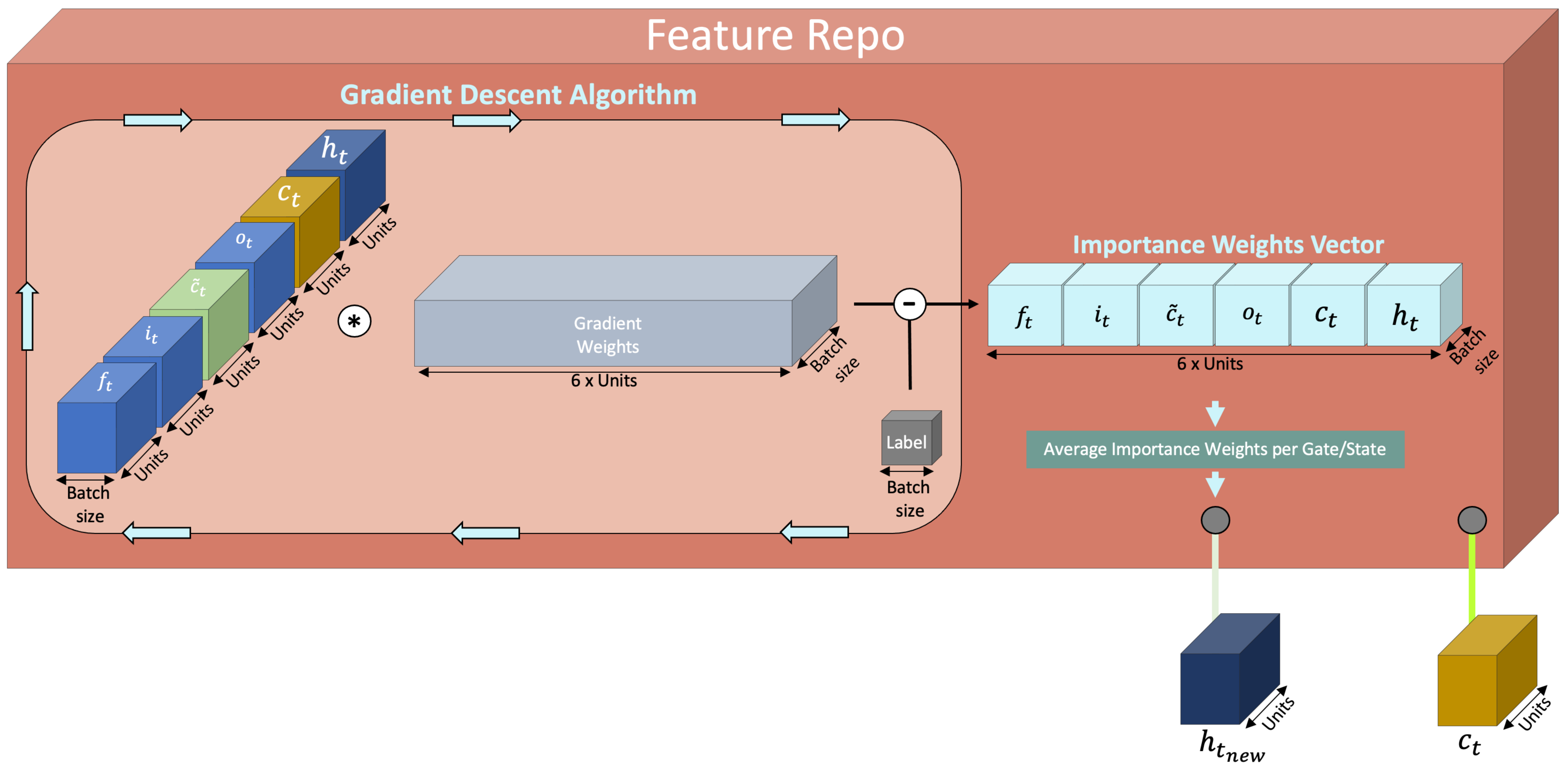}
    \caption{From left to right: the Feature Repo is a block within the OPTM-LSTM cell and contains the mechanism that will decide which state (or gate) will replace the default LSTM's cell output. It is developed in three stages: the first stage is the GD method that contains the features (i.e., LSTM's cell states and gates) and the guarantor (i.e., the current and already known mid-price that acts as a non-forecasting label) which are calculated based on matrix multiplications and vector differences. After a few iterations, the GD will achieve convergence and the final vector of the learned weights, which we call importance vector, will then be stored for the next phase. The next phase is taking place outside the GD algorithm and is related to the importance vector which contains the learned weights per number of units for every LSTM's states and gates. Every state (or gate) has $1 \times U$ number of weights and by calculating their average we can identify which state (or gate) is more important. This leads to the final step which is OPTM-LSTM's cell final output. Now that we know which gate (or state) is the most important based on their average weight we will replace the final $h_t$ state with the $h_{t_{new}}$ which can be any of the $f_t$, $i_t$, $\tilde{c}_t$, $o_t$, $c_t$, and $h_t$. The final cell state $c_t$ (see bottom right corner) remains the same since we did not notice any significant overall forecasting improvements by replacing it with the second most important state or gate. Note: The Feature Repo is designed in a three-dimensional space. More specifically, every gate (or state), the gradient weights block, the label, or the importance weights vector, are presented as three-dimensional tensors but this is only for convenience to demonstrate the different building blocks. In reality, these are two-dimensional tensors with a batch size of 1.}
    \label{fig:repo}
\end{figure*}

A closer look inside the Feature Repo (i.e., feature repository) block can be seen in \hyperref[fig:repo]{Fig. \ref{fig:repo}}. The Feature Repo block contains several critical components which are: the collection of the six internal gates and states, the gradient weights which are updated based on the GD learning algorithm, the labels that reflect only the current mid-price (i.e., non-forecasting regression problem), and the input time-step tensor $x_t$. The combination of these components will highlight which state or gate is the most important feature by averaging the highest price in terms of the learned weights per internal gate or state. We need to mention that the GD weights are not part of the backpropagation through time training method (BPTT) - the mechanism that optimizes the LSTM's trainable parts $W_f$, $W_i$, $W_{\tilde{c}}$, $W_o$, $U_f$, $U_i$, $U_{\tilde{c}}$, and $U_o$ - see \hyperref[sec:FirstAppendix]{Appendix \ref{sec:FirstAppendix}} for a detailed derivation of the BPTT process. 

The mechanics of the Feature Repo block are expressed mathematically based on the following three steps:\\

\noindent 1) For the input $\textit{x}_{t} \in \mathbb{R}^{1 \times D}$, where the first dimension represents the batch size and $D$ the number of input features (e.g., current LOB state) and $U$ the number of hidden units, we concatenate all the internal LSTM gates and states which are $\textit{f}_t$, $\textit{i}_t$, $\tilde{\textit{c}}_t$, $\textit{o}_t$, $\textit{c}_t$, and $\textit{h}_t$, in one vector $\textit{r}_t \in \mathbb{R}^{1 \times C}$, where $C = 6 \ \times  \ U$ is the total number of the most recently calculated internal gates and states in relation to the number of hidden units, as follows:

    \begin{equation}
        \textit{r}_{t} = [\ \textit{f}_t \ | \ \textit{i}_t \ | \ \tilde{\textit{c}}_t \ | \ \textit{o}_t \ | \ \textit{c}_t \ | \ \textit{h}_t]
    \end{equation}

\medskip
\noindent 2) For the non-forecasting label $y_t \ \in \mathbb{R}^{1 \times 1}$ that represents the present and already known mid-price and the vector $r_t$ the iterative process of the GD algorithm (see \hyperref[algo:gd]{Algorithm \ref{algo:gd}}) unfolds, as follows:

\begin{algorithm}
  \caption{Online Gradient Descent Algorithm}
  \begin{algorithmic}[1]
    \REQUIRE Learning rate $\alpha$, number of iterations $I$, initial parameter $\theta$, non-forecasting labels $y_{t}$
    \FOR{$i=1$ to $I$}
      \STATE Compute the predicted labels $\hat{y_t}$ = $\theta^T \cdot r_t$
      \STATE Compute the $error$ = $\hat{y_t} - y_{t}$
      \STATE Compute the gradient $\nabla J(\theta)$ = $2 \cdot (r_t^T \cdot error)$
      \STATE Update the parameter $\theta \gets \theta - \alpha \cdot \nabla J(\theta)$
    \ENDFOR
    \ENSURE The optimized parameter vector $\theta$
  \end{algorithmic}
  \label{algo:gd}
\end{algorithm}

\noindent with the parameter $\theta \in \mathbb{R}^{C \times 1}$. 

\medskip    
\noindent 3) The trained parameter $\theta$ is the \textit{Importance Weights Vector} or $IWV$ (see \hyperref[fig:repo]{Fig. \ref{fig:repo}}) and it will be partitioned based on the six gates and states. Each of the components of this partition represents one of the six gates and states with respect to the number of hidden units $U$. We then calculate the average importance per component $AI_c$, as follows:

\begin{equation}
AI_{c} = \frac{1}{U} \cdot \sum_{j= (c \times U) - (U -1)}^{c \times U} IWV_j 
\end{equation}

\noindent where $c=$1, 2, 3, 4, 5, 6 and then we select the component with the highest $AI_c$ value. This component will be the optimum output (or $h_{t_{new}}$) of the OPTM-LSTM cell.

\subsection{Complexity}
\noindent The LSTM cell is very efficient in terms of time and space complexity. More specifically, the LSTM cell is local in time and space, which means that every update per timestep is independent of the input sequence length. LSTM's cell time and space complexity per timestep is $\textit{O(W)}$ where $\textit{W}$ is the number of its parameters and it is equal to $\textit{W} = 4 \times U^2 + 4 \times U \times I + U \times O + 3 \times U$, where $\textit{U}$ is the number of hidden cells, $\textit{I}$ is the number of input units, and $\textit{O}$ is the number of the output units. The newly introduced OPTM-LSTM cell contains, in terms of computational complexity, an additional term that represents the GD's online updates. As a result, OPTM-LSTM's time and space complexities are equal to $O(W + N \times D)$ and $O(W + N)$, respectively, where $\textit{D} = 6 \times U$ and $N$ is the number of input sample points. 

The input sample points are based on a two-dimensional tensor with dimensions equal to \textit{the number of hidden cell units} $\times$ \textit{the six gates/states} for the first dimension and \textit{the number of LOB features} for the second dimension. An additional critical point is that GD's complexity analysis is based on a batch size with a single item since the OPTM-LSTM performs online reading of the input data. An additional consideration for the OPTM-LSTM complexity analysis is the complexity (time and space) of the backward pass which is based on the BPTT (see \hyperref[sec:FirstAppendix]{Appendix \ref{sec:FirstAppendix}}) and has $O(N^2 \times L)$ and $O(N \times L)$ time and space complexities, respectively, with N being the number of network units and $L$ being the full timestep length. The BPTT is clear from the GD step since the Feature Repo block (see \hyperref[fig:revlstmcell]{Fig. \ref{fig:revlstmcell}}) is not participating in the backward learning process. 

\subsection{Motivation} 
\noindent Now that we have a deeper understanding of what the OPTM-LSTM does, we will provide the motivate behind the development of the optimum output LSTM cell. Two key components led us to the creation of the OPTM-LSTM, one is theoretical and the other one is practical:

\begin{itemize}
    \setlength\itemsep{0.6em}
    \item \textit{Theoretical}: The main advantage of the LSTM architecture is to retain the provided information and delay (or avoid) the vanishing and exploding gradients. If we want to reduce even further the probability of the vanishing (or exploding) gradients we have to pay attention to the derivative of the cell state $\tilde{c}$ at the trading event $t$ (see \hyperref[eq:eq27]{Eq. \ref{eq:eq27}} in \hyperref[sec:FirstAppendix]{Appendix \ref{sec:FirstAppendix}}) which contains the forget gate with a range between $0$ and $1$. The vanishing and exploding gradient problem refer to the case where the RNN/LSTM during the BPTT process is governed by small or large derivative elements - a process that is directly connected to the length of the input sequence (i.e., look-back period). As a result, the forget gate just slows down \cite{vani}, but does not eliminate, the vanishing of gradients phenomenon where the rest of the cell state's gradients are responsible for the exploding gradients. There are always methods that can hold the gradients within boundaries (e.g., gradient clipping \cite{phd_1} or restricted initialization \cite{init}) but these methods are not native to the LSTM cell and further experimental calibration is required. The OPTM-LSTM by shuffling strategically LSTM's cell final output delays (or avoids) even further the possibility of uncontrolled gradients.    
    
    \item \textit{Experimental}: An in vitro experimental approach of LSTM's cell internal gates and states exhibited a remarkable characteristic of their behaviour. More specifically, by extracting the internal states and gates from their LSTM cell (i.e., in vitro approach) per trading event and treating them as trained features, we noticed a constant feature importance alternation while trying to forecast, in an online manner, LOB's mid-price target price. More specifically, we extracted LSTM's cell components and we considered them as trading signals. Instead of relying on the full input feature vector length, we performed a feature importance approach based on the GD learning algorithm. The result was that the feature importance (i.e., LSTM's internal gates and states) frequency alternation was constant and independent of the stock profile. 
    Based on that evidence, we developed the Feature Repo block (see \hyperref[fig:repo]{Fig. \ref{fig:repo}}) within the existing LSTM cell.
\end{itemize}

\section{Experiments}\label{section:Expe}
\noindent In this section, we present details and performance results of the OPTM-LSTM cell based on an online HFT forecasting experimental protocol. In that protocol, we predict the value of the next mid-price. The OPTM-LSTM will compete against several key RNN developments such as prototype LSTM, bidirectional LSTMs, LSTMs with attention layer, GRUs, a hybrid model based on LSTM and CNN, and two naive models (i.e., naive/baseline regressor and the persistence algorithm which considers a flat tick-by-tick mid-price prediction). More specifically, the naive/baseline regressor is based on the idea that the targets in the training set will be a constant value and that value will be utilized for the MSE calculation for the targets in the test set. The inputs to this naive model are the full LOB data but they do not have any contribution to the forecasting process. The second baseline model (i.e., persistence algorithm) is developed according to a forecasting function that predicts (i.e., flat prediction) that the next target price is the same as the current target price $S_{t+1}$ = $S_{t}$, where $S$ represents the target price. The input values to this model will be just the mid-price (i.e., single feature) that we will formally define in the following section. The experimental section unfolds as follows: Initially, we introduce some general concepts about the input data which is based on HFT LOBs, then we describe details of the online experimental protocol and we conclude the section based on tables with performance measures in terms of MSE scores. We present MSE score results based on raw data and two normalization settings (i.e., MinMax and Zscore - two methods that are based on the minimum and maximum values and zero mean and unit standard deviation per input, respectively). 

\subsection{HFT LOB}
\noindent LOB is a vital component for the ML trader that wants to extract information about the interplay between liquidity and price for both the ask and bid sides. Every trading side (i.e., ask or bid) is divided into several price levels (e.g., in our case every LOB has 10 price levels for every side) that change asynchronously. The most sensitive and desirable prices, in terms of the current trading activity, are the best ask and bid prices that form the so-called spread. The spread is the distance between the highest bid and the lowest ask price. Averaging these two prices (i.e., best ask and best bid price) we form the mid-price (MP), which is an artificial indicator (i.e., it is not a traded index) but can be utilized as a sensitive proxy for forecasting tasks. The mid-price is extracted for every trading event. Formally, the mid-price at trading event $t$ can be calculated as follows: 

\begin{equation}
MP_t = \frac{P_{A_t} + P_{B_t}}{2},
\end{equation}

\noindent where $MP_t$ is the mid-price at the trading event $t$, $P_{A_t}$ and $P_{B_t}$ are the best ask and bid prices, respectively. A critical point here is the perception of the trading event $t$. More specifically, the time perception in the HFT space should be avoided and instead, only the trading events should be considered as measurements of sequential trading progression in an intra-day trading setting. This is due to the time-frequency abnormalities in the HFT universe (i.e., trading events that arrive simultaneously or exhibit extensive trading inactivity for several milliseconds). To tackle these time irregularities we perceive every trading event as independent information that is part of the LOB inventory. LOB's current ask and bid price levels and their corresponding inventory states (i.e., volume per LOB level) will be the forces that will trigger the changes in the OPTM-LSTM output information. That information will be utilized to predict the next mid-price at the trading event $t+1$. 

\subsection{Experimental Protocol and Datasets}
\noindent The main objective of our experimental protocol is to forecast the next LOB's mid-price based on tick-level HFT data without any information lag. This is an online regression forecasting task that considers every trading event in LOB. That means that we do not sample the data and the forecasting of the next mid-price considers only the latest information arrival (i.e., only the current state of the LOB). Although the ML trader will perform only the aforementioned forecasting regression task, an additional internal and fully-automated non-forecasting regression task (i.e., the guarantor mechanism) is taking place within the OPTM-LSTM cell. This internal non-forecasting regression problem will arrange and verify the importance order of the cell's internal gates and states. Our experimental ML protocol is based on HFT LOB datasets tested on two high-liquid US stocks, Amazon and Google, and two less-liquid Nordic stocks, Kesko and Wartsila obtained from NASDAQ and followed the ITCH protocol - a protocol that offers advanced visibility of NASDAQ's equities markets, and it ensures ultra-low latency for the market data feeds. To avoid memory loss due to decimal points we multiplied the LOB price levels by 10,000. The historical period that we cover is the first two trading months of the year 2015 for the two US stocks and two trading months of the year 2010 (i.e., from the 1st of June until the end of July) for the two Nordic stocks.

We use a progressive training scenario which is up to 20,000,000 trading events for the selected US stocks (i.e., this is approximately two months' worth of trading data for the two high-liquid stocks US stocks) and up to 5,000,000 trading events for the two Nordic stocks (i.e., 5,000,000 trading events is approximately two months worth of trading data for the two less-liquid Nordic stocks). The motivation for selecting a training set of up to 20,000,000 for the two US stocks and 2,000,000 trading events for the two Nordic stocks (i.e., despite the availability of 5,000,000 trading events for Wartsila) is the fact that the selected models reached a plateau or in some cases, there was a performance drop after that specific input data size, respectively. Since we deal with ultra-fast phase executions we need to find every model's plateau performance. For the testing set, we use 1,000 trading events that we also evaluate progressively. The motivation for the selected testing length is the fact that the span of these 1,000 trading events can vary from several events that arrive simultaneously (i.e., same timestamp registration) to events that are up to eight minutes apart. The progressive testing framework covers 1,000 trading events with every iteration (i.e., new testing trading event) acting as the latest training event in the next forecasting step. In simple terms, the training set absorbs progressively/sequentially every new testing event. The testing performance is measured based on a sequentially-stored MSE score that we average at the end of that testing period. 

The reported models' performance results are based on an extensive topological and hyper-parameter grid search for optimal MSE scores. More specifically, we limited the grid search to up to four hidden layers for the five RNN competitors (i.e., prototype LSTM, LSTM with attention, bidirectional, GRU, and hybrid model) and up to two hidden layers for OPTM-LSTM. The reason for these depth limitations stems from the fact that we operate under an HFT setting and the topological grid search for the five RNN competitors is quite extensive. This grid search considers as hyperparameters numerous combinations in terms of the number of hidden units, optimizers, dropout levels, a look-back period, and batch sizes. The topological grid search limitations to the OPTM-LSTM model were motivated by the reported MSE performance of competitors' results. That means that we applied an early stopping performance mechanism when we achieved an overall (i.e., against the five RNN) lower MSE score. We report in detail the optimal topologies in \hyperref[sec:results]{Section \ref{sec:results}}. The results reported progressively based on several data size scenarios The motivation for this type of performance reporting is that HFT requires ultra-fast data digestion/learning and we need to know what is the optimal minimum data size per model. Based on the concept of ultra-fast data digestion/learning we provide an additional layer of reporting which is related to the number of training epochs and we name them Long Training (i.e., training up to 60 epochs) and Short Training (i.e., training up to 5 epochs). The Long Training and the Short Training settings will give us an insight into every model's behaviour with respect to data requirements and learning time.  

Lastly, to scrutinize the effectiveness of the OPTM-LSTM model we provide MSE scores based on raw data (i.e., for several data size scenarios) and two normalization settings (i.e., MinMax and Zscore based on specific and critical for the MSE performance data size scenarios) for two different input feature sets: the first feature set is the full LOB data (i.e., 40 features which are the ask and bid price levels and their corresponding volume levels) and the second feature set considers only the mid-price. These two different features groups are suitable for employing two naive regressors such as the naive/baseline regressor (i.e., based on the full LOB data input) and the persistence algorithm (i.e., based only on the mid-price) for a flat tick-by-tick forecasting comparison against the five RNN competitors and the OPTM-LSTM model. We conclude our experimental setting with a self-comparison of the OPTM-LSTM performance. That means that we compare the Long Training with the Short Training MSE scores for our model and check its ability to learn fast and based on smaller datasets. We complement our result tables with plots for a visual interpretation of our findings.  

The motivation for these two progressive training settings is to investigate whether the OPTM-LSTM cell is robust across higher and lower trading volume stock examples and also to check whether it provides better performance compared to the other RNN developments. Additionally, this two-phased experimental protocol tries to answer a critical question for the HFT ML trader: can a small (i.e., in terms of size), but well-trained (i.e., a high number of epochs), dataset compete with a larger, but less trained, dataset? The restrictions in the previous questions stem from the fact that the HFT ML trader requires the fastest forecasting ability (i.e., model's performance plateau) together with the shortest training time (i.e., minimum optimal data size). In simple terms, we want to cover the minimum distance between the model's performance, input data size, and the number of epochs.    

\subsection{Results}\label{sec:results}
\noindent The experimental objective of our OPTM-LSTM cell is to find its best/plateau performance by relying only on a very small portion of the given data and achieve lower MSE scores against its well-established competitors (i.e., five RNN models based on the prototype LSTM, LSTM with attention, bidirectional LSTM, GRU and a hybrid model based on an LSTM-CNN architecture) and the two naive forecasting models (i.e., naive/baseline regressor and persistence algorithm). We also offer an additional self-comparison between the Short and Long Training settings for the OPTM-LSTM model. The Short and Long Training experiments refer to the process where we use just five epochs (i.e., Short Training) against a longer training process of up to 60 epochs (i.e., Long Training). The batch size is selected to be minimal for the OPTM-LSTM and equal to 1. This is possible since our optimum output cell allows for optimized communication between the connected OPTM-LSTM cells. This optimized cell communication gives us an additional line of customization which is the selection of the look-back period length. In our experiments, we also select the look-back period to be 1 (i.e., only the current LOB data). 

We employ these five RNN competitors against our new cell and we curate them based on a wide and fully-automated topological grid search range. That means that we run an extensive topological grid search for these models and report the best candidates in terms of the lowest MSE score. The reported models consider the following as hyper-parameters: the depth which is limited to four hidden layers, the number of hidden units that varies from 8 to 512, the type of optimizer (e.g., Adam, Nadam, RMSProp, and Stochastic Gradient Descent or SGD), the batch size which varies from 1 to 64, the use (or not) of Dropout percentage (e.g., 0.20 or 0.50), and a look-back period of 1, 5 or 10 past trading LOB states. Our OPTM-LSTM model hyper-parametrization considers the following, as hyperparameters: the depth is limited to two hidden layers, the number of hidden units varies from 4 to 512, and the type of optimizer (e.g., Adam, Nadam, RMSProp, and SGD). Our model considers a batch size and a look-back period of 1. We need to mention that the optimal models' topologies are stock specific and that can be seen in \hyperref[tab:DeepModelsUS]{Table \ref{tab:DeepModelsUS}} and \hyperref[tab:DeepModelsFinnish]{Table \ref{tab:DeepModelsFinnish}}. The five RNN models and the baseline/naive models can be found in the tables as: 'LSTM' for the prototype LSTM, 'Attention' for the LSTM with the attention layer, 'Bidirectional' for the bidirectional LSTM, 'GRU' for the GRU RNN, 'Hybrid' for the hybrid LSTM-CNN model, 'Baseline' for the naive/baseline model based on the LOB data input, and 'Persistence' for the persistence model based on the mid-price as input.  

\begin{table*}[!hbtp]
\centering
\captionsetup{width=1.00\textwidth}
\caption{Best-performing candidates based on a topological grid search for the US stocks.}
\scalebox{0.750}{
\begin{tabular}{cclccccl}
\cmidrule[2pt]{1-3}\cmidrule[2pt]{6-8}
\multirow{1}{*}{\textbf{Stock}}&{\textbf{Model}}&\multirow{1}{*}{\textbf{Topology}}& \qquad & \qquad &\textbf{Stock} & \textbf{Model} & \textbf{Topology}\\
\cmidrule{1-3}\cmidrule{6-8}
Amazon & LSTM & $\bullet$ LSTM layer with 32 units  &\qquad & \qquad &Google & LSTM & $\bullet$ LSTM layer with 32 units \\
&&              $\bullet$ Dropout 50\%              &\qquad &  \qquad &&            & $\bullet$ Dropout 50\% \\ 
&&              $\bullet$ Dense layer with 1 unit   &\qquad &  \qquad &&            & $\bullet$ Dense layer with 1 unit \\ 
&&              $\bullet$ RMSProp optimizer         &\qquad &  \qquad &&            & $\bullet$ Adam optimizer \\ 
&&              $\bullet$ Batch size of 32 samples  &\qquad &  \qquad &&            & $\bullet$ Batch size of 32 samples \\ 
\cmidrule{2-3}\cmidrule{7-8}
       & LSTM       & $\bullet$ LSTM layer with 40 units          &\qquad &\qquad & & LSTM      & $\bullet$ LSTM layer with 64 units \\ 
       & with       & $\bullet$ PReLU                             &\qquad &\qquad & & with      & $\bullet$ PReLU \\
       & Attention  & $\bullet$ Attention layer                   &\qquad &\qquad & & Attention & $\bullet$ Attention layer \\
       &            & $\bullet$ Dense layer with 40 units         &\qquad &\qquad & &           & $\bullet$ Dense layer with 32 units\\
       &            & $\bullet$ Dense layer with 1 unit           &\qquad &\qquad & &           & $\bullet$ Dense layer with 1 unit\\
       &            & $\bullet$ Nadam optimizer                   &\qquad &\qquad & &           & $\bullet$ Nadam optimizer\\
       &            & $\bullet$ Batch size of 64 samples          &\qquad &\qquad & &           & $\bullet$ Batch size of 64 samples\\
\cmidrule{2-3}\cmidrule{7-8}
& Bidirectional & $\bullet$ Bidirectional LSTM layer with 32 units &\qquad & \qquad & & Bidirectional & $\bullet$ Bidirectional LSTM layer with 32 units \\ 
& RNN           & $\bullet$ Dense layer with 32 units              &\qquad & \qquad & & RNN           & $\bullet$ Dense layer with 32 units \\ 
&               & $\bullet$ Dropout 50\%                           &\qquad & \qquad & &               & $\bullet$ Dropout 50\%   \\ 
&               & $\bullet$ Dense layer with 4 units               &\qquad & \qquad & &               & $\bullet$ Dense layer with 4 units  \\ 
&               & $\bullet$ Dense layer with 1 unit                &\qquad & \qquad & &               & $\bullet$ Dense layer with 1 unit  \\ 
&               & $\bullet$ Adam optimizer                         &\qquad & \qquad & &               & $\bullet$ Adam optimizer \\ 
&               & $\bullet$ Batch size of 32 samples               &\qquad & \qquad & &               & $\bullet$ Batch size of 64 samples \\ 
\cmidrule{2-3}\cmidrule{7-8}
& GRU & $\bullet$ GRU with 32 units                                &\qquad & \qquad & & GRU & $\bullet$ GRU with 32 units  \\ 
&     & $\bullet$ Dense layer with 32 units                        &\qquad & \qquad & &  & $\bullet$ GRU with 32 units   \\ 
&     & $\bullet$ Dense layer with 1 unit                          &\qquad & \qquad & &  & $\bullet$ Dense layer with 32 units   \\ 
&     & $\bullet$ Adam Optimizer                                   &\qquad & \qquad & &  & $\bullet$ Dense layer with 1 unit  \\ 
&     & $\bullet$ Batch size of 32 samples                         &\qquad & \qquad & &  & $\bullet$ Adam Optimizer      \\ 
&     &                                                   &\qquad & \qquad & &  & $\bullet$ Batch size of 32 samples      \\ 
\cmidrule{2-3}\cmidrule{7-8}
& Hybrid & $\bullet$ 1D Convolution layer with with 60 filters, 6 as kernel size &\qquad & \qquad & & Hybrid & $\bullet$ 1D Convolution layer with with 60 filters, 6 as kernel size  \\ 
&        & $\bullet$ RelU activation function                                    &\qquad & \qquad & &        & $\bullet$ MaxPooling1D \\ 
&        & $\bullet$ LSTM layer with 64 units with Tanh activation function      &\qquad & \qquad & &        & $\bullet$ LSTM layer with 64 units with Tanh activation function\\ 
&        & $\bullet$ LSTM layer with 64 units with Tanh activation function      &\qquad & \qquad & &        & $\bullet$ LSTM layer with 64 units with Tanh activation function \\ 
&        & $\bullet$ Dense layer with 30 units with ReLU activation function     &\qquad & \qquad & &        & $\bullet$ Dense layer with 30 units with ReLU activation function \\ 
&        & $\bullet$ Dense layer with 10 units with ReLU activation function     &\qquad & \qquad & &        & $\bullet$ Dense layer with 10 units with ReLU activation function \\ 
&        & $\bullet$ Dense layer with 1 unit                                     &\qquad & \qquad & &        & $\bullet$ Dense layer with 1 unit \\ 
&        & $\bullet$ Nadam Optimizer                                             &\qquad & \qquad & &       & $\bullet$ Nadam Optimizer\\ 
&        & $\bullet$ Batch size of 32 samples                                    &\qquad & \qquad & &        & $\bullet$ Batch size of 64 samples      \\ 
\cmidrule{2-3}\cmidrule{7-8}
& OPTM-LSTM & $\bullet$ OPTM-LSTM layer with 64 units  &\qquad &  \qquad &        & OPTM-LSTM & $\bullet$ OPTM-LSTM layer with 8 units \\
&           & $\bullet$ Dense layer with 4 units       &\qquad &  \qquad &        &     & $\bullet$ Dense layer with 4 units  \\ 
&           & $\bullet$ Dense layer with 1 unit        &\qquad &  \qquad &        &         & $\bullet$ Dense layer with 1 unit \\ 
&           & $\bullet$ Adam optimizer                 &\qquad &  \qquad &        &         & $\bullet$ Adam optimizer \\ 
&           & $\bullet$ Batch size of 1 sample         &\qquad &  \qquad &        &         & $\bullet$ Batch size of 1 sample \\ 
\cmidrule[2pt]{1-3}\cmidrule[2pt]{6-8}
\end{tabular}}
\medskip
\label{tab:DeepModelsUS}
\end{table*}

We see that the OPTM-LSTM requires fewer building blocks compared to its well-established competitors. The effectiveness of the OPTM-LSTM cell is highlighted based on the development of a two-phased experimental setting that unfolds, as follows: the first phase, based on Short Training, our optimum output LSTM cell competes with the best candidate models (see \hyperref[tab:DeepModelsUS]{Table \ref{tab:DeepModelsUS}} and \hyperref[tab:DeepModelsFinnish]{Table \ref{tab:DeepModelsFinnish}}) via a progressive training process which incrementally goes from 1,000 trading events up to 20,000,000 trading events for the two high-liquid stocks (i.e., Amazon and Google) and from 1,000 trading events up to 2,000,000 trading events for the less-liquid Nordic stocks. The second phase is based on Long Training equipped with an active Early Stopping mechanism, the same models, and the same progressive training process that goes from 1,000 to 15,000 trading events for both sets of stocks. 

Considering this question, we present our findings (i.e., MSE scores) per stock in \hyperref[tab:GoogleShort]{Table \ref{tab:GoogleShort}} - \hyperref[tab:WartsilaBenchmark]{Table \ref{tab:WartsilaBenchmark}} in the following order: 

\begin{itemize}
    \item Two MSE score tables (i.e., Short and Long experimental protocols) that contain several data size scenarios based on raw LOB data as input for the OPTM-LSTM and the five RNN competitors. It is important to mention that these data size scenarios will highlight the optimal minimal sample length that is required for the models to operate effectively in HFT. 
     \vspace{0.1cm} 
    \item According to the optimal minimal sample length above, two additional multidimensional MSE score tables (i.e., Benchmark Training and Benchmark Testing) will summarize/compare the OPTM-LSTM against the two baseline models and the five RNN competitors, under two different input settings (i.e., LOB data and mid-price) and two normalization settings (i.e., MinMax and Zscore). 
\end{itemize}

\begin{figure*}[hbt!]
    \centering
  \subfloat[OPTM-LSTM training MSE scores \label{1a}]{%
       \scalebox{0.55}{\includegraphics[width=0.65\linewidth]{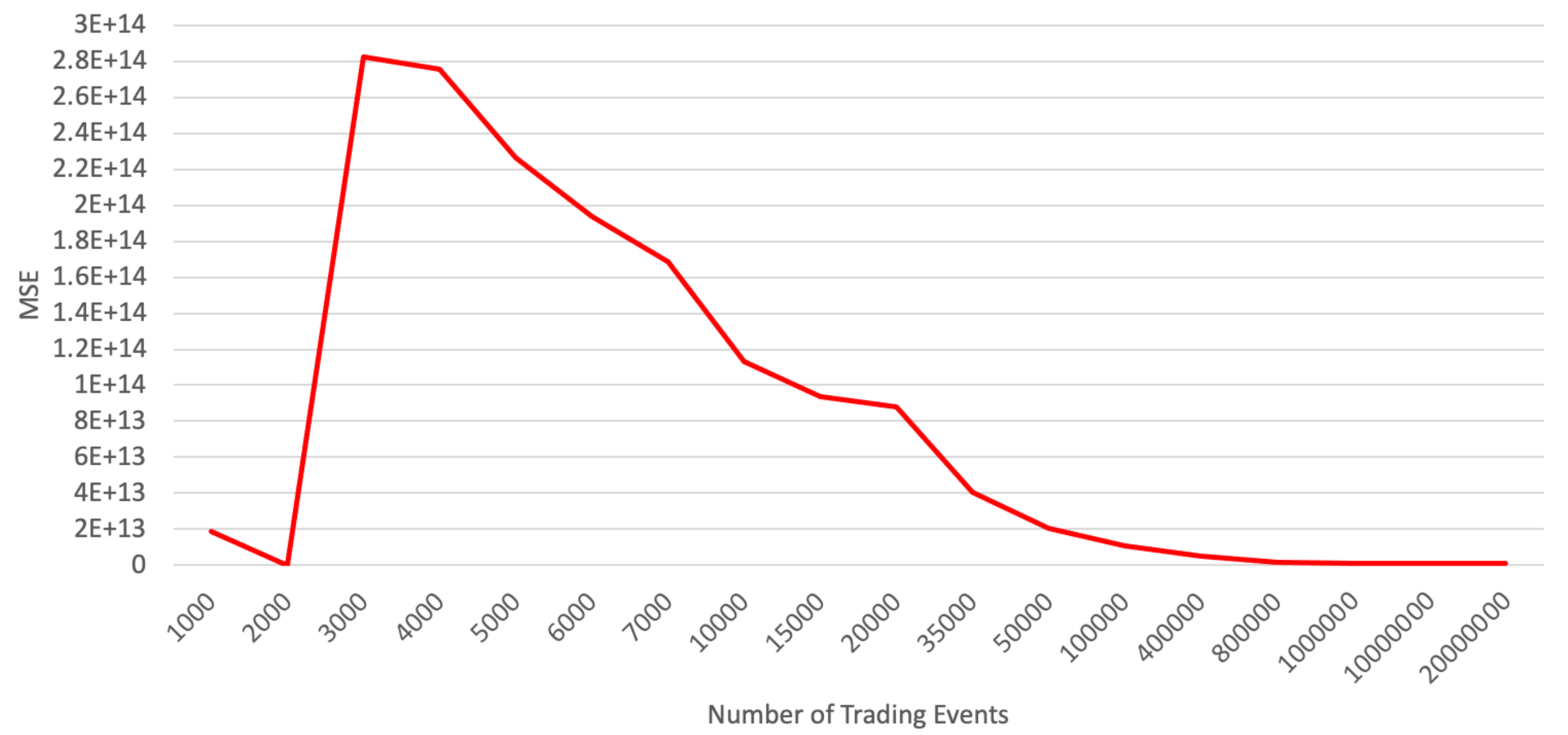}}}
  \subfloat[OPTM-LSTM testing MSE scores \label{1b}]{%
        \scalebox{0.55}{\includegraphics[width=0.65\linewidth]{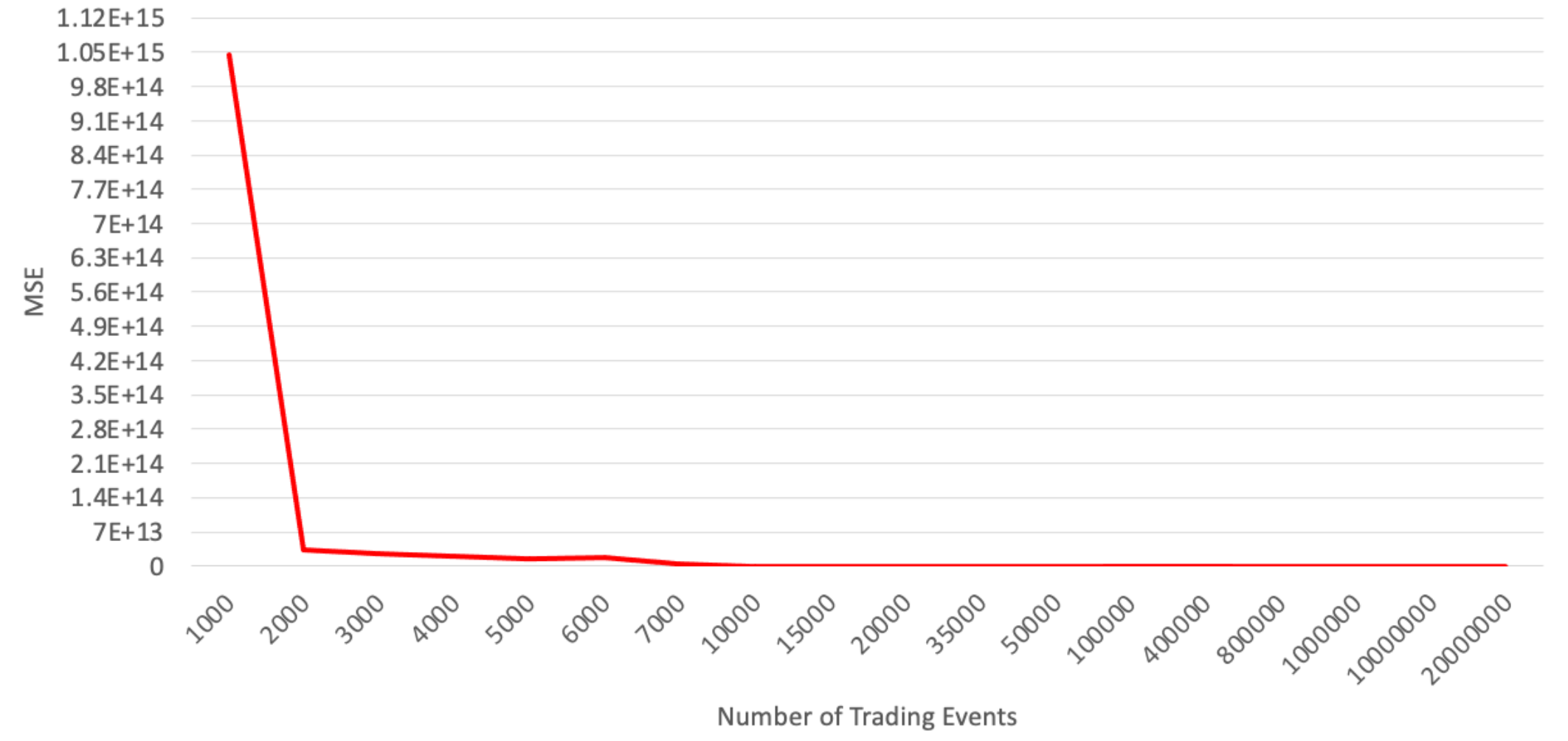}}}
    \\
  \subfloat[LSTM training MSE scores \label{1c}]{%
        \scalebox{0.55}{\includegraphics[width=0.65\linewidth]{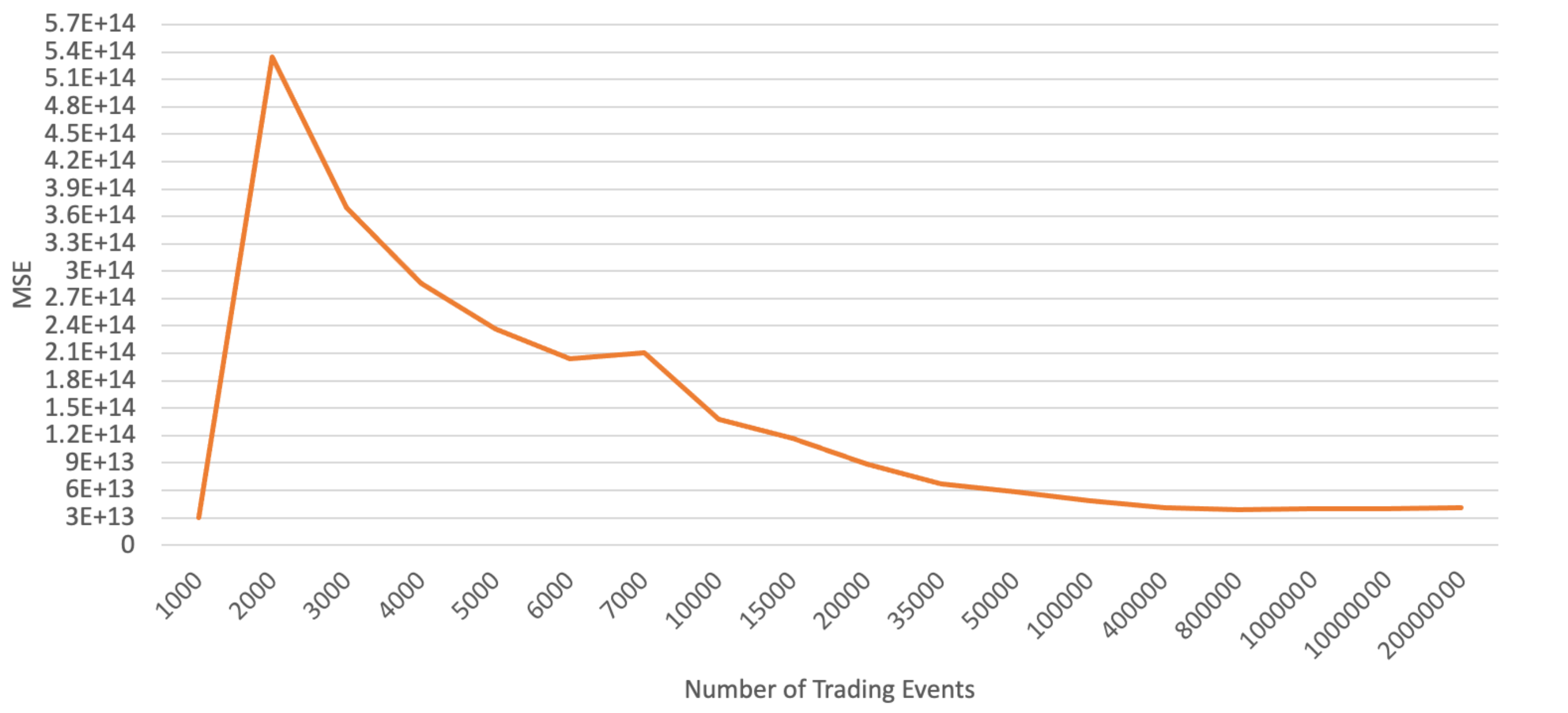}}}
  \subfloat[LSTM testing MSE scores \label{1d}]{%
        \scalebox{0.55}{\includegraphics[width=0.65\linewidth]{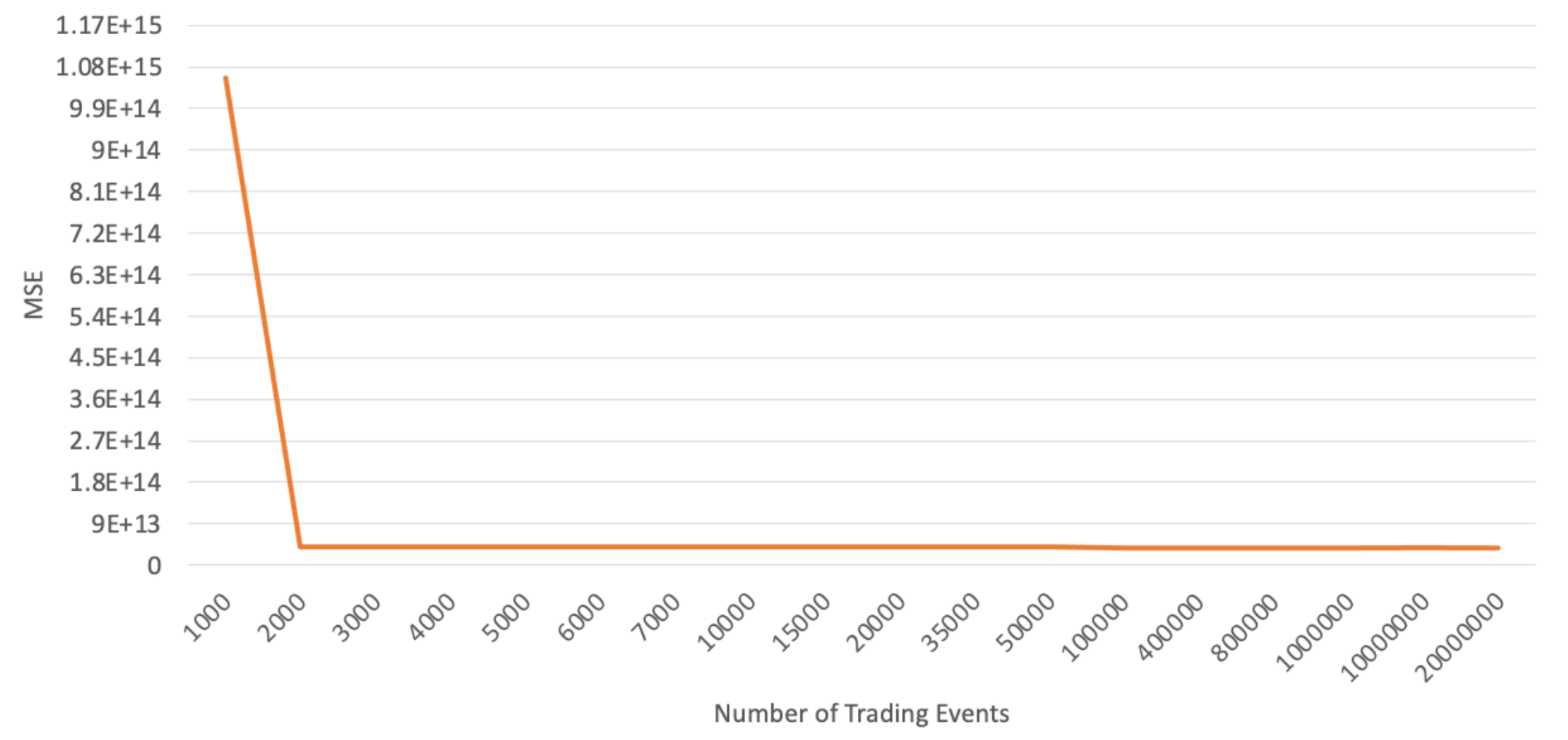}}}
    \\
  \subfloat[Attention LSTM training MSE scores\label{1c}]{%
        \scalebox{0.55}{\includegraphics[width=0.65\linewidth]{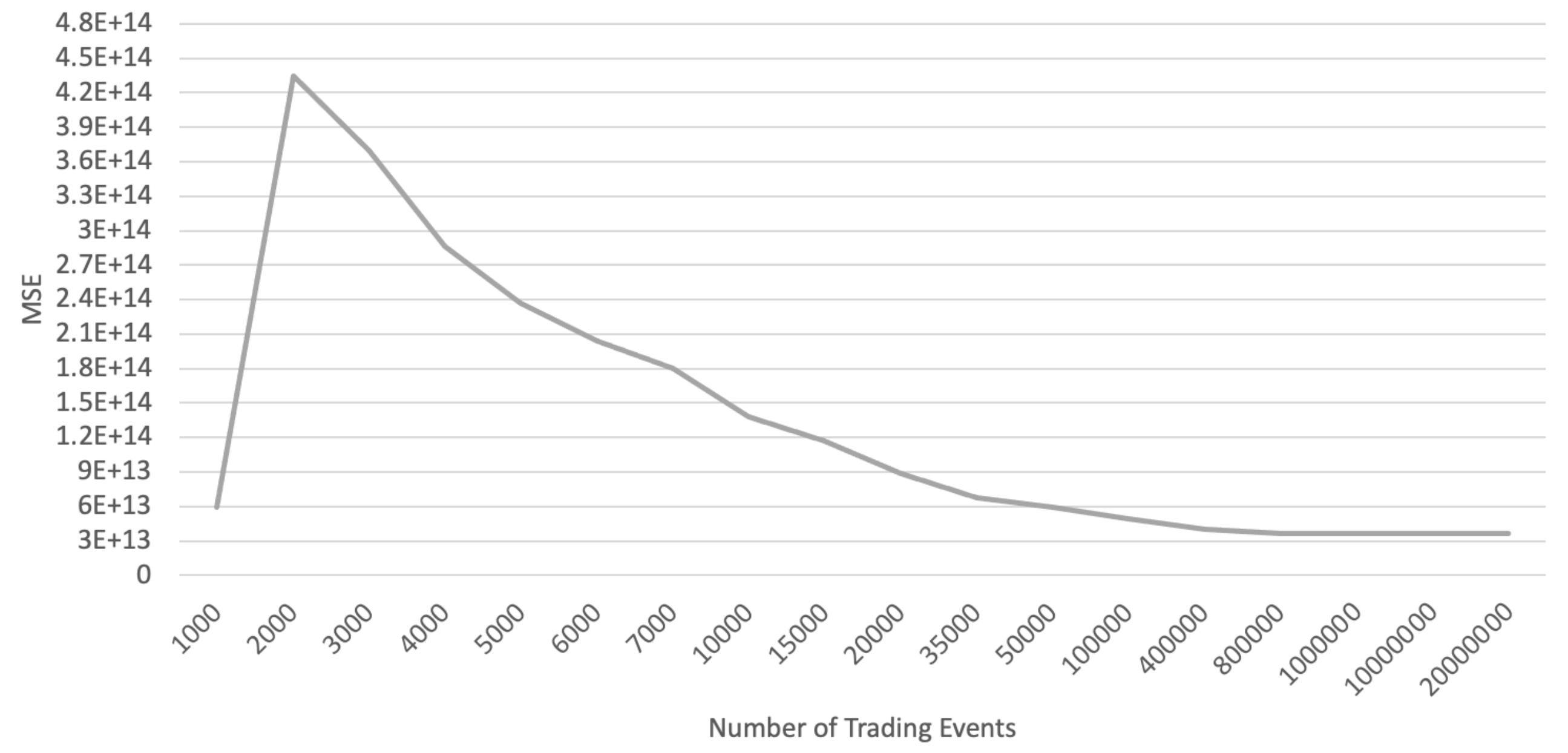}}}
  \subfloat[Attention LSTM testing MSE scores \label{1d}]{%
        \scalebox{0.55}{\includegraphics[width=0.65\linewidth]{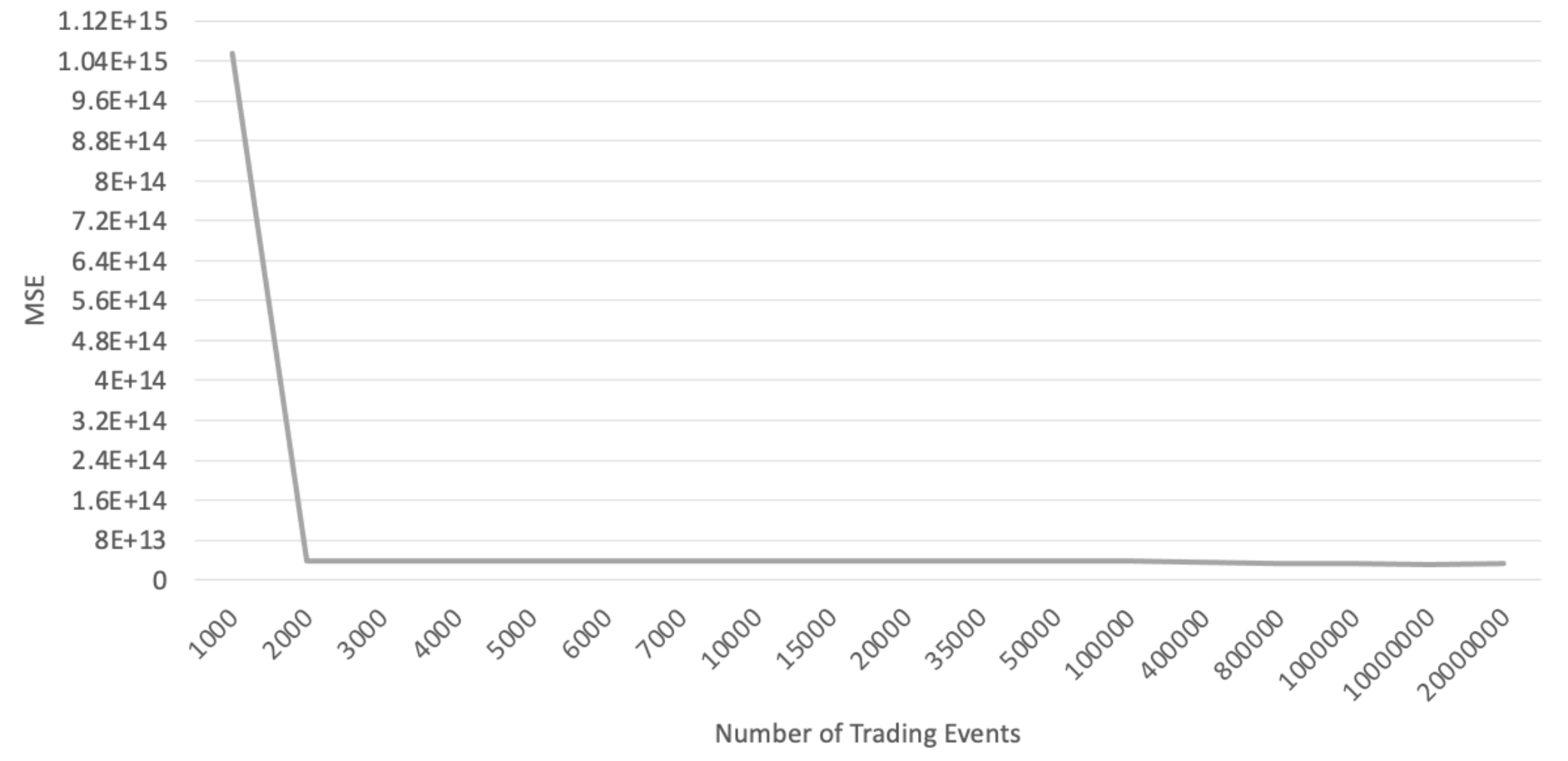}}}  
    \\
  \subfloat[Bidirectional training MSE scores \label{1a}]{%
       \scalebox{0.55}{\includegraphics[width=0.65\linewidth]{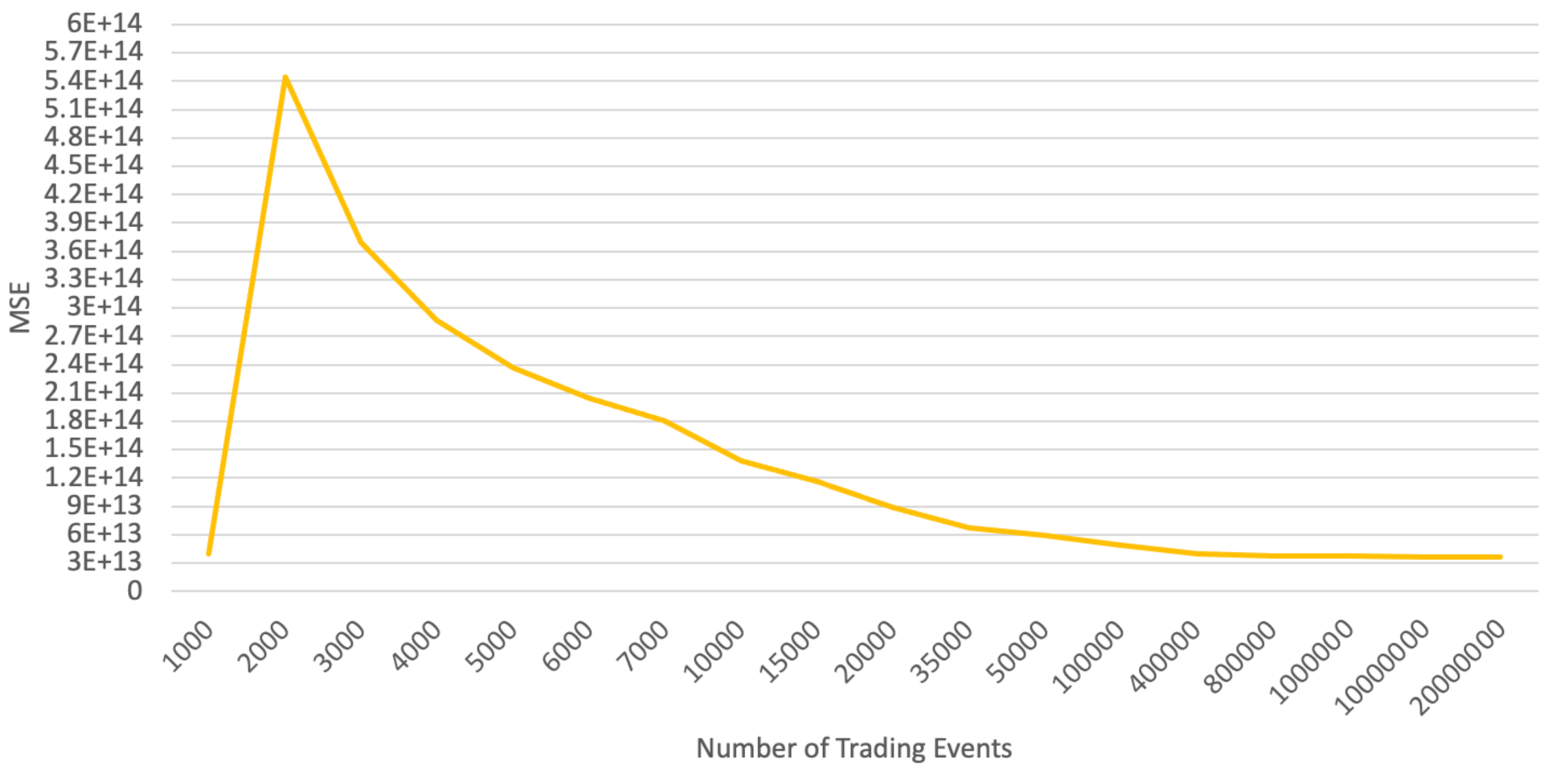}}}
  \subfloat[Bidirectional testing MSE scores \label{1b}]{%
        \scalebox{0.55}{\includegraphics[width=0.65\linewidth]{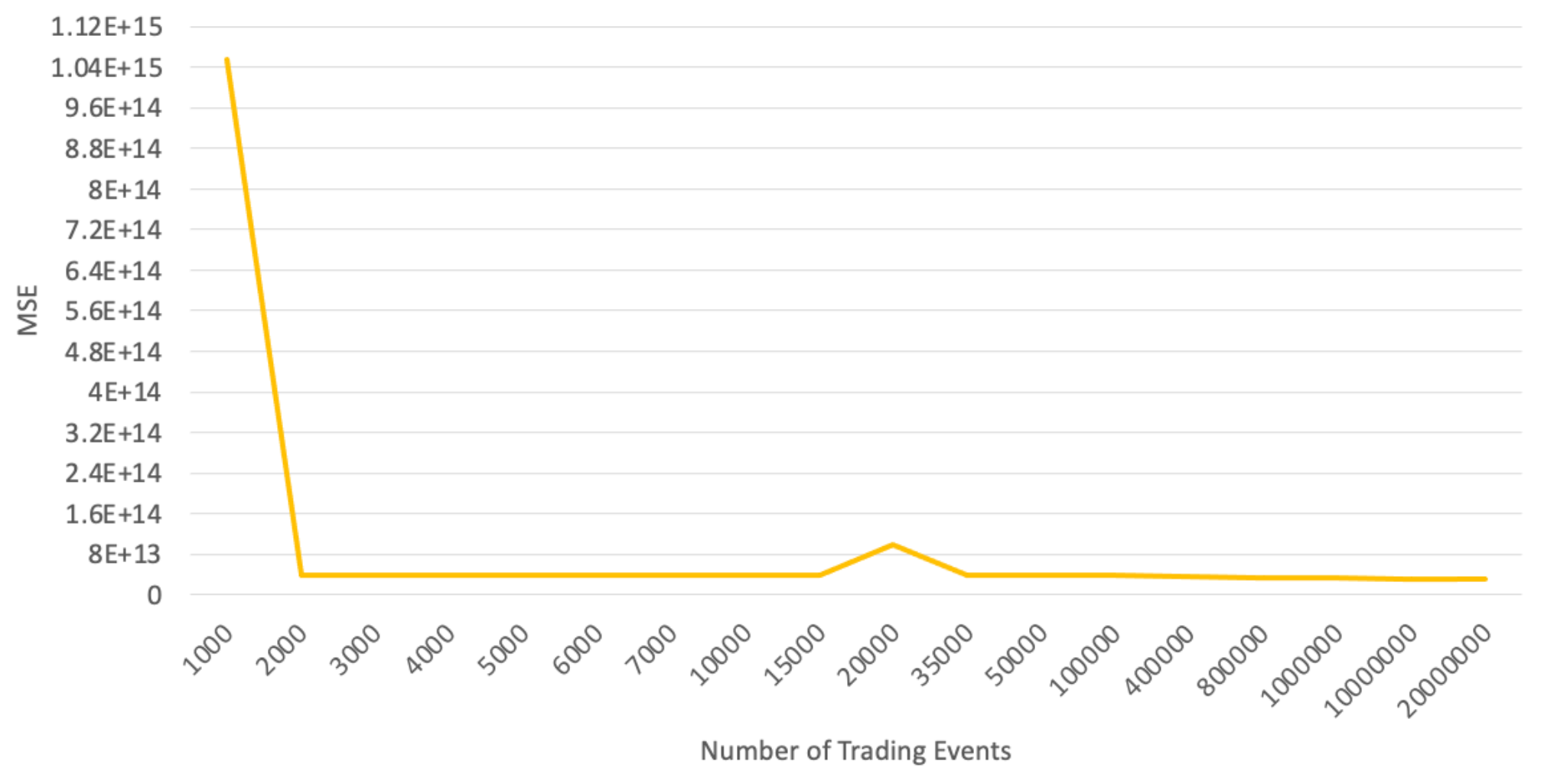}}}
    \\
  \subfloat[GRU training MSE scores \label{1a}]{%
       \scalebox{0.55}{\includegraphics[width=0.65\linewidth]{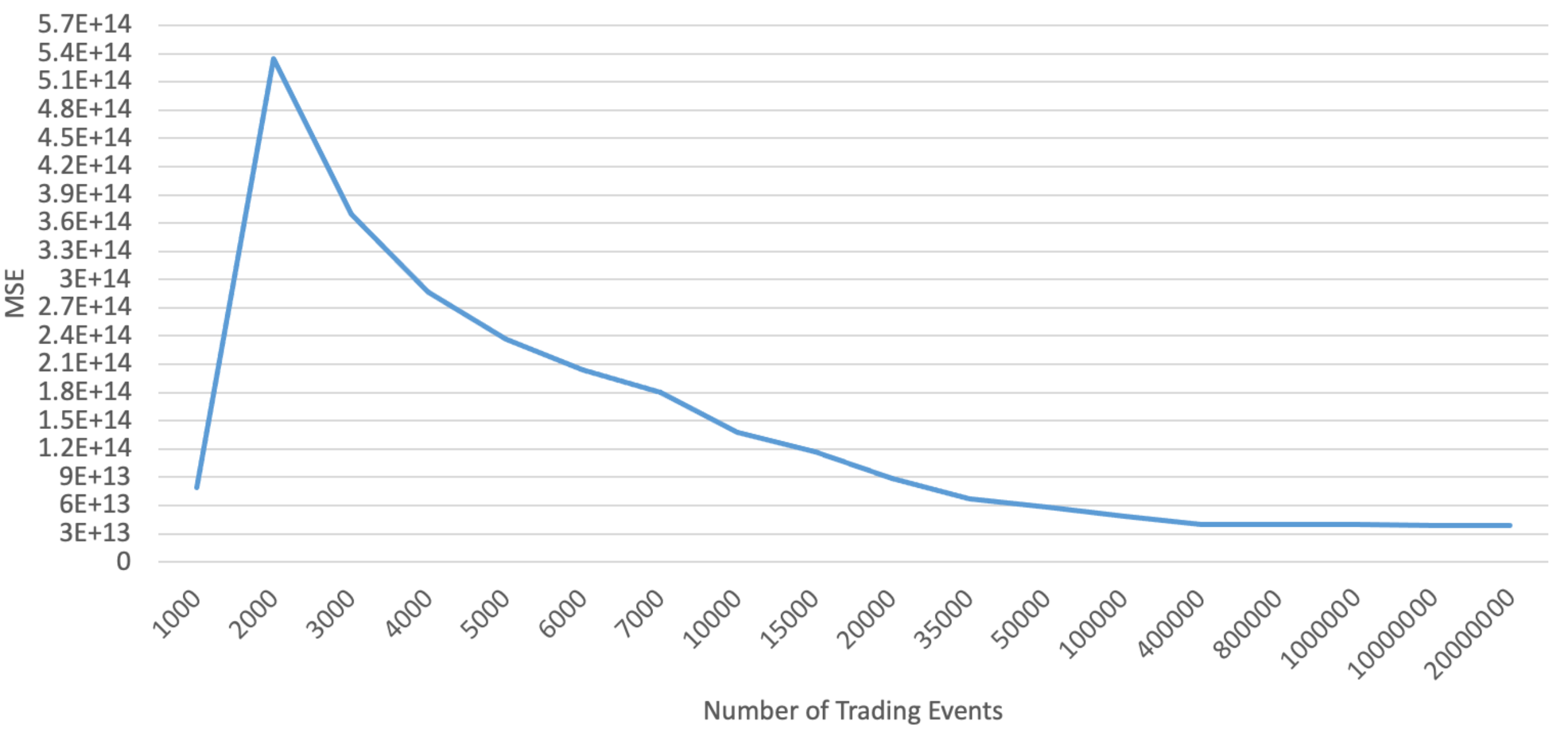}}}
  \subfloat[GRU testing MSE scores \label{1b}]{%
        \scalebox{0.55}{\includegraphics[width=0.65\linewidth]{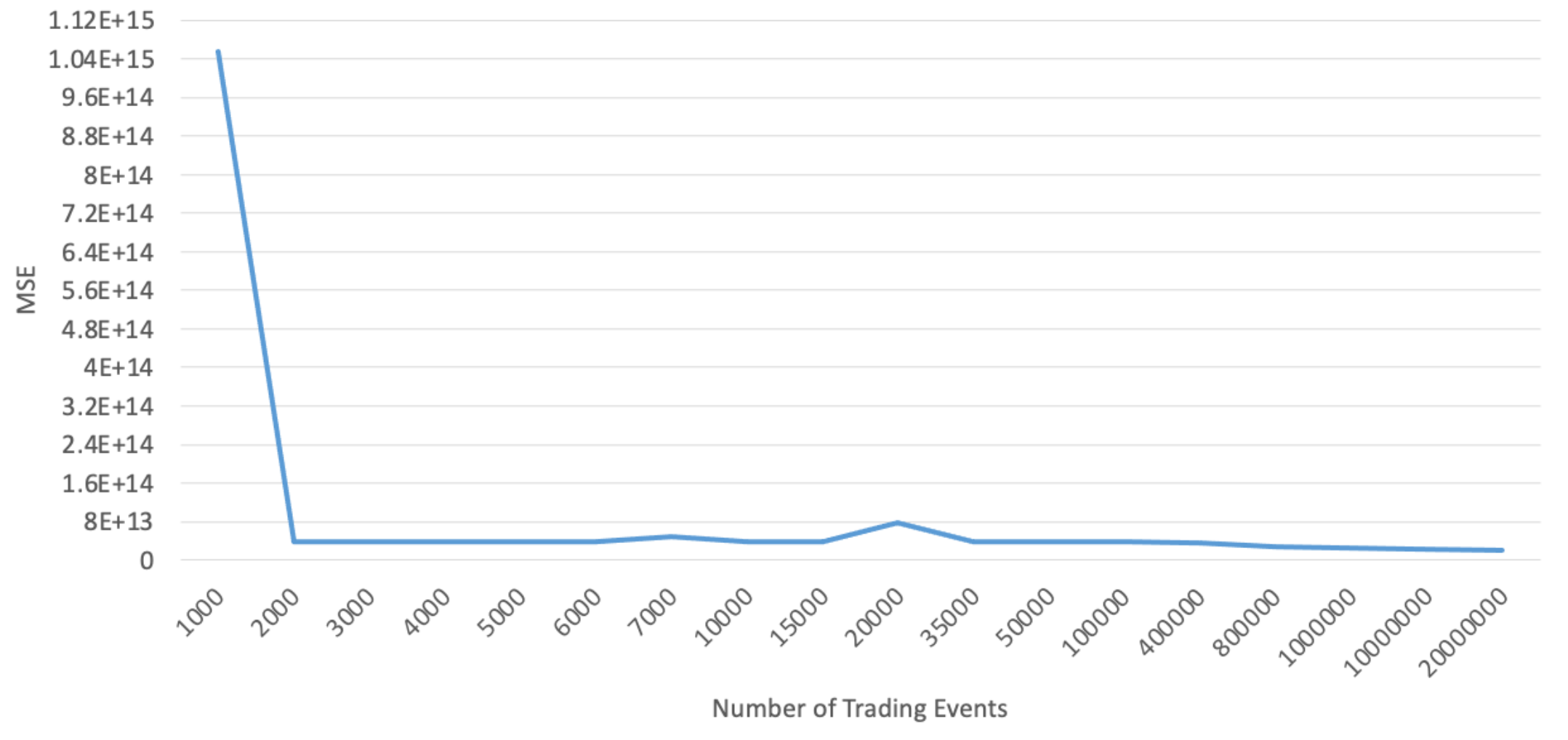}}}
    \\
  \subfloat[Hybrid training MSE scores \label{1a}]{%
       \scalebox{0.55}{\includegraphics[width=0.65\linewidth]{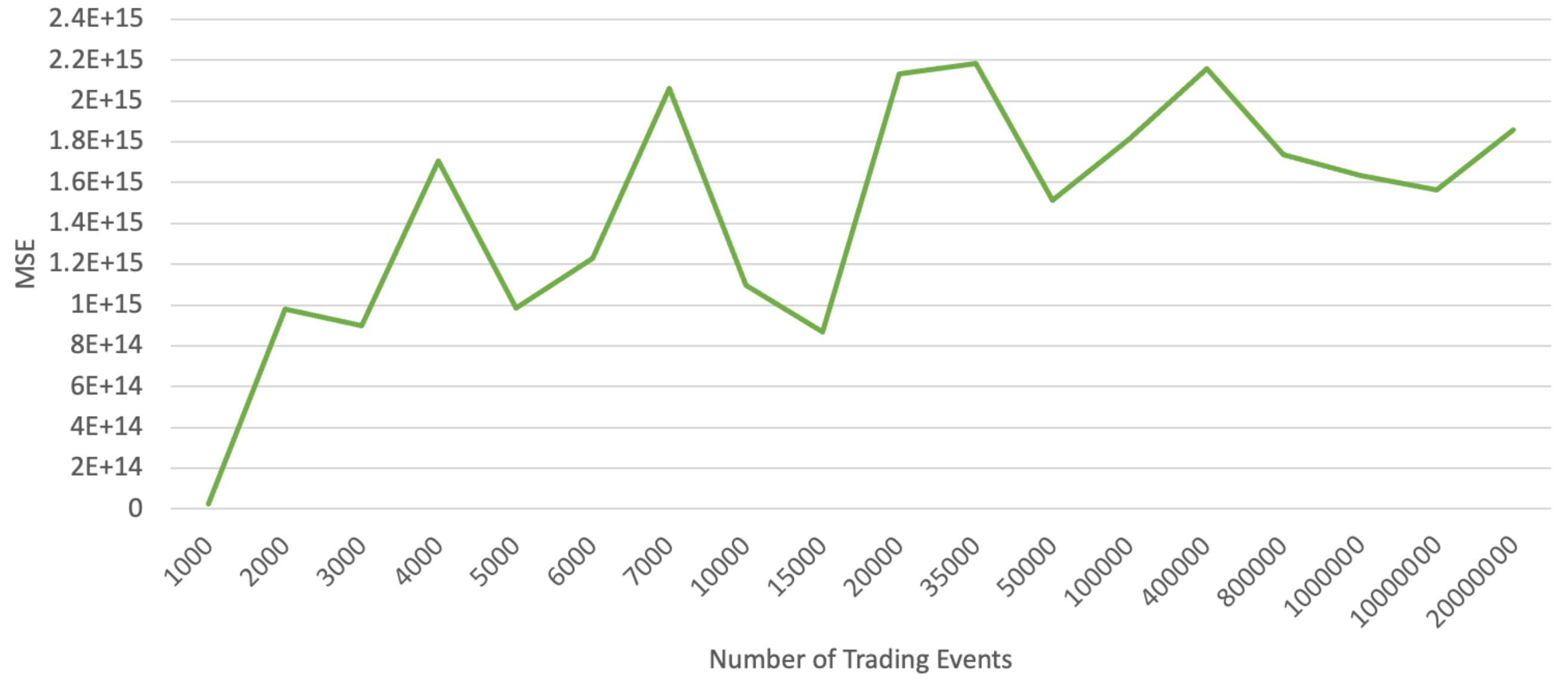}}}
  \subfloat[Hybrid testing MSE scores \label{1b}]{%
        \scalebox{0.55}{\includegraphics[width=0.65\linewidth]{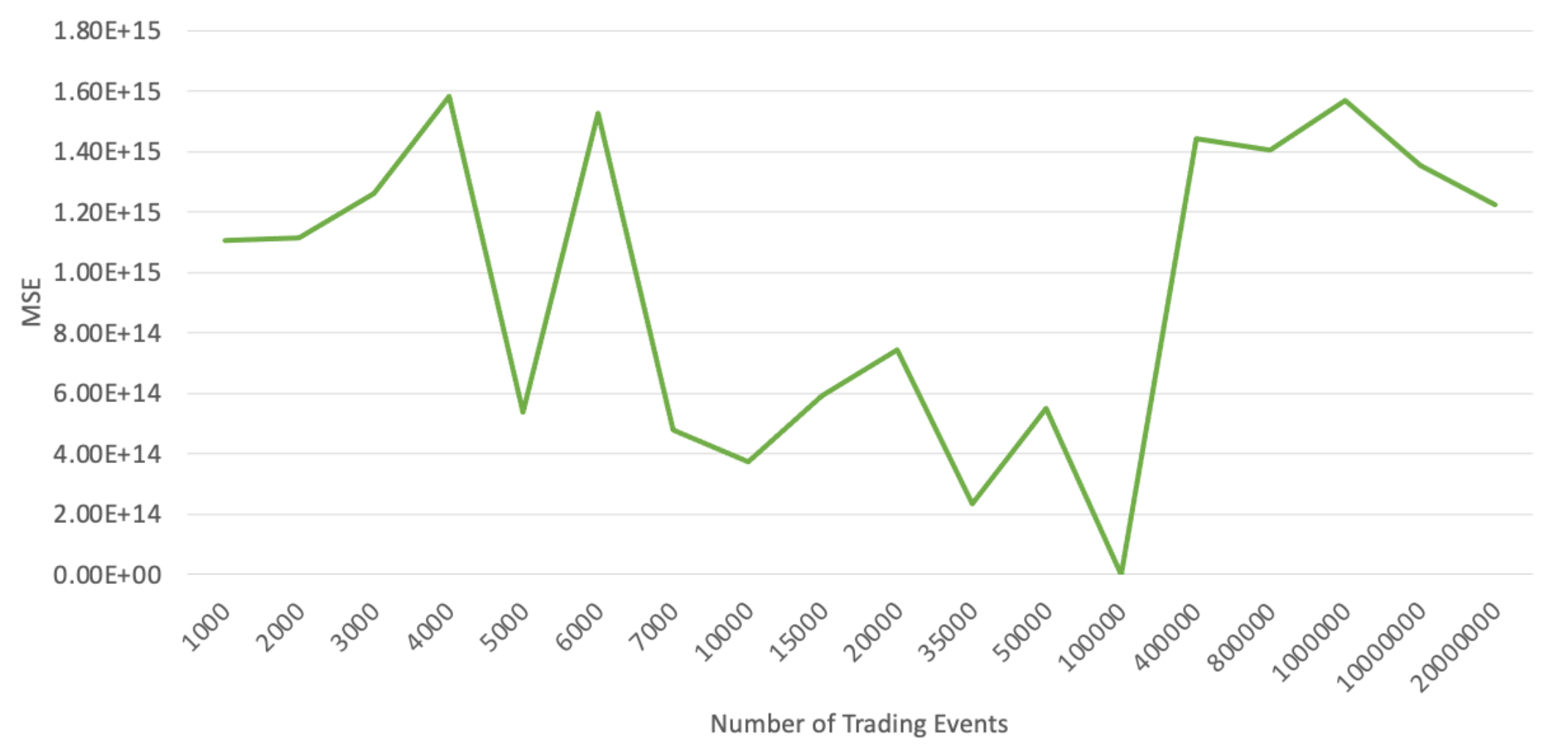}}}  
  \caption{Google Short MSE scores based on \hyperref[tab:GoogleShort]{Table \ref{tab:GoogleShort}}.}
  \label{fig:GoogleShort} 
\end{figure*}

\begin{figure*}[hbt!]
    \centering
  \subfloat[OPTM-LSTM training MSE scores \label{1a}]{%
       \scalebox{0.55}{\includegraphics[width=0.65\linewidth]{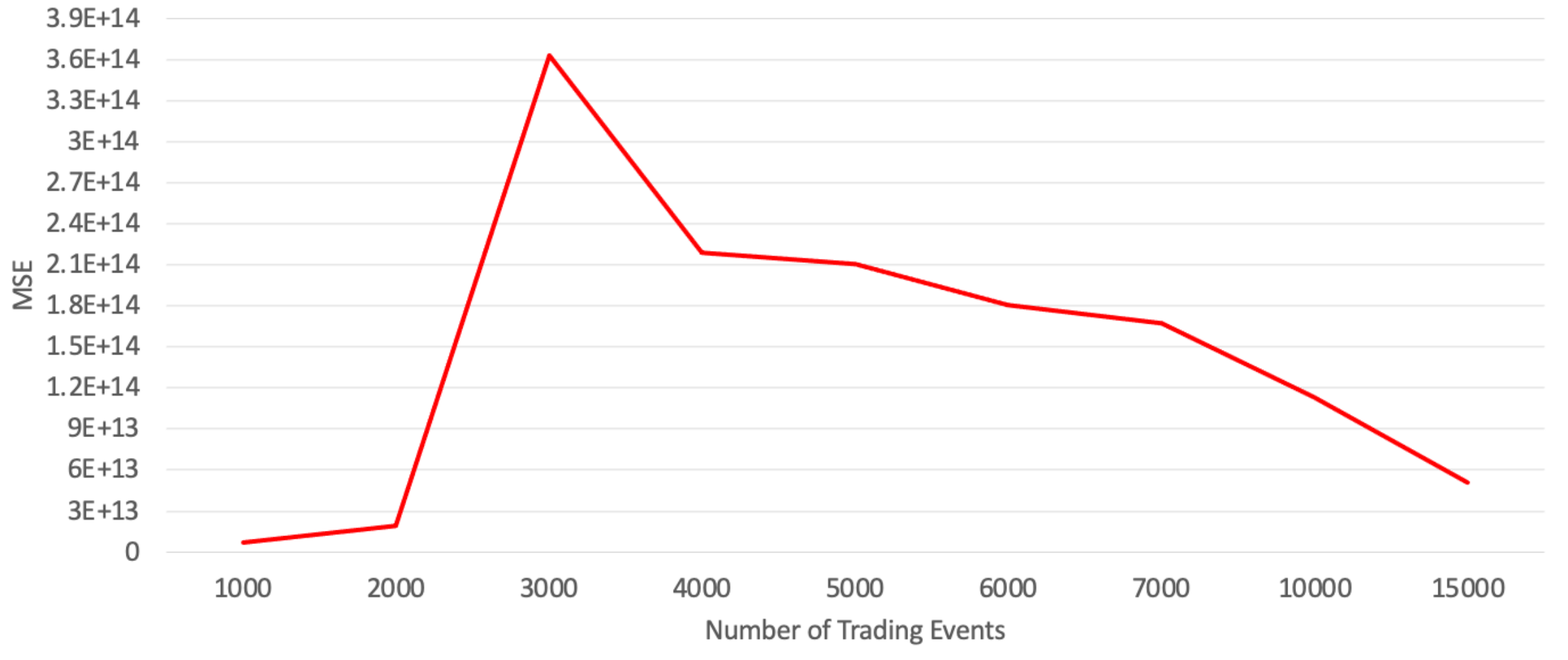}}}
  \subfloat[OPTM-LSTM testing MSE scores \label{1b}]{%
        \scalebox{0.55}{\includegraphics[width=0.65\linewidth]{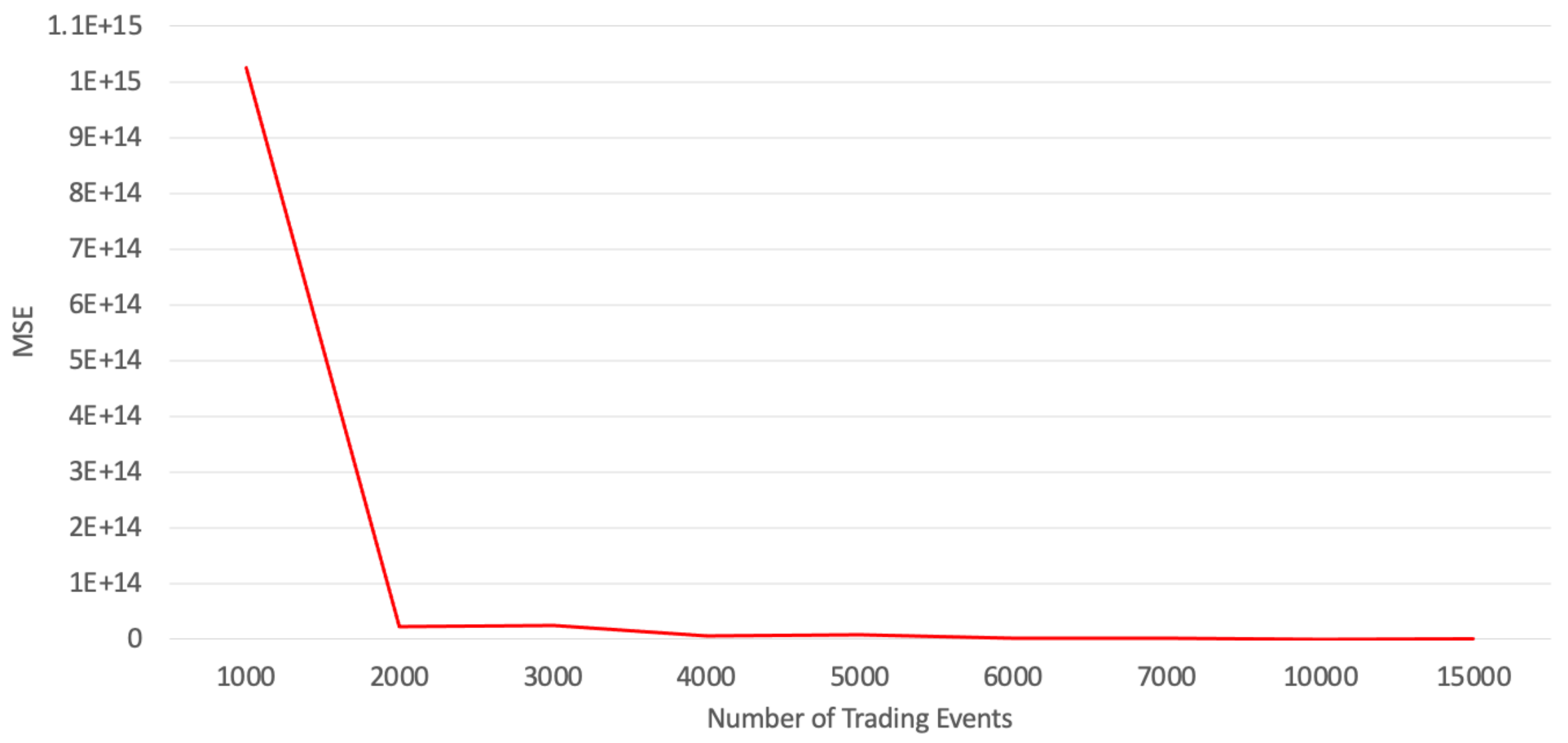}}}
    \\
  \subfloat[LSTM training MSE scores \label{1c}]{%
        \scalebox{0.55}{\includegraphics[width=0.65\linewidth]{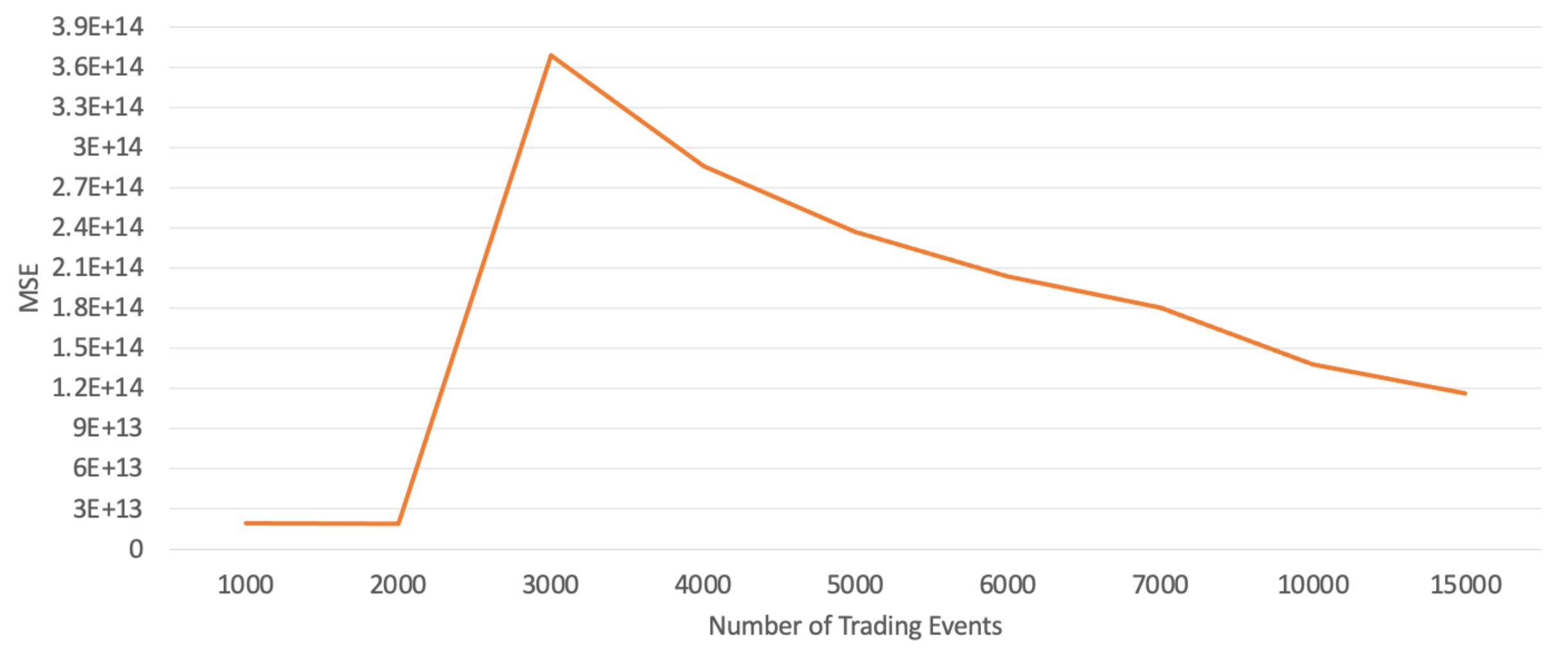}}}
  \subfloat[LSTM testing MSE scores \label{1d}]{%
        \scalebox{0.55}{\includegraphics[width=0.65\linewidth]{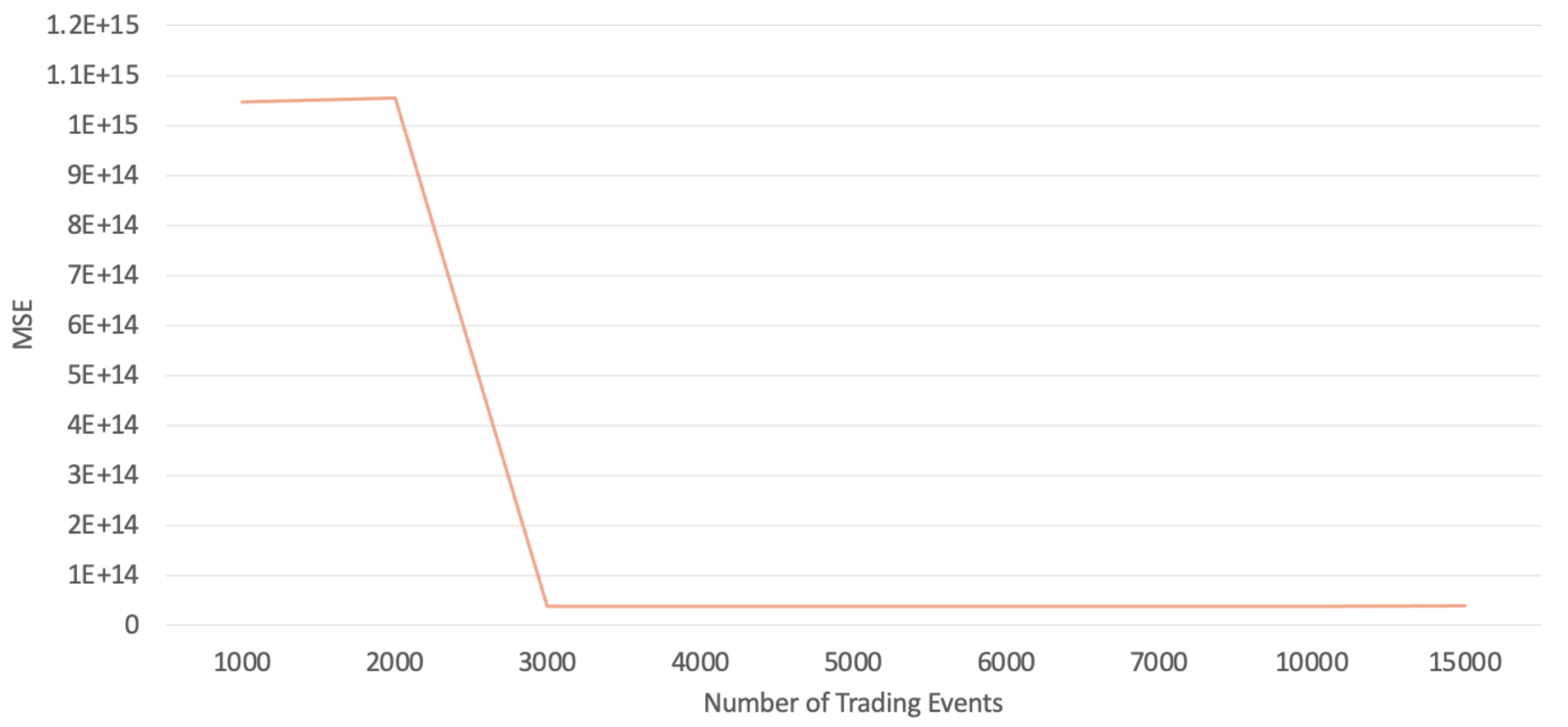}}}
    \\
  \subfloat[Attention LSTM training MSE scores\label{1c}]{%
        \scalebox{0.55}{\includegraphics[width=0.65\linewidth]{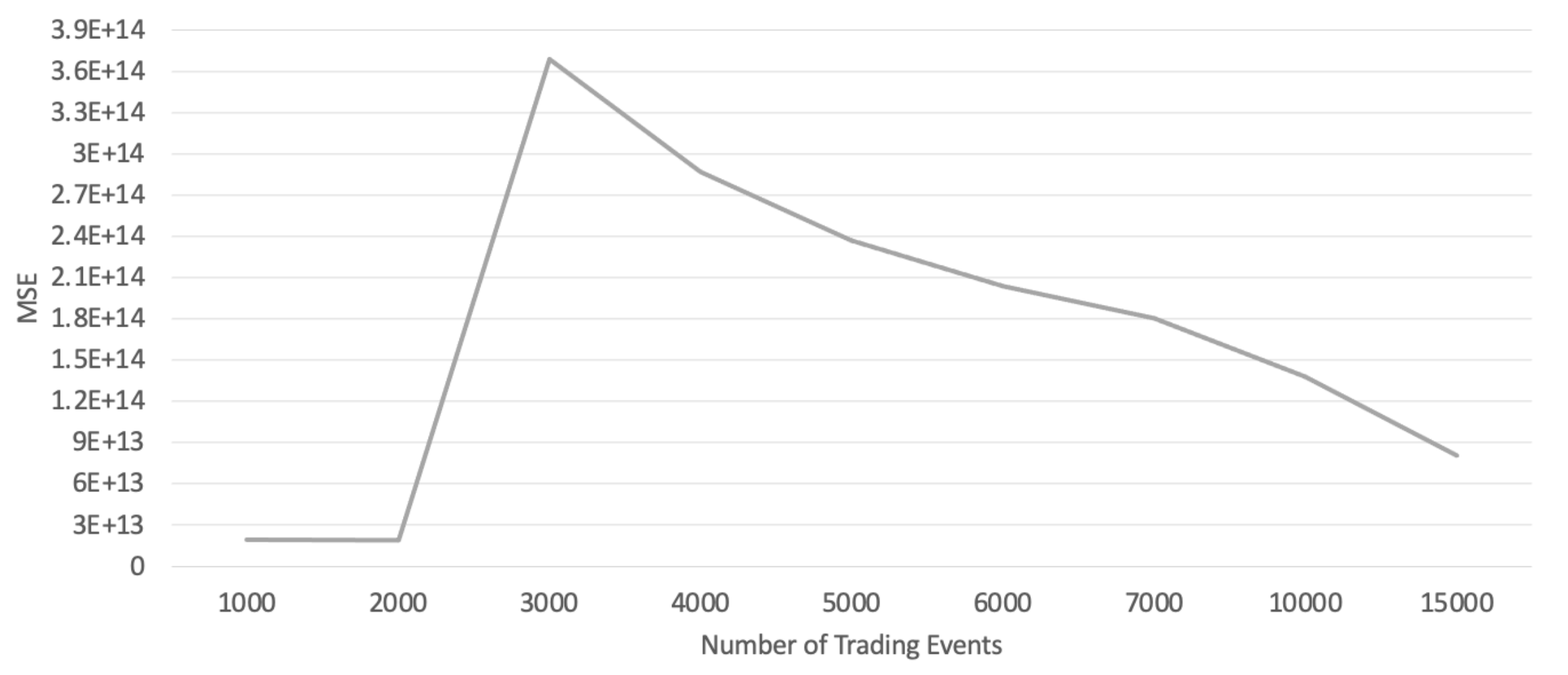}}}
  \subfloat[Attention LSTM testing MSE scores \label{1d}]{%
        \scalebox{0.55}{\includegraphics[width=0.65\linewidth]{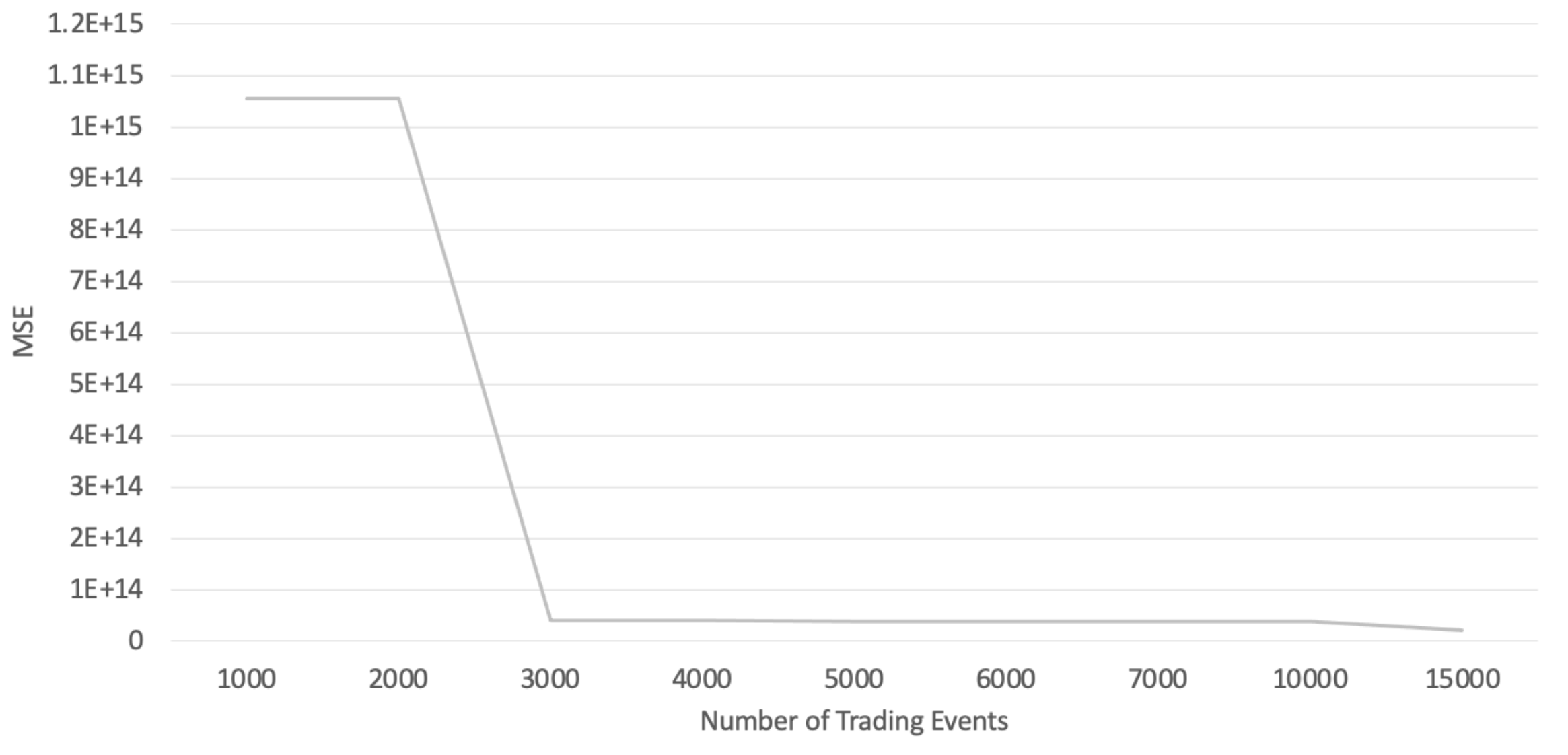}}}  
    \\
  \subfloat[Bidirectional training MSE scores \label{1a}]{%
       \scalebox{0.55}{\includegraphics[width=0.65\linewidth]{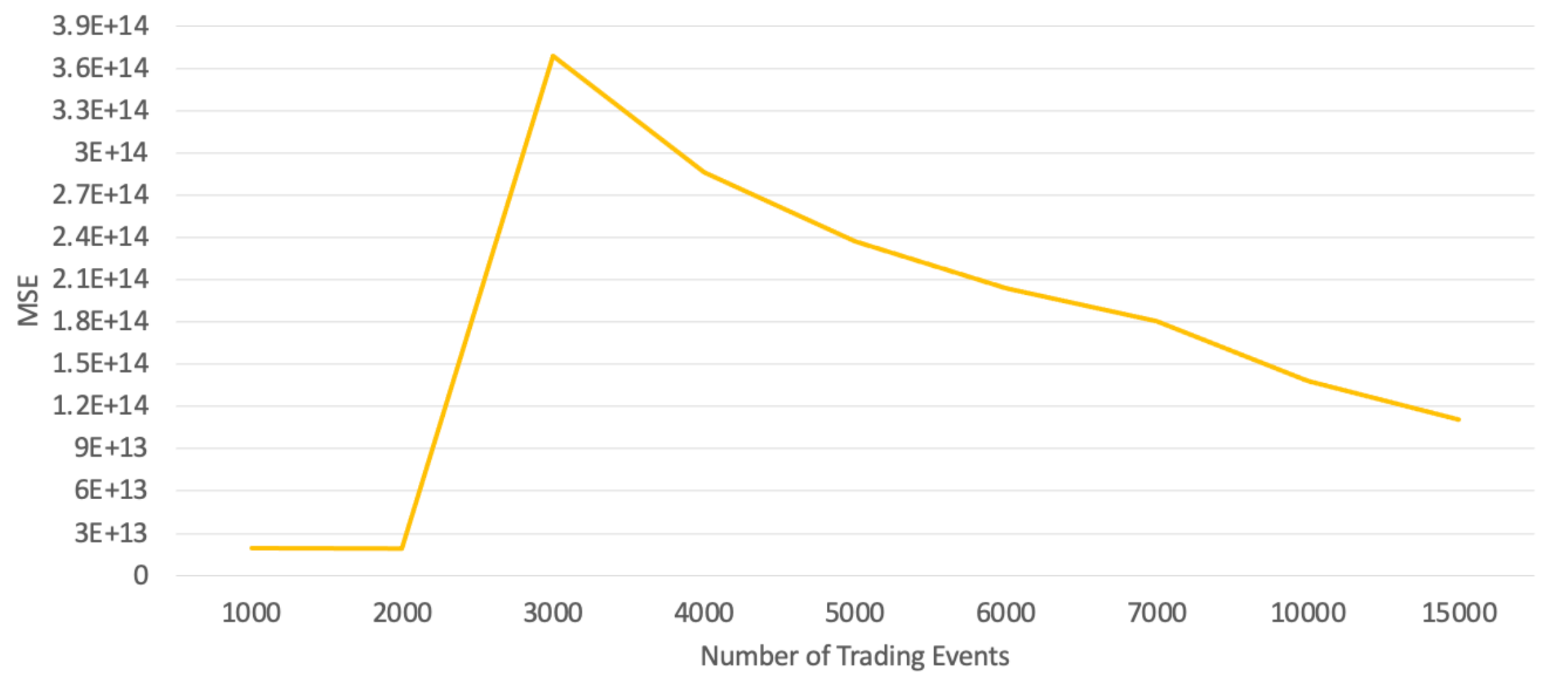}}}
  \subfloat[Bidirectional testing MSE scores \label{1b}]{%
        \scalebox{0.55}{\includegraphics[width=0.65\linewidth]{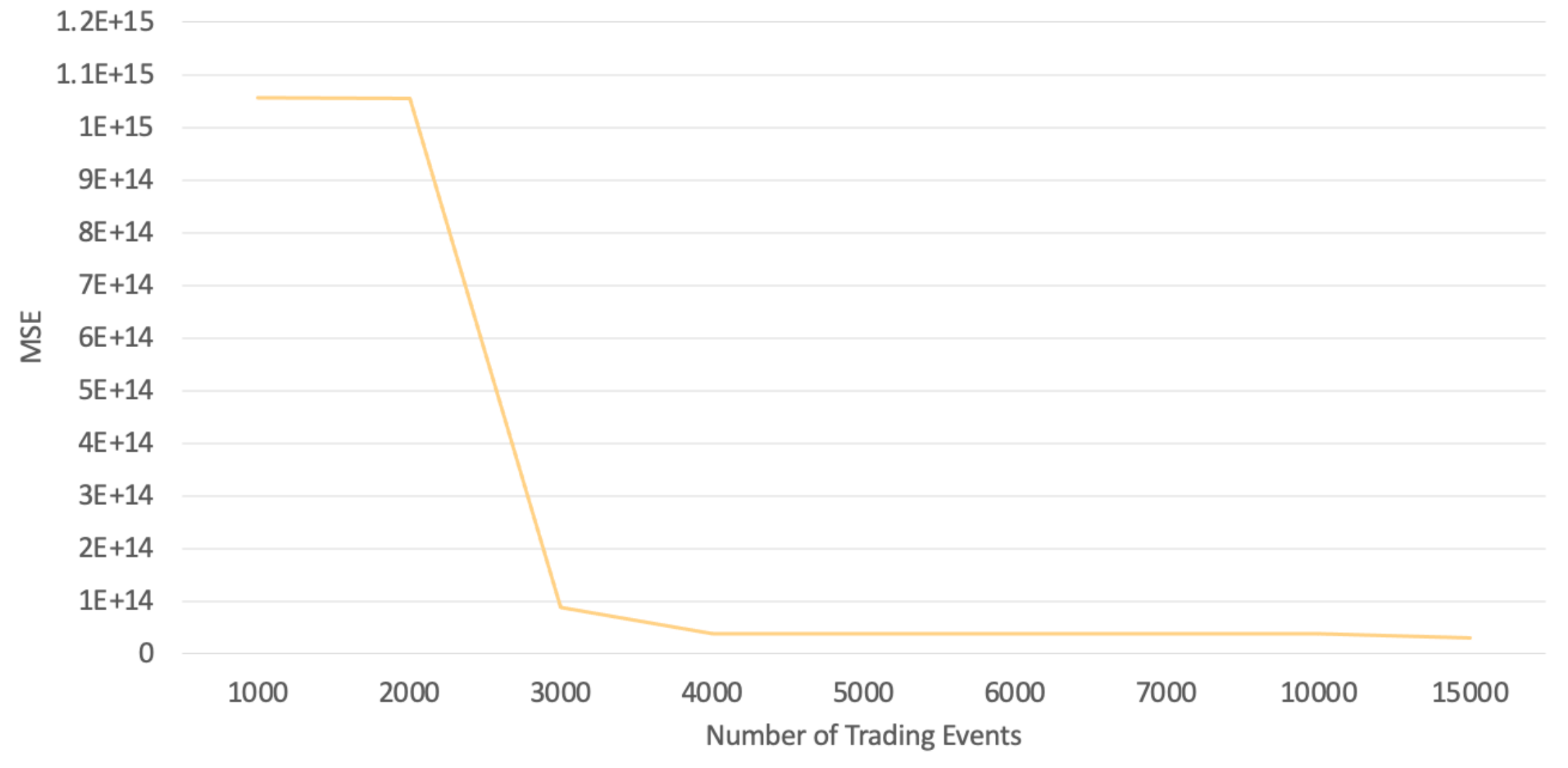}}}
    \\
  \subfloat[GRU training MSE scores \label{1a}]{%
       \scalebox{0.55}{\includegraphics[width=0.65\linewidth]{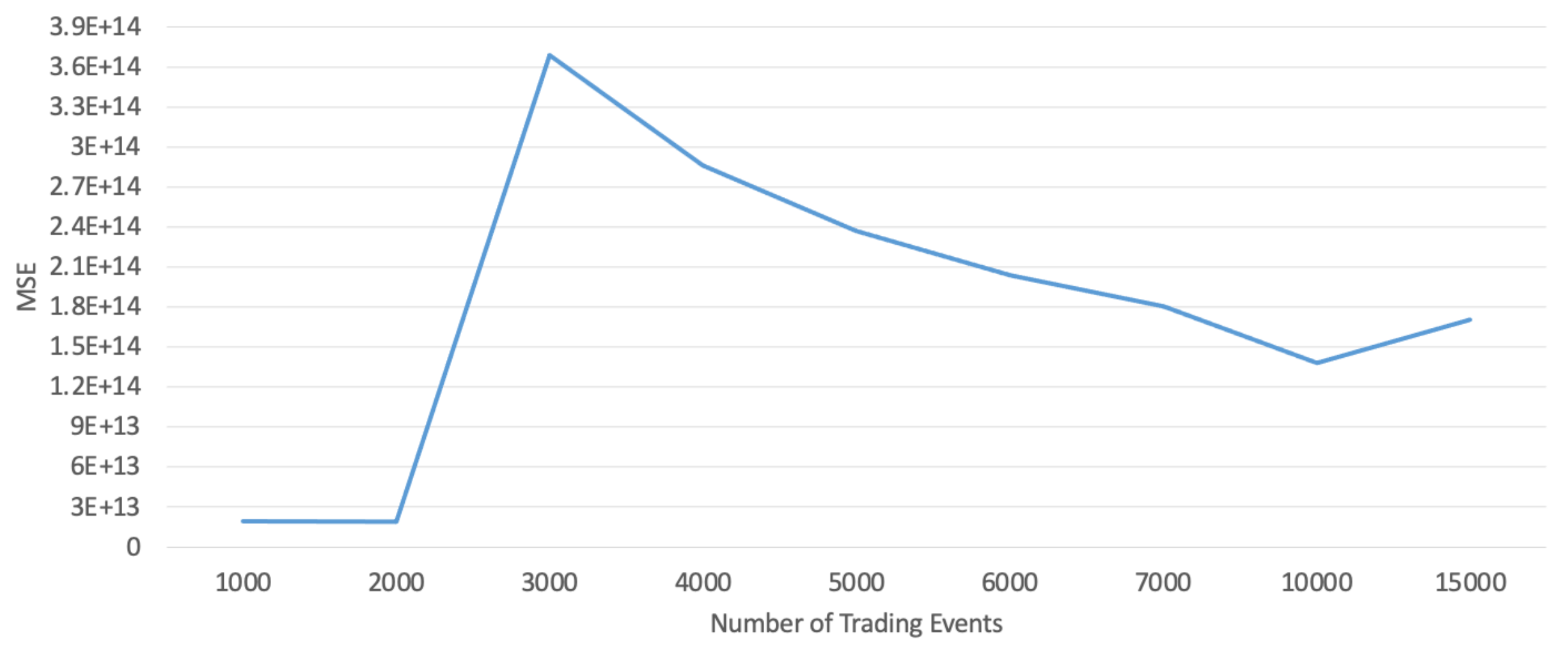}}}
  \subfloat[GRU testing MSE scores \label{1b}]{%
        \scalebox{0.55}{\includegraphics[width=0.65\linewidth]{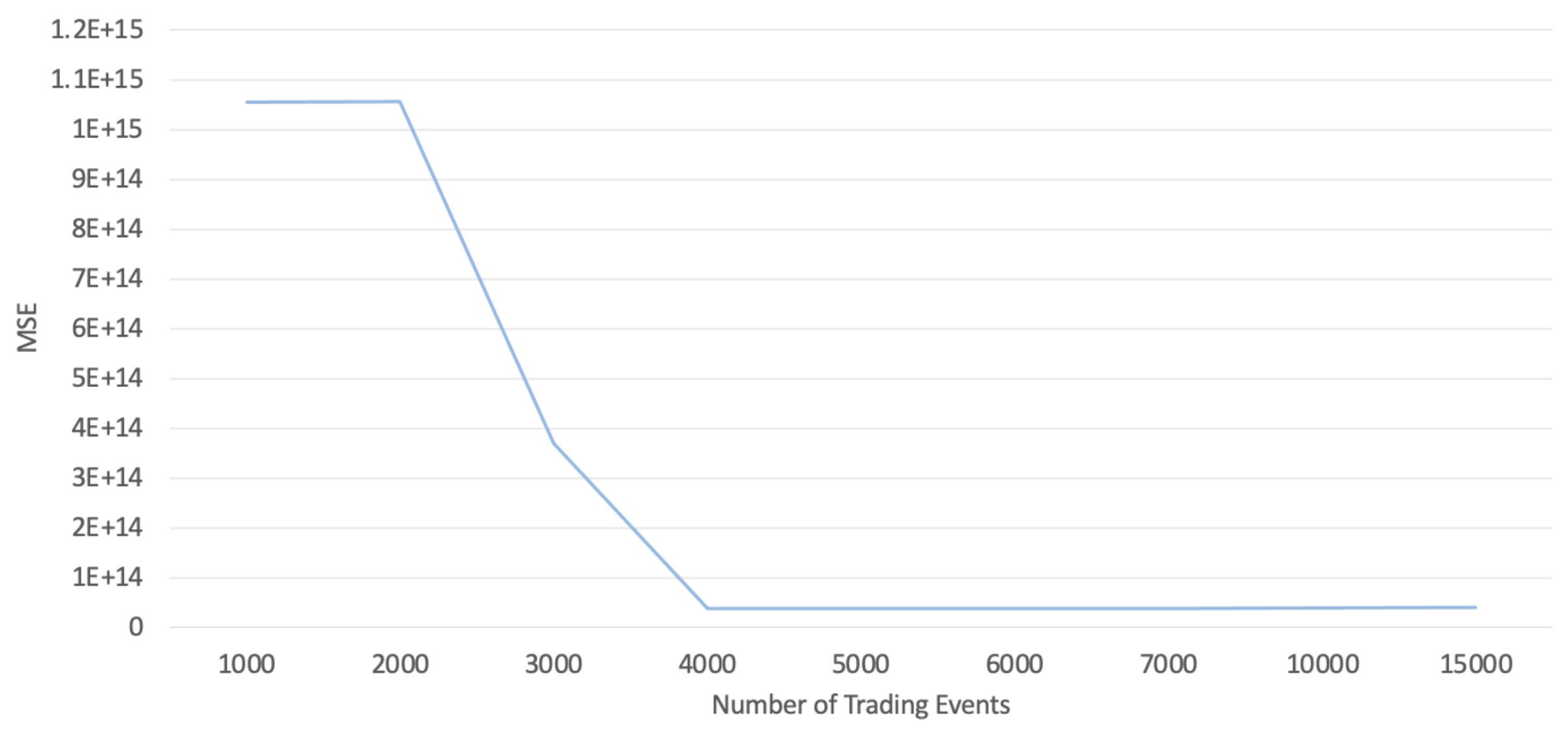}}}
    \\
  \subfloat[Hybrid training MSE scores \label{1a}]{%
       \scalebox{0.55}{\includegraphics[width=0.65\linewidth]{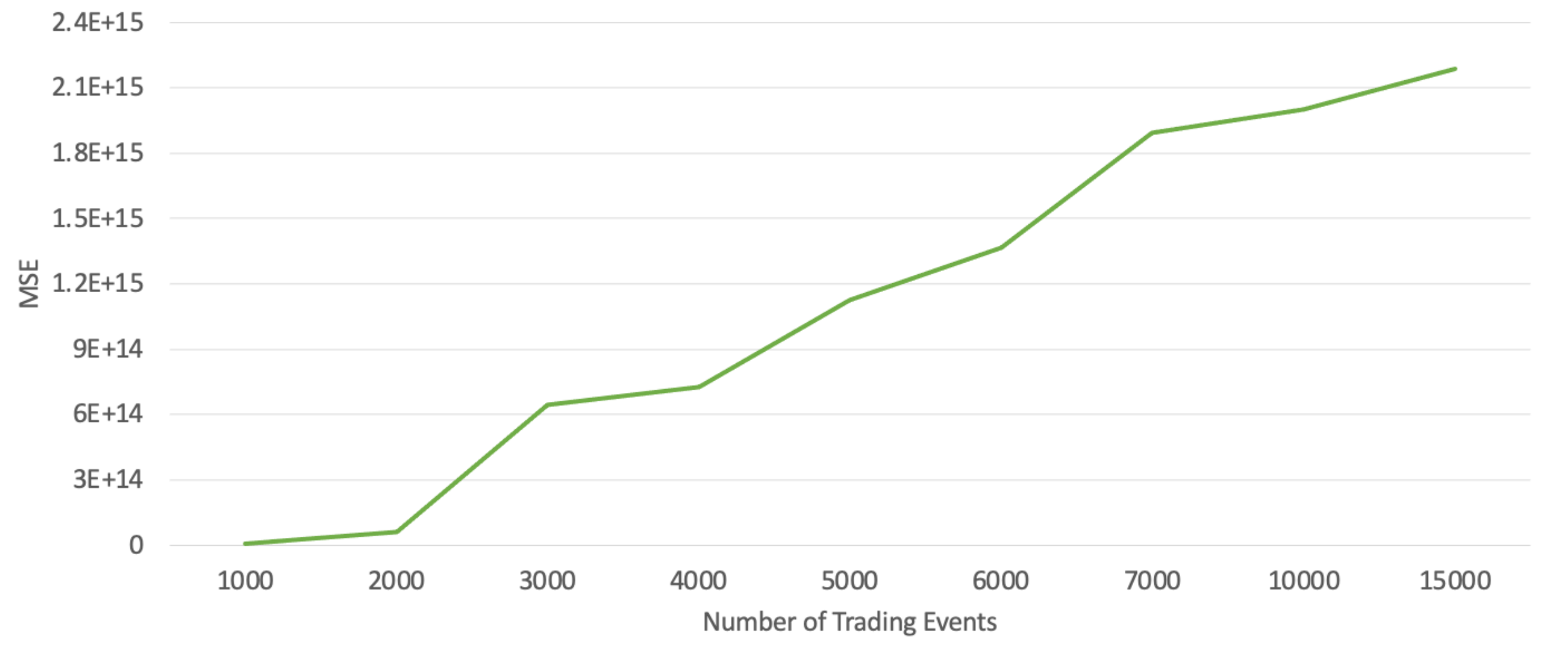}}}
  \subfloat[Hybrid testing MSE scores \label{1b}]{%
        \scalebox{0.55}{\includegraphics[width=0.65\linewidth]{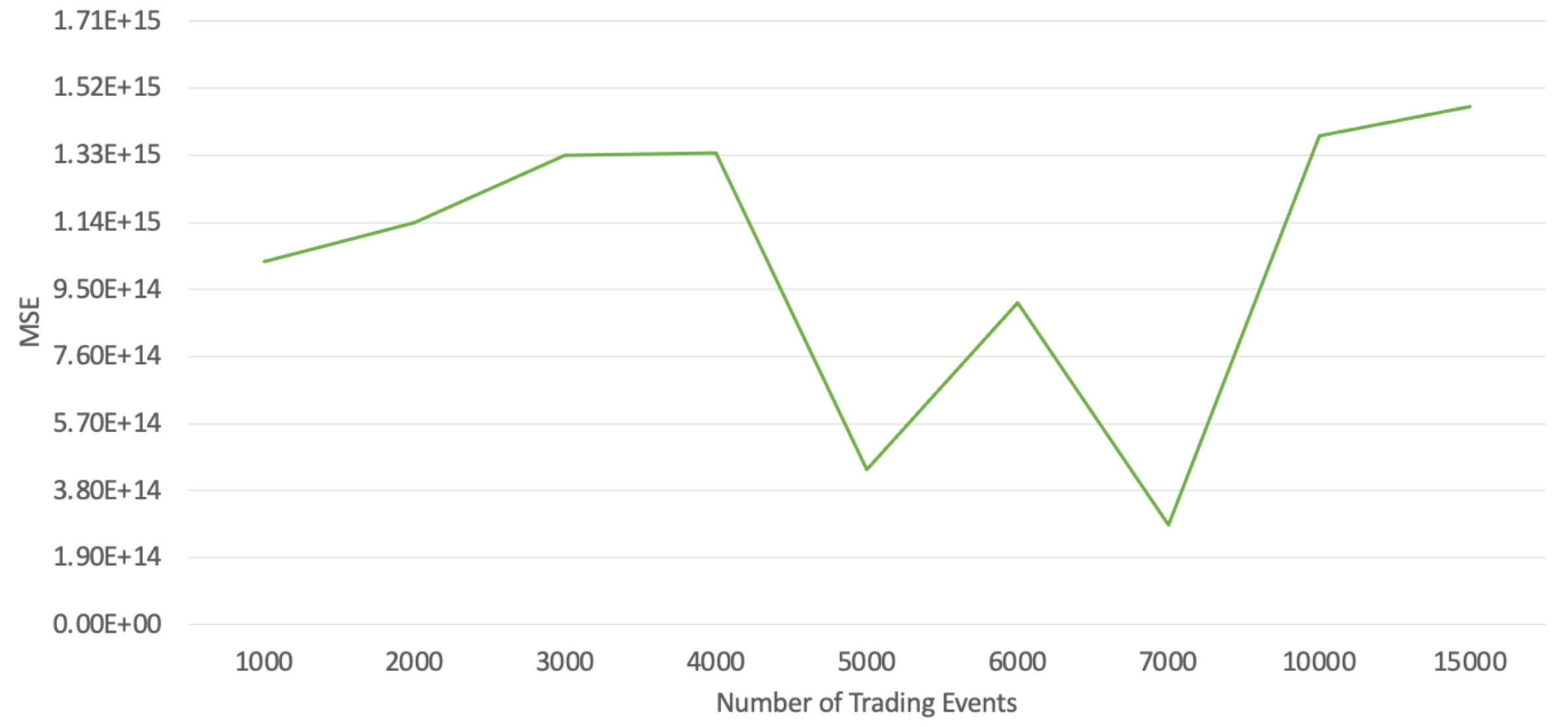}}}  
  \caption{Google Long MSE scores based on \hyperref[tab:GoogleLong]{Table \ref{tab:GoogleLong}}.}
  \label{fig:GoogleLong} 
\end{figure*}

For space economy, we present the case of Google in \hyperref[tab:GoogleShort]{Table \ref{tab:GoogleShort}} and \hyperref[tab:GoogleLong]{Table \ref{tab:GoogleLong}} in the the main body of the text and the rest of the stocks (i.e., Amazon, Kesko, and Wartsila) can be found in \hyperref[sec:SecondAppendix]{Appendix \ref{sec:SecondAppendix}}. The Short and Long experimental results tables are combined with time-series plots (e.g., \hyperref[fig:GoogleShort]{Fig. \ref{fig:GoogleShort}} and \hyperref[fig:GoogleLong]{Fig. \ref{fig:GoogleLong}}). 

\begin{table*}[!htpb]
\centering
\captionsetup{width=.70\textwidth}
\caption{Google MSE scores under the Short experimental protocol.}
\scalebox{0.60}{
\begin{tabular}{rcrlcrcrlcrcrl}
\cmidrule[2pt]{1-6}\cmidrule[2pt]{6-9}\cmidrule[2pt]{9-14}
\textbf{Stock} & \textbf{Size} & \textbf{Model} & \textbf{MSE - Train} & \qquad & \textbf{Stock} & \textbf{Size} & \textbf{Model} & \textbf{MSE - Train} & \qquad & \textbf{Stock} & \textbf{Size} & \textbf{Model} & \textbf{MSE - Train}\\
\cmidrule{1-4}\cmidrule{6-9}\cmidrule{11-14}
Google & 1,000 & \textbf{OPTM-LSTM}& \textbf{1.86157E+13} & \qquad &   Google    &  2,000 & \textbf{OPTM-LSTM}    & \textbf{6.81980E+04} & \qquad &  Google & 3,000 & \textbf{OPTM-LSTM}       & \textbf{2.82656E+14}\\   
 &             & LSTM          & 3.04603E+13 & \qquad &            &         & LSTM                         & 5.34948E+14     & \qquad &       &        & LSTM          & 3.69276E+14\\ 
 &             & Attention     & 5.94603E+13 & \qquad &            &         & Attention                    & 4.34950E+14      & \qquad &       &       & Attention     & 3.69279E+14\\   
 &             & Bidirectional & 3.94602E+13 & \qquad &            &         & Bidirectional                & 5.44951E+14     & \qquad &       &        & Bidirectional & 3.69279E+14\\   
 &             & GRU           & 7.94603E+13 & \qquad &            &         & GRU                          & 5.34951E+14     & \qquad &       &        & GRU           & 3.69279E+14\\   
 &             & Hybrid        & 2.42938E+13 & \qquad &            &         & Hybrid                       & 9.79887E+14      & \qquad &       &       & Hybrid        & 9.00659E+14\\
\cmidrule{2-4}\cmidrule{7-9}\cmidrule{12-14}
 & 4,000 & \textbf{OPTM-LSTM}& \textbf{2.75419E+14} & \qquad &     &  5,000 & \textbf{OPTM-LSTM} & \textbf{2.26636E+14} & \qquad &  & 6,000 & \textbf{OPTM-LSTM} & \textbf{1.93647E+14}\\   
 &             & LSTM          & 2.86605E+14& \qquad &             &         & LSTM                    & 2.37062E+14& \qquad &       &        & LSTM          & 2.04036E+14\\ 
 &             & Attention     & 2.86606E+14& \qquad &             &         & Attention               & 2.37062E+14& \qquad &       &        & Attention     & 2.04038E+14\\   
 &             & Bidirectional & 2.86606E+14& \qquad &             &         & Bidirectional           & 2.37062E+14& \qquad &       &        & Bidirectional & 2.04038E+14\\   
 &             & GRU           & 2.86606E+14& \qquad &             &         & GRU                     & 2.37062E+14& \qquad &       &        & GRU           & 2.04038E+14\\   
 &             & Hybrid        & 1.70827E+15& \qquad &             &         & Hybrid                  & 9.83038E+14& \qquad &      &      & Hybrid           & 1.22734E+15\\
\cmidrule{2-4}\cmidrule{7-9}\cmidrule{12-14}
  & 7,000 & \textbf{OPTM-LSTM}& \textbf{1.68468E+14} & \qquad &     &  10,000 & \textbf{OPTM-LSTM}  & \textbf{1.13063E+14} & \qquad &  & 15,000 & \textbf{OPTM-LSTM} & \textbf{9.39333E+13}\\   
 &              & LSTM          & 2.10449E+14 & \qquad &            &         & LSTM         & 1.38019E+14 & \qquad &        &        & LSTM          & 1.16983E+14\\ 
 &              & Attention     & 1.80449E+14 & \qquad &            &         & Attention    & 1.38012E+14 & \qquad &        &        & Attention     & 1.16973E+14\\   
 &              & Bidirectional & 1.80449E+14 & \qquad &            &         & Bidirectional& 1.38012E+14 & \qquad &        &        & Bidirectional & 1.16973E+14\\   
 &              & GRU           & 1.80449E+14 & \qquad &            &         & GRU          & 1.38012E+14 & \qquad &        &        & GRU           & 1.16973E+14\\   
 &              & Hybrid        & 2.06190E+15 & \qquad &            &         & Hybrid      & 1.09899E+15  & \qquad &        &       & Hybrid        & 8.66925E+14\\
\cmidrule{2-4}\cmidrule{7-9}\cmidrule{12-14}
 & 20,000 & \textbf{OPTM-LSTM} & \textbf{8.79208E+13} & \qquad &    &  35,000  & \textbf{OPTM-LSTM} & \textbf{4.04241E+13} & \qquad &  & 50,000 & \textbf{OPTM-LSTM} & \textbf{2.03210E+13}\\   
 &              & LSTM          & 8.85471E+13 & \qquad &            &         & LSTM          & 6.73828E+13 & \qquad &        &        & LSTM          & 5.88844E+13\\ 
 &              & Attention     & 8.85388E+13 & \qquad &            &         & Attention    & 6.73811E+13 & \qquad &        &        & Attention     & 5.88940E+13\\   
 &              & Bidirectional & 8.85380E+13 & \qquad &            &         & Bidirectional& 6.73812E+13 & \qquad &        &        & Bidirectional & 5.88937E+13\\   
 &              & GRU           & 8.85379E+13 & \qquad &            &         & GRU           & 6.73805E+13 & \qquad &        &        & GRU           & 5.88918E+13\\   
 &              & Hybrid        & 2.13339E+15  & \qquad &           &         & Hybrid        & 2.18678E+15 & \qquad & &               & Hybrid        & 1.51111E+15\\
\cmidrule{2-4}\cmidrule{7-9}\cmidrule{12-14}
 & 100,000 & \textbf{OPTM-LSTM}& \textbf{1.09730E+13} & \qquad &     &  400,000  & \textbf{OPTM-LSTM} & \textbf{5.00414E+12} & \qquad &  & 800,000 & \textbf{OPTM-LSTM} & \textbf{1.55909E+12}\\   
 &                & LSTM          & 4.88444E+13& \qquad &            &         & LSTM         & 4.08671E+13 & \qquad &         &        & LSTM          & 3.91895E+13\\ 
 &                & Attention     & 4.88383E+13& \qquad &            &         & Attention    & 4.01958E+13 & \qquad &         &        & Attention     & 3.62271E+13\\   
 &                & Bidirectional & 4.88300E+13  & \qquad &          &         & Bidirectional& 3.98711E+13  & \qquad &         &     & Bidirectional & 3.70604E+13\\   
 &                & GRU           & 4.88339E+13& \qquad &            &         & GRU          & 3.97324E+13 & \qquad &         &        & GRU           & 3.95741E+13\\   
 &                & Hybrid        & 1.81421E+15& \qquad &            &         & Hybrid       & 2.15967E+15 & \qquad &         &        & Hybrid        & 1.73484E+15\\
\cmidrule{2-4}\cmidrule{7-9}\cmidrule{12-14}
 & 1,000,000 & \textbf{OPTM-LSTM}       & \textbf{1.25909E+12} & \qquad &   &  10,000,000 & \textbf{OPTM-LSTM}  & \textbf{1.05909E+12} & \qquad &  & 20,000,000 & \textbf{OPTM-LSTM} & \textbf{9.59093E+11}\\   
 &                  & LSTM         & 3.99895E+13 & \qquad &             &         & LSTM         & 3.99995E+13& \qquad &        &        & LSTM          & 4.06759E+13\\ 
 &                  & Attention    & 3.60522E+13 & \qquad &             &         & Attention    & 3.60009E+13& \qquad &        &        & Attention     & 3.62271E+13\\   
 &                  & Bidirectional& 3.70007E+13 & \qquad &             &         & Bidirectional& 3.67804E+13& \qquad &        &        & Bidirectional & 3.64675E+13\\   
 &                  & GRU          & 3.97990E+13  & \qquad &             &         & GRU         & 3.92581E+13& \qquad &        &        & GRU           & 3.90696E+13\\   
 &                  & Hybrid       & 1.63679E+15 & \qquad &             &         & Hybrid       & 1.56326E+15& \qquad &        &        & Hybrid        & 1.85684E+15\\
\cmidrule[2pt]{1-6}\cmidrule[2pt]{6-9}\cmidrule[2pt]{9-14}
\textbf{Stock} & \textbf{Size} & \textbf{Model} & \textbf{MSE - Test} & \qquad & \textbf{Stock} & \textbf{Size} & \textbf{Model} & \textbf{MSE - Test} & \qquad & \textbf{Stock} & \textbf{Size} & \textbf{Model} & \textbf{MSE - Test}\\
\cmidrule{1-4}\cmidrule{6-9}\cmidrule{11-14}
 Google & 1,000 & \textbf{OPTM-LSTM} & \textbf{1.04441E+15} & \qquad &  Google     &  2,000 & \textbf{OPTM-LSTM}& \textbf{3.32290E+13} & \qquad & Google & 3,000 & \textbf{OPTM-LSTM} & \textbf{2.60755E+13}\\   
 &             & LSTM          & 1.05494E+15 & \qquad &            &        & LSTM         & 3.94267E+13            & \qquad &       &        & LSTM          & 3.93324E+13\\ 
 &             & Attention     & 1.05495E+15 & \qquad &            &        & Attention    & 3.94266E+13            & \qquad &       &        & Attention     & 3.93324E+13\\   
 &             & Bidirectional & 1.05495E+15 & \qquad &            &        & Bidirectional& 3.94266E+13            & \qquad &       &        & Bidirectional & 3.93324E+13\\   
 &             & GRU           & 1.05494E+15 & \qquad &            &        & GRU          & 3.94267E+13            & \qquad &       &        & GRU           & 3.93324E+13\\   
 &             & Hybrid        & 1.11003E+15 & \qquad &            &        & Hybrid       & 1.12736E+15            & \qquad &       &        & Hybrid        & 1.26887E+15\\
\cmidrule{2-4}\cmidrule{7-9}\cmidrule{12-14}
 & 4,000  & \textbf{OPTM-LSTM}& \textbf{2.02484E+13} & \qquad &    &  5,000  & \textbf{OPTM-LSTM}  & \textbf{1.39428E+13} & \qquad &  & 6,000 & \textbf{OPTM-LSTM} & \textbf{1.75926E+13}\\   
 &             & LSTM           & 3.93318E+13& \qquad &                       &         & LSTM         & 3.92138E+13 & \qquad &           &        & LSTM          & 3.91251E+13\\ 
 &             & Attention      & 3.93315E+13& \qquad &                       &         & Attention    & 3.92136E+13 & \qquad &          &        & Attention     & 3.91260E+13\\   
 &             & Bidirectional  & 3.93316E+13& \qquad &                       &         & Bidirectional& 3.92136E+13 & \qquad &          &        & Bidirectional & 3.91256E+13\\   
 &             & GRU            & 3.93318E+13& \qquad &                       &         & GRU          & 3.92139E+13 & \qquad &          &        & GRU           & 3.91256E+13\\   
 &             & Hybrid         & 1.58088E+15& \qquad &                       &         & Hybrid       & 5.37368E+14 & \qquad &          &        & Hybrid        & 1.52489E+15\\

\cmidrule{2-4}\cmidrule{7-9}\cmidrule{12-14}
& 7,000 & \textbf{OPTM-LSTM}& \textbf{4.60990E+12} & \qquad &      &  10,000 & \textbf{OPTM-LSTM}  & \textbf{5.14485E+10} & \qquad &  & 15,000 & \textbf{OPTM-LSTM} & \textbf{3.40823E+11}\\   
 &              & LSTM          & 3.91154E+13 & \qquad &            &         & LSTM         & 3.91224E+13 & \qquad &        &                                    & LSTM   & 3.91048E+13\\ 
 &              & Attention     & 3.91166E+13 & \qquad &            &         & Attention    & 3.91234E+13 & \qquad &        &                                    & Attention & 3.91057E+13\\ 
 &              & Bidirectional & 3.91162E+13 & \qquad &            &         & Bidirectional& 3.91234E+13 & \qquad &       &                                     & Bidirectional & 3.91058E+13\\   
 &              & GRU           & 4.91163E+13 & \qquad &            &         & GRU          & 3.91234E+13 & \qquad &        &                                    & GRU           & 3.91058E+13\\   
 &              & Hybrid        & 4.78848E+14 & \qquad &            &         & Hybrid       & 3.71959E+14 & \qquad &        &                                    & Hybrid & 5.93297E+14\\

\cmidrule{2-4}\cmidrule{7-9}\cmidrule{12-14}
 & 20,000 & \textbf{OPTM-LSTM}& \textbf{5.39592E+09} & \qquad &      &  35,000 & \textbf{OPTM-LSTM}  & \textbf{2.05396E+12} & \qquad &  & 50,000 & \textbf{OPTM-LSTM} & \textbf{3.13280E+10}\\   
 &              & LSTM          & 3.90233E+13 & \qquad &            &         & LSTM         & 3.91767E+13 & \qquad &        &        & LSTM          & 3.89640E+13\\ 
 &              & Attention     & 3.90265E+13 & \qquad &            &         & Attention    & 3.91775E+13 & \qquad &        &        & Attention     & 3.89657E+13\\   
 &              & Bidirectional & 9.90264E+13 & \qquad &            &         & Bidirectional& 3.91775E+13 & \qquad &        &        & Bidirectional & 3.89653E+13\\   
 &              & GRU           & 7.90257E+13 & \qquad &            &         & GRU          & 3.91765E+13 & \qquad &        &        & GRU           & 3.89625E+13\\   
 &              & Hybrid        & 7.41977E+14 & \qquad &            &         & Hybrid       & 2.33358E+14 & \qquad &        &        & Hybrid        & 5.49750E+14\\

\cmidrule{2-4}\cmidrule{7-9}\cmidrule{12-14}
 & 100,000 & \textbf{OPTM-LSTM}& \textbf{1.20071E+09} & \qquad &  &  400,000 & \textbf{OPTM-LSTM}  & \textbf{3.62520E+09} & \qquad &  & 800,000 & \textbf{OPTM-LSTM} & \textbf{4.74797E+09}\\   
 &                & LSTM          & 3.84952E+13& \qquad &          &         & LSTM         & 3.83070E+13 & \qquad &         &        & LSTM          & 3.82016E+13\\ 
 &                & Attention     & 3.84821E+13& \qquad &          &         & Attention    & 3.70009E+13& \qquad &         &        & Attention     & 3.31423E+13\\   
 &                & Bidirectional & 3.84672E+13& \qquad &          &         & Bidirectional& 3.63139E+13& \qquad &         &        & Bidirectional & 3.44671E+13\\   
 &                & GRU           & 3.84753E+13& \qquad &          &         & GRU          & 3.60069E+13& \qquad &         &        & GRU           & 2.86685E+13\\   
 &                & Hybrid        & 1.83510E+09& \qquad &          &         & Hybrid       & 1.44407E+15& \qquad &         &        & Hybrid        & 1.41009E+15\\

\cmidrule{2-4}\cmidrule{7-9}\cmidrule{12-14}
& 1,000,000 & \textbf{OPTM-LSTM} & \textbf{4.85674E+09} & \qquad &  &  10,000,000 & \textbf{OPTM-LSTM} & \textbf{4.94797E+09} & \qquad & & 20,000,000  & \textbf{OPTM-LSTM} & \textbf{5.67455E+09}\\   
 &                  & LSTM         & 3.82005E+13 & \qquad &            &         & LSTM            & 3.80002E+13       & \qquad &        &            & LSTM          & 3.82016E+13\\ 
 &                  & Attention    & 3.20579E+13 & \qquad &            &         & Attention       & 3.20015E+13       & \qquad &        &            & Attention     & 3.21794E+13\\   
 &                  & Bidirectional& 3.35623E+13 & \qquad &            &         & Bidirectional   & 3.14658E+13       & \qquad &        &            & Bidirectional & 3.02094E+13\\   
 &                  & GRU          & 2.51040E+13 & \qquad &            &         & GRU             & 2.27702E+13       & \qquad &        &            & GRU           & 2.05398E+13\\   
 &                  & Hybrid       & 1.57883E+15 & \qquad &            &         & Hybrid          & 1.35003E+15       & \qquad &        &            & Hybrid        & 1.22934E+15\\
\cmidrule[2pt]{1-6}\cmidrule[2pt]{6-9}\cmidrule[2pt]{9-14}
\end{tabular}}
\medskip
\label{tab:GoogleShort}
\end{table*}

\begin{table*}[!htbp]
\centering
\captionsetup{width=.70\textwidth}
\caption{Google MSE scores under the Long experimental protocol.}
\scalebox{0.62}{
\begin{tabular}{rcrlcrcrlcrcrl}
\cmidrule[2pt]{1-6}\cmidrule[2pt]{6-9}\cmidrule[2pt]{9-14}
\textbf{Size} & \textbf{Stock} & \textbf{Model} & \textbf{MSE - Train} & \qquad & \textbf{Size} & \textbf{Stock} & \textbf{Model} & \textbf{MSE - Train} & \qquad & \textbf{Size} & \textbf{Stock} & \textbf{Model} & \textbf{MSE - Train}\\
\cmidrule{1-4}\cmidrule{6-9}\cmidrule{11-14}
 Google & 1,000 & \textbf{OPTM-LSTM} & \textbf{7.13393E+12} & \qquad & Google &  2,000 & \textbf{OPTM-LSTM}& \textbf{1.95355E+13}  & \qquad & Google & 3,000 & \textbf{OPTM-LSTM}& \textbf{3.62966E+14}\\   
 &             & LSTM          & 1.92849E+13           & \qquad &       &        & LSTM         & 1.95598E+13            & \qquad &       &        & LSTM          & 3.69272E+14\\ 
 &             & Attention     & 1.94585E+13           & \qquad &       &        & Attention    & 1.95592E+13            & \qquad &       &        & Attention     & 3.69269E+14\\   
 &             & Bidirectional & 1.94585E+13           & \qquad &       &        & Bidirectional& 1.97603E+13            & \qquad &       &        & Bidirectional & 3.69270E+14\\   
 &             & GRU           & 1.94583E+13           & \qquad &       &        & GRU          & 1.96591E+13            & \qquad &       &        & GRU           & 3.69269E+14\\   
 &             & Hybrid        & 5.15419E+13           & \qquad &       &        & Hybrid       & 6.22681E+13            & \qquad &       &       & Hybrid         & 6.43307E+14\\
\cmidrule{2-4}\cmidrule{7-9}\cmidrule{12-14}
 & 4,000 & \textbf{OPTM-LSTM} & \textbf{2.18488E+14} & \qquad &        &  5,000  & \textbf{OPTM-LSTM}  & \textbf{2.70888E+14} & \qquad &  & 6,000 & \textbf{OPTM-LSTM} & \textbf{1.80689E+14}\\   
 &             & LSTM           & 2.86598E+14 & \qquad &                       &         & LSTM         & 2.37052E+14 & \qquad &          &        & LSTM          & 2.04031E+14\\ 
 &             & Attention      & 2.86591E+14 & \qquad &                       &         & Attention    & 2.37038E+14 & \qquad &          &        & Attention     & 2.03984E+14\\   
 &             & Bidirectional  & 2.86589E+14 & \qquad &                       &         & Bidirectional& 2.37032E+14 & \qquad &          &        & Bidirectional & 2.03993E+14\\   
 &             & GRU            & 2.86549E+14 & \qquad &                       &         & GRU          & 2.37033E+14  & \qquad &          &        & GRU           & 2.03987E+14\\   
 &             & Hybrid         & 7.24462E+14 & \qquad &                       &         & Hybrid       & 1.12468E+15 & \qquad &          &        & Hybrid        & 1.36507E+15\\
\cmidrule{2-4}\cmidrule{7-9}\cmidrule{12-14}
  & 7,000 & \textbf{OPTM-LSTM}& \textbf{1.67232E+14} & \qquad &      &  10,000 & \textbf{OPTM-LSTM}  & \textbf{1.13227E+14} & \qquad &  & 15,000 & \textbf{OPTM-LSTM} & \textbf{5.04511E+13}\\  
 &              & LSTM          & 1.80438E+14 & \qquad &            &         & LSTM         & 1.37966E+14 & \qquad &        &                            & LSTM          & 1.16907E+14\\ 
 &              & Attention     & 1.80375E+14 & \qquad &            &         & Attention    & 1.37812E+14 & \qquad &        &                            & Attention     & 8.07426E+13\\ 
 &              & Bidirectional & 1.80392E+14 & \qquad &            &         & Bidirectional& 1.37815E+14 & \qquad &        &                            & Bidirectional & 1.10559E+14\\   
 &              & GRU           & 1.80364E+14 & \qquad &            &         & GRU          & 1.37761E+14 & \qquad &        &                            & GRU           & 1.70163E+14\\   
 &              & Hybrid        & 1.89508E+15 & \qquad &            &         & Hybrid       & 1.99974E+15 & \qquad &        &                            & Hybrid        & 2.19058E+15\\
\cmidrule[2pt]{1-6}\cmidrule[2pt]{6-9}\cmidrule[2pt]{9-14}
\textbf{Stock} & \textbf{Size} & \textbf{Model} & \textbf{MSE - Test} & \qquad & \textbf{Stock} & \textbf{Size} & \textbf{Model} & \textbf{MSE - Test} & \qquad & \textbf{Stock} & \textbf{Size} & \textbf{Model} & \textbf{MSE - Test}\\
\cmidrule{1-4}\cmidrule{6-9}\cmidrule{11-14}
 Google & 1,000 & \textbf{OPTM-LSTM} & \textbf{1.02565E+15} & \qquad & Google &  2,000 & \textbf{OPTM-LSTM}& \textbf{2.21936E+13}  & \qquad & Google & 3,000 & \textbf{OPTM-LSTM}& \textbf{2.42504E+13}\\   
 &             & LSTM          & 1.04581E+15           & \qquad &       &        & LSTM         & 1.05493E+15            & \qquad &       &        & LSTM          & 3.93302E+13\\ 
 &             & Attention     & 1.05495E+15           & \qquad &       &        & Attention    & 1.05495E+15            & \qquad &       &        & Attention     & 3.93206E+13\\   
 &             & Bidirectional & 1.05495E+15           & \qquad &       &        & Bidirectional& 1.05495E+15            & \qquad &       &        & Bidirectional & 8.93262E+13\\   
 &             & GRU           & 1.05494E+15           & \qquad &       &        & GRU          & 1.05493E+15            & \qquad &       &        & GRU           & 3.69269E+14\\   
 &             & Hybrid        & 1.03878E+15           & \qquad &       &        & Hybrid       & 1.14449E+15            & \qquad &       &        & Hybrid        & 1.33994E+15\\
\cmidrule{2-4}\cmidrule{7-9}\cmidrule{12-14}
 & 4,000 & \textbf{OPTM-LSTM} & \textbf{6.48726E+12} & \qquad &        &  5,000  & \textbf{OPTM-LSTM}  & \textbf{7.15408E+12} & \qquad &  & 6,000 & \textbf{OPTM-LSTM} & \textbf{1.67019E+12}\\   
 &             & LSTM           & 3.93236E+13& \qquad &                       &         & LSTM         & 3.92020E+13& \qquad &          &        & LSTM          & 3.91175E+13\\ 
 &             & Attention      & 3.93125E+13& \qquad &                       &         & Attention    & 3.91765E+13& \qquad &          &        & Attention     & 3.90502E+13\\   
 &             & Bidirectional  & 3.93107E+13& \qquad &                       &         & Bidirectional& 3.91813E+13& \qquad &          &        & Bidirectional & 3.90623E+13\\   
 &             & GRU            & 3.93068E+13& \qquad &                       &         & GRU          & 3.91734E+13& \qquad &          &        & GRU           & 3.90501E+13\\   
 &             & Hybrid         & 1.33515E+15 & \qquad &                      &         & Hybrid       & 4.39233E+14& \qquad &          &        & Hybrid        & 9.10956E+14\\
\cmidrule{2-4}\cmidrule{7-9}\cmidrule{12-14}
& 7,000 & \textbf{OPTM-LSTM}& \textbf{2.64051E+12} & \qquad &      &  10,000 & \textbf{OPTM-LSTM}  & \textbf{3.48733E+11} & \qquad &  & 15,000 & \textbf{OPTM-LSTM} & \textbf{1.74703E+09}\\  
 &              & LSTM          & 3.90955E+13 & \qquad &            &         & LSTM         & 3.90843E+13 & \qquad &        &                           & LSTM                         & 3.90295E+13\\ 
 &              & Attention     & 3.90155E+13 & \qquad &            &         & Attention    & 3.88844E+13 & \qquad &        &                           & Attention                    & 2.17723E+13\\ 
 &              & Bidirectional & 3.90497E+13 & \qquad &            &         & Bidirectional& 3.88841E+13 & \qquad &        &                           & Bidirectional                & 3.01027E+13\\   
 &              & GRU           & 3.90018E+13 & \qquad &            &         & GRU          & 3.88263E+13 & \qquad &        &                           & GRU                          & 3.94140E+13\\   
 &              & Hybrid        & 2.82755E+14 & \qquad &            &         & Hybrid       & 1.38452E+15 & \qquad &        &                           & Hybrid                       & 1.46894E+15\\
\cmidrule[2pt]{1-6}\cmidrule[2pt]{6-9}\cmidrule[2pt]{9-14}
\end{tabular}}
\medskip
\label{tab:GoogleLong}
\end{table*}

\hyperref[tab:GoogleShort]{Table \ref{tab:GoogleShort}} and \hyperref[tab:GoogleLong]{Table \ref{tab:GoogleLong}} and their corresponding time-series plots reveal several critical performance advantages of the OPTM-LSTM cell. The first point is that this new cell achieves better performance across all the selected stock examples. Its performance is stable for the majority of the provided data sizes for both stock categories (i.e., high-liquid and less-liquid stocks). More specifically, there were very few cases for Amazon stock (see \hyperref[tab:AmazonShort]{Table \ref{tab:AmazonShort}}) that the hybrid model, despite its huge score fluctuations, approached the performance of the OPTM-LSTM NN. This performance similarity (i.e., the similarity between OPTM-LSTM and the hybrid model) stems from the fact that the hybrid model was oscillating heavily across many MSE scores. That oscillation was probably due to the fact that CNNs are non-linear functions which means that small changes in the input source can cause significant changes in the output. In terms of testing performance, the OPTM-LSTM NN achieved better results compared to the rest of the models in both experimental protocol (i.e., protocols based on a lower and higher number of epochs) settings and all the selected stocks. The majority of the models reached their plateau performance before the 20,000,000 trading events for the high-liquid stocks and 2,000,000 for the less-liquid stocks, respectively. 

OPTM-LSTM was able to achieve lower MSE scores compared to its competitors because it can immediately recognize every next price or volume fluctuation by the internal guarantor (i.e., non-forecasting internal mechanism) that the cell is equipped and it is able to online re-adjust/optimize its cell output. Another important point is that the OPTM-LSTM NN has a very simple topology. The winning topology was based on just one hidden layer, with only a few units, with minimal look-back history and batch size. This is important since the HFT ML trader will utilize a simple and light NN structure. Despite the fact that the OPTM-LSTM has a higher computational and time complexity per cell unit compared to the prototype LSTM, it processes the data flow in a similar or even faster manner since its structural elements (i.e., the look-back period, number of layers, and units) require less training parameters. Additionally, the internal optimization mechanism and the gates/states selection are part only of the forward step and no additional computational cost is attached to the backpropagation step since these parameters are trained locally (i.e., they do not participate in the overall chained partial derivation rule). 

An additional interpretation of the lower MSE scores that the OPTM-LSTM achieved arrives from the fact that it can capitalize on the LOB structure better compare to the rest of the RNN family. In particular, every LOB time instance is a collection of current and previous price and volume levels. For instance, we utilized for our experiments LOBs with 10 price levels, which means that there are 10 price and volume levels for the bid and ask sides, respectively. The best levels, which form the mid-price, are more informed compared to the levels deeper in the book which potentially carry past information. These deeper LOB levels can act as a look-back period for our OPTM-LSTM cell. The same information is also available to the other models but they have already trained and developed a stale topology that may carry noise.

On a different note, the OPTM-LSTM cell performs an internal cell feature selection based on the current and previous LOB's state without exposure to the disadvantages of the vanishing and exploding gradients. The gradient problem is a problem that RNNs suffer from due to the use of sequential inputs (i.e., time-steps or look-back period). LSTMs, despite the common perception that they do not suffer from that problem, are exposed to that risk since they just prevent or delay the gradients from a fast-paced vanishing (see \hyperref[eq:cellstate]{Eq. \ref{eq:cellstate}} and \hyperref[eq:eq27]{Eq. \ref{eq:eq27}}). It is also very difficult to identify the ideal length of the look-back period but the OPTM-LSTM cell solves that problem by viewing the input space in different dimensions and free for the batch size. In simple terms, it discards the previous timesteps and focuses on the present LOB state (i.e., present ask and bid LOB levels). That way every batch of size 1 is updated according to the present LOB's supply and demand and hedges the HFT ML trader from a potential vanishing gradient risk. As a result, the backpropagation process treats every OPTM-LSTM cell as a separate NN and not as a part of a chain rule calculation. 

\begin{figure*}[hbt!]
    \centering
  \subfloat[OPTM-LSTM training MSE scores \label{1a}]{%
       \scalebox{0.55}{\includegraphics[width=0.65\linewidth]{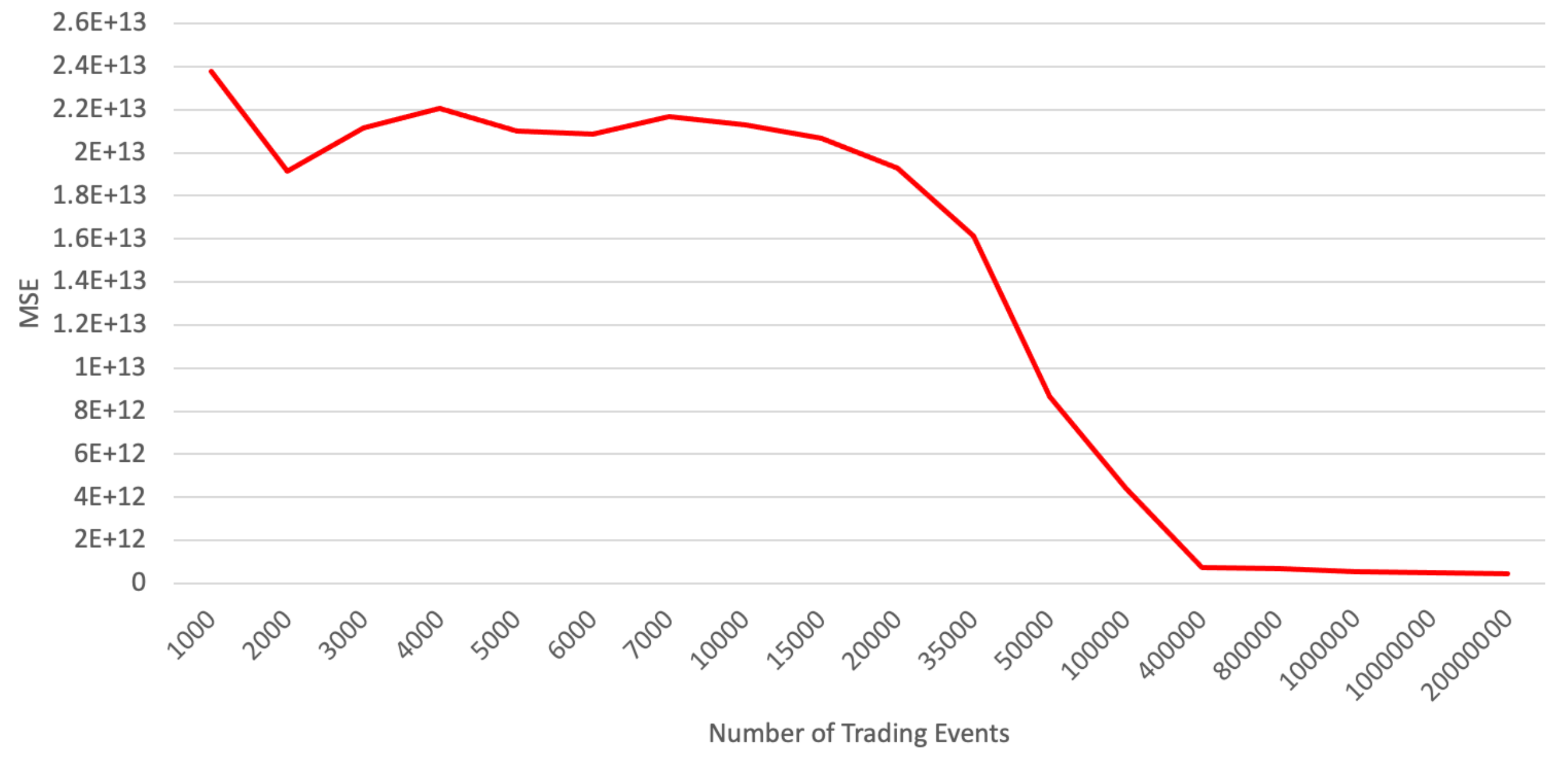}}}
  \subfloat[OPTM-LSTM testing MSE scores \label{1b}]{%
        \scalebox{0.55}{\includegraphics[width=0.65\linewidth]{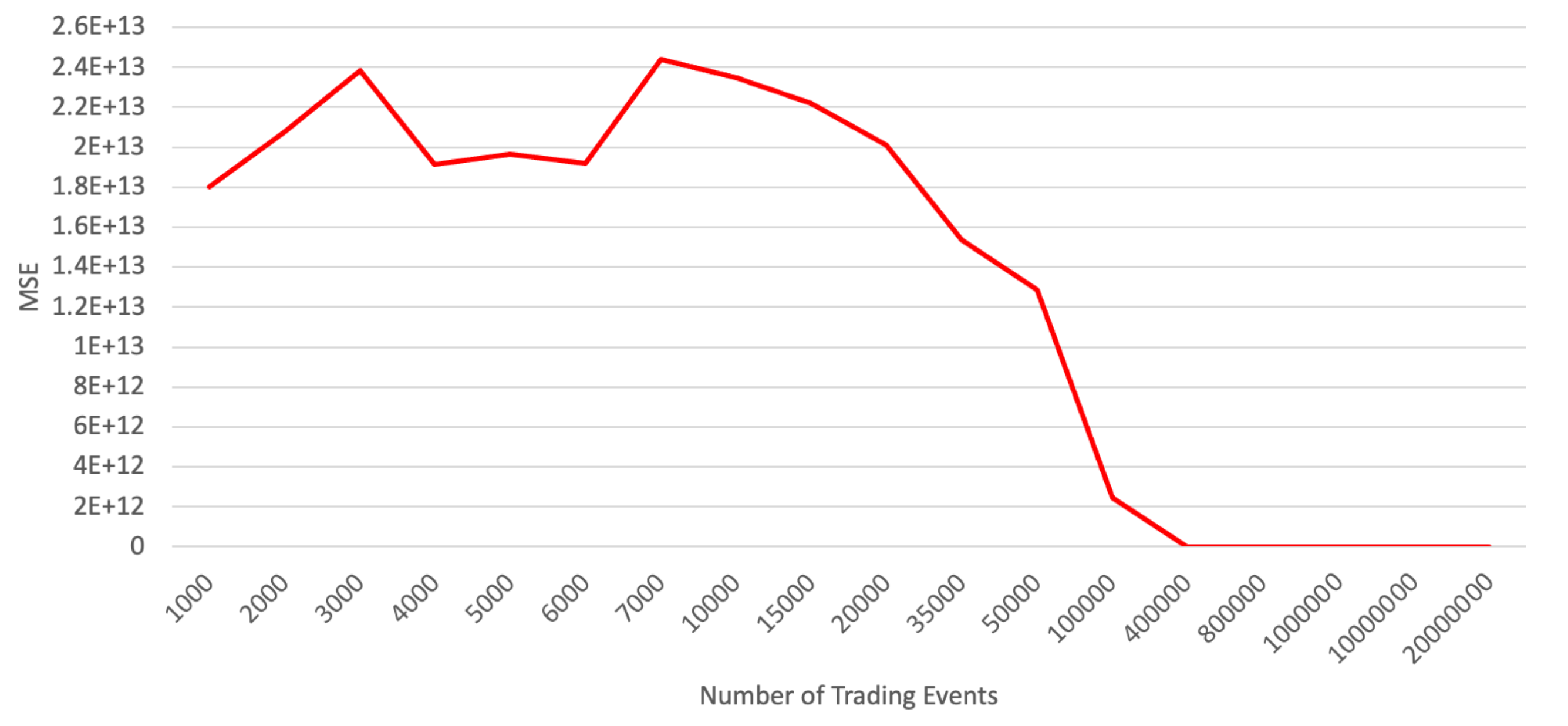}}}
    \\
  \subfloat[LSTM training MSE scores \label{1c}]{%
        \scalebox{0.55}{\includegraphics[width=0.65\linewidth]{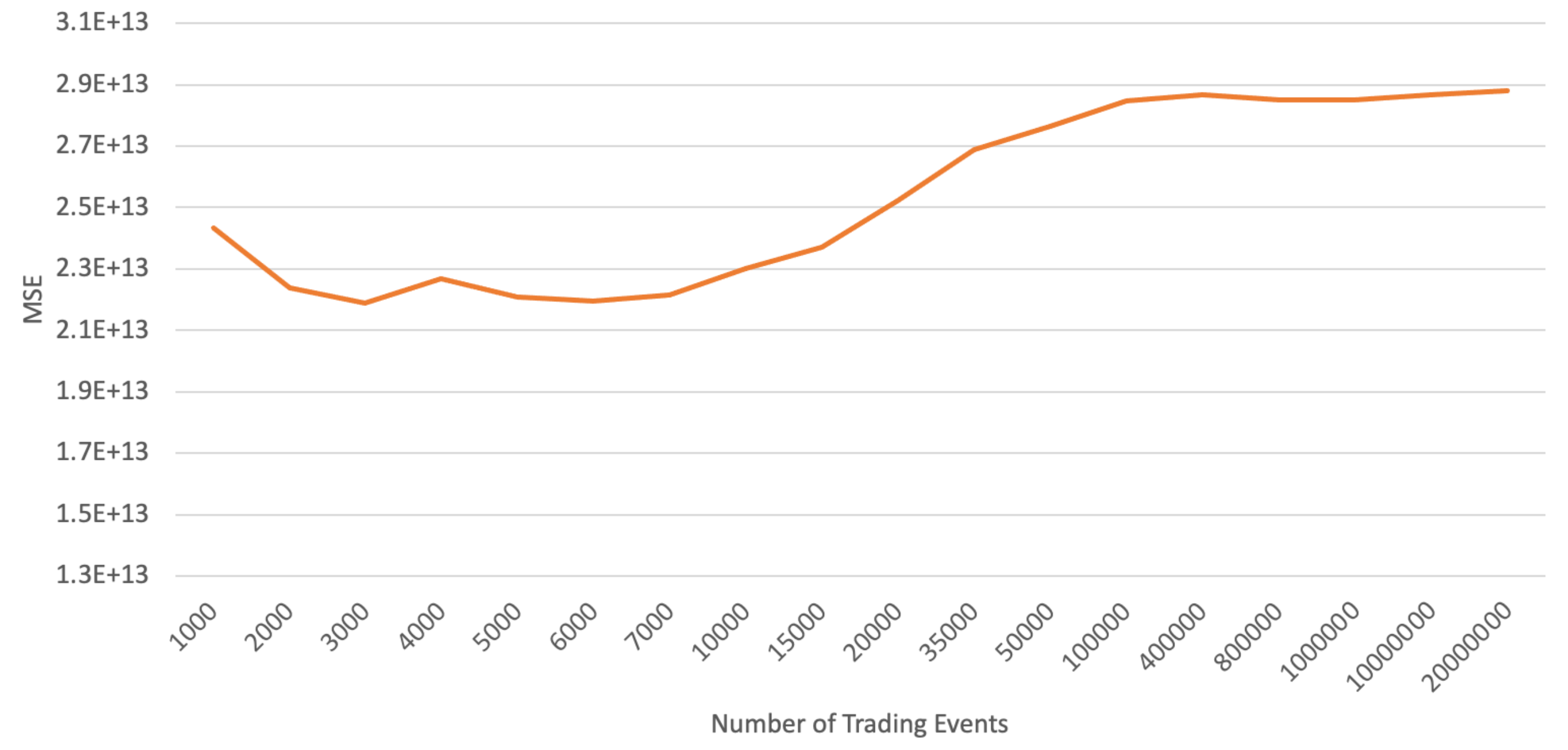}}}
  \subfloat[LSTM testing MSE scores \label{1d}]{%
        \scalebox{0.55}{\includegraphics[width=0.65\linewidth]{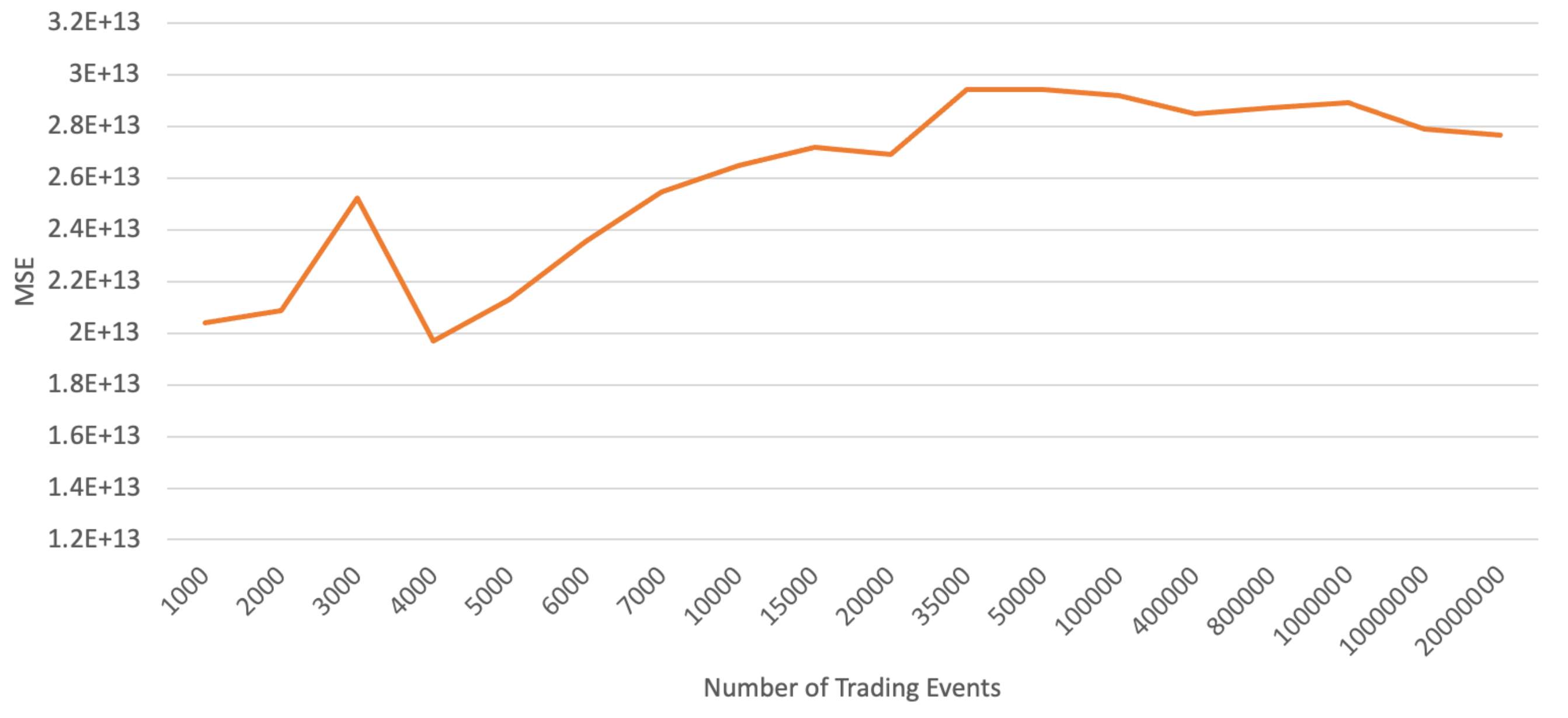}}}
    \\
  \subfloat[Attention LSTM training MSE scores\label{1c}]{%
        \scalebox{0.55}{\includegraphics[width=0.65\linewidth]{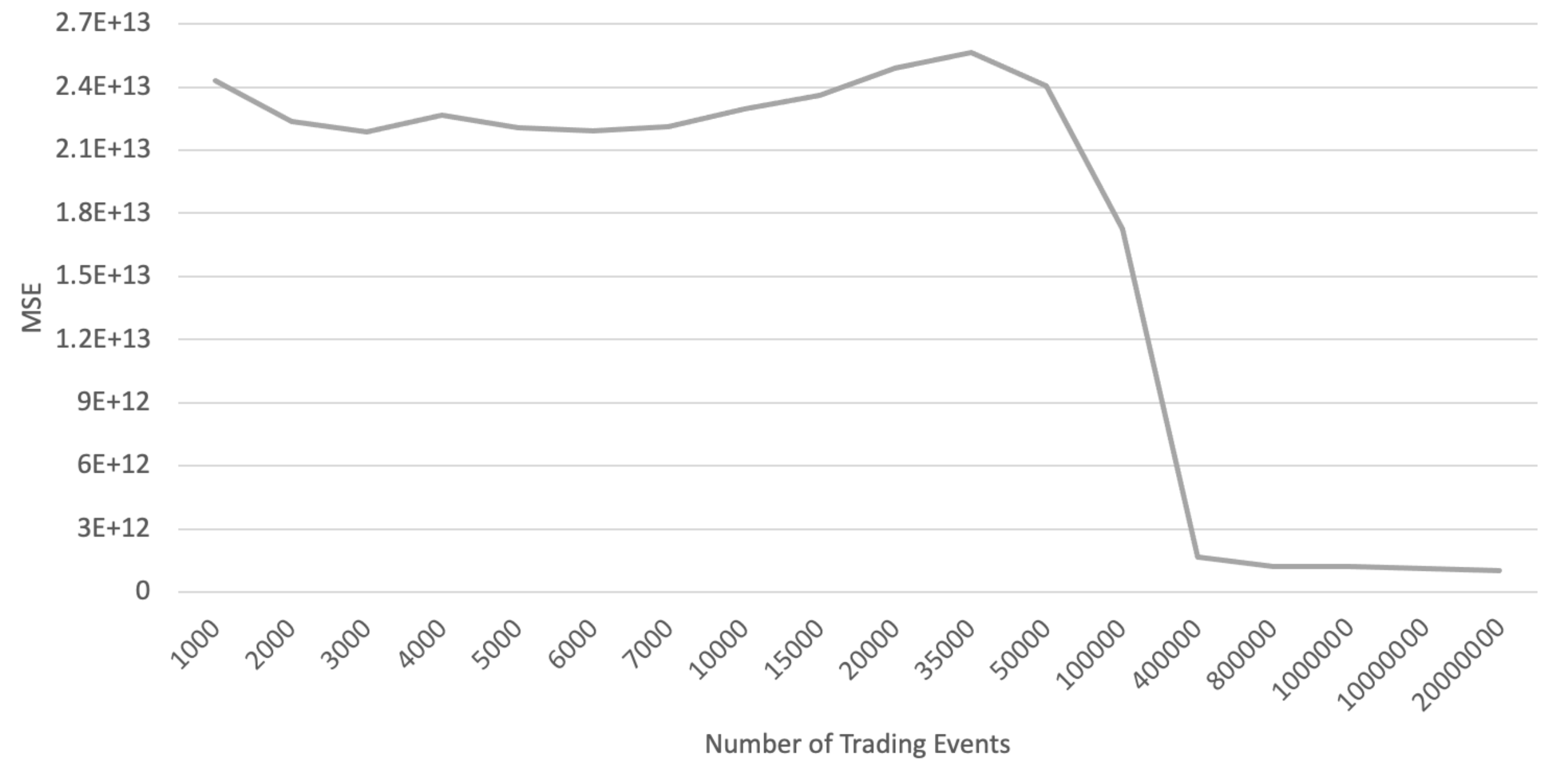}}}
  \subfloat[Attention LSTM testing MSE scores \label{1d}]{%
        \scalebox{0.55}{\includegraphics[width=0.65\linewidth]{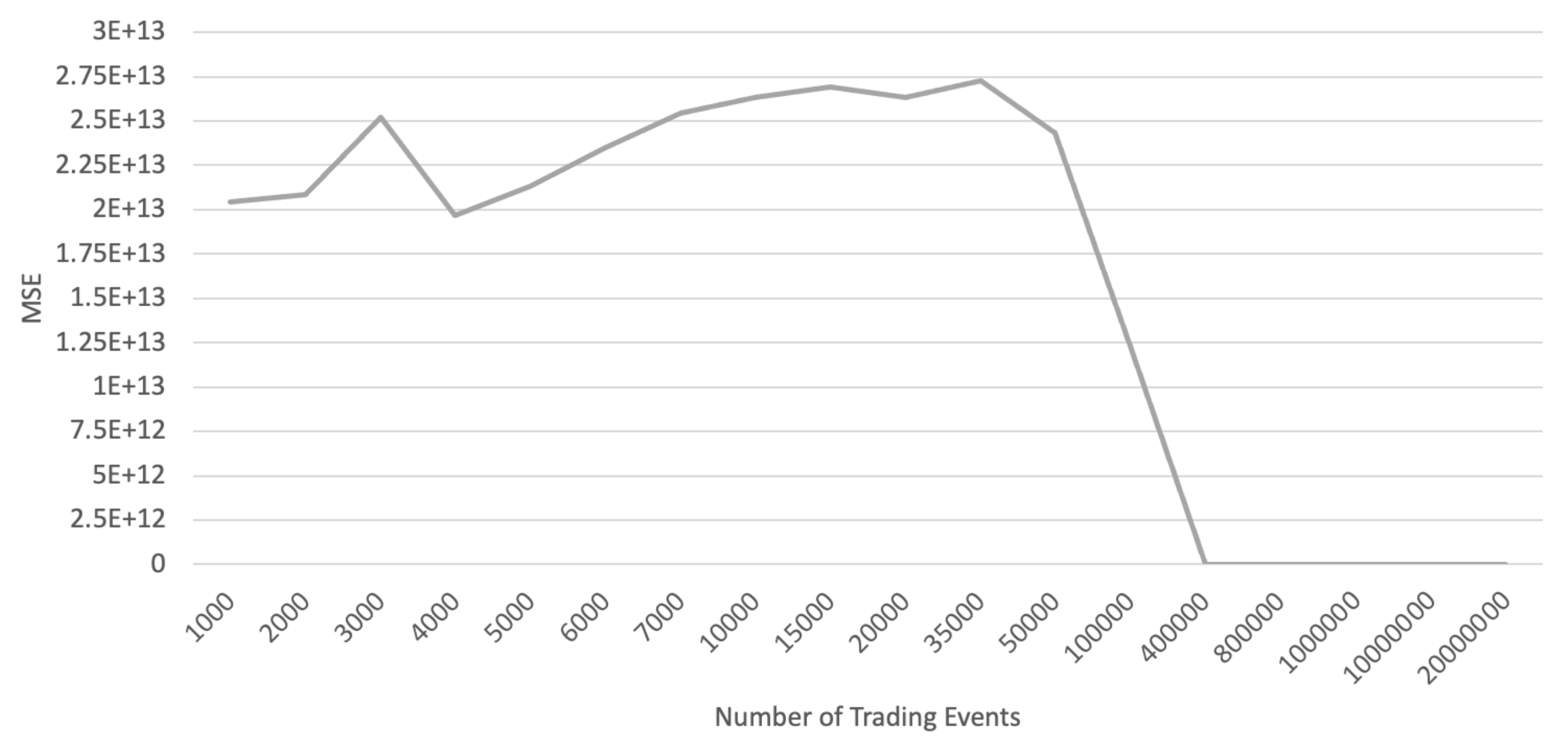}}}  
    \\
  \subfloat[Bidirectional training MSE scores \label{1a}]{%
       \scalebox{0.55}{\includegraphics[width=0.65\linewidth]{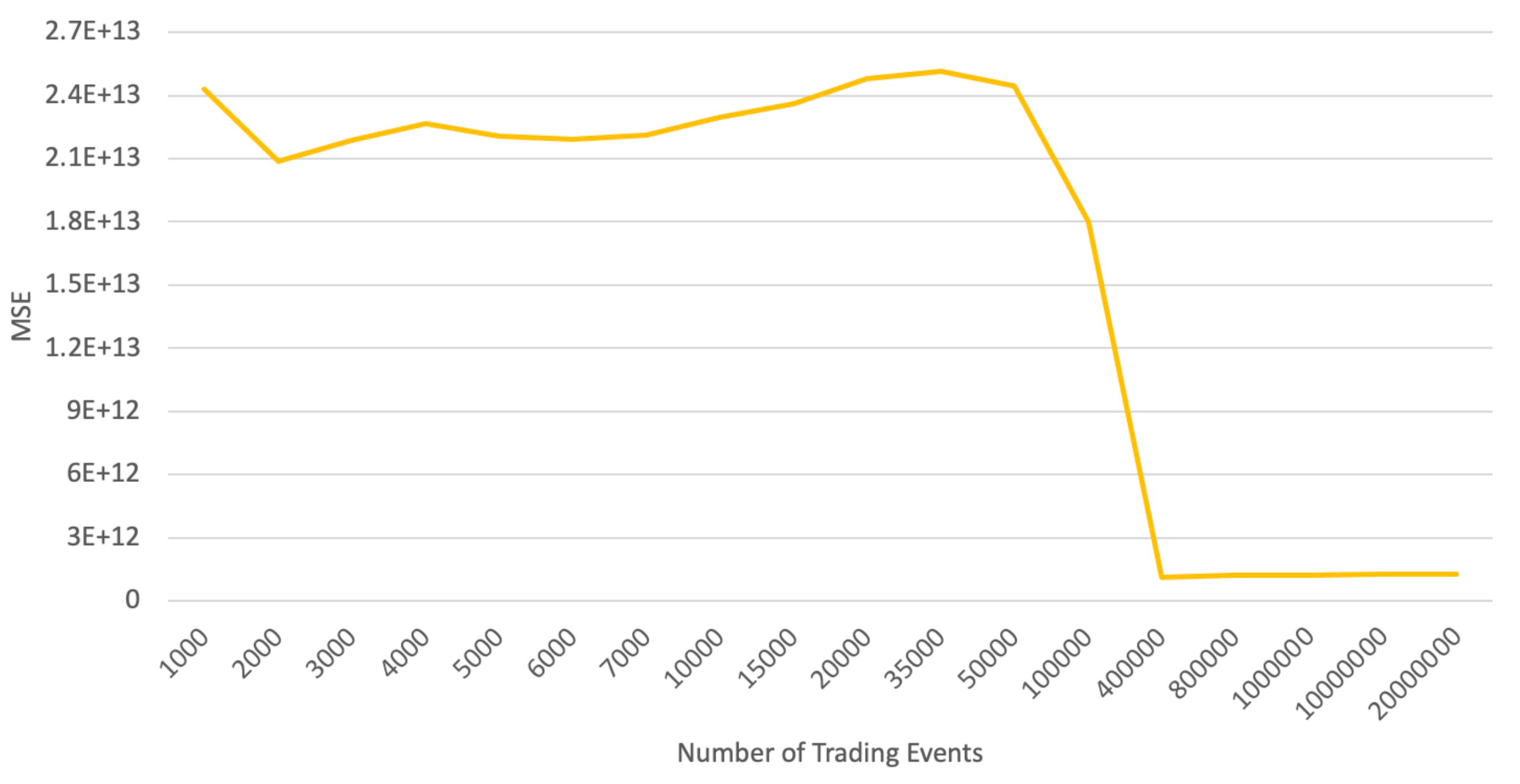}}}
  \subfloat[Bidirectional testing MSE scores \label{1b}]{%
        \scalebox{0.55}{\includegraphics[width=0.65\linewidth]{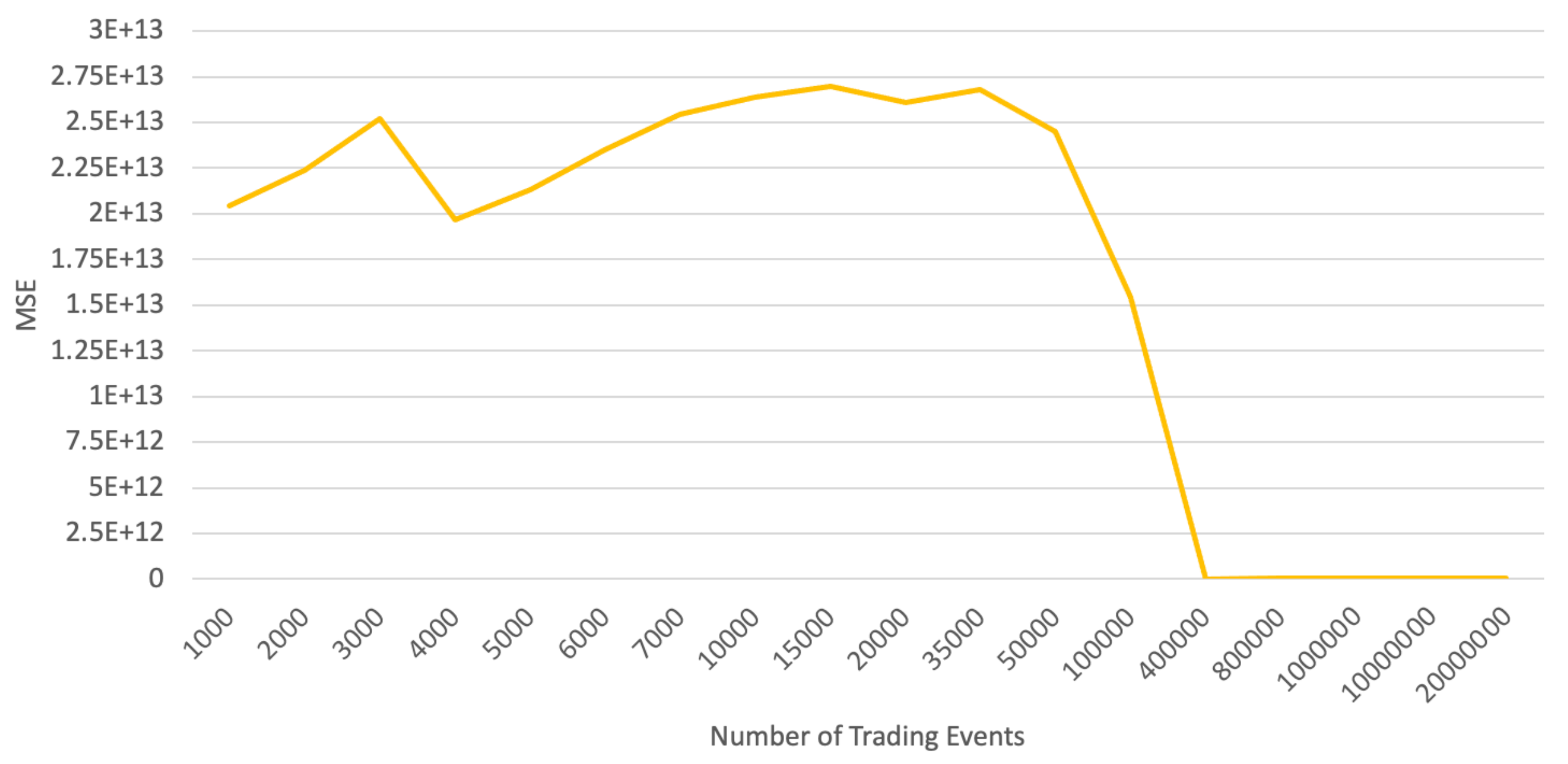}}}
    \\
  \subfloat[GRU training MSE scores \label{1a}]{%
       \scalebox{0.55}{\includegraphics[width=0.65\linewidth]{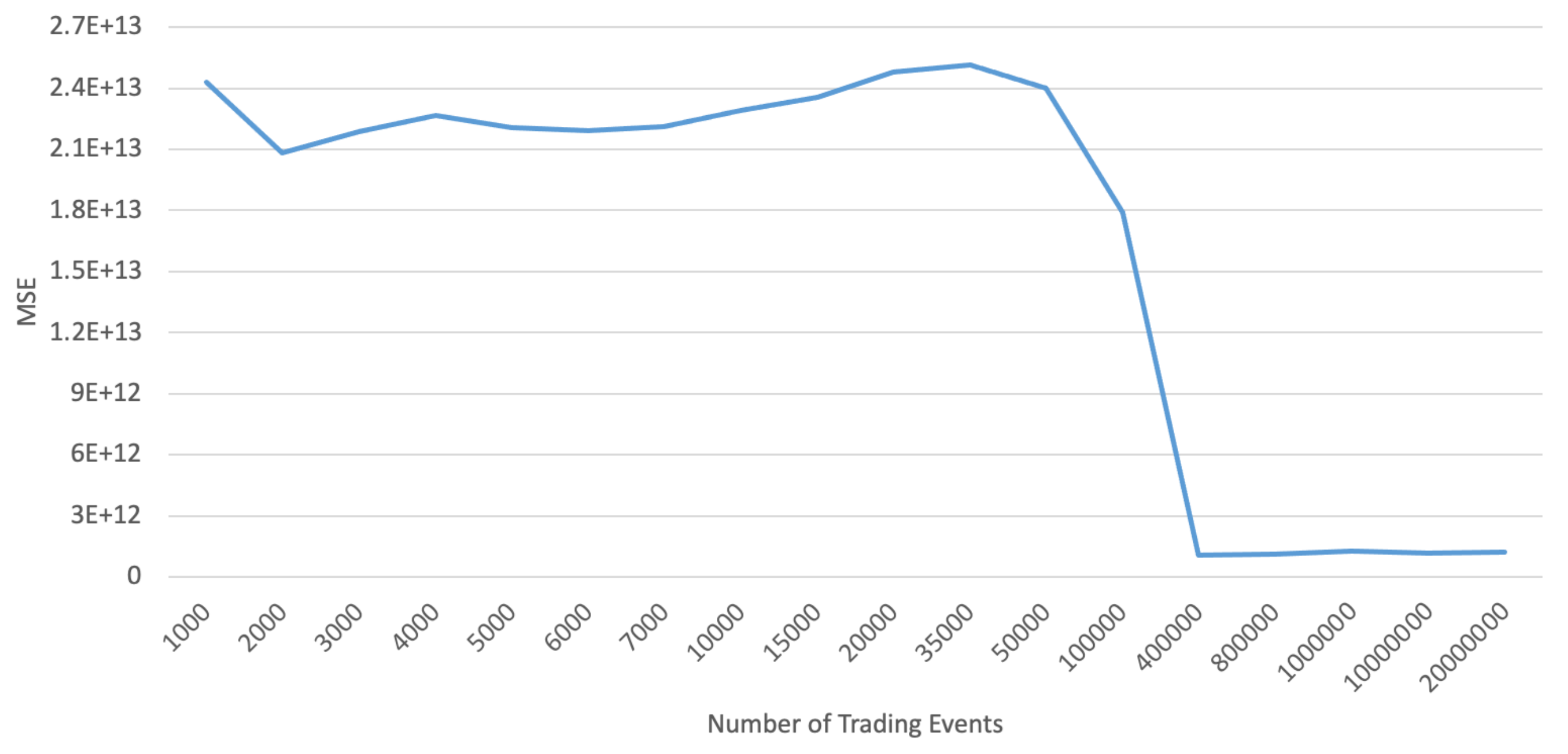}}}
  \subfloat[GRU testing MSE scores \label{1b}]{%
        \scalebox{0.55}{\includegraphics[width=0.65\linewidth]{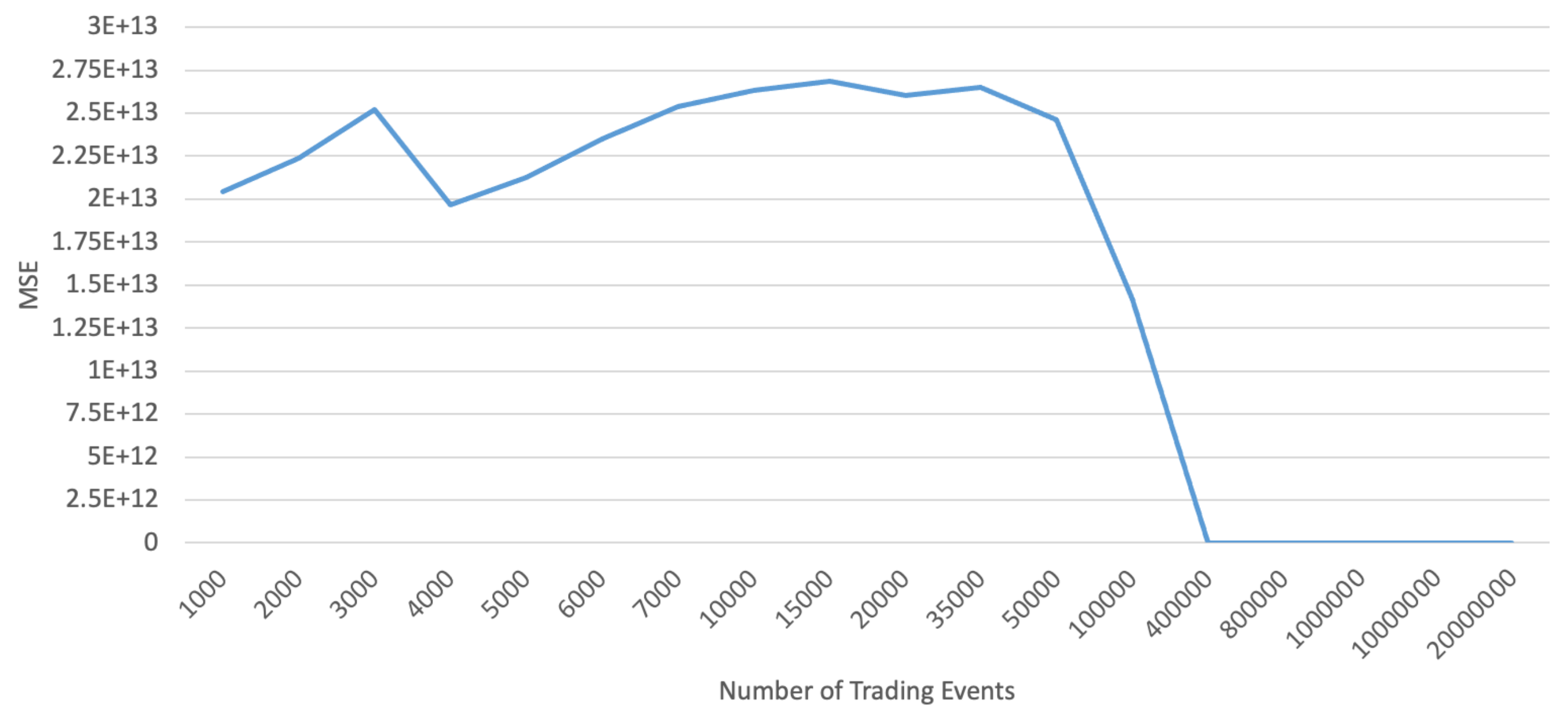}}}
    \\
  \subfloat[Hybrid training MSE scores \label{1a}]{%
       \scalebox{0.55}{\includegraphics[width=0.65\linewidth]{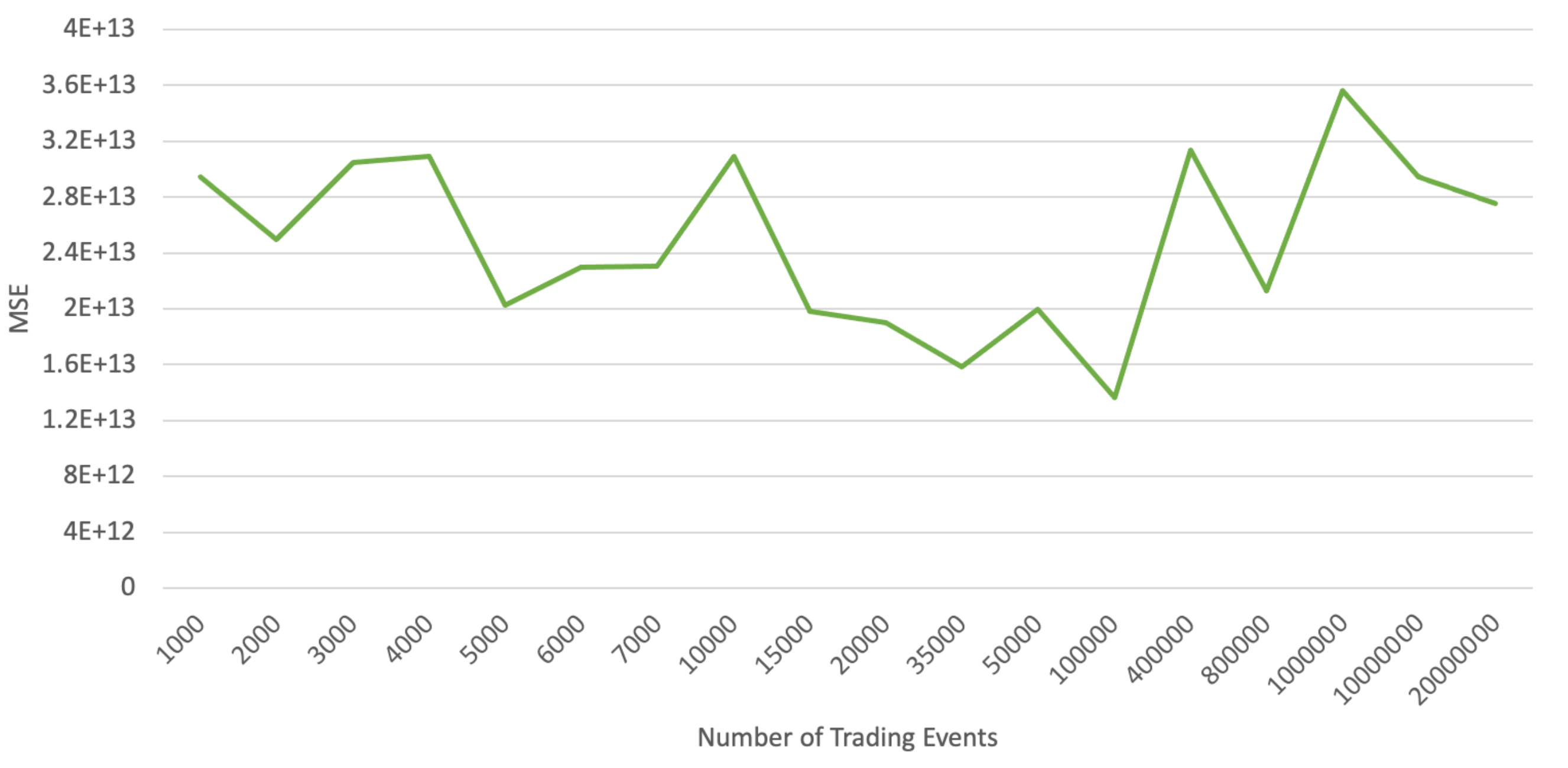}}}
  \subfloat[Hybrid testing MSE scores \label{1b}]{%
        \scalebox{0.55}{\includegraphics[width=0.65\linewidth]{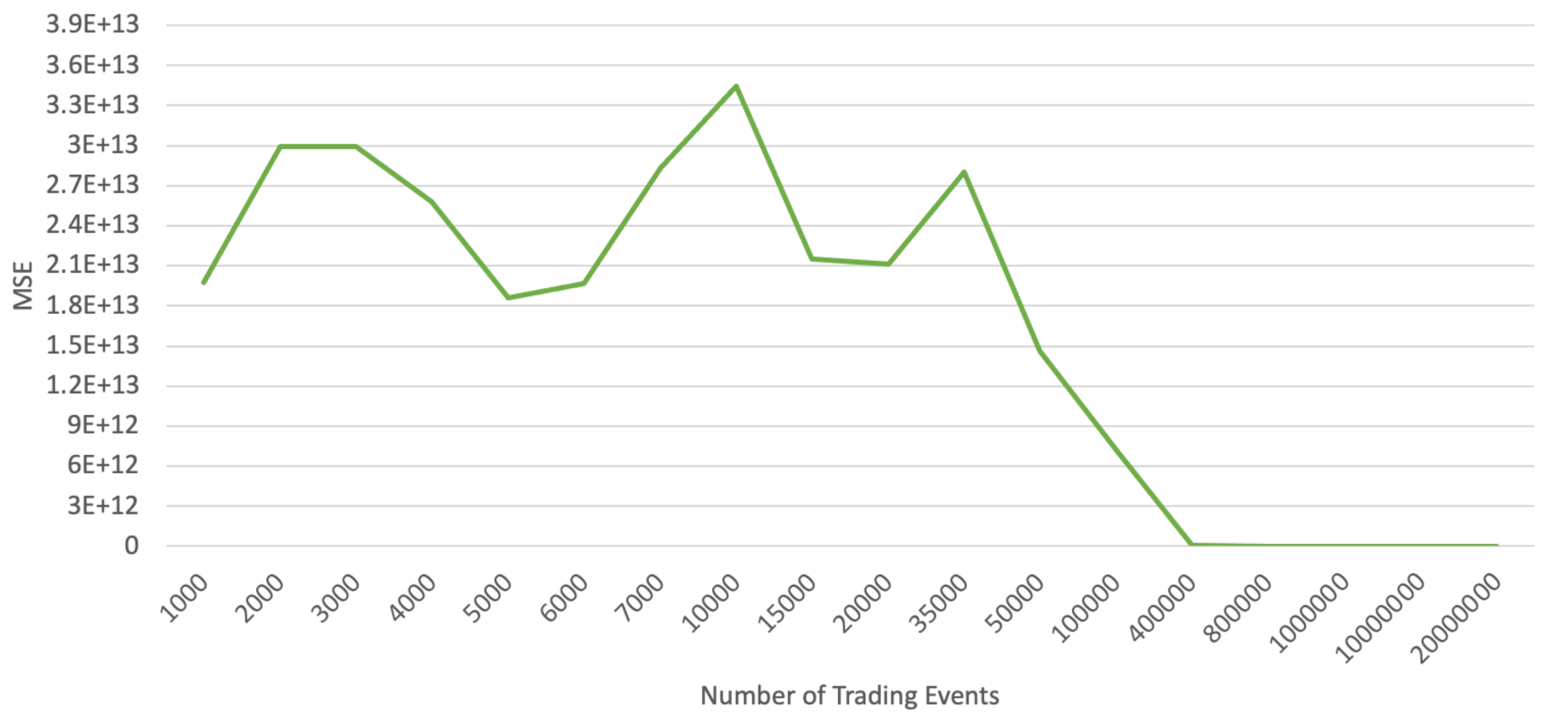}}}  
  \caption{Amazon Short MSE scores based on \hyperref[tab:AmazonShort]{Table \ref{tab:AmazonShort}}.}
  \label{fig:AmazonShort} 
\end{figure*}

\begin{figure*}[hbt!]
    \centering
  \subfloat[OPTM-LSTM training MSE scores \label{1a}]{%
       \scalebox{0.52}{\includegraphics[width=0.67\linewidth]{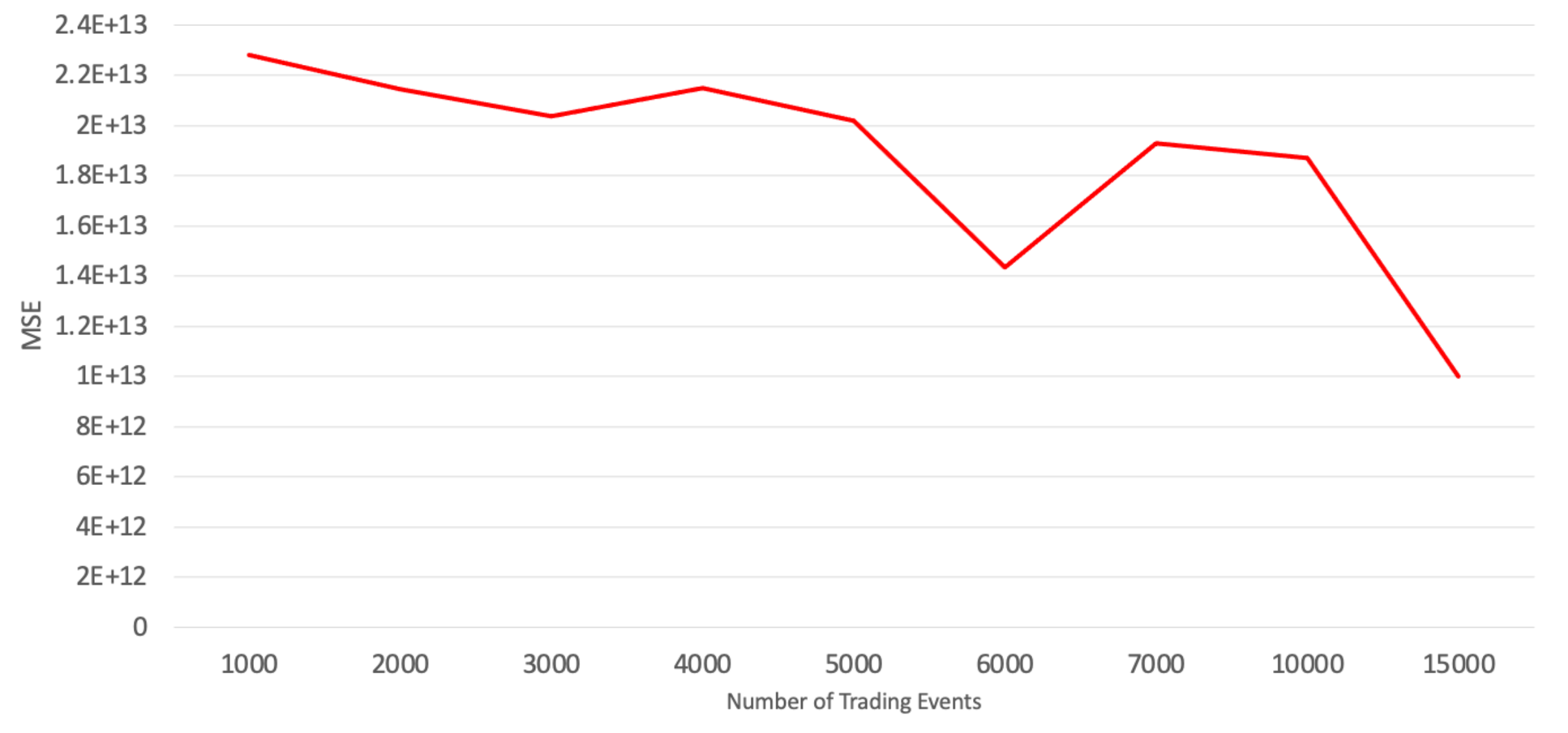}}}
  \subfloat[OPTM-LSTM testing MSE scores \label{1b}]{%
        \scalebox{0.52}{\includegraphics[width=0.67\linewidth]{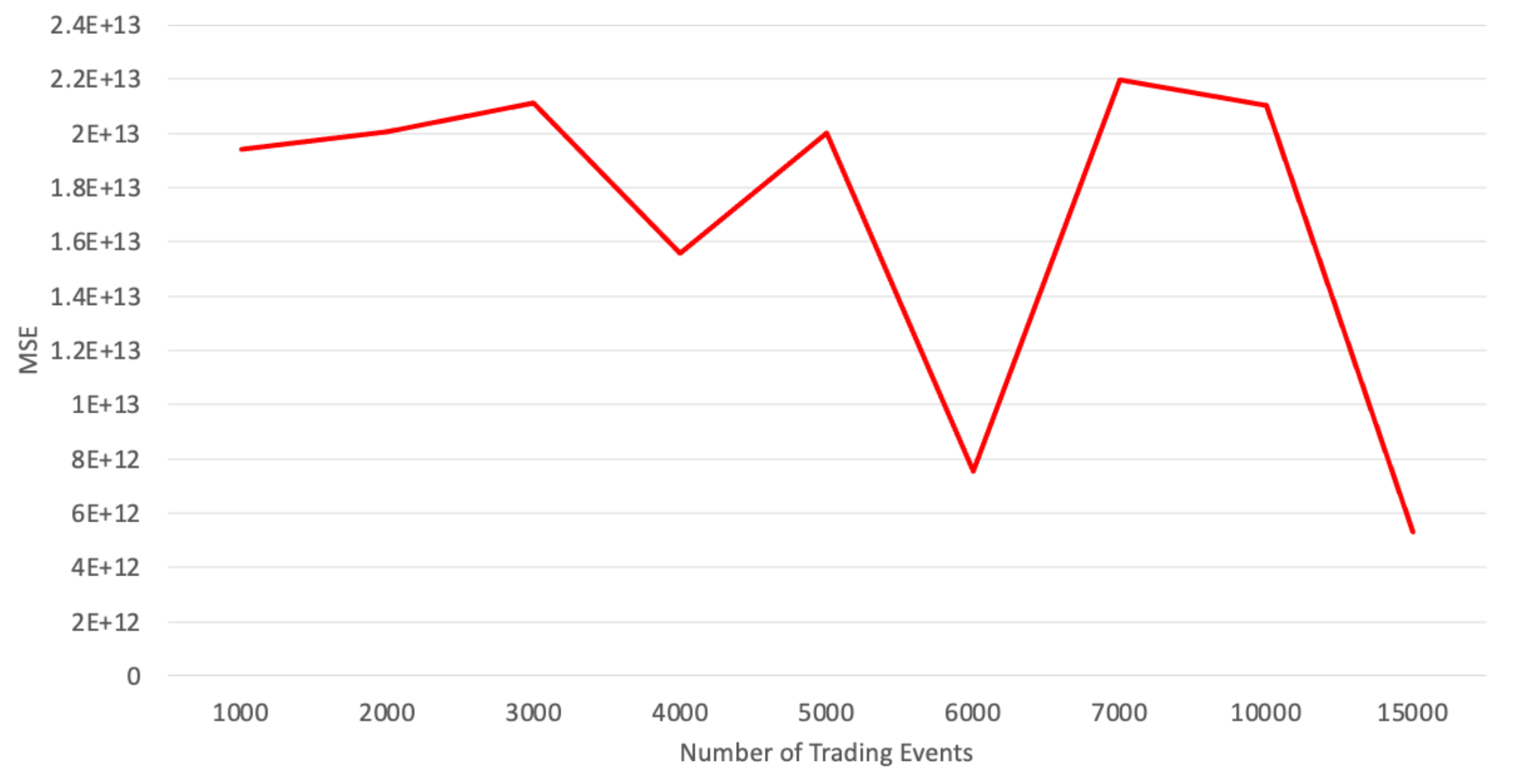}}}
    \\
  \subfloat[LSTM training MSE scores \label{1c}]{%
        \scalebox{0.52}{\includegraphics[width=0.67\linewidth]{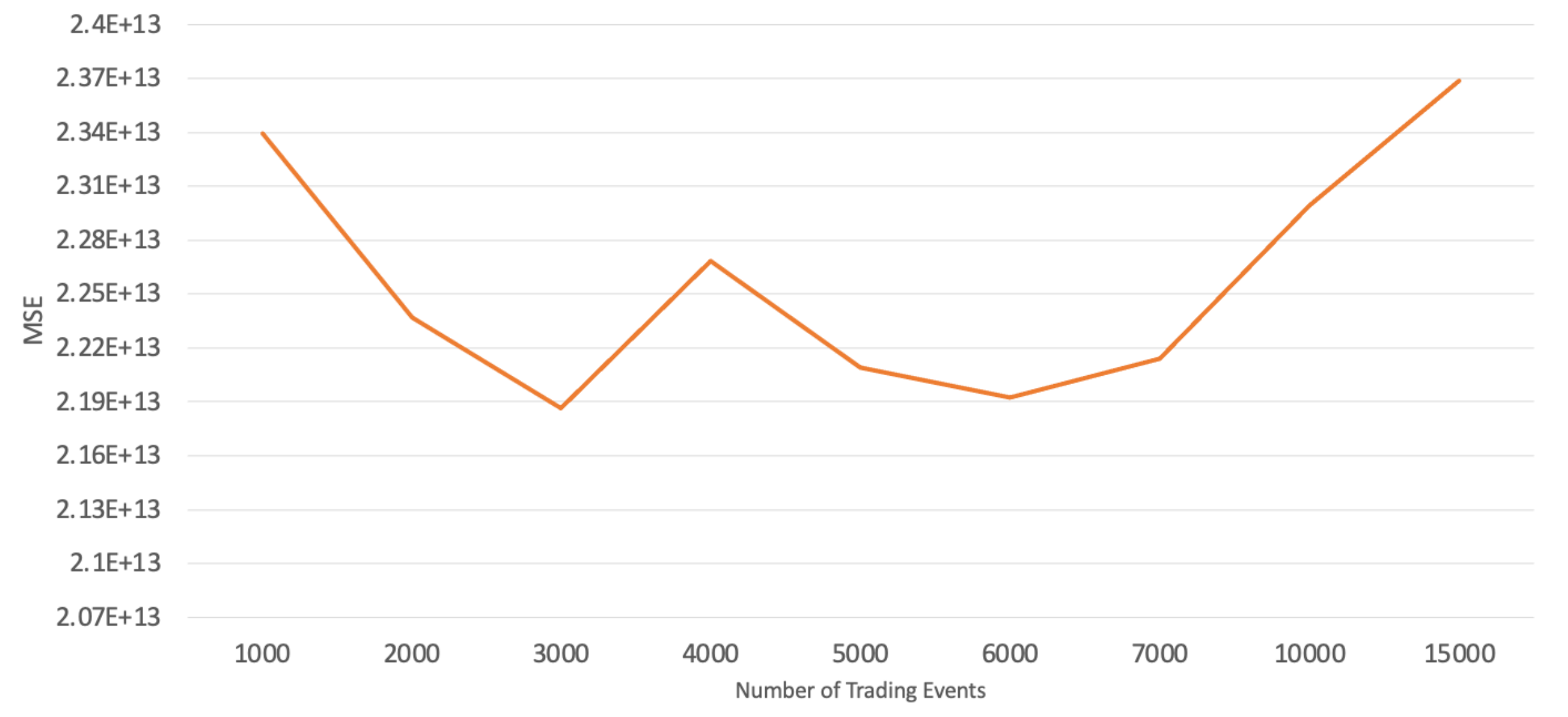}}}
  \subfloat[LSTM testing MSE scores \label{1d}]{%
        \scalebox{0.52}{\includegraphics[width=0.67\linewidth]{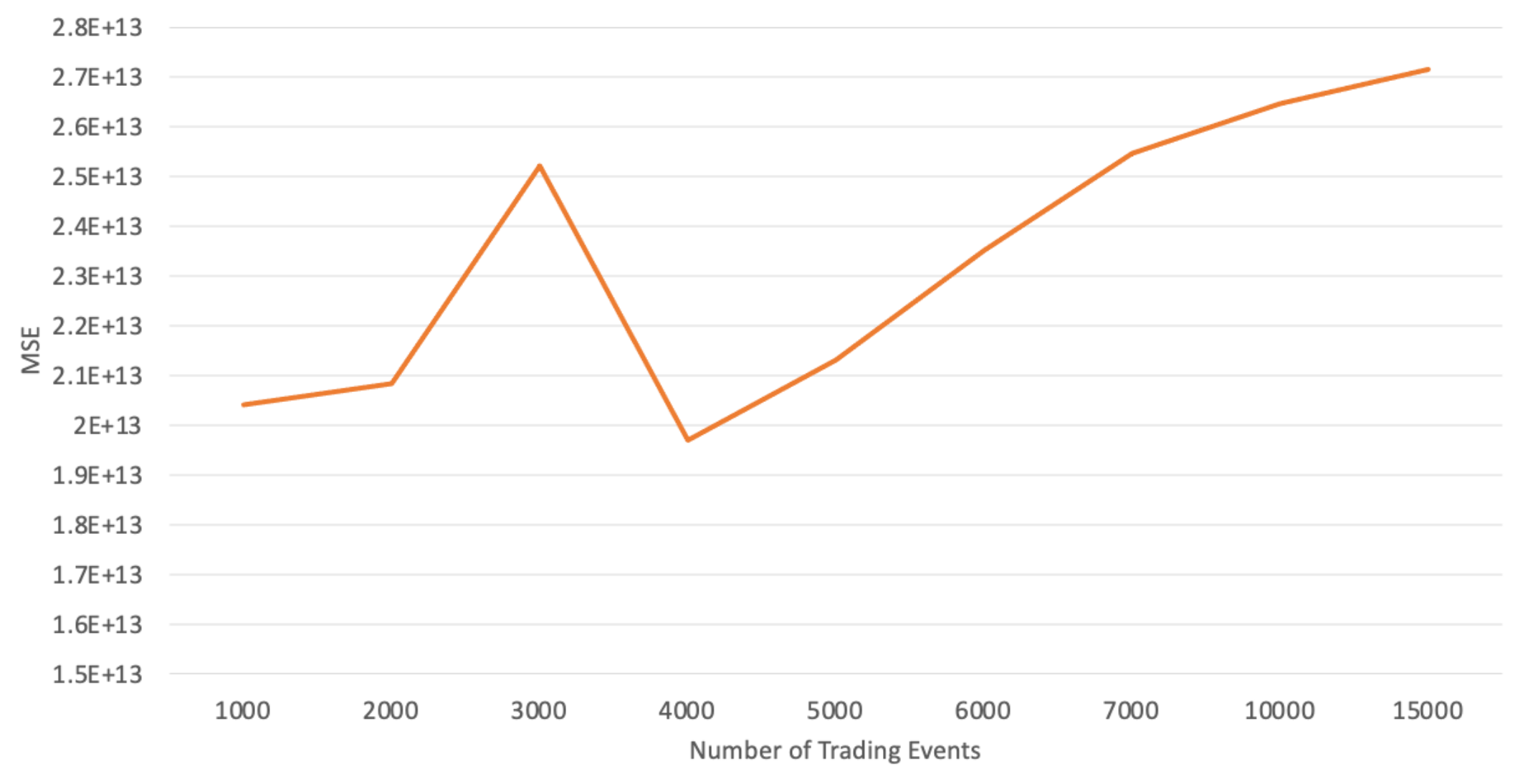}}}
    \\
  \subfloat[Attention LSTM training MSE scores\label{1c}]{%
        \scalebox{0.52}{\includegraphics[width=0.67\linewidth]{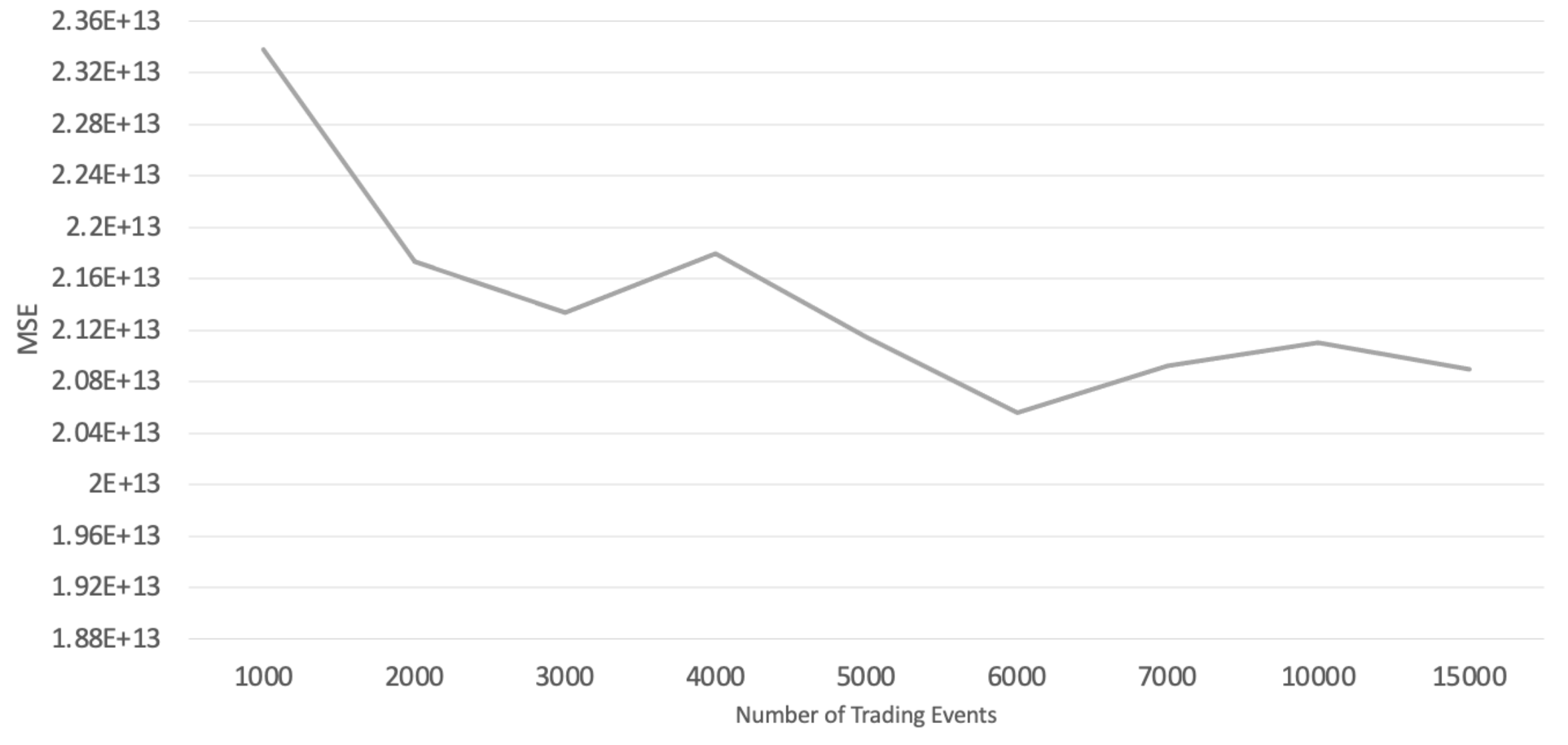}}}
  \subfloat[Attention LSTM testing MSE scores \label{1d}]{%
        \scalebox{0.52}{\includegraphics[width=0.67\linewidth]{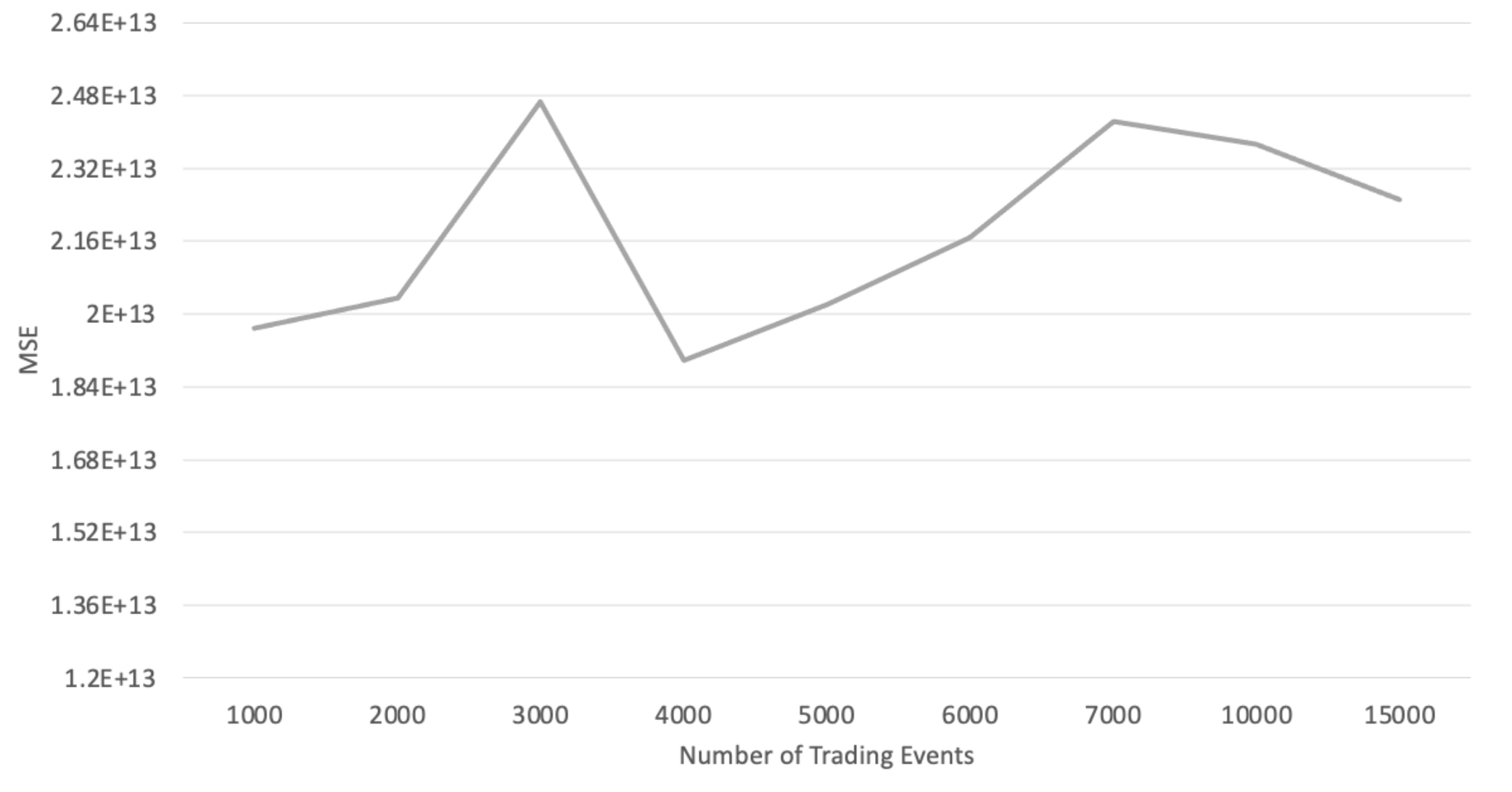}}}  
    \\
  \subfloat[Bidirectional training MSE scores \label{1a}]{%
       \scalebox{0.52}{\includegraphics[width=0.67\linewidth]{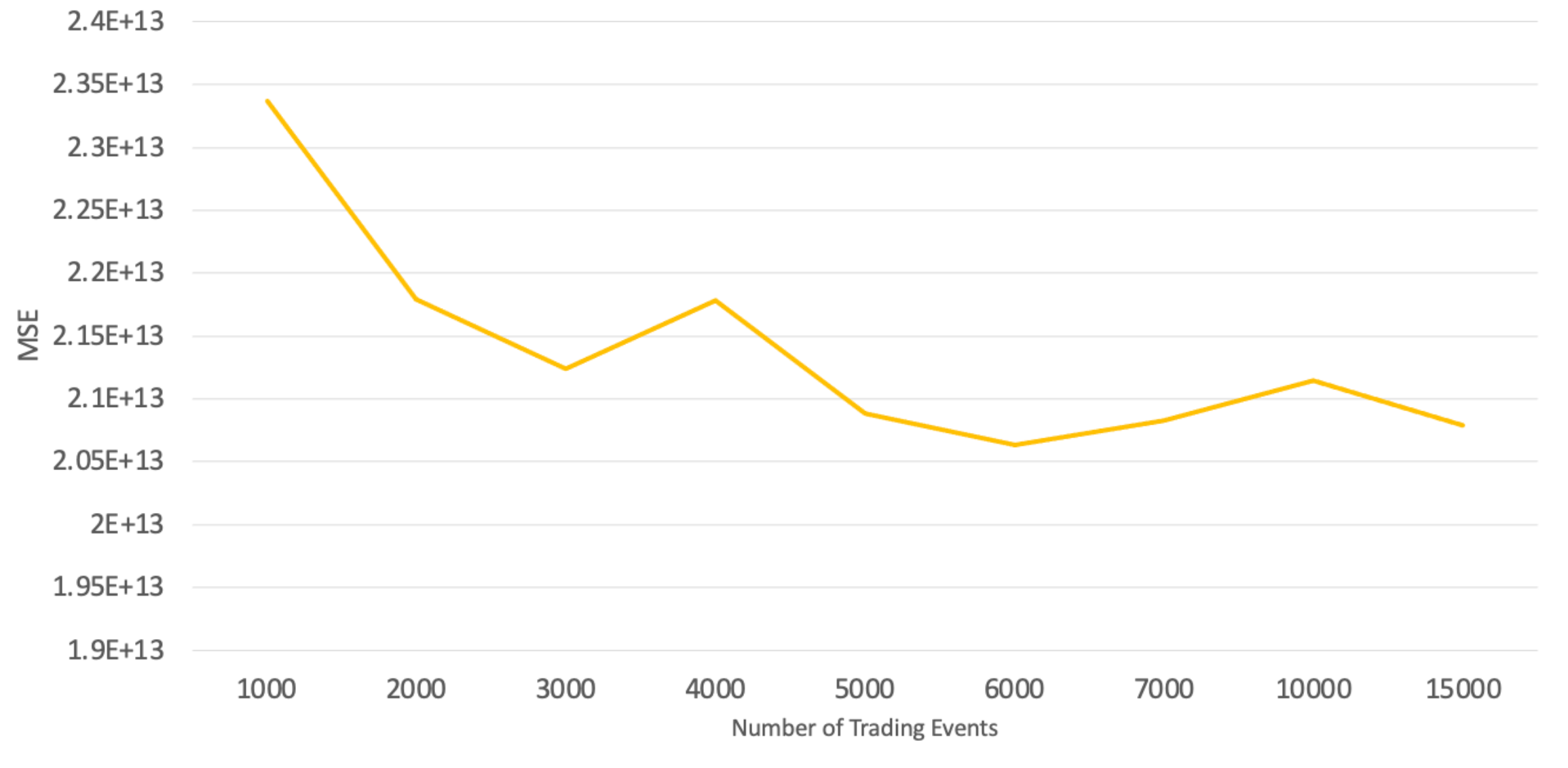}}}
  \subfloat[Bidirectional testing MSE scores \label{1b}]{%
        \scalebox{0.52}{\includegraphics[width=0.67\linewidth]{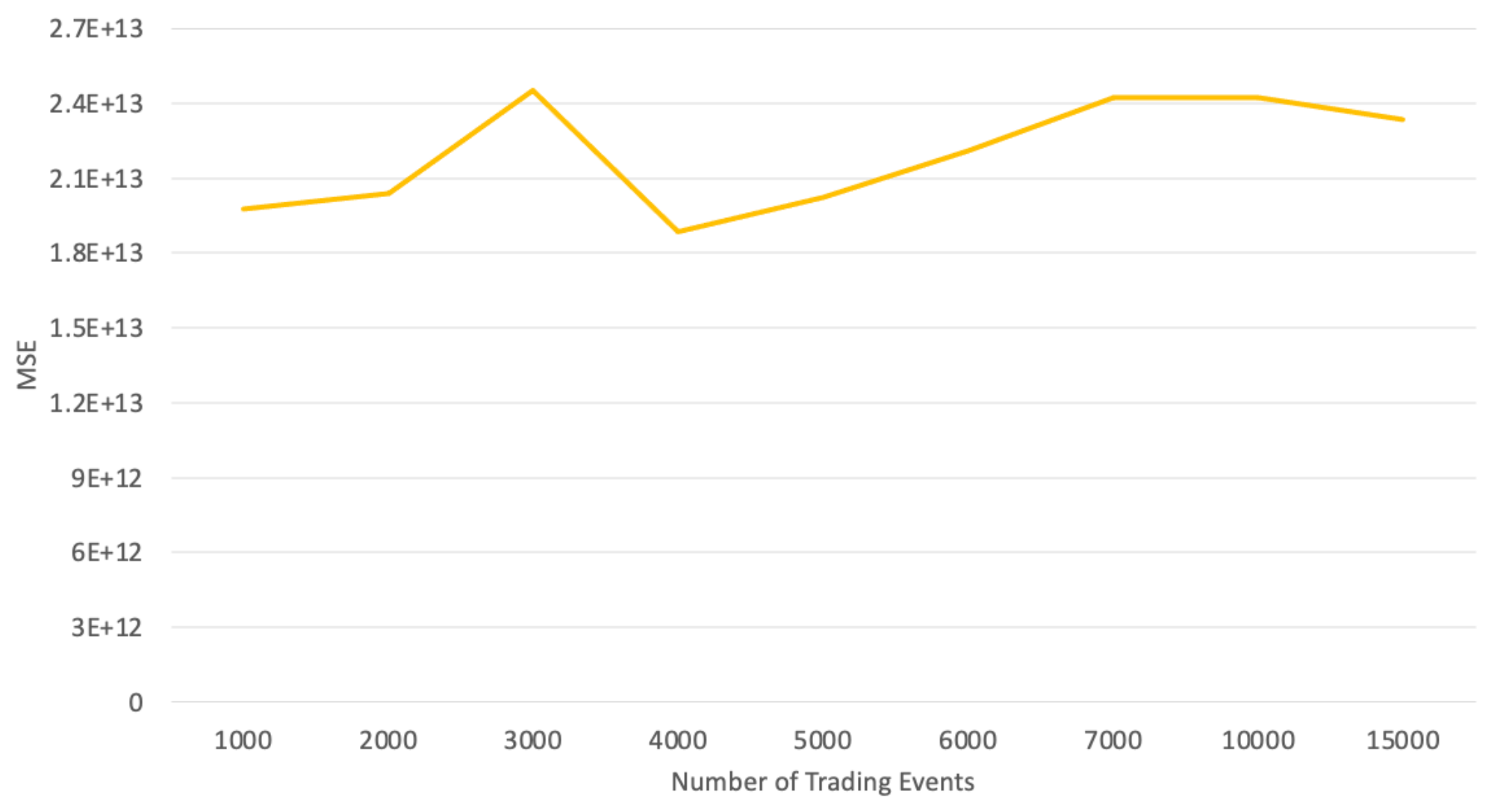}}}
    \\
  \subfloat[GRU training MSE scores \label{1a}]{%
       \scalebox{0.52}{\includegraphics[width=0.67\linewidth]{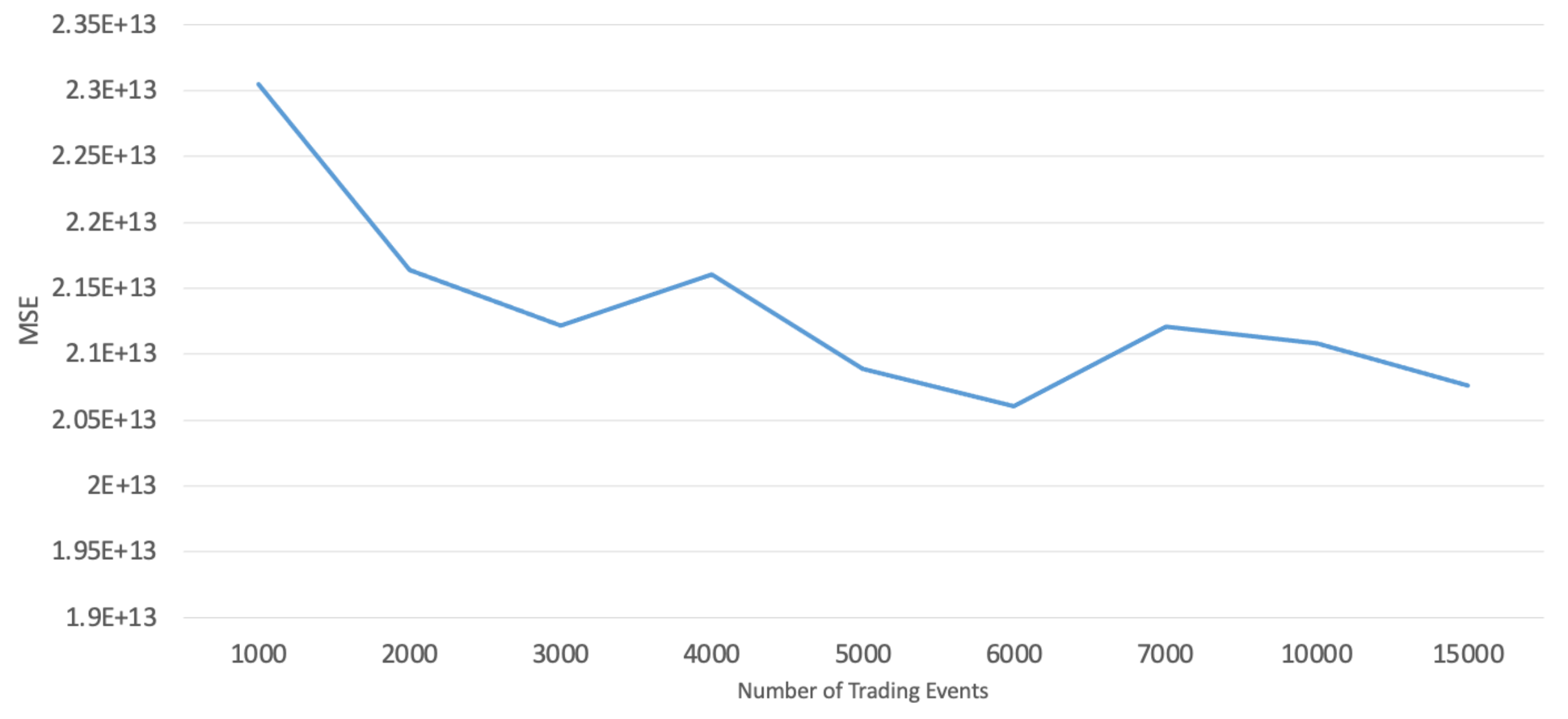}}}
  \subfloat[GRU testing MSE scores \label{1b}]{%
        \scalebox{0.52}{\includegraphics[width=0.67\linewidth]{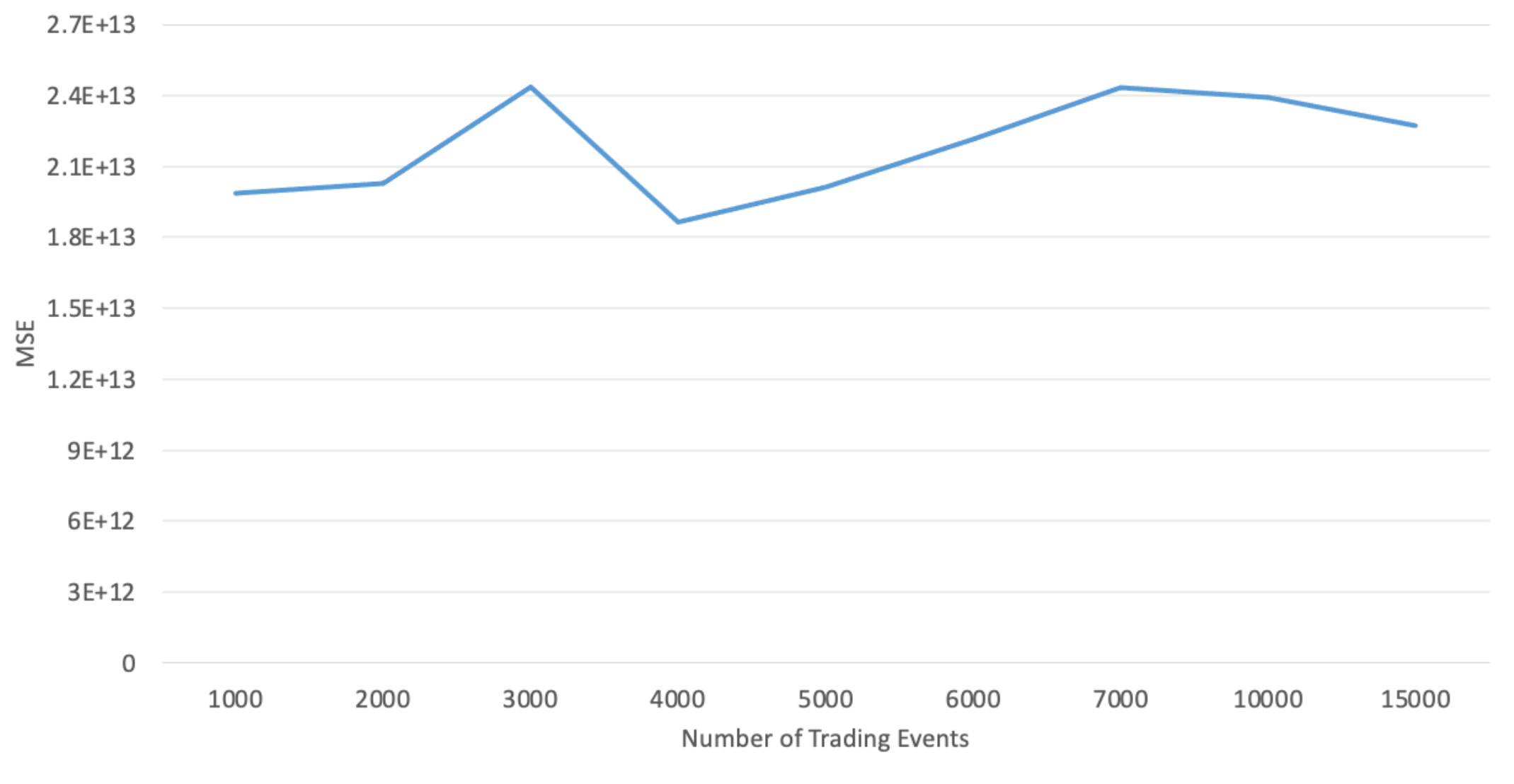}}}
    \\
  \subfloat[Hybrid training MSE scores \label{1a}]{%
       \scalebox{0.52}{\includegraphics[width=0.67\linewidth]{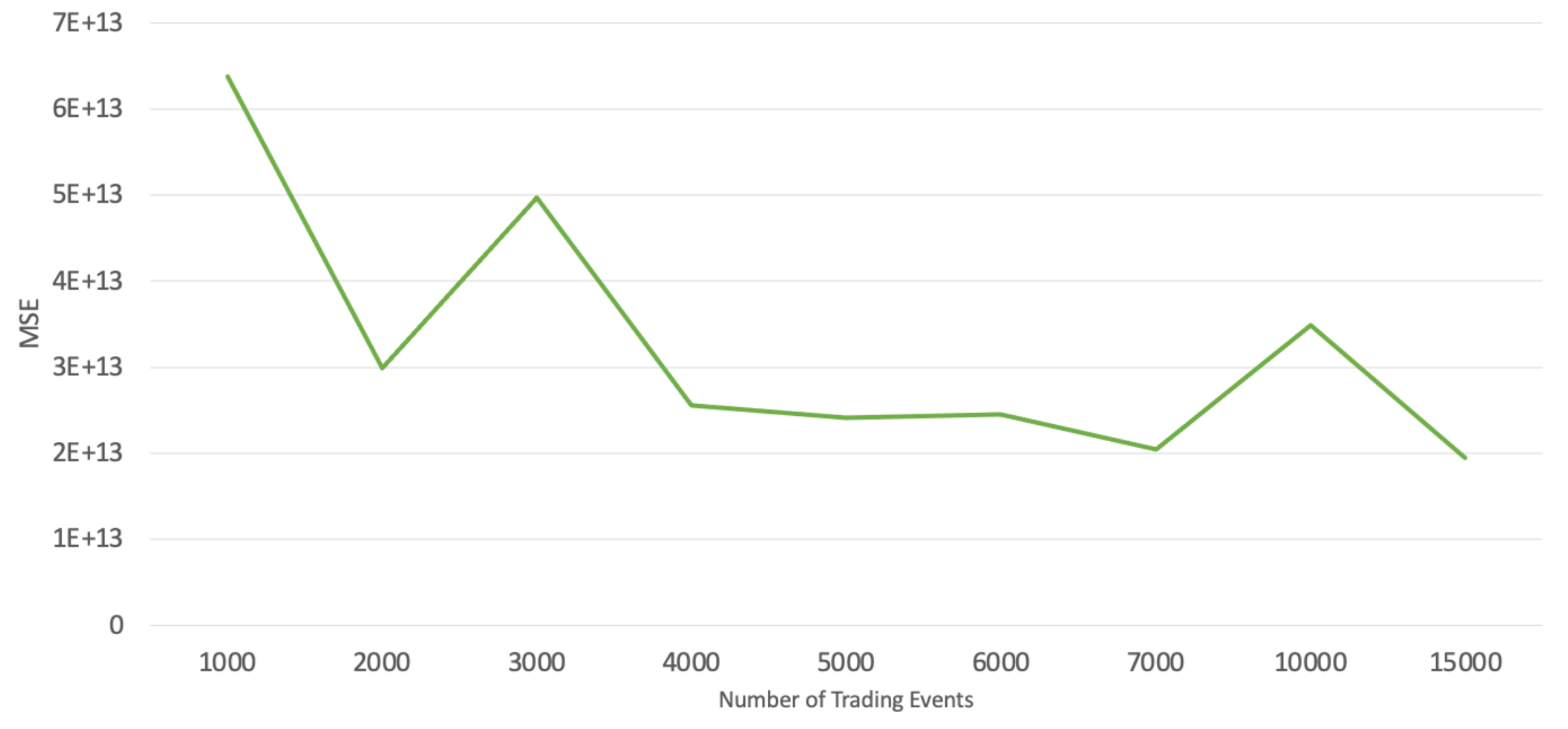}}}
  \subfloat[Hybrid testing MSE scores \label{1b}]{%
        \scalebox{0.52}{\includegraphics[width=0.67\linewidth]{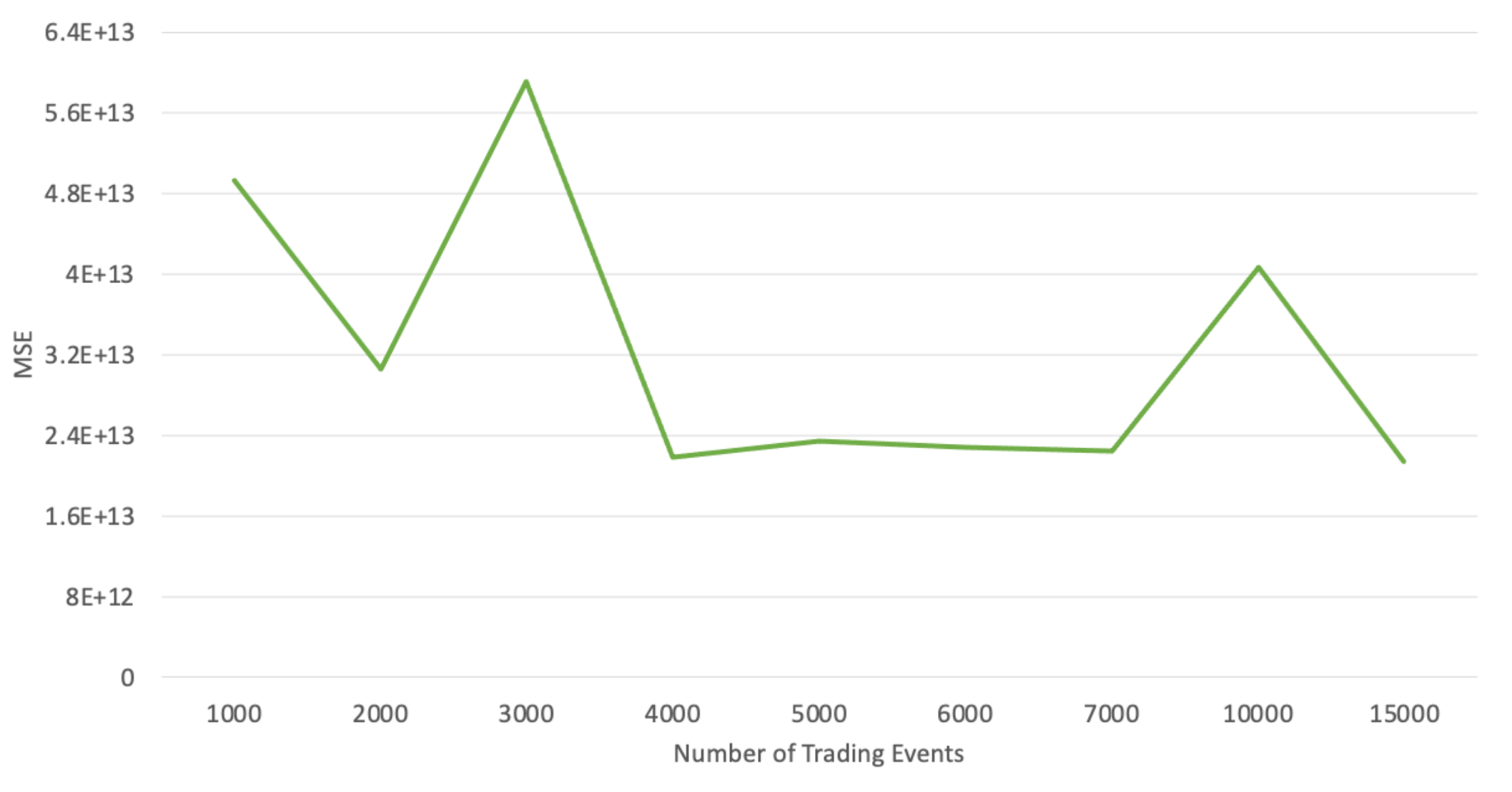}}}  
  \caption{Amazon Long MSE scores based on \hyperref[tab:AmazonLong]{Table \ref{tab:AmazonLong}}.}
  \label{fig:AmazonLong} 
\end{figure*}

We also noticed that all RNN models, including the OPTM-LSTM, are sensitive to the higher mid-price variance. For instance, for Google stock, we see in \hyperref[fig:GoogleShort]{Fig. \ref{fig:GoogleShort}} and \hyperref[fig:GoogleLong]{Fig. \ref{fig:GoogleLong}} that all models exhibit higher MSE scores between 2,000 and 3,000 data samples compare to Amazon (see \hyperref[fig:AmazonShort]{Fig. \ref{fig:AmazonShort}} and \hyperref[fig:AmazonLong]{Fig. \ref{fig:AmazonLong}}) that for the same data size samples the models performed better. This is due to a combination of two things: the short data size length and the variability of the mid-price. More specifically, Google and Amazon exhibit different mid-price variances under the same short data size. For instance, the mid-price variance for 3,000 data samples for Google and Amazon are 7.37E+14 and 5.78E+12 (without any data transformation), respectively. Under the same comparison for 5,000 data samples, the variances are 2.03E+14 (Google) and 5.66E+12 (Amazon) which is a drop in variance of 72\% and 2\%, respectively. The variance acts as a threshold for all the models' performance during training, especially in the case of shorter trading horizons (i.e., smaller data samples). This rapid variance drop plays a significant role in the testing performance which follows the training set. That means that in the case of Google stock, the models will be tested in a set with 72\% less mid-price variability (the rest of the LOB price levels follow a similar variance profile). 
 
Next, \hyperref[tab:GoogleBenchmark]{Table \ref{tab:GoogleBenchmark}} contains the MSE score of the OPTM-LSTM NN against two baseline methods and the five RNN models under two normalization settings and the raw data. The OPTM-LSTM exhibited the same stable behaviour as in the Short Training and Long Training experimental protocols. The newly introduced cell was also able to achieve better results against the two baseline models. To get the reported MSE scores we run again a smaller scale topological grid search for all the RNN and the OPTM-LSTM models. Among the three different input settings (i.e., raw, MinMax, and Zscore) we found that the Zscore normalization was the most challenging setting. Under that setting the differences across the majority of the models were very close to each other but the OPTM-LSTM cell performed better than the rest. The reason is that the cell is very adaptive in terms of learning and in particular, we refer to the learning rate within the cell. The cell is equipped with an internal GD method with a fixed learning rate that can be adjusted according to the selected stock. For instance, in the case of Google, the learning rate was adjusted to 0.0001 and that hyperparameter tuning helped us to achieve better results.

\begin{figure*}[hbt!]
    \centering
  \subfloat[Google: Self-comparison Training Performance \label{1a}]{%
        \scalebox{0.63}{\includegraphics[width=0.72\linewidth]{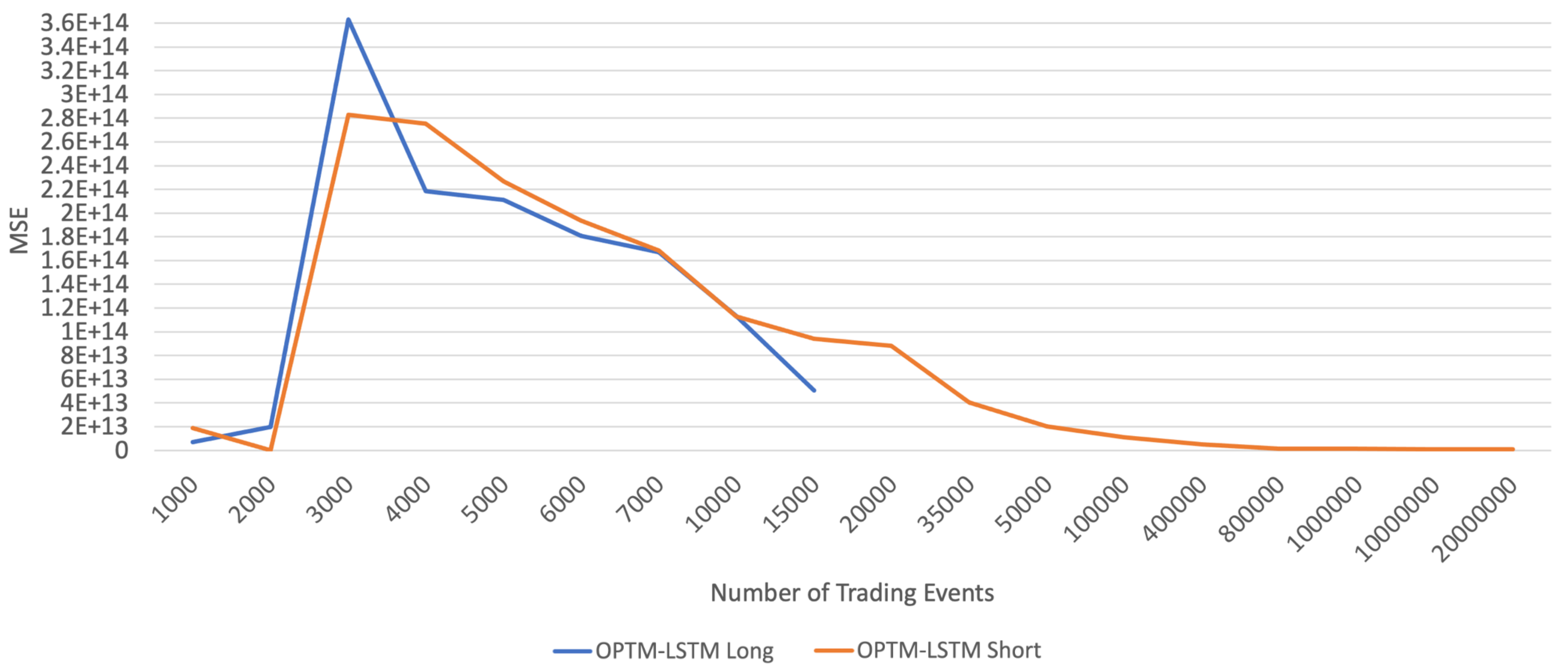}}}
    \hfill
  \subfloat[Google: Self-comparison Testing Performance \label{1b}]{%
         \scalebox{0.63}{\includegraphics[width=0.72\linewidth]{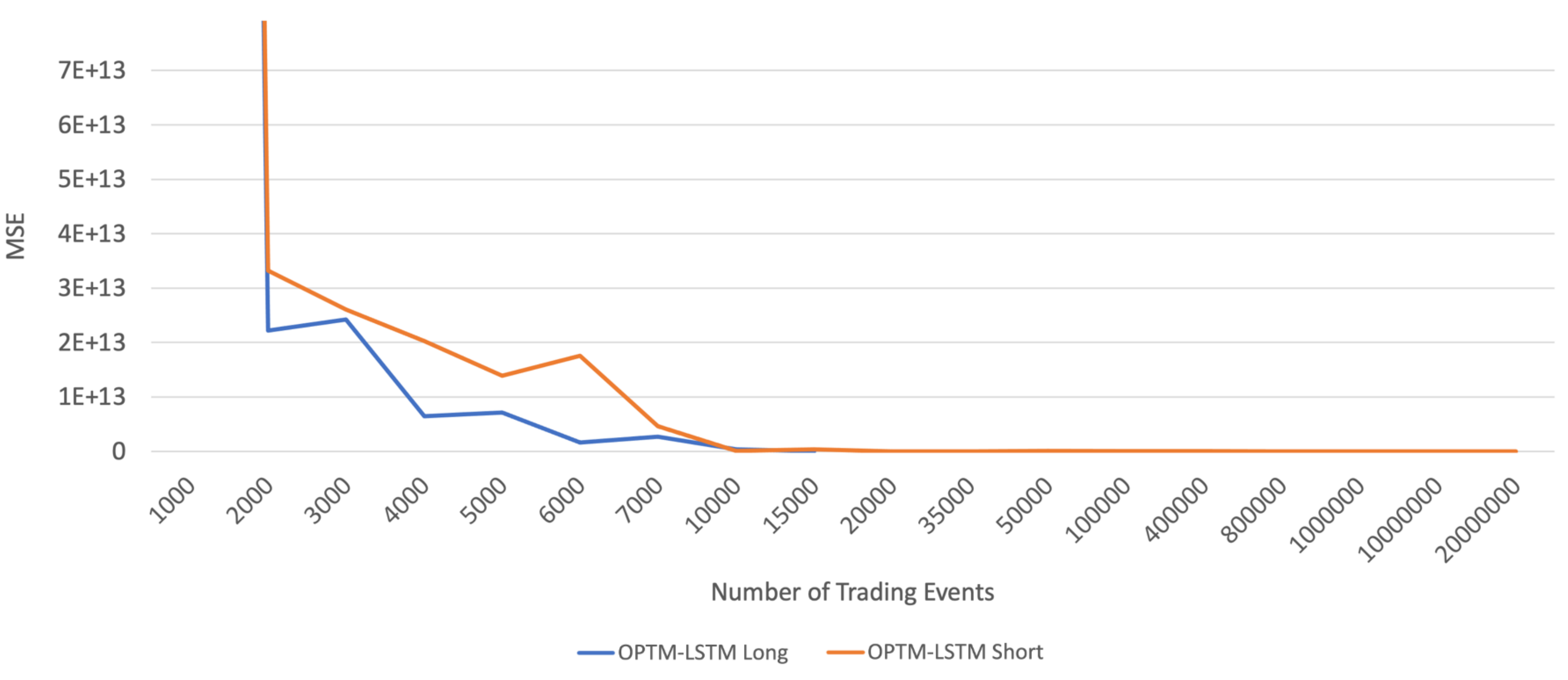}}}
        \\
   \subfloat[Amazon: Self-comparison Training Performance \label{2a}]{%
        \scalebox{0.63}{\includegraphics[width=0.72\linewidth]{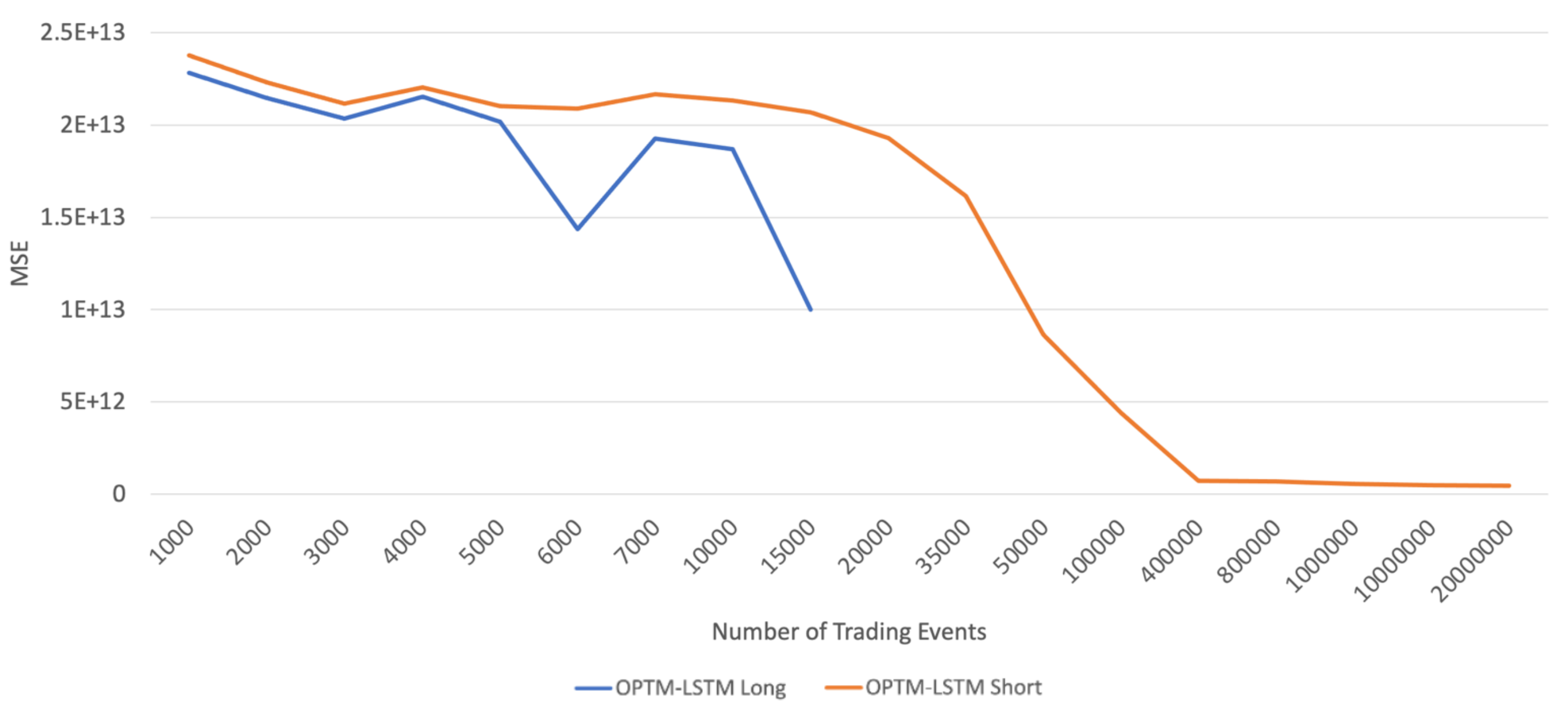}}}
    \hfill
  \subfloat[Amazon: Self-comparison Testing Performance \label{2b}]{%
         \scalebox{0.63}{\includegraphics[width=0.72\linewidth]{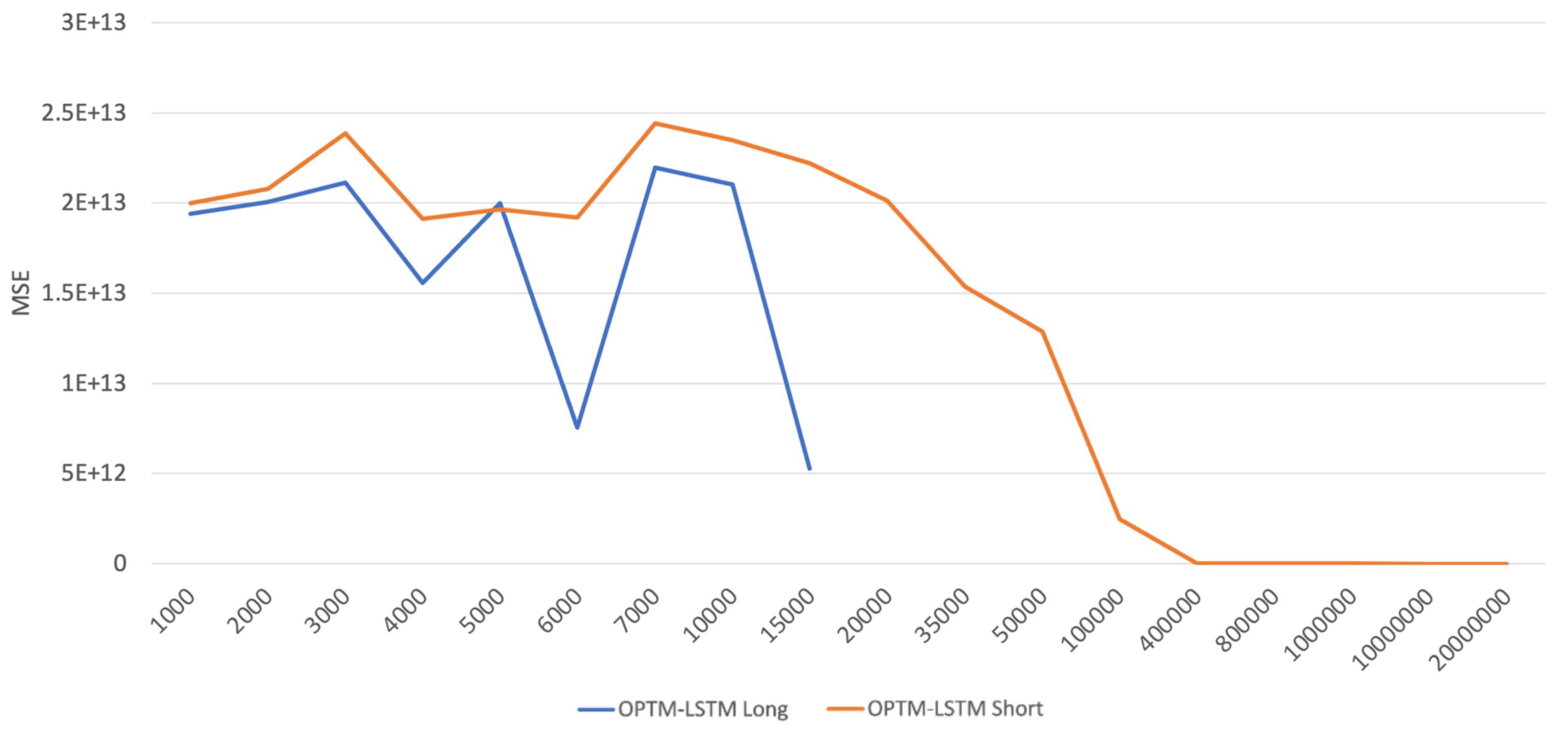}}}
        \\
   \subfloat[Kesko: Self-comparison Training Performance \label{3a}]{%
        \scalebox{0.63}{\includegraphics[width=0.72\linewidth]{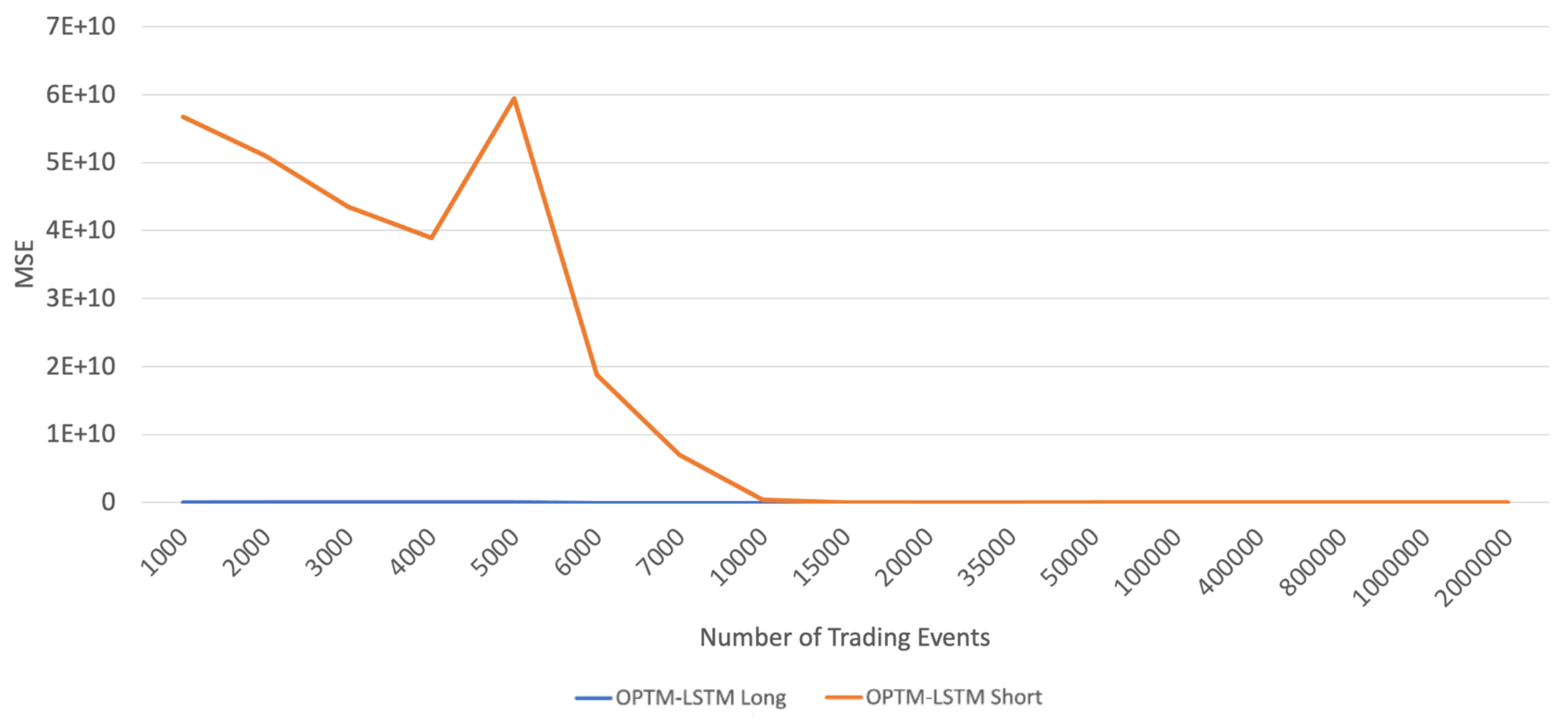}}}
    \hfill
  \subfloat[Kesko: Self-comparison Testing Performance \label{3b}]{%
         \scalebox{0.63}{\includegraphics[width=0.72\linewidth]{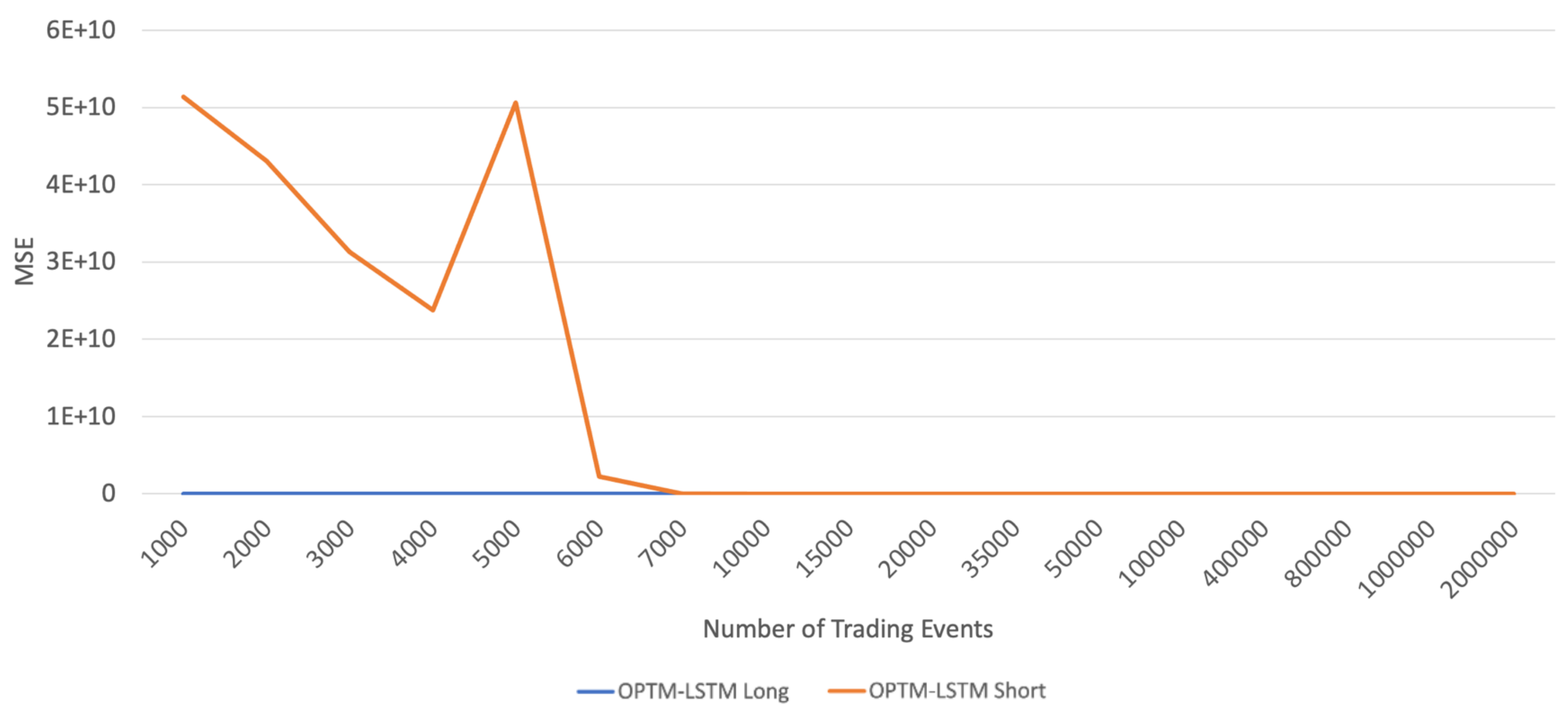}}}
        \\
   \subfloat[Wartsila: Self-comparison Training Performance \label{4a}]{%
        \scalebox{0.63}{\includegraphics[width=0.72\linewidth]{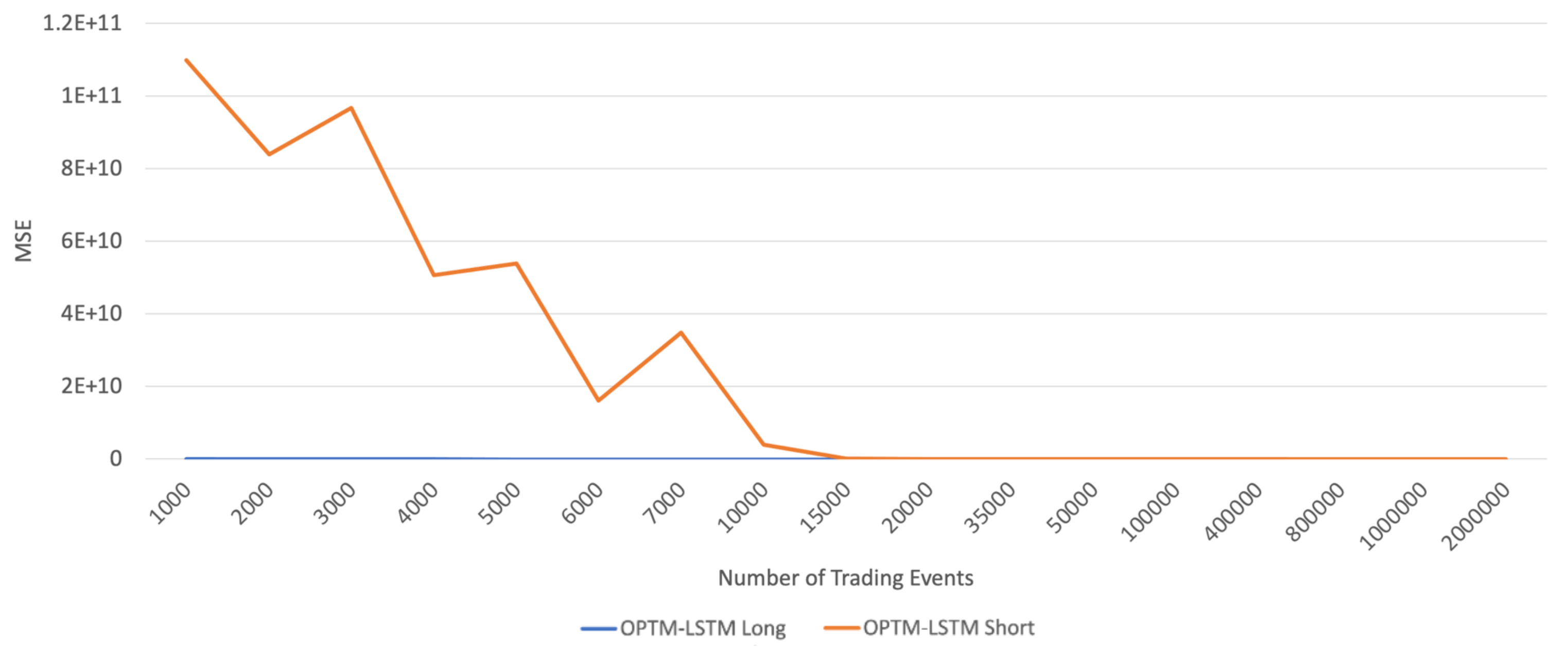}}}
    \hfill
  \subfloat[Wartsila: Self-comparison Testing Performance \label{4b}]{%
         \scalebox{0.63}{\includegraphics[width=0.72\linewidth]{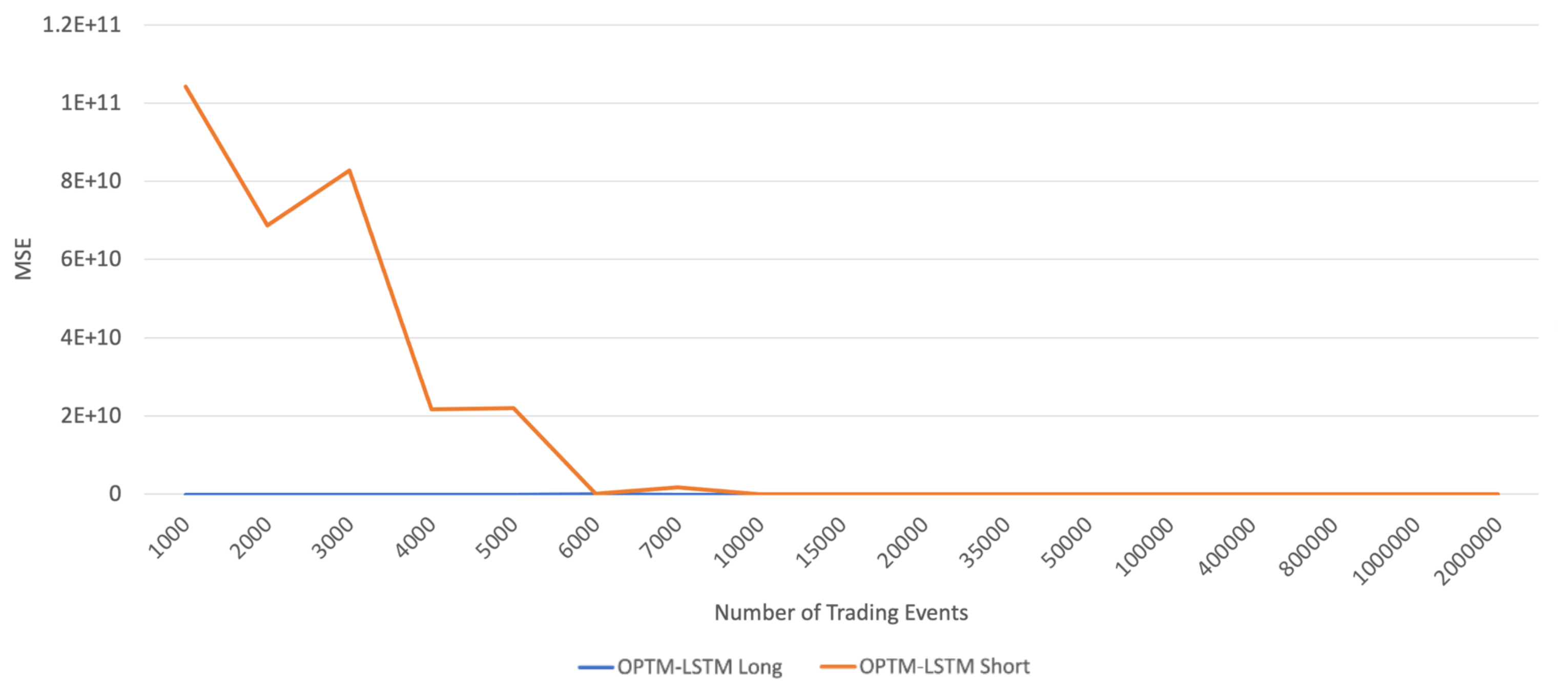}}}  
  \caption{OPTM-LSTM cell self-comparison performance between the Short and Long experimental settings.}
  \label{fig:Self} 
\end{figure*}

The last part of our experimental analysis belongs to the self-comparison between the Short Training against the Long Training experimental protocols. To get a better understanding of that comparison we utilize again the case of Google and we isolate the MSE scores per data size for both protocols - \hyperref[fig:Self]{Fig. \ref{fig:Self}}. It is obvious from the MSE scores that the higher the number of epochs the lower the MSE scores. Amazon, Kesko, and Wartsila exhibited a similar response (see \hyperref[sec:SecondAppendix]{Appendix \ref{sec:SecondAppendix}}). If the ML trader can use up to 15,000 trading events and additional time capacity for the higher number of epochs (e.g., approximately 0.4 seconds per epoch for the 15,000 trading events according to a single GPU machine based on the NVIDIA Ampere A100) then the Long Training is more suitable for the forecasting task otherwise if there is a capacity for additional data samples but limited time then the Short Training protocol is better. 

Finally, we should mention that the overall forecasting behaviour, apart from the OPTM-LSTM which achieved better results, of the five RNN competitors and the two baseline models follows the literature in terms of forecasting performance. For instance, authors in \cite{greff} conclude that the prototype LSTM is robust across a wide range of datasets and any modifications to its internal topology (e.g., removal of gates) did not improve the performance of the prototype LSTM. This is something that we also noticed in the behaviour of the five RNN competitors. The performance of the RNN family models exhibited similar performance behaviour across a wide range of data size, normalization, and feature scenarios to the baseline models and in particular the persistence algorithm which can be identified as a reliable predictor. We noticed a similar behaviour of this baseline model against the prototype LSTM as the authors in \cite{dynamic}. Similarly, authors in \cite{mlops} compared the prototype LSTM against the persistence algorithm with the LSTM outperforming the baseline model. 

\subsection{Limitations and Future Research}
Although our OPTM-LSTM NN offers better forecasting performance in the online HFT universe there are a few limitations that we need to mention. One limitation is the restricted number of stocks and trading horizons. Despite the fact that we utilized indicative examples of high-liquid and less-liquid stocks we believe that a wider selection of stocks will provide more insights into the behaviour of the OPTM-LSTM model. Another limitation of our study is the selected trading horizon. We limited the selected number of trading events only when we realize a steady decrease in the MSE score performance. Potentially, an even lower MSE score could have been achieved by incorporating a larger trading horizon. Last but not least, we believe that a more advanced optimization method can be utilized as a part of the OPTM-LSTM cell. Apart from the aforementioned limitations, the development of the OPTM-LSTM cell opens additional research avenues. For instance, the same internal cell architecture can be extended for classification tasks. The current internal regression supervised problem is connected directly to the final forecasting objective which is a regression problem. It will be very interesting to utilize several classification scenarios as the cell's internal supervised problem and connect them to a final classification objective. The present work is classified as a narrow artificial intelligence (AI) application. We believe that our research developments here can be utilized for other online tasks such as vision and fully autonomous or full-self driving tasks. 

\section{Conclusion}\label{section:Con}
\noindent The task of online learning and forecasting in the HFT space is challenging and we believe to the best of our knowledge that the existing literature has not addressed that challenge fully. In particular, we propose an optimum output LSTM cell architecture for the task of HFT LOB's mid-price forecasting via an online experimental protocol. The motivation for developing this optimum output cell is that several machine learning and deep learning models utilize a stale approach to training and learning. That means that the ML trader builds a predictive model which has a stale topology during the training/learning process and is disconnected from the actual forecasting objective. We named the new cell Optimum Output LSTM (OPTM-LSTM) cell because we capitalize on the existing LSTM cell architecture by reshuffling strategically and online the final cell's output. This new cell is equipped with an internal supervised and non-forecasting regression task, which acts as a guarantor for the ranking of the internal states and gates. This process is taking place online which means that the optimum output cell is directly connected to the latest information in relation to the predicted value. We tested the effectiveness of our new cell on two high-liquid stocks (i.e., Amazon and Google) and two less-liquid stocks (i.e., Kesko and Wartsila) via several data size scenarios and training routines. Our method achieved better performance in terms of MSE scores against a wide range of RNN and baseline models. We believe that our work opens additional research avenues not only in the space of HFT online AI learning but hopefully in other fields such as vision and fully-autonomous driving. 

\section*{Acknowledgments}
\noindent The authors would like to thank CSC-IT Center for Science, Finland, for the generous computational resources.

\newpage
\appendices
\section{}
\label{sec:FirstAppendix}

\noindent LSTM's backward pass, named Backpropagation Through Time (BPTT), is the learning/update mechanism of all the trainable parameters which are $W_f$, $W_i$, $W_{\tilde{c}}$, $W_o$, $U_f$, $U_i$, $U_{\tilde{c}}$, and $U_o$ including biases. Despite the fact that we mainly utilize a many-to-one approach in our experimental protocol we explain below the general idea that the loss is calculated and then summed up for the entire look-back period. In the many-to-one case, the remaining steps of each individual timestep that the temporal is not considered can be assumed by the ML trader that they are equal to zero. We will explain the BPTT process based on the partial derivatives w.r.t. (i.e., with respect to): (1) internal states (i.e., candidate gate, cell state, and hidden state), (2) internal gates (i.e., output gate, forget gate, and input gate), and (3) shared weights (i.e., $V$, $W$, and $U$). At this stage we should introduce the notation V which represents the shared weights between LSTM's individual timestep prediction and the selected loss function - in our case will be the MSE. We will also use for our partial derivations the linear activation transformation which is suitable for our regression task and comes as a default option for the Dense layer in TensorFlow. Dimensions of the internal gates and states and their equivalent partial derivatives retain their dimensional profile throughout BPTT. 

\begin{itemize}
    \item \textbf{Hidden state} $h_t$:
    \begin{equation}\label{eq:1}
        \frac{\partial L_t}{\partial{h}_t} = \frac{\partial L_t}{\partial{\hat{y}_t}} \cdot \frac{\partial{\hat{y}_t}}{\partial{h_t}},
    \end{equation}
    \begin{equation}\label{eq2}
        \frac{\partial L_t}{\partial \hat{y}_t} = \frac{\partial}{\partial{\hat{y}_t}}[\frac{1}{2} \cdot (y_t - \hat{y}_t)^2] = \hat{y}_t - y_t,
    \end{equation}
    \begin{equation}\label{eq3}
        \frac{\partial{\hat{y}_t}}{\partial{h_t}} = \frac{\partial}{\partial h_t} [V \cdot h_t] = V,
    \end{equation}
    \begin{equation}\label{eq4}
        \frac{\partial L_t}{\partial{h}_t} = (\hat{y}_t - y_t) \cdot V^T.
    \end{equation}
\end{itemize}

\begin{itemize}
    \item \textbf{Cell state} $c_t$:
    \begin{equation}\label{eq5}
        \frac{\partial{L_t}}{\partial{c_t}} = \frac{\partial L_t}{\partial{\hat{y}_t}} \cdot \frac{\partial{\hat{y}_t}}{\partial{h_t}} \cdot \frac{\partial{h_t}}{\partial{c_t}},
    \end{equation}
    where:
    \begin{equation}\label{eq6}
        \frac{\partial{h_t}}{\partial{c_t}} = \frac{\partial}{\partial{c_t}}[o_t \cdot tanh(c_t)] = o_t \cdot (1-tanh^2(c_t)),
    \end{equation}
    and based on (\hyperref[eq4]{\ref{eq4}}) and (\hyperref[eq6]{\ref{eq6}}) the loss $L$ w.r.t. the cell state $c$ at time $t$ is:
    \begin{equation}
         \frac{\partial{L_t}}{\partial{c_t}} = (\hat{y}_t - y_t) \cdot V^T \odot o_t \cdot (1-tanh^2(c_t)).
    \end{equation}
\end{itemize}
\begin{itemize}
    \item \textbf{Candidate gate} $\tilde{c}_t$:
    \begin{equation}\label{eq7}
        \frac{\partial{L_t}}{\partial{\tilde{c}_t}} = \frac{\partial{L}_t}{\partial{\hat{y}_t}} \cdot \frac{\partial{\hat{y}_t}}{\partial{h_t}} \cdot \frac{\partial{h_t}}{\partial{c_t}} \cdot \frac{\partial{c_t}}{\partial{\tilde{c}_t}},
    \end{equation}
    where: 
    \begin{equation}\label{eq8}
        \frac{\partial{c_t}}{\partial{\tilde{c}_t}} = \frac{\partial}{\partial{\tilde{c}_t}}[f_t \cdot c_{t-1} + i_t \cdot \tilde{c}_t] = i_t,
    \end{equation}
    and based on (\hyperref[eq4]{\ref{eq4}}) and (\hyperref[eq5]{\ref{eq5}}) the loss $L$ w.r.t. the candidate gate $\tilde{c}$ at time $t$ is:
    \begin{equation}\label{eq9}
        \frac{\partial{L_t}}{\partial{\tilde{c}_t}} = (\hat{y}_t - y_t)^2 \cdot V^T \odot o_t \cdot (1-tanh^2(c_t)) \odot i_t,
    \end{equation}
\end{itemize}
\begin{itemize}
    \item \textbf{Output gate} $o_t$:
    \begin{equation}\label{eq10}
        \frac{\partial{L_t}}{\partial{o_t}} = \frac{\partial L_t}{\partial{\hat{y}_t}} \cdot \frac{\partial{\hat{y}_t}}{\partial{h_t}} \cdot \frac{\partial h_t}{\partial{o_t}},
    \end{equation}
    where:
    \begin{equation}\label{eq11}
        \frac{\partial h_t}{\partial{o_t}} = \frac{\partial}{\partial{o}_t}[ \textit{o}_t \odot tanh(c_t)] = tanh(c_t),
    \end{equation}
    and based on (\hyperref[eq4]{\ref{eq4}}) and (\hyperref[eq11]{\ref{eq11}}) the loss $L$ w.r.t. the output gate $o$ at time $t$ is:
    \begin{equation}
        \frac{\partial{L_t}}{\partial{o_t}} = (\hat{y}_t - y_t)^2 \cdot V^T \odot tanh(c_t).
    \end{equation}
\end{itemize}

\begin{itemize}
    \item \textbf{Input gate} $i_t$:
    \begin{equation}\label{eq12}
        \frac{\partial{L_t}}{\partial{i_t}} = \frac{\partial{L_t}}{\partial{\hat{y}_t}} \cdot \frac{\partial{\hat{y}_t}}{\partial{h_t}} \cdot \frac{\partial{h_t}}{\partial{c_t}} \cdot \frac{\partial{c_t}}{\partial{i_t}},
    \end{equation}
    where:
    \begin{equation}\label{eq13}
        \frac{\partial{c_t}}{\partial{i_t}} = \frac{\partial}{\partial{i_t}}[f_t \odot c_{t-1} + i_t \odot \tilde{c}_{t}] = \tilde{c}_t,
    \end{equation}
    
    and based on (\hyperref[eq7]{\ref{eq7}}) and (\hyperref[eq13]{\ref{eq13}}) the loss $L$ w.r.t. the input gate $i$ at time $t$ is:
    
    \begin{equation}\label{eq14}
        \frac{\partial{L_t}}{\partial{i_t}} = (\hat{y}_t - y_t) \cdot V^T \odot o_t \cdot (1-tanh^2(c_t)) \odot \tilde{c}_t.
    \end{equation}
\end{itemize}

\begin{itemize}
    \item \textbf{Forget gate} $f_t$:
    \begin{equation}\label{eq15}
        \frac{\partial{L_t}}{\partial{f_t}} = \frac{\partial{L_t}}{\partial{\hat{y}_t}} \cdot \frac{\partial{\hat{y}_t}}{\partial{h_t}} \cdot \frac{\partial{h_t}}{\partial{c_t}} \cdot \frac{\partial{c_t}}{\partial{f_t}},
    \end{equation}
    where:
    \begin{equation}\label{eq16}
        \frac{\partial{c_t}}{\partial{f_t}} = \frac{\partial{}}{\partial{f_t}}[f_t \odot c_{t-1} + i_t \odot \tilde{c}_{t}] = c_{t-1},
    \end{equation}

    and based on (\hyperref[eq7]{\ref{eq7}}) and (\hyperref[eq16]{\ref{eq16}}) the loss $L$ w.r.t. the forget gate $f$ at time $t$ is:
    
    \begin{equation}\label{eq17}
        \frac{\partial{L_t}}{\partial{f_t}} = (\hat{y}_t - y_t) \cdot V^T \odot o_t \cdot (1-tanh^2(c_t)) \odot c_{t-1}.
    \end{equation}
    where $L_t$ is the temporal loss function MSE $= \frac{1}{2}(y_t - \hat{y}_t)^2$ at timestep t, $y_t$ is the provided label, $\hat{y}_t$ is the predicted value, and $\cdot$ is the dot product.
\end{itemize}  

\noindent Next, we present the partial derivatives w.r.t. the trainable parameters/weights V, W, and U, where V represents the shared weights between LSTM's cell output $h_t$ and the predicted value $\hat{y}_t$ at time instance, $W$ represents the weights between the input tensor $x_t$ and the internal gates, and $U$ represents the weights between the previous hidden state $h_{t-1}$ and the internal gates at time instance $t$. More specifically:

\begin{sloppypar}

\begin{itemize}
    \item \textbf{Weights V}: For the entire length of the look-back period the loss function w.r.t. $V$ is:
    \begin{equation}\label{eq18}
        \frac{\partial{L}}{\partial{V}} =  \mathlarger{\mathlarger{\sum}}_{t=1}^{T} \frac{\partial{L}_t}{\partial{\hat{y}_t}}  \cdot \frac{\partial{\hat{y}_t}}{\partial{V}} = \mathlarger{\mathlarger{\sum}}_{t=1}^{T} h_t^T \cdot (\hat{y}_t - y_t).
    \end{equation}
\end {itemize}

\begin{itemize}
    \item \textbf{Weights U}: $W$ is a concatenation of the LSTM's cell matrices $W_i$, $W_f$, $W_o$, and $W_{\tilde{c}}$. The loss function $L$ will be updated w.r.t. these individual weights, as follows:
    \begin{strip}   
    \begin{equation}\label{eq19}
        \frac{\partial{L}}{\partial{U_i}} =  \mathlarger{\mathlarger{\sum}}_{t=1}^{T}
        \frac{\partial{L_t}}{\partial{i_t}} \cdot \frac{\partial{i_t}}{\partial{U_i}}\\
        =\mathlarger{\mathlarger{\sum}}_{t=1}^{T} \Bigg\{ \Big[ (\hat{y}_t - y_t)\cdot V^T \odot o_t \cdot (1-tanh^2(c_t))\odot \tilde{c}_t) \Big]^T\cdot i_t \cdot (1-i_t) \Bigg\} \odot h_{t-1},
    \end{equation}

    \begin{equation}\label{eq20}
    \frac{\partial{L}}{\partial{U_f}} =  \mathlarger{\mathlarger{\sum}}_{t=1}^{T}
    \frac{\partial{L_t}}{\partial{f_t}} \cdot \frac{\partial{f_t}}{\partial{U_f}}\\
    =\mathlarger{\mathlarger{\sum}}_{t=1}^{T} \Bigg\{ \Big[ (\hat{y}_t - y_t)\cdot V^T \odot o_t \cdot (1-tanh^2(c_t)) \odot c_{t-1}) \Big]^T \cdot f_t \cdot (1-f_t) \Bigg\} \odot h_{t-1}, 
    \end{equation}

    \begin{equation}\label{eq21}
    \frac{\partial{L}}{\partial{U_o}} =  \mathlarger{\mathlarger{\sum}}_{t=1}^{T}
    \frac{\partial{L_t}}{\partial{o_t}} \cdot \frac{\partial{o_t}}{\partial{U_o}}\\
    =\mathlarger{\mathlarger{\sum}}_{t=1}^{T} \Bigg\{ \Big[ (\hat{y}_t - y_t)\cdot V^T \odot (tanh(c_t)) \Big]^T \cdot o_t \cdot (1-o_t) \Bigg\} \odot h_{t-1},
    \end{equation}

    \begin{equation}\label{eq22}
    \frac{\partial{L}}{\partial{U_{\tilde{c}_t}}} =  \mathlarger{\mathlarger{\sum}}_{t=1}^{T}  
    \frac{\partial{L_t}}{\partial{\tilde{c}_t}} \cdot \frac{\partial{\tilde{c}}}{\partial{U_{\tilde{c}}}}\\
    =\mathlarger{\mathlarger{\sum}}_{t=1}^{T} \Bigg\{ \Big[ (\hat{y}_t - y_t)\cdot V^T \odot o_t \cdot (1-tanh^2(c_t)) \odot i_t \Big]^T \cdot (1-\tilde{c}_t^2) \Bigg\} \odot h_{t-1}.
    \end{equation}

    \item \textbf{Weights W}: The \ partial \ derivative \ of \ the \ loss \ function \\
        $L$ \ w.r.t. the \ components \ of \ the \ weight \ matrix $W$ is:\\
        \begin{equation}\label{eq23}
        \frac{\partial{L}}{\partial{W_i}} =  \mathlarger{\mathlarger{\sum}}_{t=1}^{T}
        \frac{\partial{L_t}}{\partial{i_t}} \cdot \frac{\partial{i_t}}{\partial{W_{i}}}\\
        =\mathlarger{\mathlarger{\sum}}_{t=1}^{T} \Bigg\{x_t^T \cdot (\hat{y}_t - y_t) \cdot V^T \odot o_t \cdot (1-tanh^2(c_t)) \odot \tilde{c}_t   \Bigg\} \odot i_t \cdot (1-i_t),
        \end{equation}

        \begin{equation}\label{eq24}
        \frac{\partial{L}}{\partial{W_f}} =  \mathlarger{\mathlarger{\sum}}_{t=1}^{T}
        \frac{\partial{L_t}}{\partial{f_t}} \cdot \frac{\partial{f_t}}{\partial{W_{f}}}\\
        =\mathlarger{\mathlarger{\sum}}_{t=1}^{T} \Bigg\{x_t^T \cdot (\hat{y}_t - y_t) \cdot V^T \odot o_t \cdot (1-tanh^2(c_t)) \odot c_{t-1} \Bigg\} \odot f_t \cdot (1-f_t),
        \end{equation}

        \begin{equation}\label{eq25}
        \frac{\partial{L}}{\partial{W_o}} =  \mathlarger{\mathlarger{\sum}}_{t=1}^{T}
        \frac{\partial{L_t}}{\partial{o_t}} \cdot \frac{\partial{o_t}}{\partial{W_{o}}}\\
        =\mathlarger{\mathlarger{\sum}}_{t=1}^{T} \Bigg\{x_t^T \cdot (\hat{y}_t - y_t)^2 \cdot V^T \odot tanh(c_t) \Bigg\} \odot o_t \cdot (1-o_t),
        \end{equation}

        \begin{equation}\label{eq26}
        \frac{\partial{L}}{\partial{W_{\tilde{c}}}} =  \mathlarger{\mathlarger{\sum}}_{t=1}^{T}  
        \frac{\partial{L_t}}{\partial{\tilde{c}_t}} \cdot \frac{\partial{\tilde{c}_t}}{\partial{W_{\tilde{c}}}}\\
        =\mathlarger{\mathlarger{\sum}}_{t=1}^{T} \Bigg\{x_t^T \cdot (\hat{y}_t - y_t)^2 \cdot V^T \odot o_t \cdot (1-tanh^2(c_t)) \odot i_t \Bigg\} \odot \tilde{c}_t \cdot (1-\tilde{c}_t).
        \end{equation}

        \item The generalized temporal loss $L$ w.r.t. to $U$ and $W$ at \\
        any time instance $t$ 
        cell in the BPTT process, is:

        \begin{equation}\label{eq:eq27}
        \frac{\partial{L_t}}{\partial{U}} = \frac{\partial{L_t}}{\partial{\hat{y}_t}} \cdot \frac{\partial{\hat{y}_t}}{\partial{h_t}} \cdot \frac{\partial{h_t}}{\partial{c_t}} \cdot \Bigg[ \prod_{t=2}^{T} \frac{\partial{c_t}}{\partial{c_{t-1}}} \Bigg] \cdot \frac{\partial{c_1}}{\partial{U}}, \text{ and }
        \frac{\partial{L_t}}{\partial{W}} = \frac{\partial{L_t}}{\partial{\hat{y}_t}} \cdot \frac{\partial{\hat{y}_t}}{\partial{h_t}} \cdot \frac{\partial{h_t}}{\partial{c_t}} \cdot \Bigg[ \prod_{t=2}^{T} \frac{\partial{c_t}}{\partial{c_{t-1}}} \Bigg] \cdot \frac{\partial{c_1}}{\partial{W}}.
        \end{equation}
        \end{strip}
\end{itemize}
\end{sloppypar}

\clearpage
\newpage
\section{}
\subsection*{}
\label{sec:SecondAppendix}

\begin{table*}[hbt!]
\centering
\captionsetup{width=1.00\textwidth}
\caption{Best-performing candidates based on a topological grid search for the Finnish stocks.}
\scalebox{0.710}{
\begin{tabular}{cclccccl}
\cmidrule[2pt]{1-3}\cmidrule[2pt]{6-8}
\multirow{1}{*}{\textbf{Stock}}&{\textbf{Model}}&\multirow{1}{*}{\textbf{Topology}}& \qquad & \qquad &\textbf{Stock} & \textbf{Model} & \textbf{Topology}\\
\cmidrule{1-3}\cmidrule{6-8}
Kesko & LSTM & $\bullet$ LSTM layer with 32 units  &\qquad & \qquad &Wartsila & LSTM & $\bullet$ LSTM layer with 32 units \\
&&              $\bullet$ Dense layer with 1 unit   &\qquad &  \qquad &&            & $\bullet$ Dense layer with 1 unit \\ 
&&              $\bullet$ Adam optimizer         &\qquad &  \qquad &&            & $\bullet$ Adam optimizer \\ 
&&              $\bullet$ Batch size of 32 samples  &\qquad &  \qquad &&            & $\bullet$ Batch size of 32 samples \\ 
\cmidrule{2-3}\cmidrule{7-8}
       & LSTM       & $\bullet$ LSTM layer with 32 units          &\qquad &\qquad & & LSTM      & $\bullet$ LSTM layer with 32 units \\ 
       & with       & $\bullet$ PReLU                             &\qquad &\qquad & & with      & $\bullet$ PReLU \\
       & Attention  & $\bullet$ Attention layer                   &\qquad &\qquad & & Attention & $\bullet$ Attention layer \\
       &            & $\bullet$ Dense layer with 16 units         &\qquad &\qquad & &           & $\bullet$ Dense layer with 16 units\\
       &            & $\bullet$ Dense layer with 1 unit           &\qquad &\qquad & &           & $\bullet$ Dense layer with 1 unit\\
       &            & $\bullet$ Adam optimizer                    &\qquad &\qquad & &           & $\bullet$ Adam optimizer\\
       &            & $\bullet$ Batch size of 32 samples          &\qquad &\qquad & &           & $\bullet$ Batch size of 32 samples\\
\cmidrule{2-3}\cmidrule{7-8}
& Bidirectional & $\bullet$ Bidirectional LSTM layer with 32 units &\qquad & \qquad & & Bidirectional & $\bullet$ Bidirectional LSTM layer with 32 units \\ 
& RNN           & $\bullet$ Dense layer with 32 units              &\qquad & \qquad & & RNN           & $\bullet$ Dense layer with 32 units \\ 
&               & $\bullet$ Dropout 50\%                           &\qquad & \qquad & &               & $\bullet$ Dropout 50\%   \\ 
&               & $\bullet$ Dense layer with 4 units               &\qquad & \qquad & &               & $\bullet$ Dense layer with 4 units  \\ 
&               & $\bullet$ Dense layer with 1 unit                &\qquad & \qquad & &               & $\bullet$ Dense layer with 1 unit  \\ 
&               & $\bullet$ Adam optimizer                         &\qquad & \qquad & &               & $\bullet$ Adam optimizer \\ 
&               & $\bullet$ Batch size of 32 samples               &\qquad & \qquad & &               & $\bullet$ Batch size of 32 samples \\ 
\cmidrule{2-3}\cmidrule{7-8}
& GRU & $\bullet$ GRU with 32 units                                &\qquad & \qquad & & GRU & $\bullet$ GRU with 32 units  \\ 
&     & $\bullet$ Dense layer with 32 units                        &\qquad & \qquad & &     & $\bullet$ GRU with 32 units   \\ 
&     & $\bullet$ Dense layer with 1 unit                          &\qquad & \qquad & &     & $\bullet$ Dense layer with 32 units   \\ 
&     & $\bullet$ Adam Optimizer                                   &\qquad & \qquad & &     & $\bullet$ Dense layer with 1 unit  \\ 
&     & $\bullet$ Batch size of 32 samples                         &\qquad & \qquad & &     & $\bullet$ Adam Optimizer      \\ 
&     &                                                            &\qquad & \qquad & &     & $\bullet$ Batch size of 32 samples      \\ 
\cmidrule{2-3}\cmidrule{7-8}
& Hybrid & $\bullet$ 1D Convolution layer with with 8 filters, 6 as kernel size &\qquad & \qquad & & Hybrid & $\bullet$ 1D Convolution layer with with 8 filters, 6 as kernel size  \\ 
&        & $\bullet$ RelU activation function                                    &\qquad & \qquad & &        & $\bullet$ RelU activation function \\ 
&        & $\bullet$ LSTM layer with 64 units with Tanh activation function      &\qquad & \qquad & &        & $\bullet$ LSTM layer with 64 units with Tanh activation function\\ 
&        & $\bullet$ LSTM layer with 64 units with Tanh activation function      &\qquad & \qquad & &        & $\bullet$ LSTM layer with 64 units with Tanh activation function \\ 
&        & $\bullet$ Dense layer with 10 units with ReLU activation function     &\qquad & \qquad & &        & $\bullet$ Dense layer with 10 units with ReLU activation function \\ 
&        & $\bullet$ Dense layer with 1 unit                                     &\qquad & \qquad & &        & $\bullet$ Dense layer with 1 unit \\ 
&        & $\bullet$ Adam Optimizer                                              &\qquad & \qquad & &         & $\bullet$ Adam Optimizer \\ 
&        & $\bullet$ Batch size of 32 samples                                    &\qquad & \qquad & &        & $\bullet$ Batch size of 32 samples      \\ 
\cmidrule{2-3}\cmidrule{7-8}
& OPTM-LSTM & $\bullet$ OPTM-LSTM layer with 4 units  &\qquad &  \qquad &  & OPTM-LSTM & $\bullet$ OPTM-LSTM LSTM layer with 4 units \\
&        & $\bullet$ Dense layer with 4 units         &\qquad &  \qquad &        &     & $\bullet$ Dense layer with 4 units  \\ 
&        & $\bullet$ Dense layer with 1 unit          &\qquad &  \qquad &        &         & $\bullet$ Dense layer with 1 unit \\ 
&        & $\bullet$ Adam optimizer                   &\qquad &  \qquad &        &         & $\bullet$ Adam optimizer \\ 
&        & $\bullet$ Batch size of 1 sample           &\qquad &  \qquad &        &         & $\bullet$ Batch size of 1 sample \\ 
\cmidrule[2pt]{1-3}\cmidrule[2pt]{6-8}
\end{tabular}}
\medskip
\label{tab:DeepModelsFinnish}
\end{table*}

\begin{table*}[hbt!]
\centering
\captionsetup{width=.75\textwidth}
\caption{Google Benchmark Training (left) and Benchmark Testing (right). Data sample is 35,000 trading events.}
\scalebox{0.65}{
\begin{tabular}{ccrlcccrl}
\cmidrule[2pt]{1-4}\cmidrule[2pt]{6-9}
\textbf{Input} & \textbf{Normalization} & \textbf{Model} & \textbf{MSE - Train} & \qquad & \textbf{Input} & \textbf{Normalization} & \textbf{Model} & \textbf{MSE - Test} \\
\cmidrule{1-4}\cmidrule{6-9}
 LOB Data & Raw & \textbf{OPTM-LSTM} & \textbf{4.04241E+13} & \qquad & LOB Data &  Raw & \textbf{OPTM-LSTM}& \textbf{2.05396E+12}\\   
 &               & LSTM          & 6.73657E+13               & \qquad &       &        & LSTM         & 3.91887E+13            \\ 
 &               & Attention     & 6.72123E+13               & \qquad &       &        & Attention    & 3.91715E+13           \\   
 &               & Bidirectional & 6.73657E+13               & \qquad &       &        & Bidirectional& 3.91545E+13          \\   
 &               & GRU           & 6.73774E+13               & \qquad &       &        & GRU          & 3.91948E+13         \\   
 &               & Hybrid        & 2.18984E+15               & \qquad &       &        & Hybrid       & 2.33644E+14         \\
 &               & Baseline      & 5.76034E+15               & \qquad &       &        & Baseline     & 1.38235E+14 \\

\cmidrule{2-4}\cmidrule{7-9}
 & MinMax & \textbf{OPTM-LSTM} & \textbf{1.88756E-05}        & \qquad &       &  MinMax & \textbf{OPTM-LSTM}  & \textbf{1.31094E-05} \\   
 &             & LSTM           & 3.79995E-05                & \qquad &       &         & LSTM                &  3.99546E-05  \\ 
 &             & Attention      & 5.71432E-05                & \qquad &       &         & Attention           &  4.94253E-05 \\   
 &             & Bidirectional  & 3.72352E-05                & \qquad &       &         & Bidirectional       &  3.59546E-05\\   
 &             & GRU            & 4.15867E-05                & \qquad &       &         & GRU                 &  3.30894E-05 \\   
 &             & Hybrid         & 6.56392E-05                & \qquad &       &         & Hybrid              &  6.78675E-05\\
 &             & Baseline       & 6.81553E-05                & \qquad &       &         & Baseline            &  6.92356E-05 \\

\cmidrule{2-4}\cmidrule{7-9}
 & Zscore       & \textbf{OPTM-LSTM}& \textbf{1.61746E+02} & \qquad &         & Zscore & \textbf{OPTM-LSTM}  & \textbf{1.59013E+02} \\   
 &              & LSTM          & 1.63453E+02 & \qquad &            &         & LSTM         & 2.34154E+02  \\ 
 &              & Attention     & 1.64778E+02 & \qquad &            &         & Attention    & 6.94510E+02  \\ 
 &              & Bidirectional & 1.64456E+02 & \qquad &            &         & Bidirectional& 4.56984E+02 \\   
 &              & GRU           & 1.63756E+02 & \qquad &            &         & GRU          & 5.65567E+02 \\   
 &              & Hybrid        & 4.66898E+02 & \qquad &            &         & Hybrid       & 6.72002E+02 \\
 &              & Baseline      & 6.32211E+02 & \qquad &            &         & Baseline     & 6.40224E+02 \\
 \cmidrule{1-4}\cmidrule{6-9}
Mid-price & Raw & \textbf{OPTM-LSTM} & \textbf{4.31765E+13} & \qquad & Mid-price &  Raw & \textbf{OPTM-LSTM}& \textbf{4.36219E+13}\\   
 &               & LSTM          &  6.74887E+13             & \qquad &       &        & LSTM         &   7.22948E+13         \\ 
 &               & Attention     &  6.74443E+13             & \qquad &       &        & Attention    &   7.42857E+13         \\   
 &               & Bidirectional &  6.70889E+13             & \qquad &       &        & Bidirectional&   7.88225E+13        \\   
 &               & GRU           &  6.70366E+13             & \qquad &       &        & GRU          &   7.90564E+13       \\   
 &               & Hybrid        &  6.68133E+13             & \qquad &       &        & Hybrid       &   8.98765E+13           \\
 &               & Persistence   &  9.32445E+13             & \qquad &       &        & Persistence  &   9.92332E+13           \\

\cmidrule{2-4}\cmidrule{7-9}
 & MinMax & \textbf{OPTM-LSTM} & \textbf{5.72756E-03}        & \qquad &       &  MinMax & \textbf{OPTM-LSTM}  & \textbf{5.95884E-03} \\   
 &             & LSTM           & 2.50001E-02                & \qquad &       &         & LSTM                & 2.51443E-02 \\ 
 &             & Attention      & 5.05911E-02                & \qquad &       &         & Attention           & 5.21542E-02 \\   
 &             & Bidirectional  & 5.45328E-02                & \qquad &       &         & Bidirectional       & 6.12984E-02 \\   
 &             & GRU            & 3.42819E-02                & \qquad &       &         & GRU                 & 3.01012E-02 \\   
 &             & Hybrid         & 7.69334E-02                & \qquad &       &         & Hybrid              & 8.95335E-02\\
 &             & Persistence    & 5.75984E-01                & \qquad &       &         & Persistence         & 6.36985E-01 \\

\cmidrule{2-4}\cmidrule{7-9}
 & Zscore       & \textbf{OPTM-LSTM}& \textbf{1.19774E+02} & \qquad &         &  Zscore & \textbf{OPTM-LSTM}  & \textbf{1.20403E+02} \\   
 &              & LSTM          & 1.22889E+02 & \qquad &            &         & LSTM         & 1.25776E+02  \\ 
 &              & Attention     & 1.22980E+02 & \qquad &            &         & Attention    & 1.26503E+02 \\ 
 &              & Bidirectional & 1.24435E+02 & \qquad &            &         & Bidirectional& 1.28245E+02 \\   
 &              & GRU           & 1.22025E+02 & \qquad &            &         & GRU          & 1.27121E+02 \\   
 &              & Hybrid        & 1.29885E+02 & \qquad &            &         & Hybrid       & 2.35335E+02 \\
 &              & Persistence   & 2.90325E+02 & \qquad &            &         & Persistence  & 2.99894E+02 \\
\cmidrule[2pt]{1-4}\cmidrule[2pt]{6-9}
\end{tabular}}
\medskip
\label{tab:GoogleBenchmark}
\end{table*}

\begin{table*}[hbt!]
\centering
\captionsetup{width=.70\textwidth}
\caption{Amazon MSE scores under the Short experimental protocol.}
\scalebox{0.58}{
\begin{tabular}{rcrlcrcrlcrcrl}
\cmidrule[2pt]{1-6}\cmidrule[2pt]{6-9}\cmidrule[2pt]{9-14}
\textbf{Size} & \textbf{Stock} & \textbf{Model} & \textbf{MSE - Train} & \qquad & \textbf{Size} & \textbf{Stock} & \textbf{Model} & \textbf{MSE - Train} & \qquad & \textbf{Size} & \textbf{Stock} & \textbf{Model} & \textbf{MSE - Train}\\
\cmidrule{1-4}\cmidrule{6-9}\cmidrule{11-14}
Amazon & 1,000 & \textbf{OPTM-LSTM}& \textbf{2.37755E+13} & \qquad &  Amazon    &  2,000  & \textbf{OPTM-LSTM}       & \textbf{1.91119E+13} & \qquad & Amazon & 3,000 & \textbf{OPTM-LSTM}       & \textbf{2.11565E+13}\\   
 &             & LSTM          & 2.43245E+13 & \qquad &            &         & LSTM         & 2.23752E+13 & \qquad &       &        & LSTM          & 2.18732E+13\\ 
 &             & Attention     & 2.43227E+13 & \qquad &            &         & Attention    & 2.23726E+13 & \qquad &       &        & Attention     & 2.18664E+13\\   
 &             & Bidirectional & 2.43231E+13 & \qquad &            &         & Bidirectional& 2.08533E+13 & \qquad &       &        & Bidirectional & 2.18677E+13\\   
 &             & GRU           & 2.43227E+13 & \qquad &            &         & GRU          & 2.08469E+13 & \qquad &       &        & GRU           & 2.18628E+13\\   
 &             & Hybrid        & 2.94457E+13 & \qquad &            &         & Hybrid       & 2.49406E+13 & \qquad &       &        & Hybrid        & 3.04949E+13\\
\cmidrule{2-4}\cmidrule{7-9}\cmidrule{12-14}
 & 4,000 & \textbf{OPTM-LSTM}& \textbf{2.20394E+13} & \qquad &     &  5,000 & OPTM-LSTM        & 2.10221E+13 & \qquad &  & 6,000 & \textbf{OPTM-LSTM} & \textbf{2.08704E+13}\\   
 &             & LSTM          & 2.26918E+13& \qquad &             &          & LSTM         & 2.21002E+13 & \qquad &       &        & LSTM          & 2.19382E+13\\ 
 &             & Attention     & 2.26825E+13& \qquad &             &          & Attention    & 2.20843E+13 & \qquad &       &        & Attention     & 2.19207E+13\\   
 &             & Bidirectional & 2.26816E+13& \qquad &             &          & Bidirectional& 2.20871E+13 & \qquad &       &        & Bidirectional & 2.19159E+13\\   
 &             & GRU           & 2.26779E+13& \qquad &             &          & GRU          & 2.20792E+13 & \qquad &       &        & GRU           & 2.19059E+13\\   
 &             & Hybrid        & 3.09348E+13& \qquad &             &          & \textbf{Hybrid} & \textbf{2.02846E+13}& \qquad &    && Hybrid        & 2.30043E+13\\
\cmidrule{2-4}\cmidrule{7-9}\cmidrule{12-14}
 & 7,000  & \textbf{OPTM-LSTM}& \textbf{2.16789E+13} & \qquad &     &  10,000 & \textbf{OPTM-LSTM} & \textbf{2.13176E+13} & \qquad &  &  15,000  & OPTM-LSTM & 2.06974E+13\\   
 &              & LSTM          & 2.21532E+13 & \qquad &            &         & LSTM         & 2.30184E+13 & \qquad &        &        & LSTM          & 2.37194E+13\\ 
 &              & Attention     & 2.21188E+13 & \qquad &            &         & Attention    & 2.29426E+13 & \qquad &        &        & Attention     & 2.35976E+13\\   
 &              & Bidirectional & 2.21234E+13 & \qquad &            &         & Bidirectional& 2.29432E+13 & \qquad &        &        & Bidirectional & 2.36162E+13\\   
 &              & GRU           & 2.21123E+13 & \qquad &            &         & GRU          & 2.29194E+13 & \qquad &        &        & GRU           & 2.35377E+13\\   
 &              & Hybrid        & 2.30576E+13 & \qquad &            &         & Hybrid       & 3.08916E+13 & \qquad &        &        & \textbf{Hybrid} & \textbf{1.98247E+13}\\
\cmidrule{2-4}\cmidrule{7-9}\cmidrule{12-14}
 & 20,000       & OPTM-LSTM     & 1.93105E+13 & \qquad &            &  35,000 & OPTM-LSTM    & 1.61412E+13 & \qquad &        & 50,000 & \textbf{OPTM-LSTM} & \textbf{8.65632E+12}\\   
 &              & LSTM          & 2.52202E+13 & \qquad &            &         & LSTM         & 2.68694E+13 & \qquad &        &        & LSTM          & 2.76307E+13\\ 
 &              & Attention     & 2.49042E+13 & \qquad &            &         & Attention    & 2.56509E+13 & \qquad &        &        & Attention     & 2.40673E+13\\   
 &              & Bidirectional & 2.47769E+13 & \qquad &            &         & Bidirectional& 2.51290E+13  & \qquad &       &        & Bidirectional & 2.44334E+13\\   
 &              & GRU           & 2.47721E+13 & \qquad &            &         & GRU          & 2.51239E+13 & \qquad &        &        & GRU           & 2.39928E+13\\   
 &              & \textbf{Hybrid}& \textbf{1.89910E+13}  & \qquad &    & & \textbf{Hybrid}     & \textbf{1.58455E+13}  & \qquad & & & Hybrid        & 1.99902E+13\\
\cmidrule{2-4}\cmidrule{7-9}\cmidrule{12-14}
  & 100,000 & \textbf{OPTM-LSTM}& \textbf{4.42847E+12} & \qquad &    &  400,000  & \textbf{OPTM-LSTM} & \textbf{7.48735E+11} & \qquad &  & 800,000 & \textbf{OPTM-LSTM} & \textbf{7.09238E+11}\\   
 &                & LSTM          & 2.84573E+13& \qquad &            &         & LSTM         & 2.86594E+13 & \qquad &         &        & LSTM          & 2.84907E+13\\ 
 &                & Attention     & 1.72512E+13& \qquad &            &         & Attention    & 1.67204E+12 & \qquad &         &        & Attention     & 1.22332E+12\\   
 &                & Bidirectional & 1.79991E+13& \qquad &            &         & Bidirectional& 1.11909E+12 & \qquad &         &        & Bidirectional & 1.21469E+12\\   
 &                & GRU           & 1.79168E+13& \qquad &            &         & GRU          & 1.08493E+12 & \qquad &         &        & GRU           & 1.10944E+12\\   
 &                & Hybrid        & 1.36274E+13& \qquad &            &         & Hybrid       & 3.13572E+13 & \qquad &         &        & Hybrid        & 2.12974E+13\\
\cmidrule{2-4}\cmidrule{7-9}\cmidrule{12-14}
  & 1,000,000 & \textbf{OPTM-LSTM}       & \textbf{5.57388E+11} & \qquad &   &  10,000,000 & \textbf{OPTM-LSTM}  & \textbf{5.09864E+11} & \qquad &  & 20,000,000 & \textbf{OPTM-LSTM} & \textbf{4.70387E+11}\\   
 &                  & LSTM         & 2.85123E+13 & \qquad &             &         & LSTM    & 2.86813E+13     & \qquad &        &        & LSTM          & 2.87918E+13\\ 
 &                  & Attention    & 1.20907E+12 & \qquad &             &         & Attention& 1.10192E+12    & \qquad &        &        & Attention     & 1.03817E+12\\   
 &                  & Bidirectional& 1.23501E+12 & \qquad &             &         & Bidirectional& 1.24928E+12& \qquad &        &        & Bidirectional & 1.25935E+12\\   
 &                  & GRU          & 1.24944E+12 & \qquad &             &         & GRU          & 1.17564E+12& \qquad &        &        & GRU           & 1.19487E+12\\   
 &                  & Hybrid       & 3.56355E+13 & \qquad &             &         & Hybrid       & 2.94568E+13& \qquad &        &        & Hybrid        & 2.75774E+13\\
\cmidrule[2pt]{1-6}\cmidrule[2pt]{6-9}\cmidrule[2pt]{9-14}
\textbf{Size} & \textbf{Stock} & \textbf{Model} & \textbf{MSE - Test} & \qquad & \textbf{Stock} & \textbf{Size} & \textbf{Model} & \textbf{MSE - Test} & \qquad & \textbf{Stock} & \textbf{Size} & \textbf{Model} & \textbf{MSE - Test}\\
\cmidrule{1-4}\cmidrule{6-9}\cmidrule{11-14}
 Amazon & 1,000 & \textbf{OPTM-LSTM}        & \textbf{1.80046E+13} & \qquad &  Amazon     &  2,000 & \textbf{OPTM-LSTM}& \textbf{2.07877E+13} & \qquad & Amazon & 3,000 & \textbf{OPTM-LSTM} & \textbf{2.38613E+13}\\   
 &             & LSTM          & 2.04199E+13  & \qquad &            &         & LSTM         & 2.08535E+13            & \qquad &       &        & LSTM          & 2.52190E+13\\ 
 &             & Attention     & 2.04171E+13 & \qquad &            &         & Attention    & 2.08495E+13            & \qquad &       &        & Attention     & 2.52077E+13\\   
 &             & Bidirectional & 2.04172E+13 & \qquad &            &         & Bidirectional& 2.23727E+13            & \qquad &       &        & Bidirectional & 2.52091E+13\\   
 &             & GRU           & 2.43227E+13 & \qquad &            &         & GRU          & 2.23705E+13            & \qquad &       &        & GRU           & 2.52008E+13\\   
 &             & Hybrid& 1.97743E+13 & \qquad &  &         & Hybrid       & 2.99633E+13             & \qquad &       &        & Hybrid        & 2.99564E+13\\
\cmidrule{2-4}\cmidrule{7-9}\cmidrule{12-14}
 & 4,000 & \textbf{OPTM-LSTM}& \textbf{1.91248E+13} & \qquad &       &  5,000  & OPTM-LSTM & 1.96493E+13 & \qquad &  & 6,000 & \textbf{OPTM-LSTM} & \textbf{1.92065E+13}\\   
 &             & LSTM           & 1.97085E+13& \qquad &                       &         & LSTM         & 2.13228E+13 & \qquad &          &        & LSTM          & 2.35344E+13\\ 
 &             & Attention      & 1.96953E+13& \qquad &                       &         & Attention    & 2.12964E+13 & \qquad &          &        & Attention     & 2.35050E+13\\   
 &             & Bidirectional  & 1.96927E+13& \qquad &                       &         & Bidirectional& 2.13016E+13 & \qquad &          &        & Bidirectional & 2.35021E+13\\   
 &             & GRU            & 1.96876E+13& \qquad &                       &         & GRU          & 2.12849E+13 & \qquad &          &        & GRU           & 2.34807E+13\\   
 &             & Hybrid         & 2.57867E+13& \qquad &                       &         & \textbf{Hybrid} & \textbf{1.86085E+13}& \qquad &  &     & Hybrid        & 1.96575E+13\\
\cmidrule{2-4}\cmidrule{7-9}\cmidrule{12-14}
  & 7,000 & \textbf{OPTM-LSTM}& \textbf{2.44086E+13} & \qquad &  &  10,000  & \textbf{OPTM-LSTM}  & \textbf{2.34885E+13} & \qquad &  & 15,000 & OPTM-LSTM  & 2.22071E+13\\   
 &              & LSTM          & 2.54747E+13 & \qquad &            &         & LSTM         & 2.65046E+13 & \qquad &        &                               & LSTM          & 2.71849E+13\\ 
 &              & Attention     & 2.54255E+13 & \qquad &            &         & Attention    & 2.63515E+13 & \qquad &        &                               & Attention     & 2.69329E+13\\ 
 &              & Bidirectional & 2.54299E+13 & \qquad &            &         & Bidirectional& 2.63566E+13  & \qquad &        &                               & Bidirectional & 2.69567E+13\\   
 &              & GRU           & 2.54086E+13 & \qquad &            &         & GRU          & 2.63118E+13 & \qquad &        &                               & GRU           & 2.68626E+13\\   
 &              & Hybrid        & 2.83514E+13 & \qquad &            &         & Hybrid       & 3.44564E+13 & \qquad &        &                               & \textbf{Hybrid} & \textbf{2.15106E+13}\\
\cmidrule{2-4}\cmidrule{7-9}\cmidrule{12-14}
& 20,000  & \textbf{OPTM-LSTM}& \textbf{2.01197E+13} & \qquad &     &  35,000  & \textbf{OPTM-LSTM}  & \textbf{1.53563E+13} & \qquad &  & 50,000 & \textbf{OPTM-LSTM} & \textbf{1.28607E+13}\\   
 &              & LSTM          & 2.69286E+13 & \qquad &            &         & LSTM         & 2.94158E+13 & \qquad &        &        & LSTM          & 2.94260E+13\\ 
 &              & Attention     & 2.63352E+13 & \qquad &            &         & Attention    & 2.72802E+13 & \qquad &        &        & Attention     & 2.43106E+13\\   
 &              & Bidirectional & 2.61174E+13 & \qquad &            &         & Bidirectional& 2.68018E+13 & \qquad &        &        & Bidirectional & 2.45215E+13\\   
 &              & GRU           & 2.60337E+13 & \qquad &            &         & GRU          & 2.64865E+13 & \qquad &        &        & GRU           & 2.46317E+13\\   
 &              & Hybrid        & 2.11629E+13 & \qquad &            &         & Hybrid       & 2.79983E+13 & \qquad &        &        & Hybrid        & 1.46543E+13\\

\cmidrule{2-4}\cmidrule{7-9}\cmidrule{12-14}
  & 100,000 & \textbf{OPTM-LSTM}& \textbf{2.46977E+12} & \qquad &  &  400,000 & \textbf{OPTM-LSTM}  & \textbf{6.29513E+09} & \qquad &  & 800,000 & \textbf{OPTM-LSTM} & \textbf{2.50453E+07}\\   
 &                & LSTM          & 2.92036E+13& \qquad &           &         & LSTM         & 2.84972E+13 & \qquad &         &        & LSTM          & 2.87144E+13\\ 
 &                & Attention     & 1.23114E+13& \qquad &           &         & Attention    & 2.34635E+10 & \qquad &         &        & Attention     & 2.18847E+09\\   
 &                & Bidirectional & 1.54173E+13& \qquad &           &         & Bidirectional& 1.05435E+10 & \qquad &         &        & Bidirectional & 3.53215E+10\\   
 &                & GRU           & 1.41743E+13& \qquad &           &         & GRU          & 9.97630E+09 & \qquad &         &        & GRU           & 1.57717E+09\\   
 &                & Hybrid        & 7.21288E+12& \qquad &           &         & Hybrid       & 9.82100E+10 & \qquad &         &        & Hybrid        & 8.09596E+08\\

\cmidrule{2-4}\cmidrule{7-9}\cmidrule{12-14}
  & 1,000,000 & \textbf{OPTM-LSTM} & \textbf{1.90265E+07} & \qquad &  &  10,000,000 & \textbf{OPTM-LSTM} & \textbf{1.21373E+07} & \qquad &  & 20,000,000 & \textbf{OPTM-LSTM} & \textbf{1.13554E+07}\\   
 &                  & LSTM         & 2.89144E+13 & \qquad &            &         & LSTM            & 2.79154E+13       & \qquad &        &            & LSTM          & 2.76699E+13\\ 
 &                  & Attention    & 2.00338E+09  & \qquad &           &         & Attention       & 1.98734E+09       & \qquad &        &            & Attention     & 1.93428E+09\\   
 &                  & Bidirectional& 3.43883E+10 & \qquad &            &         & Bidirectional   & 3.13243E+10       & \qquad &        &            & Bidirectional & 3.38476E+10\\   
 &                  & GRU          & 1.64125E+09  & \qquad &           &         & GRU             & 1.77395E+09       & \qquad &        &            & GRU           & 1.96355E+09\\   
 &                  & Hybrid       & 1.46733E+09  & \qquad &           &         & Hybrid          & 1.54599E+09       & \qquad &        &            & Hybrid        & 9.93424E+08\\
\cmidrule[2pt]{1-6}\cmidrule[2pt]{6-9}\cmidrule[2pt]{9-14}
\end{tabular}}
\medskip
\label{tab:AmazonShort}
\end{table*}

\begin{table*}[hbt!]
\centering
\captionsetup{width=.70\textwidth}
\caption{Amazon MSE scores under the Long experimental protocol.}
\scalebox{0.58}{
\begin{tabular}{rcrlcrcrlcrcrl}
\cmidrule[2pt]{1-6}\cmidrule[2pt]{6-9}\cmidrule[2pt]{9-14}
\textbf{Stock} & \textbf{Size} & \textbf{Model} & \textbf{MSE - Train} & \qquad & \textbf{Stock} & \textbf{Size} & \textbf{Model} & \textbf{MSE - Train} & \qquad & \textbf{Stock} & \textbf{Size} & \textbf{Model} & \textbf{MSE - Train}\\
\cmidrule{1-4}\cmidrule{6-9}\cmidrule{11-14}
 Amazon & 1,000 & \textbf{OPTM-LSTM} & \textbf{2.28218E+13} & \qquad & Amazon &  2,000 & \textbf{OPTM-LSTM}& \textbf{2.14712E+13}  & \qquad & Amazon  & 3,000 & \textbf{OPTM-LSTM}& \textbf{2.03532E+13}\\   
 &             & LSTM          & 2.33984E+13           & \qquad &       &        & LSTM         & 2.23702E+13           & \qquad &       &        & LSTM          & 2.18672E+13\\ 
 &             & Attention     & 2.33777E+13           & \qquad &       &        & Attention    & 2.17337E+13           & \qquad &       &        & Attention     & 2.13392E+13\\   
 &             & Bidirectional & 2.33644E+13           & \qquad &       &        & Bidirectional& 2.17895E+13           & \qquad &       &        & Bidirectional & 2.12358E+13\\   
 &             & GRU           & 2.30463E+13           & \qquad &       &        & GRU          & 2.16431E+13            & \qquad &       &        & GRU           & 2.12181E+13\\   
 &             & Hybrid        & 6.37836E+13           & \qquad &       &        & Hybrid       & 2.98727E+13           & \qquad &       &        & Hybrid        & 2.12181E+13\\

\cmidrule{2-4}\cmidrule{7-9}\cmidrule{12-14}
 & 4,000 & \textbf{OPTM-LSTM} & \textbf{2.15148E+13} & \qquad &       &  5,000   & \textbf{OPTM-LSTM}  & \textbf{2.01779E+13} & \qquad &  & 6,000 & \textbf{OPTM-LSTM} & \textbf{1.43550E+13}\\   
 &             & LSTM           & 2.26837E+13& \qquad &                       &         & LSTM         & 2.20931E+13 & \qquad &          &        & LSTM          & 2.19235E+13\\ 
 &             & Attention      & 2.17953E+13& \qquad &                       &         & Attention    & 2.11484E+13 & \qquad &          &        & Attention     & 2.05555E+13\\   
 &             & Bidirectional  & 2.17825E+13& \qquad &                        &         & Bidirectional& 2.08829E+13 & \qquad &          &        & Bidirectional & 2.06374E+13\\   
 &             & GRU            & 2.16035E+13& \qquad &                       &         & GRU          & 2.08893E+13 & \qquad &          &        & GRU           & 2.06077E+13\\   
 &             & Hybrid         & 2.56088E+13 & \qquad &                      &         & Hybrid       & 2.41016E+13 & \qquad &          &        & Hybrid        & 2.45141E+13\\

\cmidrule{2-4}\cmidrule{7-9}\cmidrule{12-14}
  & 7,000 & \textbf{OPTM-LSTM}& \textbf{1.92793E+13} & \qquad &     &  10,000  & \textbf{OPTM-LSTM}  & \textbf{1.87029E+13} & \qquad &  & 15,000 & \textbf{OPTM-LSTM} & \textbf{1.00104E+13}\\   
 &              & LSTM          & 2.21432E+13 & \qquad &            &         & LSTM         & 2.29951E+13 & \qquad &        &                           & LSTM   & 2.36865E+13\\ 
 &              & Attention     & 2.09194E+13 & \qquad &            &         & Attention    & 2.29951E+13 & \qquad &        &                           & Attention & 2.08998E+13\\ 
 &              & Bidirectional & 2.08263E+13 & \qquad &            &         & Bidirectional& 2.11482E+13 & \qquad &        &                           & Bidirectional & 2.07893E+13\\   
 &              & GRU           & 2.12098E+13 & \qquad &            &         & GRU          & 2.10835E+13 & \qquad &        &                           & GRU           & 2.07266E+13\\   
 &              & Hybrid        & 2.04333E+13  & \qquad &            &         & Hybrid       & 3.49398E+13 & \qquad &        &                           & Hybrid & 1.94942E+13\\
\cmidrule[2pt]{1-6}\cmidrule[2pt]{6-9}\cmidrule[2pt]{9-14}
\textbf{Stock} & \textbf{Size} & \textbf{Model} & \textbf{MSE - Test} & \qquad & \textbf{Stock} & \textbf{Size} & \textbf{Model} & \textbf{MSE - Test} & \qquad & \textbf{Stock} & \textbf{Size} & \textbf{Model} & \textbf{MSE - Test}\\
\cmidrule{1-4}\cmidrule{6-9}\cmidrule{11-14}
 Amazon & 1,000 & \textbf{OPTM-LSTM} & \textbf{1.94055E+13} & \qquad & Amazon &  2,000 & \textbf{OPTM-LSTM}& \textbf{2.00711E+13}  & \qquad & Amazon & 3,000 & \textbf{OPTM-LSTM}& \textbf{2.11361E+13}\\   
 &             & LSTM          & 2.04158E+13           & \qquad &       &        & LSTM         & 2.00849E+13            & \qquad &       &        & LSTM          & 2.52133E+13\\ 
 &             & Attention     & 1.96952E+13           & \qquad &       &        & Attention    & 2.03613E+13            & \qquad &       &        & Attention     & 2.46697E+13\\   
 &             & Bidirectional & 1.97491E+13           & \qquad &       &        & Bidirectional& 2.04502E+13            & \qquad &       &        & Bidirectional & 2.45197E+13\\   
 &             & GRU           & 1.98757E+13           & \qquad &       &        & GRU          & 2.02619E+13            & \qquad &       &        & GRU           & 2.43652E+13\\   
 &             & Hybrid        & 4.92951E+13           & \qquad &       &        & Hybrid       & 3.05593E+13            & \qquad &       &        & Hybrid        & 5.91284E+13\\

\cmidrule{2-4}\cmidrule{7-9}\cmidrule{12-14}
 & 4,000 & \textbf{OPTM-LSTM} & \textbf{1.55827E+13} & \qquad &        &  5,000  & \textbf{OPTM-LSTM}  & \textbf{2.00073E+13} & \qquad &  & 6,000 & \textbf{OPTM-LSTM} & \textbf{7.56541E+12}\\   
 &             & LSTM           & 1.97012E+13& \qquad &                       &         & LSTM         & 2.13159E+13 & \qquad &          &        & LSTM          & 2.35177E+13\\ 
 &             & Attention      & 1.89802E+13& \qquad &                       &         & Attention    & 2.02218E+13 & \qquad &          &        & Attention     & 2.16751E+13\\   
 &             & Bidirectional  & 1.88544E+13& \qquad &                       &         & Bidirectional& 2.02142E+13 & \qquad &          &        & Bidirectional & 2.21063E+13\\   
 &             & GRU            & 1.86681E+13& \qquad &                       &         & GRU          & 2.01437E+13 & \qquad &          &        & GRU           & 2.21322E+13\\   
 &             & Hybrid         & 2.17871E+13 & \qquad &                      &         & Hybrid       & 2.33849E+13 & \qquad &          &        & Hybrid        & 2.28543E+13\\
\cmidrule{2-4}\cmidrule{7-9}\cmidrule{12-14}
  & 7,000 & \textbf{OPTM-LSTM}& \textbf{2.19539E+13} & \qquad &     &  10,000  & \textbf{OPTM-LSTM}  & \textbf{2.10338E+13} & \qquad & & 15,000  & \textbf{OPTM-LSTM} & \textbf{5.28577E+12}\\   
 &              & LSTM          & 2.54651E+13 & \qquad &            &         & LSTM         & 2.64762E+13 & \qquad &        &                           & LSTM   & 2.71402E+13\\ 
 &              & Attention     & 2.42306E+13 & \qquad &            &         & Attention    & 2.37283E+13 & \qquad &        &                           & Attention & 2.24987E+13\\ 
 &              & Bidirectional & 2.42153E+13 & \qquad &            &         & Bidirectional& 2.42298E+13 & \qquad &        &                           & Bidirectional & 2.33562E+13\\   
 &              & GRU           & 2.43058E+13 & \qquad &            &         & GRU          & 2.39311E+13 & \qquad &        &                           & GRU           & 2.27396E+13\\   
 &              & Hybrid        & 2.24018E+13 & \qquad &            &         & Hybrid       & 4.06392E+13 & \qquad &        &                           & Hybrid & 2.14205E+13\\
\cmidrule[2pt]{1-6}\cmidrule[2pt]{6-9}\cmidrule[2pt]{9-14}
\end{tabular}}
\medskip
\label{tab:AmazonLong}
\end{table*}

\begin{table*}[hbt!]
\centering
\captionsetup{width=.70\textwidth}
\caption{Amazon Benchmark Training (left) and Benchmark Testing (right). Data sample is 100,000 trading events.}
\scalebox{0.58}{
\begin{tabular}{ccrlcccrl}
\cmidrule[2pt]{1-4}\cmidrule[2pt]{6-9}
\textbf{Input} & \textbf{Normalization} & \textbf{Model} & \textbf{MSE - Train} & \qquad & \textbf{Input} & \textbf{Normalization} & \textbf{Model} & \textbf{MSE - Test} \\
\cmidrule{1-4}\cmidrule{6-9}
 LOB Data & Raw & \textbf{OPTM-LSTM}& \textbf{4.42854E+12} & \qquad &  LOB Data & Raw     & \textbf{OPTM-LSTM}  & \textbf{2.46979E+12}\\   
 &              & LSTM          & 2.84573E+13             & \qquad &            &         & LSTM         & 2.92032E+13\\ 
 &              & Attention     & 1.72512E+13             & \qquad &            &         & Attention    & 1.23114E+13\\ 
 &              & Bidirectional & 1.79991E+13             & \qquad &            &         & Bidirectional& 1.54175E+13\\   
 &              & GRU           & 1.79168E+13             & \qquad &            &         & GRU          & 1.41743E+13\\   
 &              & Hybrid        & 1.36274E+13             & \qquad &            &         & Hybrid       & 7.21285E+12\\
 &              & Baseline      & 5.76229E+15             & \qquad &            &         & Baseline     & 5.98276E+15 \\
 \cmidrule{2-4}\cmidrule{7-9}
 & MinMax & \textbf{OPTM-LSTM} & \textbf{2.22887E-05} & \qquad &  &  MinMax & \textbf{OPTM-LSTM}& \textbf{2.02887E-05}\\   
 &               & LSTM          & 5.12674E-05               & \qquad &       &        & LSTM         &  4.69676E-05\\ 
 &               & Attention     & 4.75122E-05               & \qquad &       &        & Attention    &  4.20884E-05\\   
 &               & Bidirectional & 5.71983E-05               & \qquad &       &        & Bidirectional&  4.80998E-05\\   
 &               & GRU           & 4.13893E-05               & \qquad &       &        & GRU          &  4.08778E-05\\   
 &               & Hybrid        & 8.92443E-05               & \qquad &       &        & Hybrid       &  9.57998E-05\\
 &               & Baseline      & 9.99776E-05               & \qquad &       &        & Baseline     &  1.00998E-04\\

\cmidrule{2-4}\cmidrule{7-9}
 & Zscore & \textbf{OPTM-LSTM} & \textbf{6.85887E-02}       & \qquad &       &  Zscore & \textbf{OPTM-LSTM}  & \textbf{5.20034E-02}\\   
 &             & LSTM           &   8.57663E-02             & \qquad &       &         & LSTM                & 8.00990E-02\\ 
 &             & Attention      &   8.48443E-02             & \qquad &       &         & Attention           & 8.17477E-02\\   
 &             & Bidirectional  &   8.95367E-02             & \qquad &       &         & Bidirectional       & 8.03909E-02\\   
 &             & GRU            &   8.22114E-02             & \qquad &       &         & GRU                 & 7.99188E-02\\   
 &             & Hybrid         &   4.75702E-01             & \qquad &       &         & Hybrid              & 6.06009E-01\\
 &             & Baseline       &   9.86855E-01             & \qquad &       &         & Baseline            & 9.85114E-01\\
 \cmidrule{1-4}\cmidrule{6-9}
Mid-price & Raw & \textbf{OPTM-LSTM} & \textbf{5.94990E+12} & \qquad & Mid-price &  Raw & \textbf{OPTM-LSTM}& \textbf{6.20004E+12}\\   
 &               & LSTM          &  2.76338E+13             & \qquad &       &        & LSTM         &  2.94290E+13 \\ 
 &               & Attention     &  2.76912E+13             & \qquad &       &        & Attention    &  2.94904E+13  \\   
 &               & Bidirectional &  2.73117E+13             & \qquad &       &        & Bidirectional&  2.90581E+13   \\   
 &               & GRU           &  2.73403E+13             & \qquad &       &        & GRU          &  2.90554E+13  \\   
 &               & Hybrid        &  7.57040E+13             & \qquad &       &        & Hybrid       &  7.57452E+13   \\
 &               & Persistence   &  8.00184E+14             & \qquad &       &        & Persistence  &  8.01993E+14    \\

\cmidrule{2-4}\cmidrule{7-9}
 & MinMax & \textbf{OPTM-LSTM} & \textbf{1.04280E-06}      & \qquad &       &  MinMax & \textbf{OPTM-LSTM}  & \textbf{1.02934E-06} \\   
 &             & LSTM           & 1.59140E-06              & \qquad &       &         & LSTM                & 1.57829E-06 \\ 
 &             & Attention      & 1.77829E-06              & \qquad &       &         & Attention           & 2.41392E-06 \\   
 &             & Bidirectional  & 3.58820E-06              & \qquad &       &         & Bidirectional       & 3.58151E-06 \\   
 &             & GRU            & 3.64600E-06              & \qquad &       &         & GRU                 & 4.52590E-06 \\   
 &             & Hybrid         & 4.51500E-06              & \qquad &       &         & Hybrid              & 5.00910E-06\\
 &             & Persistence    & 2.05157E-06              & \qquad &       &         & Persistence         & 2.36094E-06 \\

\cmidrule{2-4}\cmidrule{7-9}
 & Zscore       & \textbf{OPTM-LSTM}& \textbf{7.79740E-03} & \qquad &         &  Zscore & \textbf{OPTM-LSTM}  & \textbf{7.94824E-03} \\   
 &              & LSTM          & 6.90000E-02 & \qquad &            &         & LSTM          & 7.50000E-02  \\ 
 &              & Attention     & 6.94000E-02 & \qquad &            &         & Attention     & 6.98876E-02\\ 
 &              & Bidirectional & 7.00000E-02 & \qquad &            &         & Bidirectional & 7.94008E-02 \\   
 &              & GRU           & 7.13324E-02 & \qquad &            &         & GRU           & 7.66547E-02\\   
 &              & Hybrid        & 1.83000E-01 & \qquad &            &         & Hybrid        & 1.63776E-01\\
 &              & Persistence   & 7.05157E-02 & \qquad &            &         & Persistence   & 7.96094E-02\\
\cmidrule[2pt]{1-4}\cmidrule[2pt]{6-9}
\end{tabular}}
\medskip
\label{tab:AmazonBenchmark}
\end{table*}

\begin{table*}[hbt!]
\centering
\captionsetup{width=.70\textwidth}
\caption{Kesko MSE scores under the Short experimental protocol.}
\scalebox{0.60}{
\begin{tabular}{rcrlcrcrlcrcrl}
\cmidrule[2pt]{1-6}\cmidrule[2pt]{6-9}\cmidrule[2pt]{9-14}
\textbf{Stock} & \textbf{Size} & \textbf{Model} & \textbf{MSE - Train} & \qquad & \textbf{Stock} & \textbf{Size} & \textbf{Model} & \textbf{MSE - Train} & \qquad & \textbf{Stock} & \textbf{Size} & \textbf{Model} & \textbf{MSE - Train}\\
\cmidrule{1-4}\cmidrule{6-9}\cmidrule{11-14}
 Kesko & 1,000 & \textbf{OPTM-LSTM}& \textbf{5.67480E+10} & \qquad &  Kesko    &  2,000  & \textbf{OPTM-LSTM}    & \textbf{5.09428E+10} & \qquad & Kesko & 3,000 & \textbf{OPTM-LSTM}       & \textbf{4.35003E+10}\\   
 &             & LSTM          & 6.88025E+10 & \qquad &            &         & LSTM                         & 6.84285E+10     & \qquad &       &        & LSTM          & 6.82272E+10\\ 
 &             & Attention     & 6.88067E+10 & \qquad &            &         & Attention                    & 6.84241E+10      & \qquad &       &        & Attention     & 6.82374E+10\\   
 &             & Bidirectional & 6.88056E+10 & \qquad &            &         & Bidirectional                & 6.84235E+10     & \qquad &       &        & Bidirectional & 6.82452E+10\\   
 &             & GRU           & 6.88063E+10 & \qquad &            &         & GRU                          & 6.84220E+10     & \qquad &       &        & GRU           & 6.82457E+10\\   
 &             & Hybrid        & 1.45781E+11 & \qquad &            &         & Hybrid                       & 1.31784E+11      & \qquad &       &        & Hybrid        & 1.59131E+11\\
\cmidrule{2-4}\cmidrule{7-9}\cmidrule{12-14}
 & 4,000 & \textbf{OPTM-LSTM}& \textbf{3.88949E+10} & \qquad &  &  5,000 & \textbf{OPTM-LSTM} & \textbf{5.93965E+10} & \qquad &  & 6,000 & \textbf{OPTM-LSTM} & \textbf{1.87280E+10}\\   
 &             & LSTM          & 6.81911E+10& \qquad &             &         & LSTM                    & 6.82883E+10& \qquad &       &        & LSTM          & 6.83506E+10\\ 
 &             & Attention     & 6.82022E+10& \qquad &             &         & Attention               & 6.82883E+10& \qquad &       &        & Attention     & 6.83441E+10\\   
 &             & Bidirectional & 6.81910E+10& \qquad &             &         & Bidirectional           & 6.82882E+10& \qquad &       &        & Bidirectional & 6.83413E+10\\   
 &             & GRU           & 6.82025E+10& \qquad &             &         & GRU                     & 6.82887E+10& \qquad &       &        & GRU           & 6.83589E+10\\   
 &             & Hybrid        & 8.82731E+10& \qquad &             &         & Hybrid                  & 8.06607E+10&\qquad &       &        & Hybrid        & 8.02223E+10\\
\cmidrule{2-4}\cmidrule{7-9}\cmidrule{12-14}
  & 7,000 & \textbf{OPTM-LSTM}& \textbf{7.01066E+09} & \qquad &      &  10,000 & \textbf{OPTM-LSTM}  & \textbf{3.36665E+08} & \qquad &  & 15,000 & \textbf{OPTM-LSTM} & \textbf{1.23322E+05}\\   
 &              & LSTM          & 6.83350E+10 & \qquad &            &         & LSTM         & 6.84798E+10 & \qquad &        &        & LSTM          & 6.84651E+10\\ 
 &              & Attention     & 6.83902E+10 & \qquad &            &         & Attention    & 6.85087E+10 & \qquad &        &        & Attention     & 6.84350E+10\\   
 &              & Bidirectional & 6.83704E+10 & \qquad &            &         & Bidirectional& 6.85370E+10 & \qquad &        &        & Bidirectional & 6.84671E+10\\   
 &              & GRU           & 6.83900E+10 & \qquad &            &         & GRU          & 6.85051E+10 & \qquad &        &        & GRU           & 6.84191E+10\\   
 &              & Hybrid        & 1.33378E+11 & \qquad &             &         & Hybrid      & 2.38552E+11  & \qquad &        &        & Hybrid & 2.94083E+11\\
\cmidrule{2-4}\cmidrule{7-9}\cmidrule{12-14}
 & 20,000 & \textbf{OPTM-LSTM} & \textbf{3.53800E+03} & \qquad &      &  35,000 & \textbf{OPTM-LSTM} & \textbf{3.08900E+03} & \qquad &  & 50,000 & \textbf{OPTM-LSTM} & \textbf{1.28150E+04}\\   
 &              & LSTM          & 6.83821E+10 & \qquad &            &         & LSTM         & 6.92214E+10 & \qquad &        &        & LSTM          & 7.00089E+10\\ 
 &              & Attention     & 6.83587E+10 & \qquad &             &         & Attention    & 6.90948E+10 & \qquad &        &        & Attention     & 7.00074E+10\\   
 &              & Bidirectional & 6.83818E+10 & \qquad &             &         & Bidirectional& 6.92188E+10 & \qquad &        &        & Bidirectional & 6.97331E+10\\   
 &              & GRU           & 6.84508E+10 & \qquad &            &         & GRU          & 6.90434E+10 & \qquad &        &        & GRU           & 7.00073E+10\\   
 &              & Hybrid        & 2.22008E+11  & \qquad &           &         & Hybrid       & 2.01700E+11 & \qquad & &               & Hybrid        & 1.62985E+11\\
\cmidrule{2-4}\cmidrule{7-9}\cmidrule{12-14}
 & 100,000 & \textbf{OPTM-LSTM}& \textbf{3.00368E+05} & \qquad &     &  400,000 & \textbf{OPTM-LSTM} & \textbf{3.01665E+05} & \qquad &  & 800,000 & \textbf{OPTM-LSTM} & \textbf{3.09889E+05}\\   
 &                & LSTM          & 7.11851E+10& \qquad &            &         & LSTM         & 7.14898E+10 & \qquad &         &        & LSTM          & 7.16874E+10\\ 
 &                & Attention     & 7.01533E+10& \qquad &            &         & Attention    & 7.01758E+10 & \qquad &         &        & Attention     & 7.10329E+10\\   
 &                & Bidirectional & 7.11849E+10  & \qquad &            &         & Bidirectional  & 7.19002E+10 & \qquad &        &        & Bidirectional & 7.21223E+10\\   
 &                & GRU           & 7.11846E+10& \qquad &            &         & GRU          & 7.12240E+10 & \qquad &         &        & GRU           & 7.17981E+10\\   
 &                & Hybrid        & 1.55687E+11& \qquad &            &         & Hybrid       & 2.08874E+11 & \qquad &         &        & Hybrid        & 1.59847E+11\\
\cmidrule{2-4}\cmidrule{7-9}\cmidrule{12-14}
 & 1,000,000 & \textbf{OPTM-LSTM}       & \textbf{3.10878E+05} & \qquad &  &  2,000,000  & \textbf{OPTM-LSTM}  & \textbf{3.12093E+05} & \qquad &        &        &  & \\   
 &                  & LSTM         & 7.20847E+10               & \qquad &             &         & LSTM    & 7.29091E+10               & \qquad &        &        &  & \\ 
 &                  & Attention    & 7.11739E+10               & \qquad &             &         & Attention& 7.19023E+10              & \qquad &        &        &  & \\   
 &                  & Bidirectional& 7.20324E+10               & \qquad &             &         & Bidirectional& 7.25881E+10          & \qquad &        &        &  & \\   
 &                  & GRU          & 7.19423E+10               & \qquad &             &         & GRU          & 7.18201E+10          & \qquad &        &        &  & \\   
 &                  & Hybrid       & 2.88737E+11               & \qquad &             &         & Hybrid       & 2.94437E+11          & \qquad &        &        &  & \\
 \cmidrule[2pt]{1-6}\cmidrule[2pt]{6-9}\cmidrule[2pt]{9-14}
\textbf{Stock} & \textbf{Size} & \textbf{Model} & \textbf{MSE - Test} & \qquad & \textbf{Stock} & \textbf{Size} & \textbf{Model} & \textbf{MSE - Test} & \qquad & \textbf{Stock} & \textbf{Size} & \textbf{Model} & \textbf{MSE - Test}\\
\cmidrule{1-4}\cmidrule{6-9}\cmidrule{11-14}
Kesko & 1,000 & \textbf{OPTM-LSTM} & \textbf{5.13944E+10} & \qquad & Kesko &  2,000 & \textbf{OPTM-LSTM}& \textbf{4.30635E+10}  & \qquad & Kesko & 3,000 & \textbf{OPTM-LSTM}& \textbf{3.12937E+10}\\   
 &             & LSTM          & 6.80401E+10           & \qquad &       &        & LSTM         & 6.78792E+10            & \qquad &       &        & LSTM          & 6.80535E+10\\ 
 &             & Attention     & 6.80458E+10           & \qquad &       &        & Attention    & 6.78726E+10            & \qquad &       &        & Attention     & 6.80665E+10\\   
 &             & Bidirectional & 6.80445E+10           & \qquad &       &        & Bidirectional& 6.78721E+10            & \qquad &       &        & Bidirectional & 6.80767E+10\\   
 &             & GRU           & 6.80452E+10           & \qquad &       &        & GRU          & 6.78698E+10            & \qquad &       &        & GRU           & 6.80774E+10\\   
 &             & Hybrid        & 1.43664E+11           & \qquad &       &        & Hybrid       & 1.33300E+11            & \qquad &       &        & Hybrid        & 1.58883E+11\\
\cmidrule{2-4}\cmidrule{7-9}\cmidrule{12-14}
 & 4,000 & \textbf{OPTM-LSTM} & \textbf{2.37974E+10} & \qquad &        &  5,000  & \textbf{OPTM-LSTM}  & \textbf{5.06404E+10} & \qquad &  & 6,000 & \textbf{OPTM-LSTM} & \textbf{2.20000E+09}\\   
 &             & LSTM           & 6.86076E+10& \qquad &                       &         & LSTM         & 6.87101E+10 & \qquad &          &        & LSTM          & 6.85633E+10\\ 
 &             & Attention      & 6.86343E+10& \qquad &                       &         & Attention    & 6.87102E+10 & \qquad &          &        & Attention     & 6.85558E+10\\   
 &             & Bidirectional  & 6.86187E+10& \qquad &                       &         & Bidirectional& 6.87102E+10 & \qquad &          &        & Bidirectional & 6.85517E+10\\   
 &             & GRU            & 6.86339E+10& \qquad &                       &         & GRU          & 6.87101E+10 & \qquad &          &        & GRU           & 6.85764E+10\\   
 &             & Hybrid         & 8.81820E+10 & \qquad &                       &         & Hybrid      & 7.00176E+10 & \qquad &          &        & Hybrid        & 8.01716E+10\\
\cmidrule{2-4}\cmidrule{7-9}\cmidrule{12-14}
 & 7,000 & \textbf{OPTM-LSTM}& \textbf{3.78775E+05} & \qquad &      &  10,000 & \textbf{OPTM-LSTM}  & \textbf{1.00700E+03} & \qquad &  & 15,000 & \textbf{OPTM-LSTM} & \textbf{1.37600E+03}\\   
 &              & LSTM          & 6.90179E+10 & \qquad &            &         & LSTM         & 6.82803E+10 & \qquad &        &                           & LSTM          & 6.82137E+10\\ 
 &              & Attention     & 6.90947E+10 & \qquad &            &         & Attention    & 6.83169E+10 & \qquad &        &                           & Attention     & 6.81673E+10\\ 
 &              & Bidirectional & 6.90667E+10 & \qquad &            &         & Bidirectional& 6.83577E+10 & \qquad &        &                           & Bidirectional & 6.82134E+10\\   
 &              & GRU           & 6.90950E+10 & \qquad &            &         & GRU          & 6.83124E+10 & \qquad &        &                           & GRU           & 6.81444E+10\\   
 &              & Hybrid        & 1.35000E+11    & \qquad &            &         & Hybrid       & 4.36224E+11 & \qquad &        &                           & Hybrid        & 4.94433E+11\\
\cmidrule{2-4}\cmidrule{7-9}\cmidrule{12-14}
 & 20,000 & \textbf{OPTM-LSTM}& \textbf{3.93000E+02} & \qquad &     &  35,000 & \textbf{OPTM-LSTM}  & \textbf{4.44000E+02} & \qquad &  & 50,000 & \textbf{OPTM-LSTM} & \textbf{9.99000E+02}\\   
 &              & LSTM          & 6.88736E+10 & \qquad &            &         & LSTM         & 7.25105E+10 & \qquad &        &        & LSTM          & 7.20523E+10\\ 
 &              & Attention     & 6.88337E+10 & \qquad &            &         & Attention    & 7.23182E+10 & \qquad &        &        & Attention     & 7.20518E+10\\   
 &              & Bidirectional & 6.88669E+10 & \qquad &            &         & Bidirectional& 7.25108E+10 & \qquad &        &        & Bidirectional & 7.16165E+10\\   
 &              & GRU           & 6.89682E+10 & \qquad &            &         & GRU          & 7.22375E+10 & \qquad &        &        & GRU           & 7.20519E+10\\   
 &              & Hybrid        & 5.98547E+11 & \qquad &            &         & Hybrid       & 2.10906E+11 & \qquad &        &        & Hybrid        & 1.67994E+11\\
\cmidrule{2-4}\cmidrule{7-9}\cmidrule{12-14}
 & 100,000 & \textbf{OPTM-LSTM}& \textbf{1.93400E+03} & \qquad &  &  400,000 & \textbf{OPTM-LSTM}  & \textbf{1.35000E+03} & \qquad &  & 800,000 & \textbf{OPTM-LSTM} & \textbf{1.26600E+03}\\   
 &                & LSTM          & 7.57863E+10& \qquad &           &         & LSTM         & 7.49856E+10  & \qquad &         &        & LSTM          & 7.53487E+10\\ 
 &                & Attention     & 7.39432E+10& \qquad &           &         & Attention    & 7.39998E+10 & \qquad &         &        & Attention     & 7.45662E+10\\   
 &                & Bidirectional & 7.57872E+10& \qquad &           &         & Bidirectional& 7.23200E+10 & \qquad &         &        & Bidirectional & 7.05777E+10\\   
 &                & GRU           & 7.57864E+10& \qquad &           &         & GRU          & 7.38821E+10 & \qquad &         &        & GRU           & 7.16384E+10\\   
 &                & Hybrid        & 1.65579E+11 & \qquad &           &         & Hybrid      & 1.90002E+11 & \qquad &         &        & Hybrid        & 2.88704E+11\\
\cmidrule{2-4}\cmidrule{7-9}\cmidrule{12-14}
 & 1,000,000 & \textbf{OPTM-LSTM} & \textbf{1.00900E+03} & \qquad &  &  2,000,000 & \textbf{OPTM-LSTM} & \textbf{1.00100E+03}  & \qquad &  &  &  & \\   
 &                  & LSTM         & 7.38745E+10 & \qquad &            &         & LSTM            & 7.09273E+10       & \qquad &        &            &     & \\ 
 &                  & Attention    & 7.49993E+10 & \qquad &            &         & Attention       & 7.10024E+10       & \qquad &        &            &     & \\   
 &                  & Bidirectional& 6.99432E+10 & \qquad &            &         & Bidirectional   & 7.02131E+10       & \qquad &        &            &     & \\   
 &                  & GRU          & 7.01342E+10  & \qquad &            &        & GRU             & 7.00800E+10       & \qquad &        &            &     & \\   
 &                  & Hybrid       & 1.87748E+11    & \qquad &            &      & Hybrid          & 2.44905E+11       & \qquad &        &            &     & \\
\cmidrule[2pt]{1-6}\cmidrule[2pt]{6-9}\cmidrule[2pt]{9-14}
\end{tabular}}
\medskip
\label{tab:KeskoShort}
\end{table*}

\begin{figure*}[hbt!]
    \centering
  \subfloat[OPTM-LSTM training MSE scores \label{1a}]{%
       \scalebox{0.53}{\includegraphics[width=0.65\linewidth]{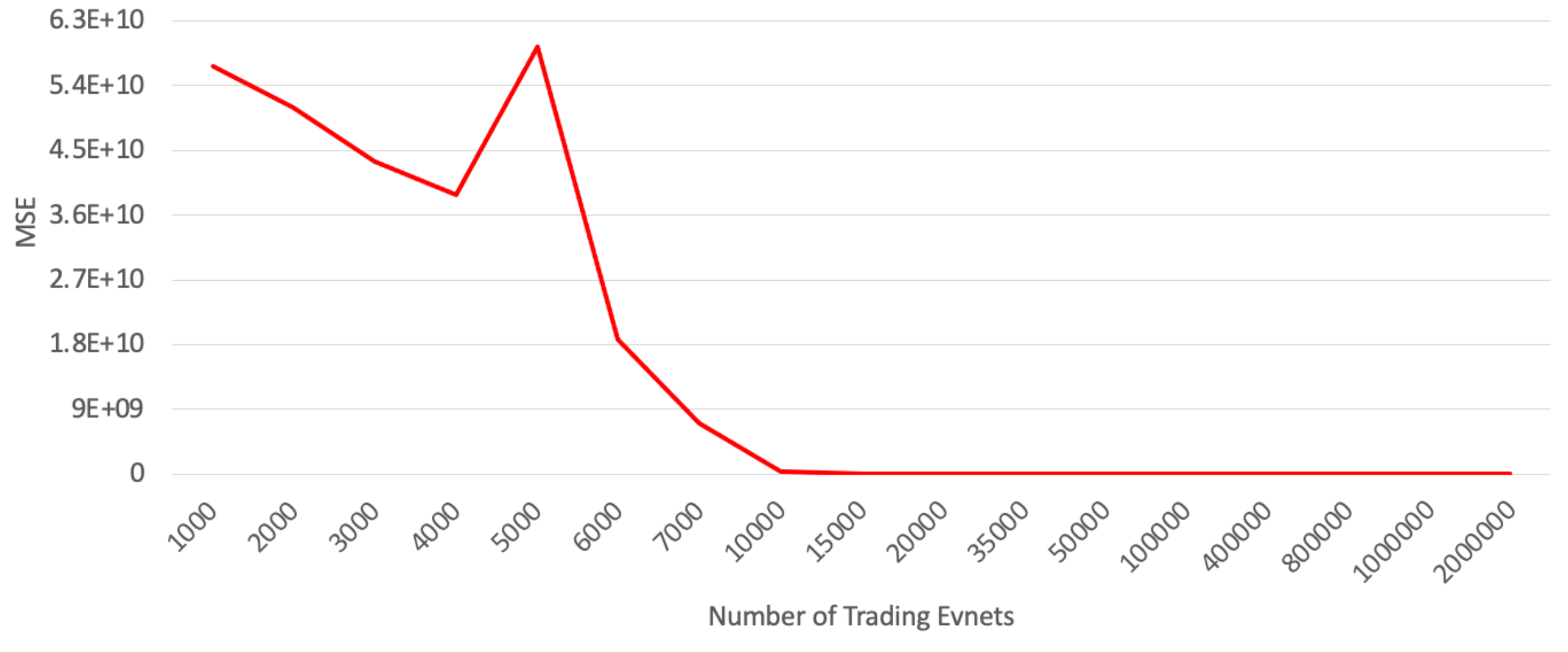}}}
  \subfloat[OPTM-LSTM testing MSE scores \label{1b}]{%
        \scalebox{0.53}{\includegraphics[width=0.65\linewidth]{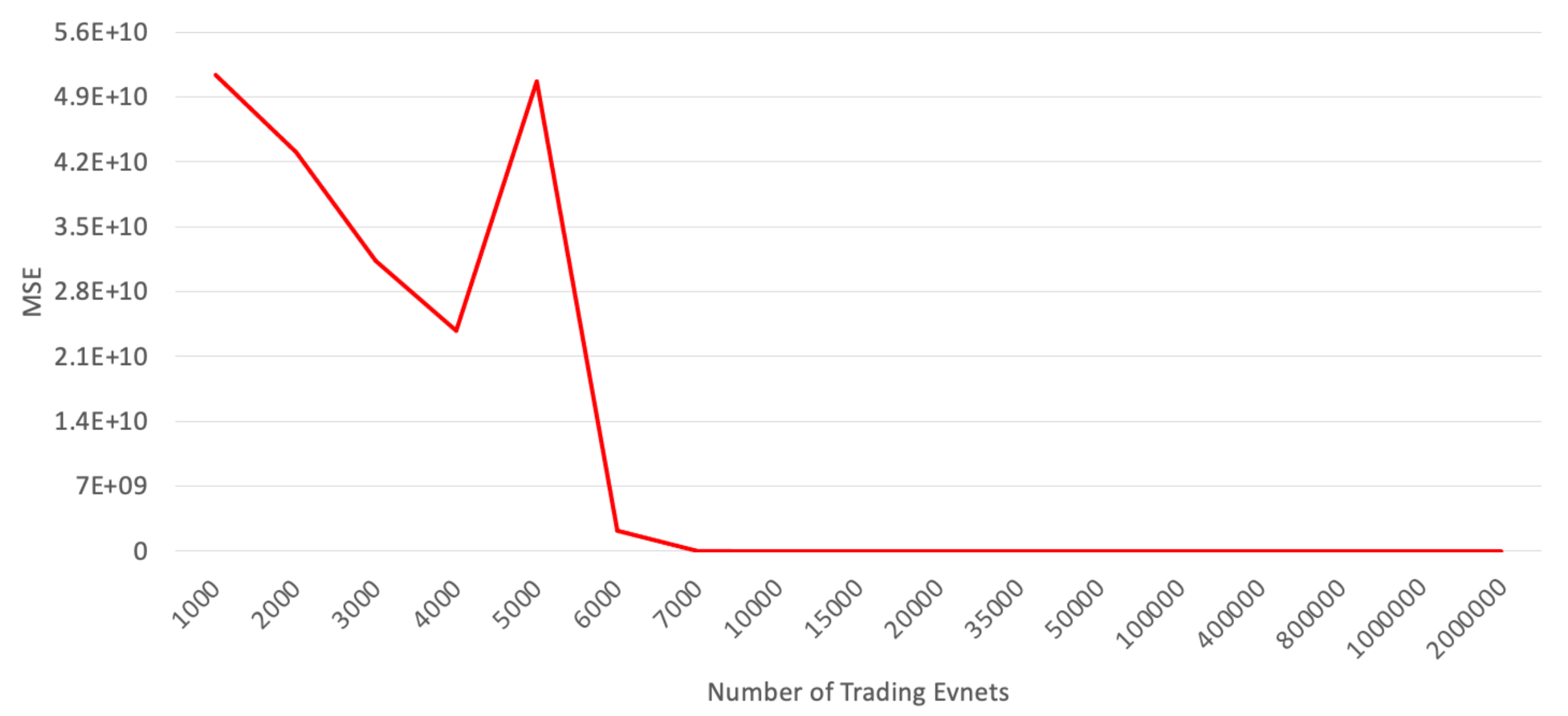}}}
    \\
  \subfloat[LSTM training MSE scores \label{1c}]{%
        \scalebox{0.53}{\includegraphics[width=0.65\linewidth]{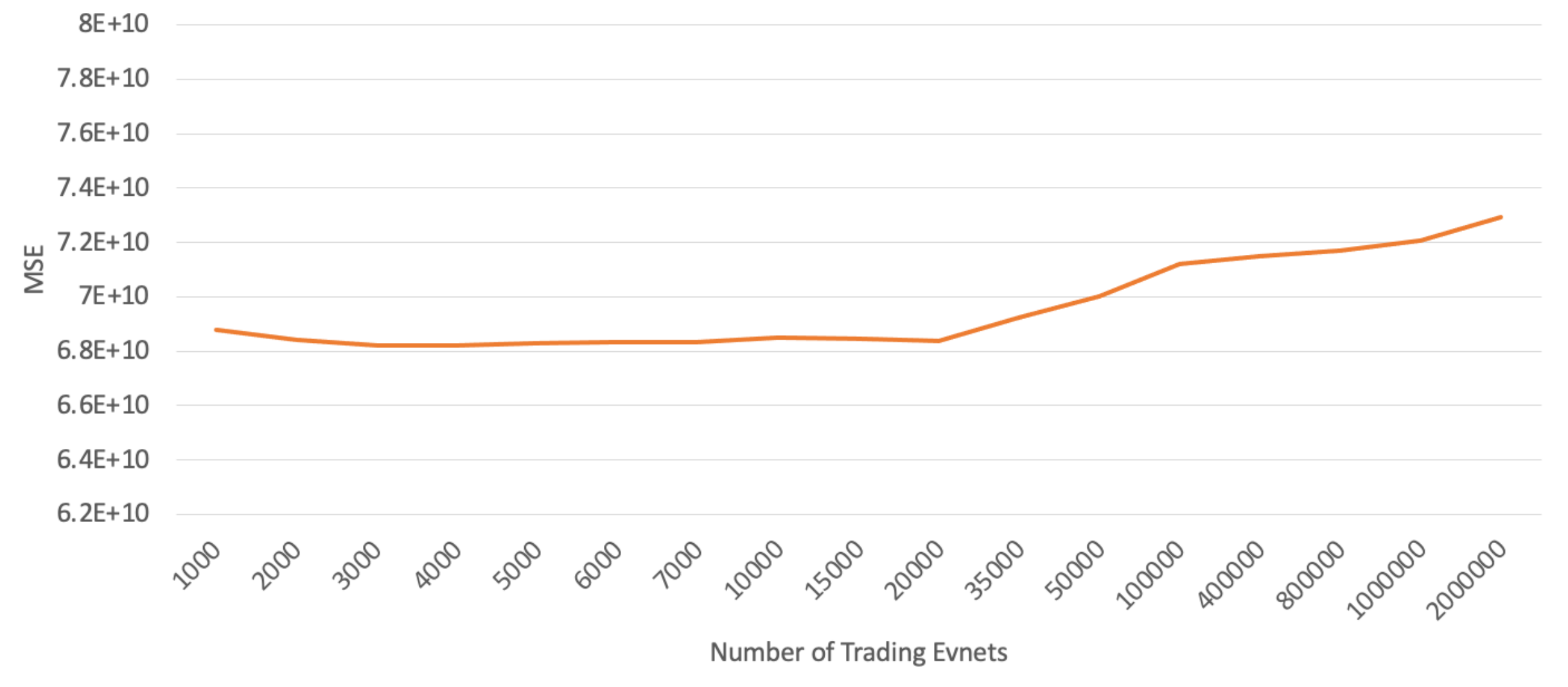}}}
  \subfloat[LSTM testing MSE scores \label{1d}]{%
        \scalebox{0.53}{\includegraphics[width=0.65\linewidth]{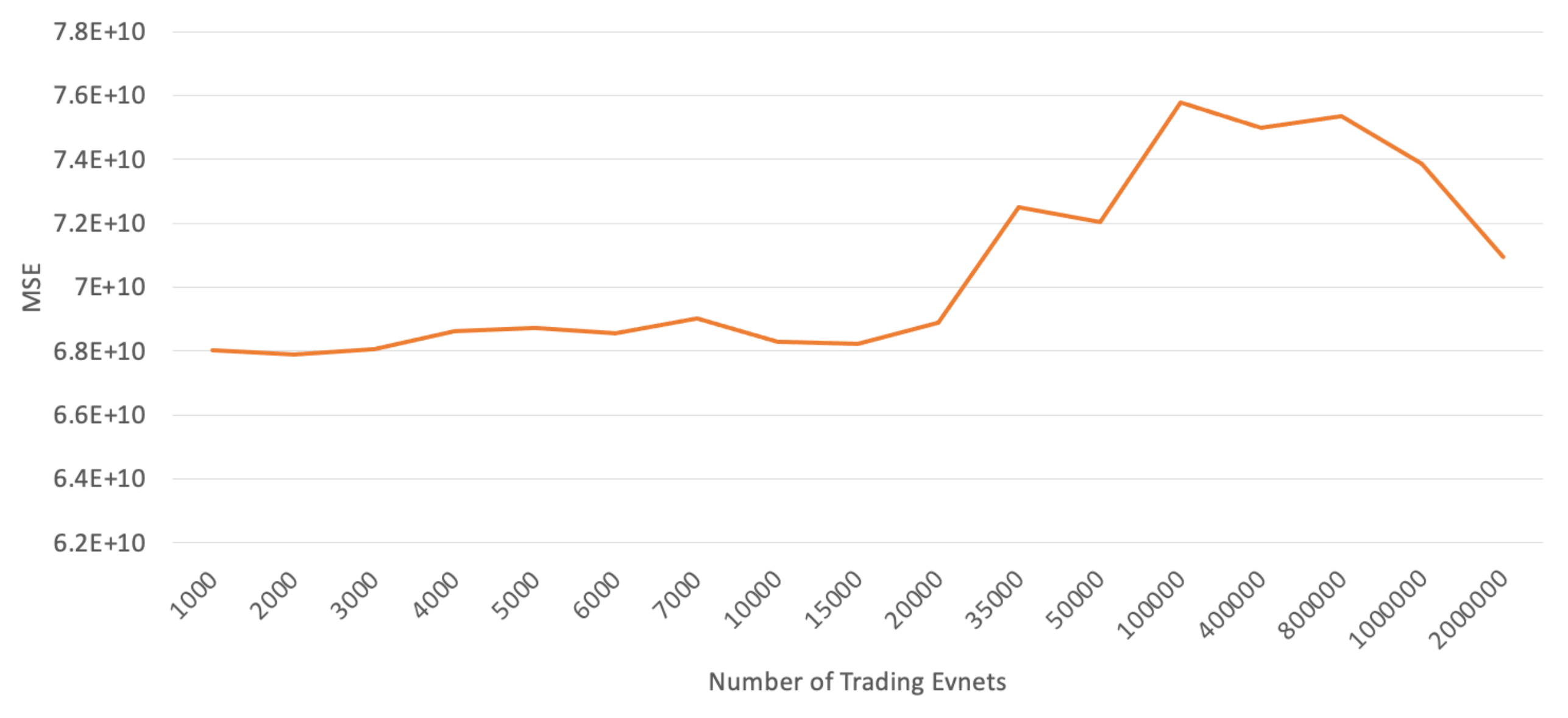}}}
    \\
  \subfloat[Attention LSTM training MSE scores\label{1c}]{%
        \scalebox{0.53}{\includegraphics[width=0.65\linewidth]{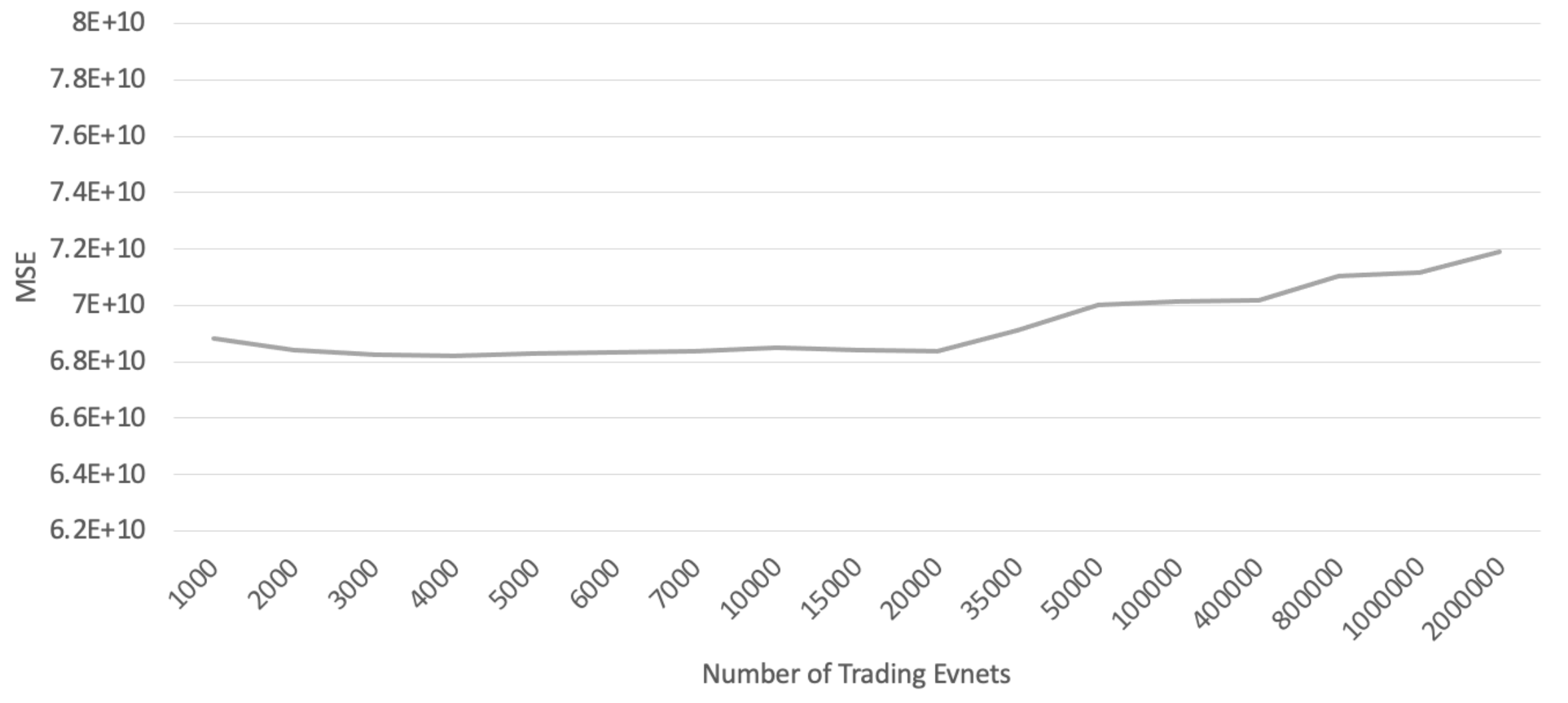}}}
  \subfloat[Attention LSTM testing MSE scores \label{1d}]{%
        \scalebox{0.53}{\includegraphics[width=0.65\linewidth]{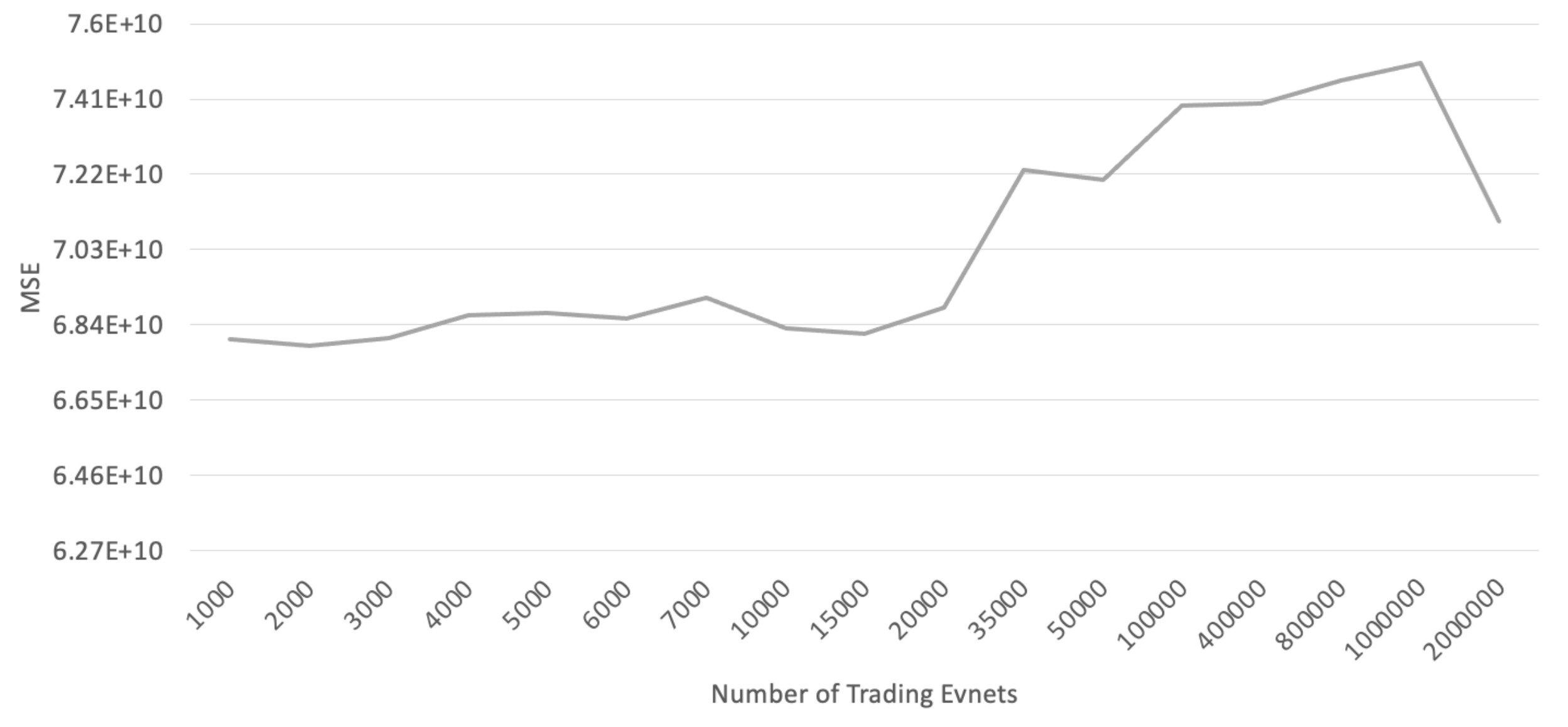}}}  
    \\
  \subfloat[Bidirectional training MSE scores \label{1a}]{%
       \scalebox{0.53}{\includegraphics[width=0.65\linewidth]{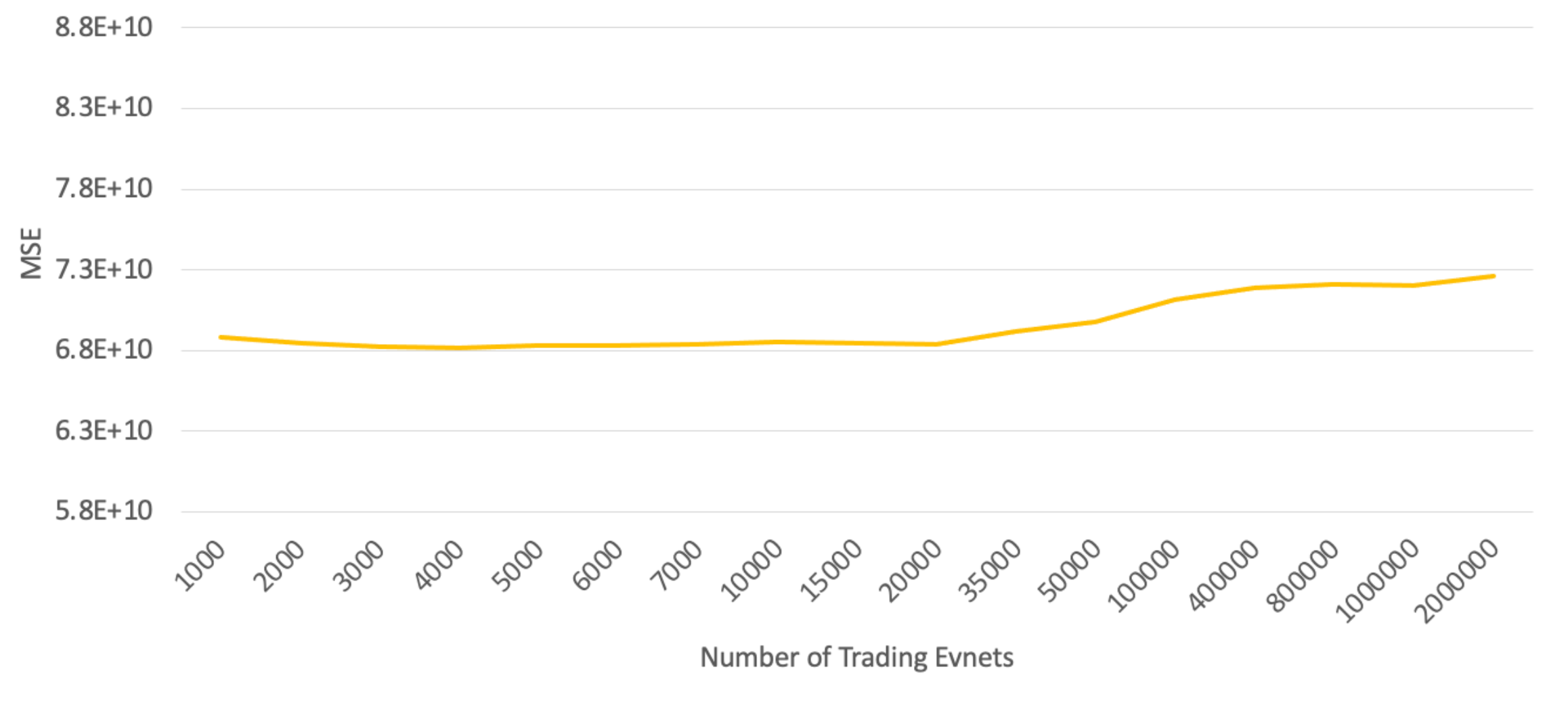}}}
  \subfloat[Bidirectional testing MSE scores \label{1b}]{%
        \scalebox{0.53}{\includegraphics[width=0.65\linewidth]{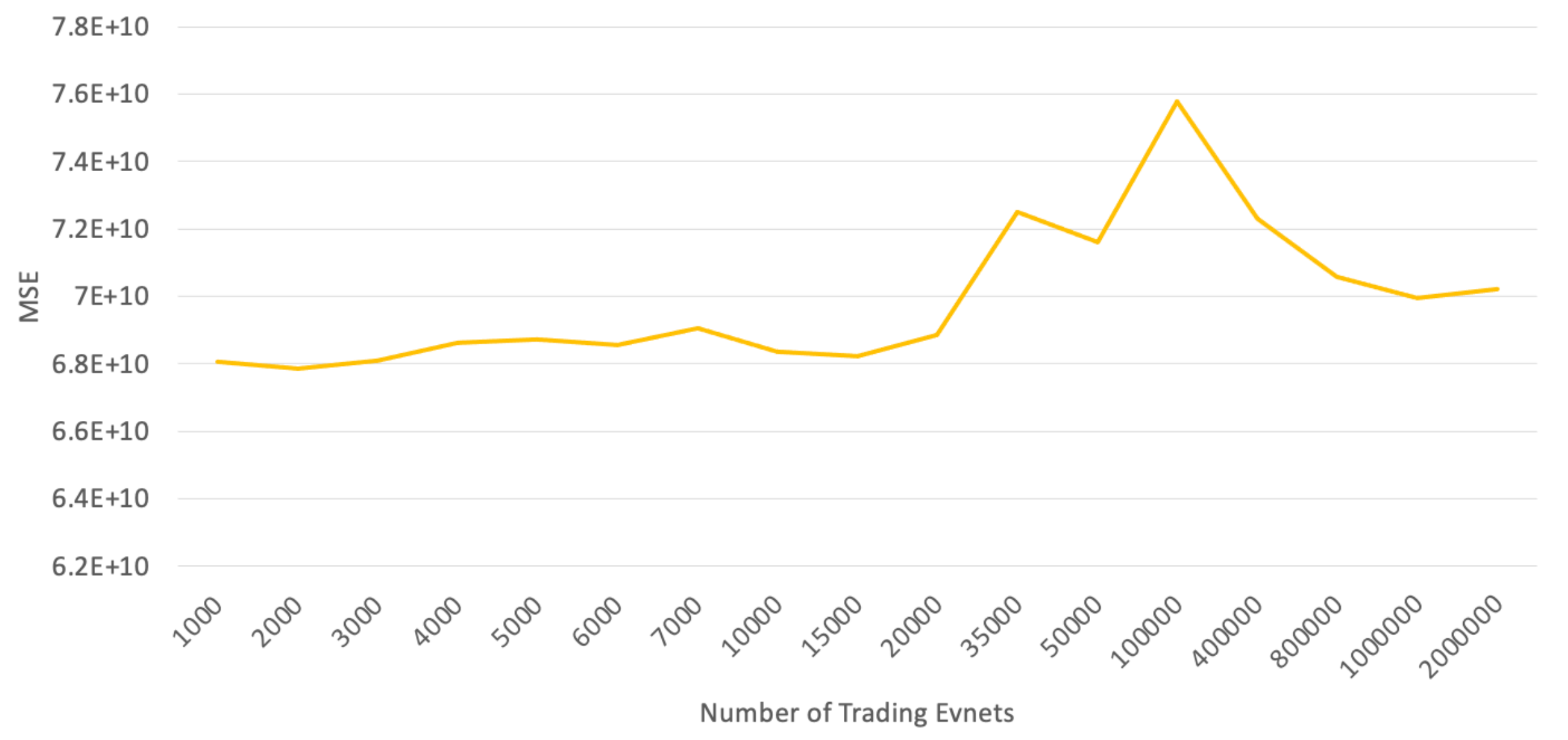}}}
    \\
  \subfloat[GRU training MSE scores \label{1a}]{%
       \scalebox{0.53}{\includegraphics[width=0.65\linewidth]{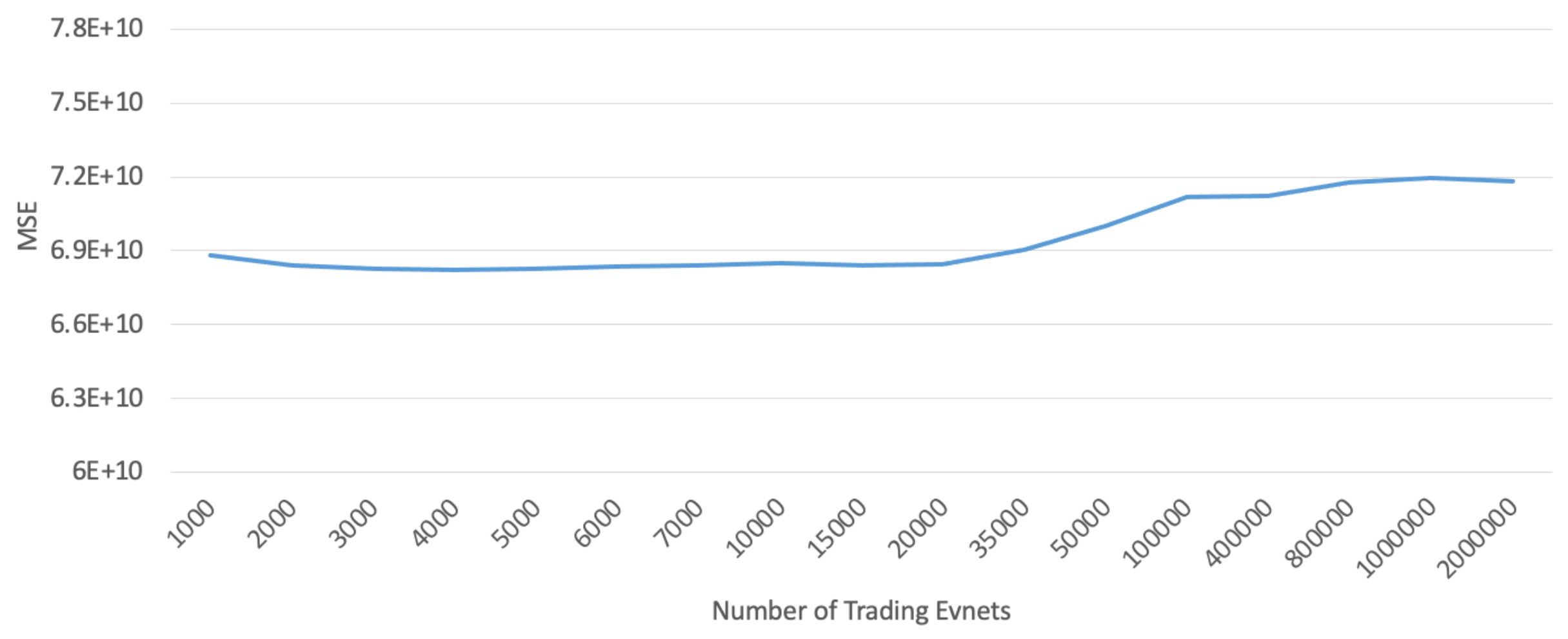}}}
  \subfloat[GRU testing MSE scores \label{1b}]{%
        \scalebox{0.53}{\includegraphics[width=0.65\linewidth]{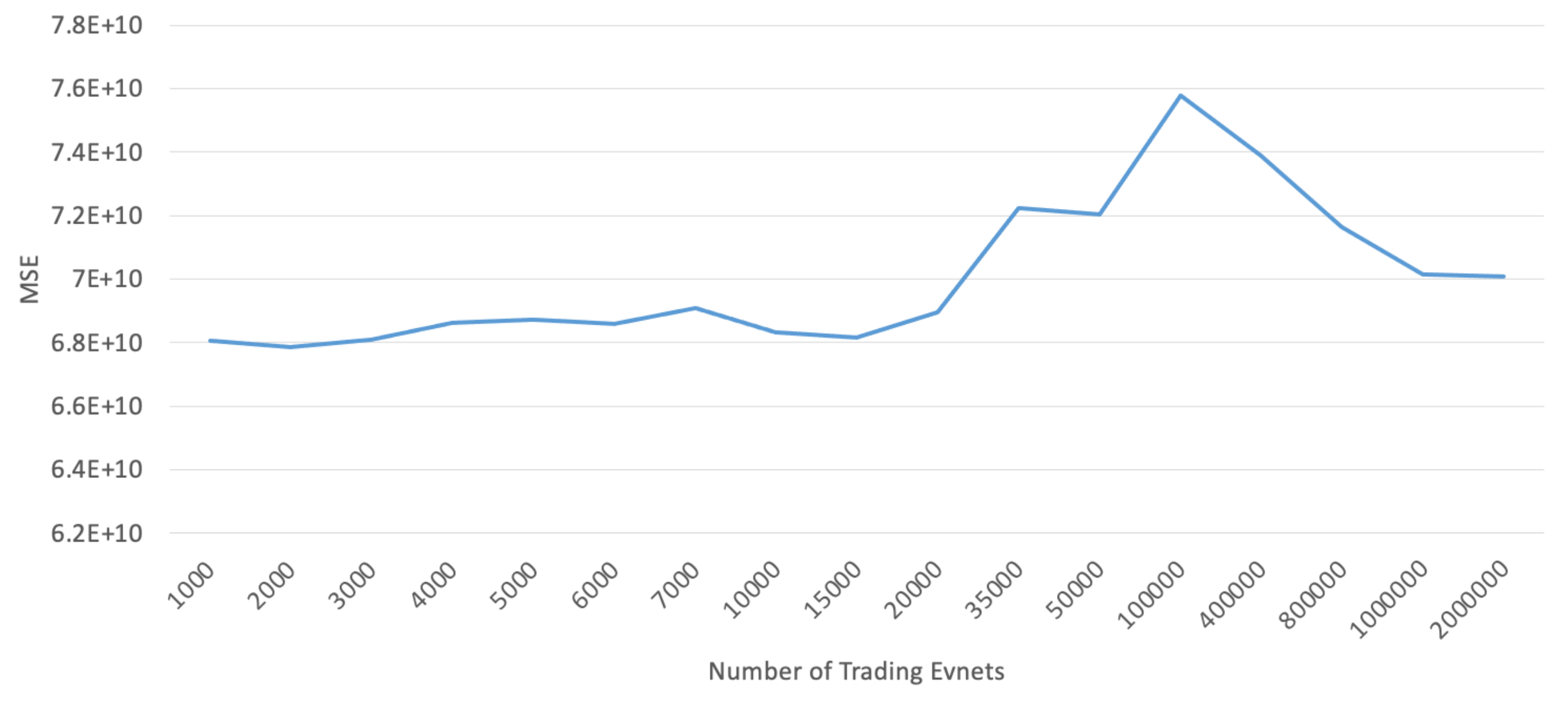}}}
    \\
  \subfloat[Hybrid training MSE scores \label{1a}]{%
       \scalebox{0.53}{\includegraphics[width=0.65\linewidth]{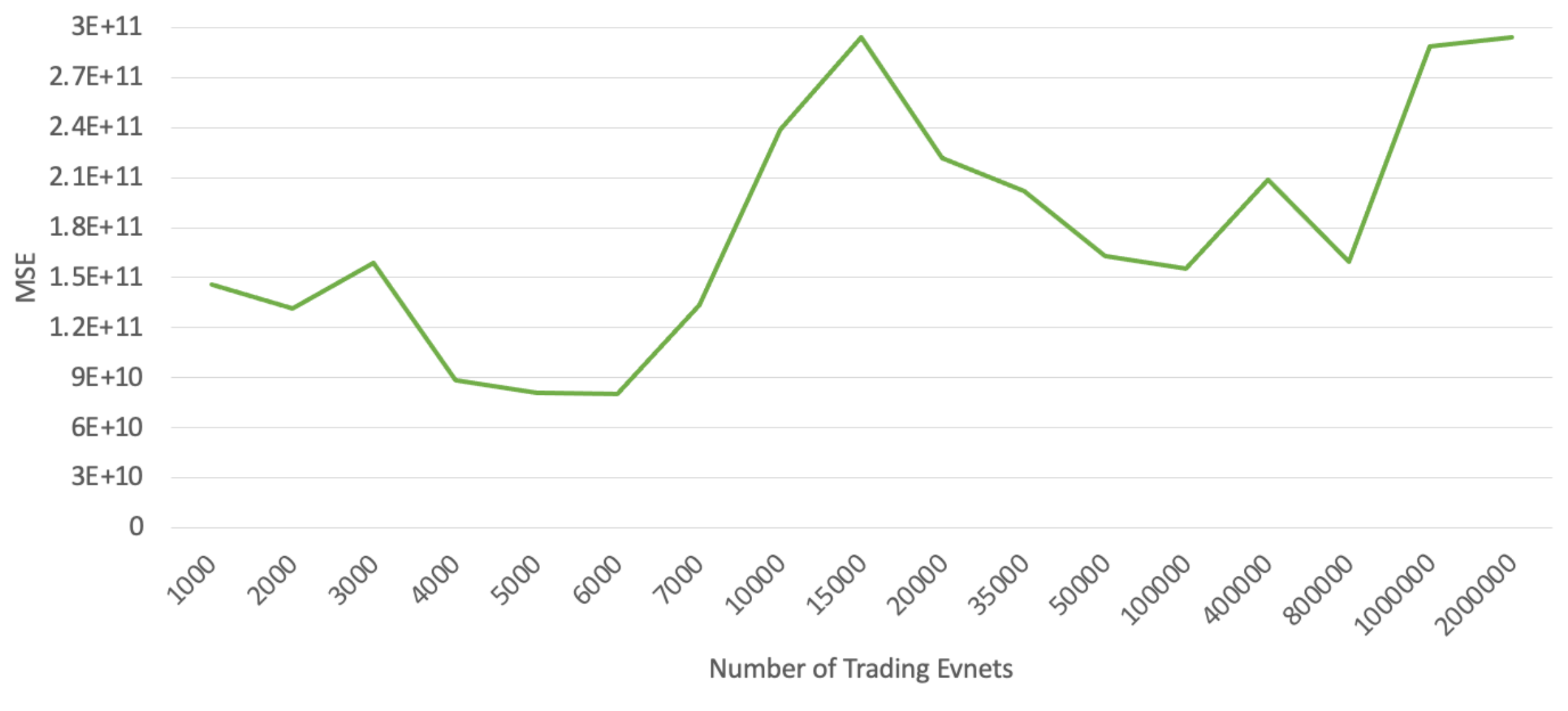}}}
  \subfloat[Hybrid testing MSE scores \label{1b}]{%
        \scalebox{0.53}{\includegraphics[width=0.65\linewidth]{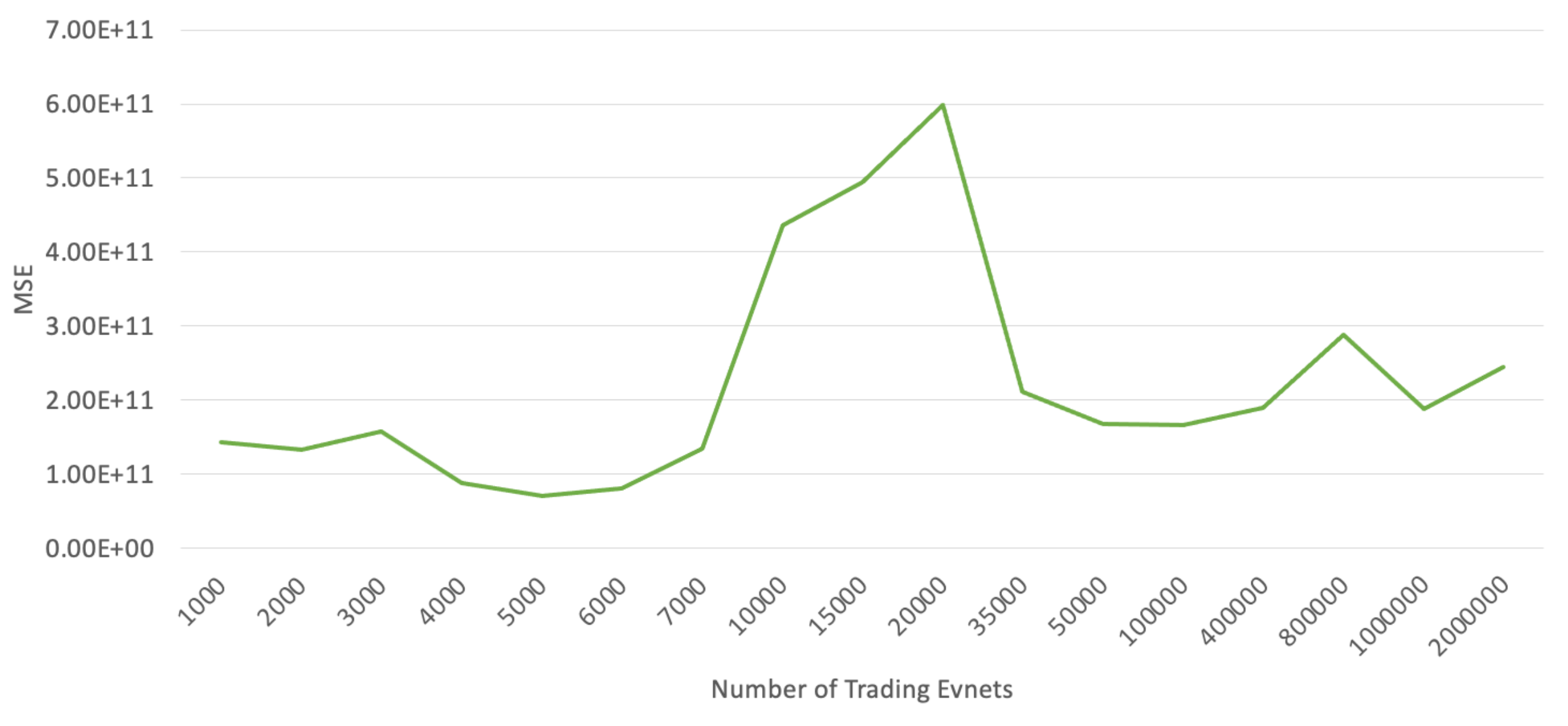}}}  
  \caption{Kesko Short MSE scores based on \hyperref[tab:KeskoShort]{Table \ref{tab:KeskoShort}}.}
  \label{fig:KeskoShort} 
\end{figure*}

\begin{table*}[hbt!]
\centering
\captionsetup{width=.70\textwidth}
\caption{Kesko MSE scores under the Long experimental protocol.}
\scalebox{0.60}{
\begin{tabular}{rcrlcrcrlcrcrl}
\cmidrule[2pt]{1-6}\cmidrule[2pt]{6-9}\cmidrule[2pt]{9-14}
\textbf{Stock} & \textbf{Size} & \textbf{Model} & \textbf{MSE - Train} & \qquad & \textbf{Stock} & \textbf{Size} & \textbf{Model} & \textbf{MSE - Train} & \qquad & \textbf{Stock} & \textbf{Size} & \textbf{Model} & \textbf{MSE - Train}\\
\cmidrule{1-4}\cmidrule{6-9}\cmidrule{11-14}
 Kesko & 1,000 & \textbf{OPTM-LSTM} & \textbf{4.68200E+04} & \qquad & Kesko &  2,000 & \textbf{OPTM-LSTM}& \textbf{2.21950E+04}  & \qquad & Kesko & 3,000 & \textbf{OPTM-LSTM}& \textbf{1.73180E+04}\\   
 &             & LSTM          & 6.88089E+10           & \qquad &       &        & LSTM         & 6.81480E+10            & \qquad &       &        & LSTM          & 6.82454E+10\\ 
 &             & Attention     & 6.88093E+10           & \qquad &       &        & Attention    & 6.81174E+10            & \qquad &       &        & Attention     & 6.82453E+10\\   
 &             & Bidirectional & 6.88091E+10           & \qquad &       &        & Bidirectional& 6.81027E+10            & \qquad &       &        & Bidirectional & 6.78117E+10\\   
 &             & GRU           & 6.88090E+10           & \qquad &       &        & GRU          & 6.84291E+10            & \qquad &       &        & GRU           & 6.82455E+10\\   
 &             & Hybrid        & 1.72457E+11              & \qquad &       &        & Hybrid    & 2.05373E+11               & \qquad &       &        & Hybrid     & 1.21925E+11\\
\cmidrule{2-4}\cmidrule{7-9}\cmidrule{12-14}
 & 4,000 & \textbf{OPTM-LSTM} & \textbf{1.49260E+04} & \qquad &        &  5,000  & \textbf{OPTM-LSTM}  & \textbf{1.23690E+04} & \qquad &  & 6,000 & \textbf{OPTM-LSTM} & \textbf{9.92900E+03}\\   
 &             & LSTM           & 6.75009E+10& \qquad &                       &         & LSTM         & 6.82883E+10 & \qquad &          &        & LSTM          & 6.77034E+10\\ 
 &             & Attention      & 6.71903E+10& \qquad &                       &         & Attention    & 6.64124E+10 & \qquad &          &        & Attention     & 6.55445E+10\\   
 &             & Bidirectional  & 6.70497E+10& \qquad &                       &         & Bidirectional& 6.63702E+10 & \qquad &          &        & Bidirectional & 6.83595E+10\\   
 &             & GRU            & 6.82024E+10& \qquad &                       &         & GRU          & 6.67256E+10 & \qquad &          &        & GRU           & 6.83590E+10\\   
 &             & Hybrid         & 2.14246E+11 & \qquad &                       &         & Hybrid      & 8.93922E+10 & \qquad &          &        & Hybrid        & 2.01985E+11\\
\cmidrule{2-4}\cmidrule{7-9}\cmidrule{12-14}
& 7,000 & \textbf{OPTM-LSTM}& \textbf{1.04300E+04} & \qquad &       &  10,000 & \textbf{OPTM-LSTM}  & \textbf{6.87700E+03} & \qquad &  & 15,000 & \textbf{OPTM-LSTM} & \textbf{5.94700E+03}\\   
 &              & LSTM          & 6.83884E+10 & \qquad &            &         & LSTM         & 6.64836E+10 & \qquad &        &                           & LSTM   & 6.58317E+10\\ 
 &              & Attention     & 6.47468E+10 & \qquad &            &         & Attention    & 5.61866E+10 & \qquad &        &                           & Attention & 6.58317E+10\\ 
 &              & Bidirectional & 6.83898E+10 & \qquad &            &         & Bidirectional& 6.06804E+10 & \qquad &        &                           & Bidirectional & 4.84266E+10\\   
 &              & GRU           & 6.83906E+10 & \qquad &            &         & GRU          & 6.85372E+10 & \qquad &        &                           & GRU           & 4.47933E+10\\   
 &              & Hybrid        & 1.27098E+11 & \qquad &            &         & Hybrid       & 2.48311E+11 & \qquad &        &                           & Hybrid       & 1.13078E+11\\
\cmidrule[2pt]{1-6}\cmidrule[2pt]{6-9}\cmidrule[2pt]{9-14}
\textbf{Stock} & \textbf{Size} & \textbf{Model} & \textbf{MSE - Test} & \qquad & \textbf{Stock} & \textbf{Size} & \textbf{Model} & \textbf{MSE - Test} & \qquad & \textbf{Stock} & \textbf{Size} & \textbf{Model} & \textbf{MSE - Test}\\
\cmidrule{1-4}\cmidrule{6-9}\cmidrule{11-14}
 Kesko & 1,000 & \textbf{OPTM-LSTM} & \textbf{3.26100E+03} & \qquad &  Kesko &  2,000 & \textbf{OPTM-LSTM}& \textbf{1.22400E+03}  & \qquad & Kesko & 3,000 & \textbf{OPTM-LSTM}& \textbf{2.64000E+03}\\   
 &             & LSTM          & 6.80489E+10           & \qquad &       &        & LSTM         & 6.76086E+10            & \qquad &       &        & LSTM          & 6.80769E+10\\ 
 &             & Attention     & 6.80492E+10           & \qquad &       &        & Attention    & 6.75622E+10            & \qquad &       &        & Attention     & 6.80767E+10\\   
 &             & Bidirectional & 6.80488E+10           & \qquad &       &        & Bidirectional& 6.75473E+10            & \qquad &       &        & Bidirectional & 6.76332E+10\\   
 &             & GRU           & 6.80489E+10           & \qquad &       &        & GRU          & 6.78792E+10            & \qquad &       &        & GRU           & 6.80770E+10\\   
 &             & Hybrid        & 1.73009E+11           & \qquad &       &        & Hybrid       & 2.81E115+11            & \qquad &       &        & Hybrid        & 1.22093E+11\\
\cmidrule{2-4}\cmidrule{7-9}\cmidrule{12-14}
 & 4,000 & \textbf{OPTM-LSTM} & \textbf{1.25600E+03} & \qquad &        &  5,000  & \textbf{OPTM-LSTM}  & \textbf{4.85000E+02} & \qquad &  & 6,000 & \textbf{OPTM-LSTM} & \textbf{1.40800E+03}\\   
 &             & LSTM           & 6.79221E+10& \qquad &                       &         & LSTM         & 6.87101E+10 & \qquad &          &        & LSTM          & 6.79125E+10\\ 
 &             & Attention      & 6.75923E+10& \qquad &                       &         & Attention    & 6.67740E+10 & \qquad &          &        & Attention     & 6.56693E+10\\   
 &             & Bidirectional  & 6.74470E+10& \qquad &                       &         & Bidirectional& 6.67399E+10 & \qquad &          &        & Bidirectional & 6.85769E+10\\   
 &             & GRU            & 6.86338E+10& \qquad &                       &         & GRU          & 6.70961E+10 & \qquad &          &        & GRU           & 6.85764E+10\\   
 &             & Hybrid         & 2.27355E+11 & \qquad &                       &         & Hybrid      & 8.95668E+10 & \qquad &          &        & Hybrid        & 3.84335E+11\\
\cmidrule{2-4}\cmidrule{7-9}\cmidrule{12-14}
 & 7,000 & \textbf{OPTM-LSTM}& \textbf{5.98900E+03} & \qquad &      &  10,000 & \textbf{OPTM-LSTM}  & \textbf{1.04000E+03} & \qquad &  & 15,000 & \textbf{OPTM-LSTM} & \textbf{4.38700E+03}\\   
 &              & LSTM          & 6.90948E+10 & \qquad &            &         & LSTM         & 6.62728E+10 & \qquad &        &                           & LSTM   & 6.55495E+10\\ 
 &              & Attention     & 6.53064E+10 & \qquad &            &         & Attention    & 5.55827E+10 & \qquad &        &                           & Attention & 6.82135E+10\\ 
 &              & Bidirectional & 6.90945E+10 & \qquad &            &         & Bidirectional& 6.02120E+10 & \qquad &        &                           & Bidirectional & 4.75339E+10\\   
 &              & GRU           & 6.90956E+10 & \qquad &            &         & GRU          & 6.83574E+10 & \qquad &        &                           & GRU           & 4.38358E+10\\   
 &              & Hybrid        & 1.27009E+11 & \qquad &            &         & Hybrid       & 3.48221E+11 & \qquad &        &                           & Hybrid & 1.13889E+11\\
\cmidrule[2pt]{1-6}\cmidrule[2pt]{6-9}\cmidrule[2pt]{9-14}
\end{tabular}}
\medskip
\label{tab:KeskoLong}
\end{table*}

\begin{figure*}[hbt!]
    \centering
  \subfloat[OPTM-LSTM training MSE scores \label{1a}]{%
       \scalebox{0.53}{\includegraphics[width=0.65\linewidth]{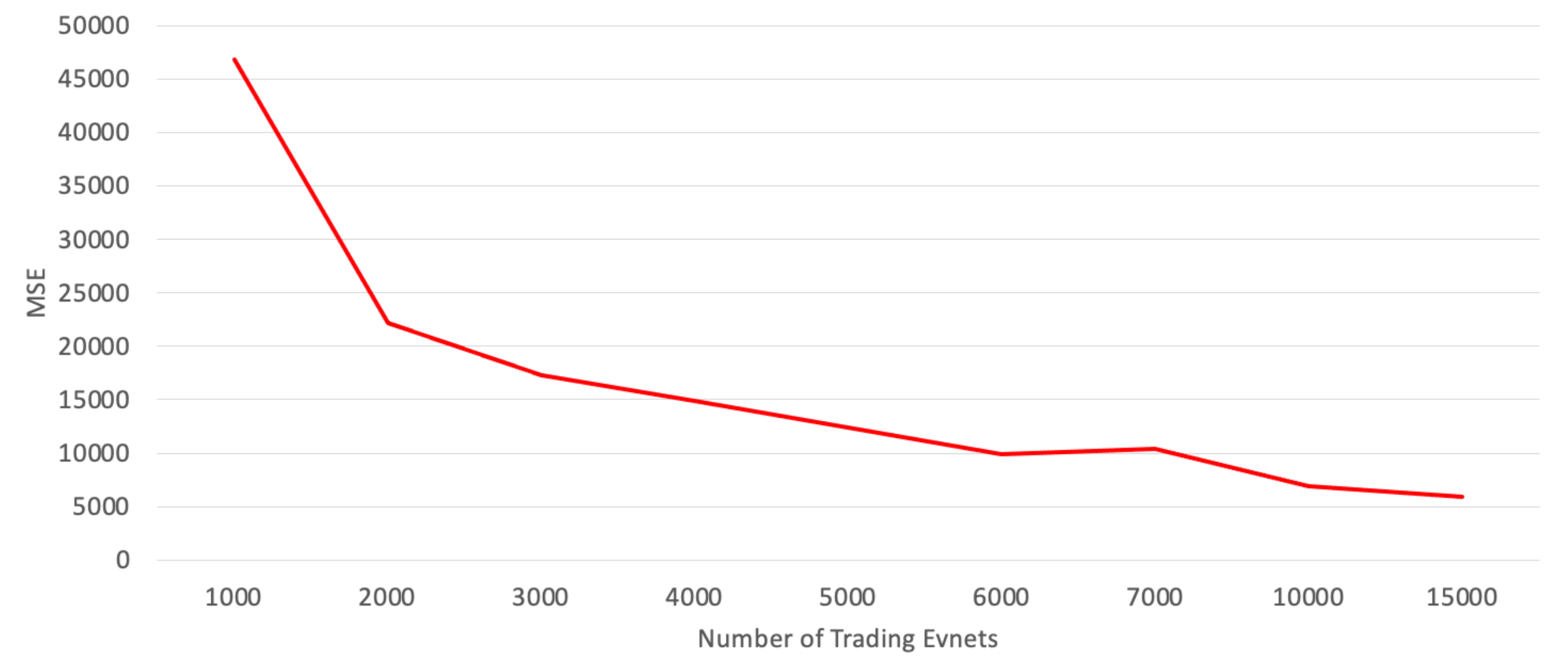}}}
  \subfloat[OPTM-LSTM testing MSE scores \label{1b}]{%
        \scalebox{0.53}{\includegraphics[width=0.65\linewidth]{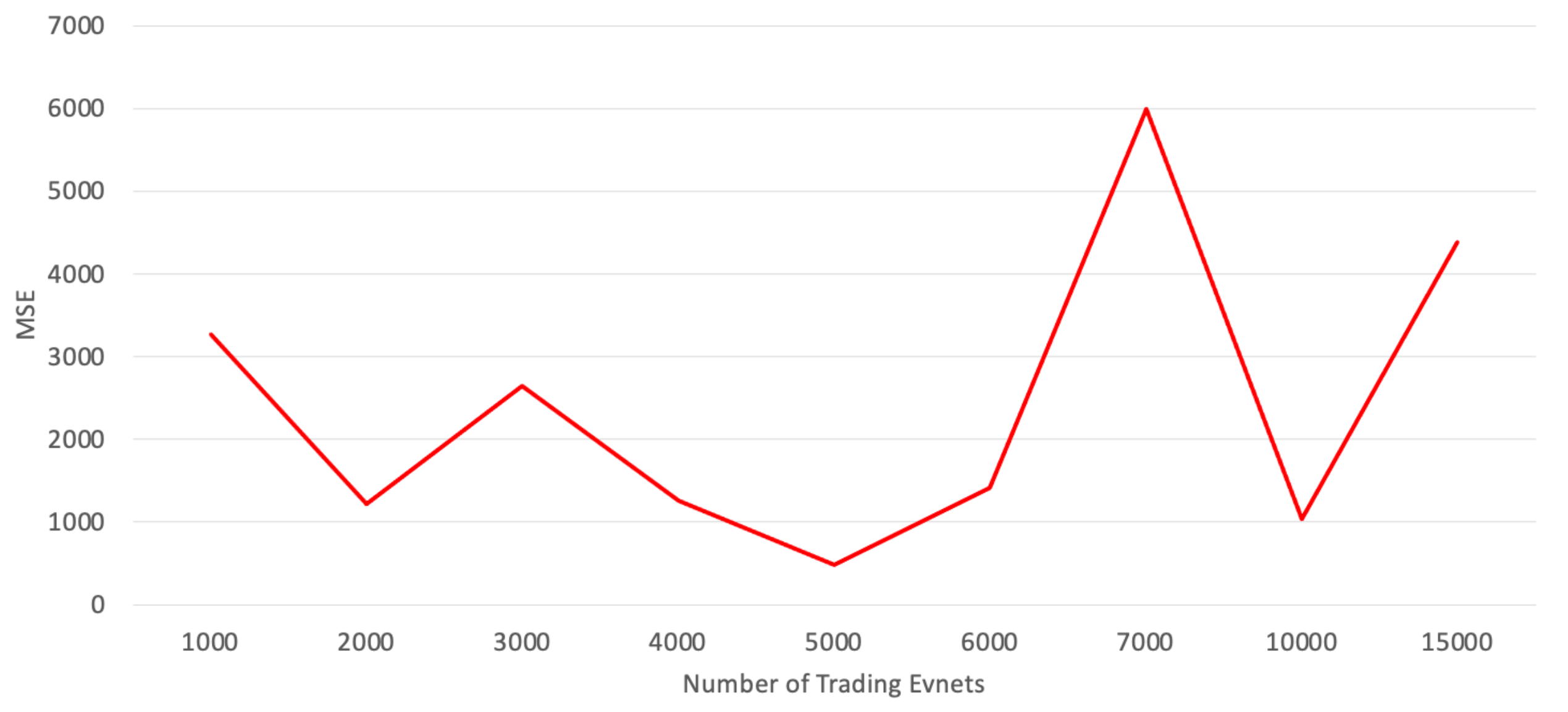}}}
    \\
  \subfloat[LSTM training MSE scores \label{1c}]{%
        \scalebox{0.53}{\includegraphics[width=0.65\linewidth]{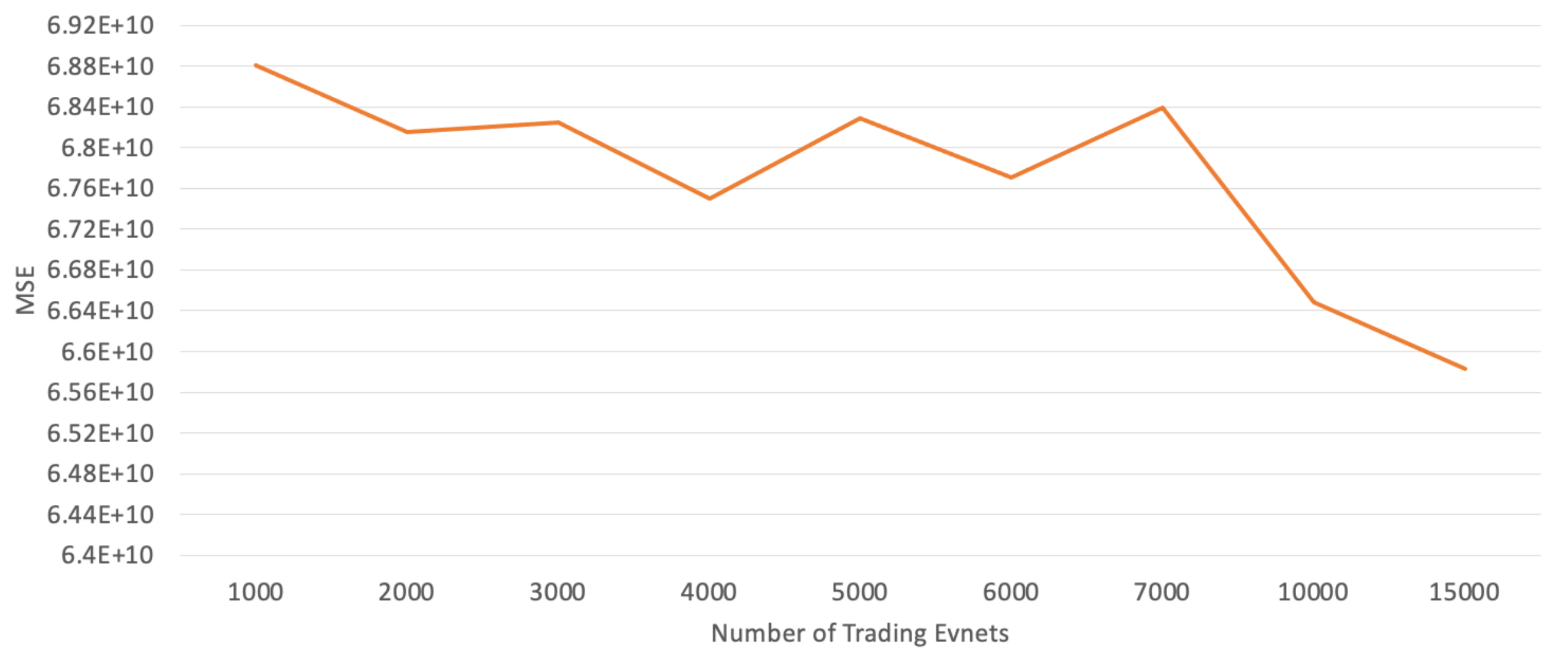}}}
  \subfloat[LSTM testing MSE scores \label{1d}]{%
        \scalebox{0.53}{\includegraphics[width=0.65\linewidth]{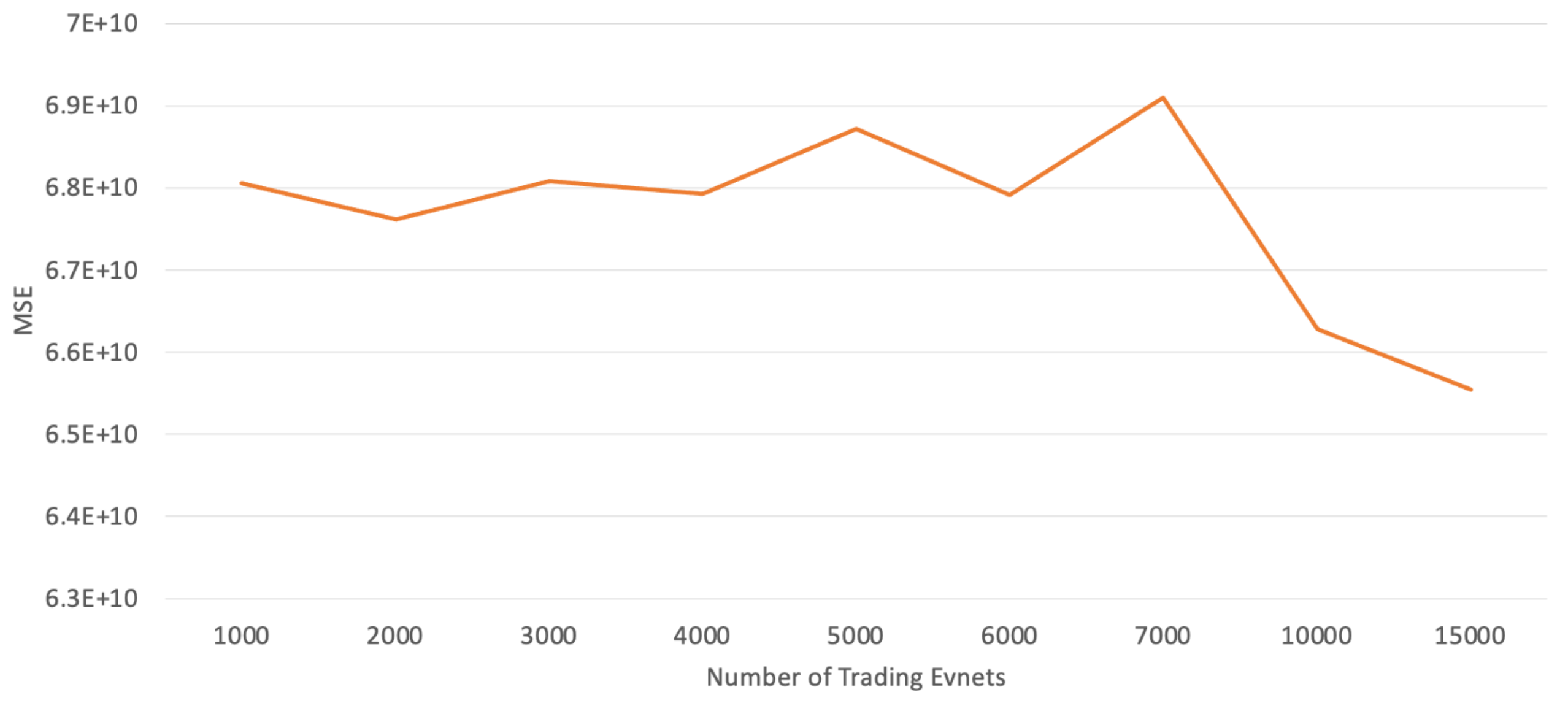}}}
    \\
  \subfloat[Attention LSTM training MSE scores\label{1c}]{%
        \scalebox{0.53}{\includegraphics[width=0.65\linewidth]{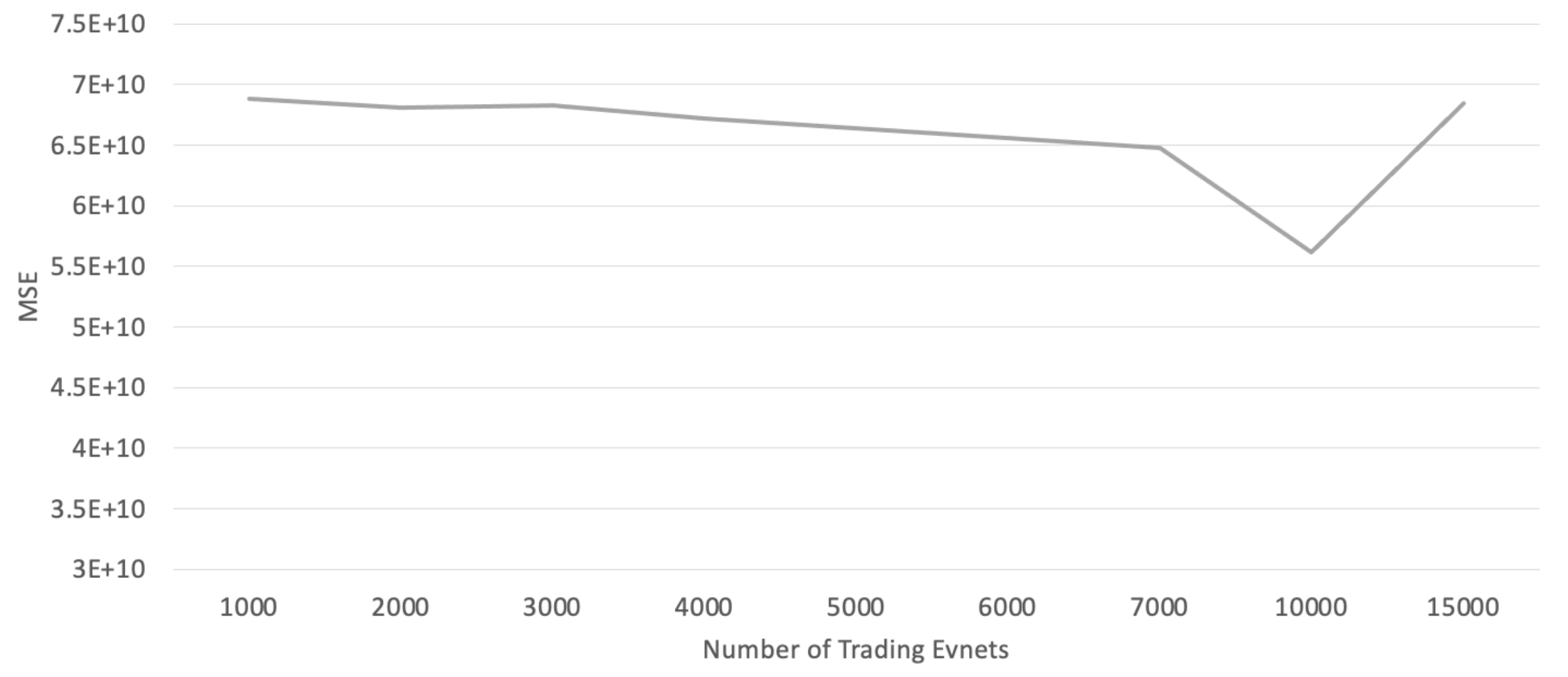}}}
  \subfloat[Attention LSTM testing MSE scores \label{1d}]{%
        \scalebox{0.53}{\includegraphics[width=0.65\linewidth]{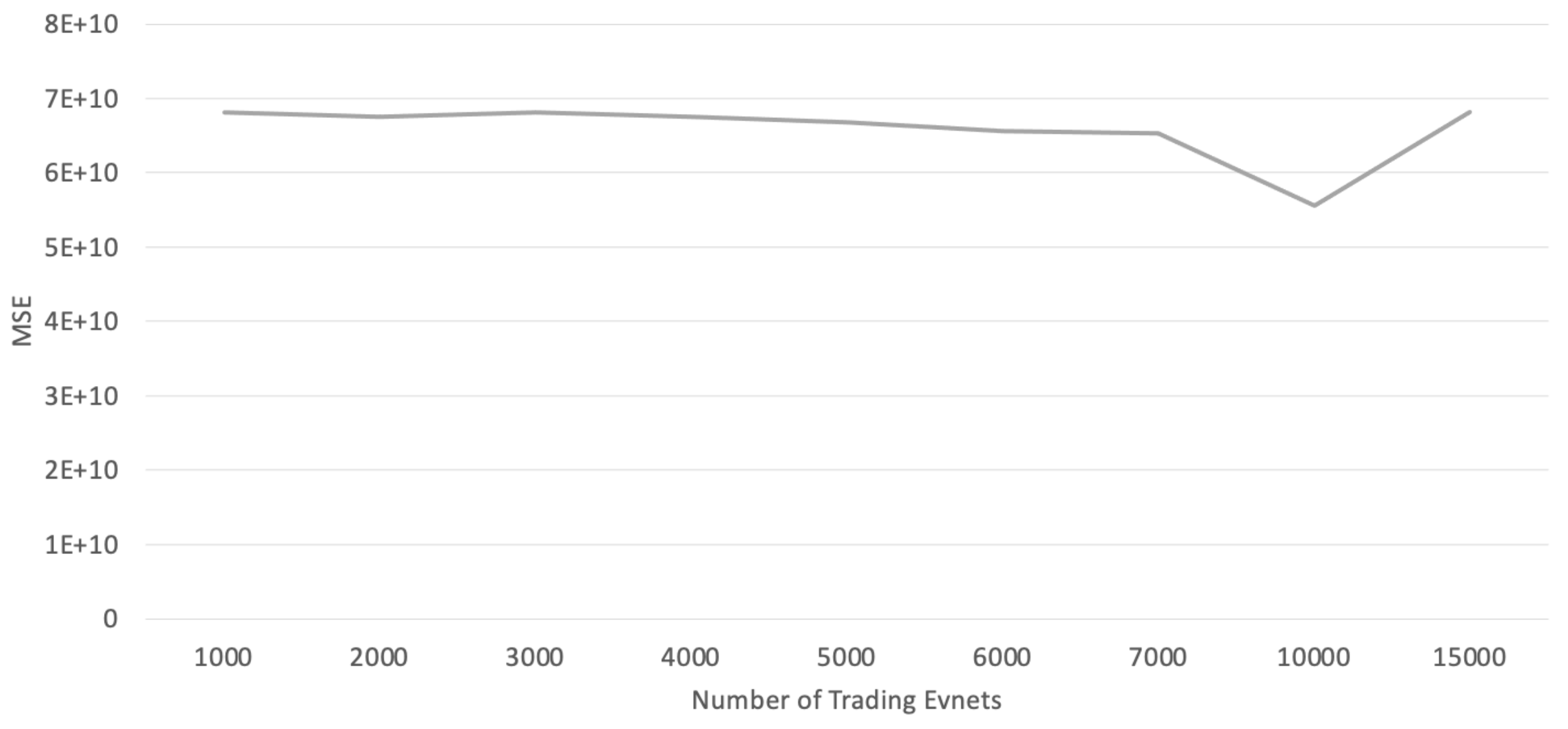}}}  
    \\
  \subfloat[Bidirectional training MSE scores \label{1a}]{%
       \scalebox{0.53}{\includegraphics[width=0.65\linewidth]{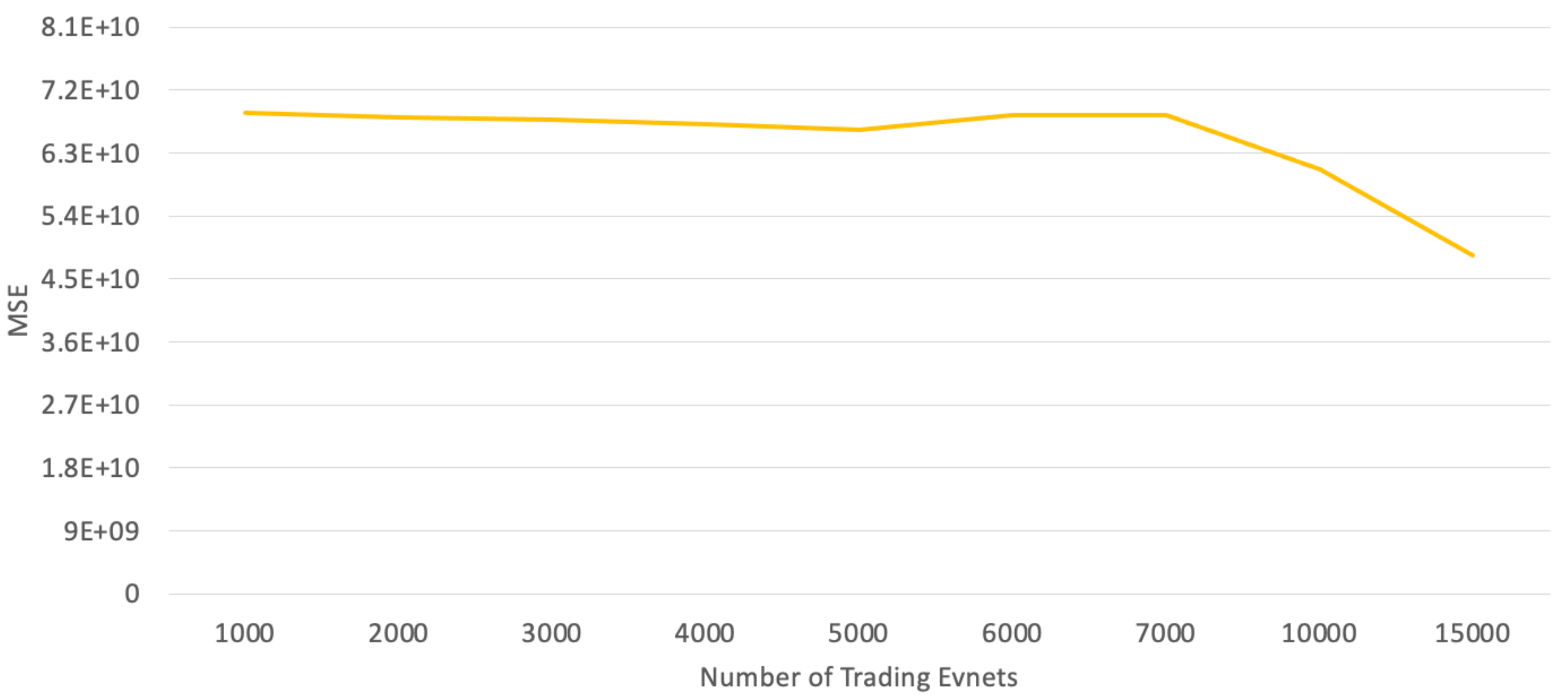}}}
  \subfloat[Bidirectional testing MSE scores \label{1b}]{%
        \scalebox{0.53}{\includegraphics[width=0.65\linewidth]{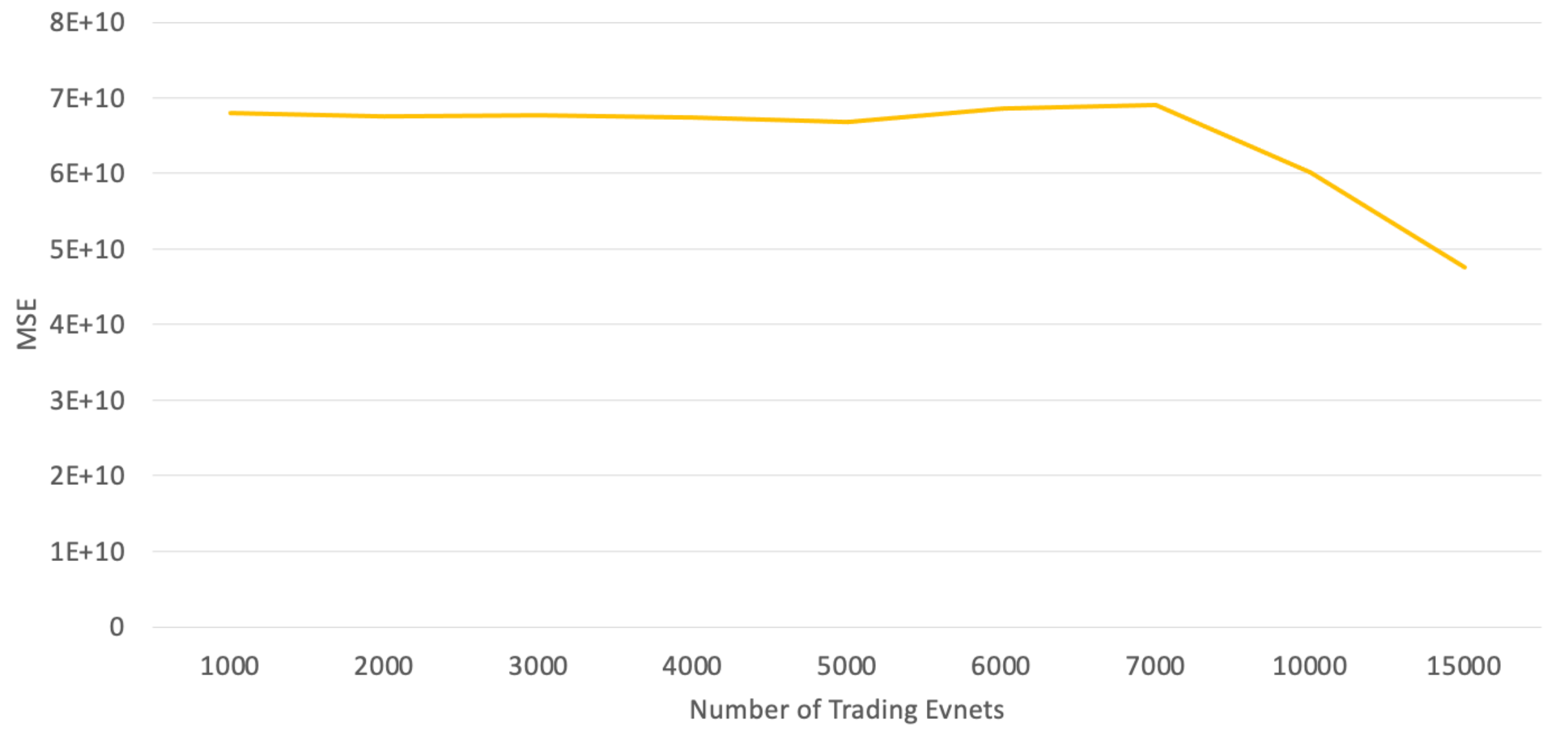}}}
    \\
  \subfloat[GRU training MSE scores \label{1a}]{%
       \scalebox{0.53}{\includegraphics[width=0.65\linewidth]{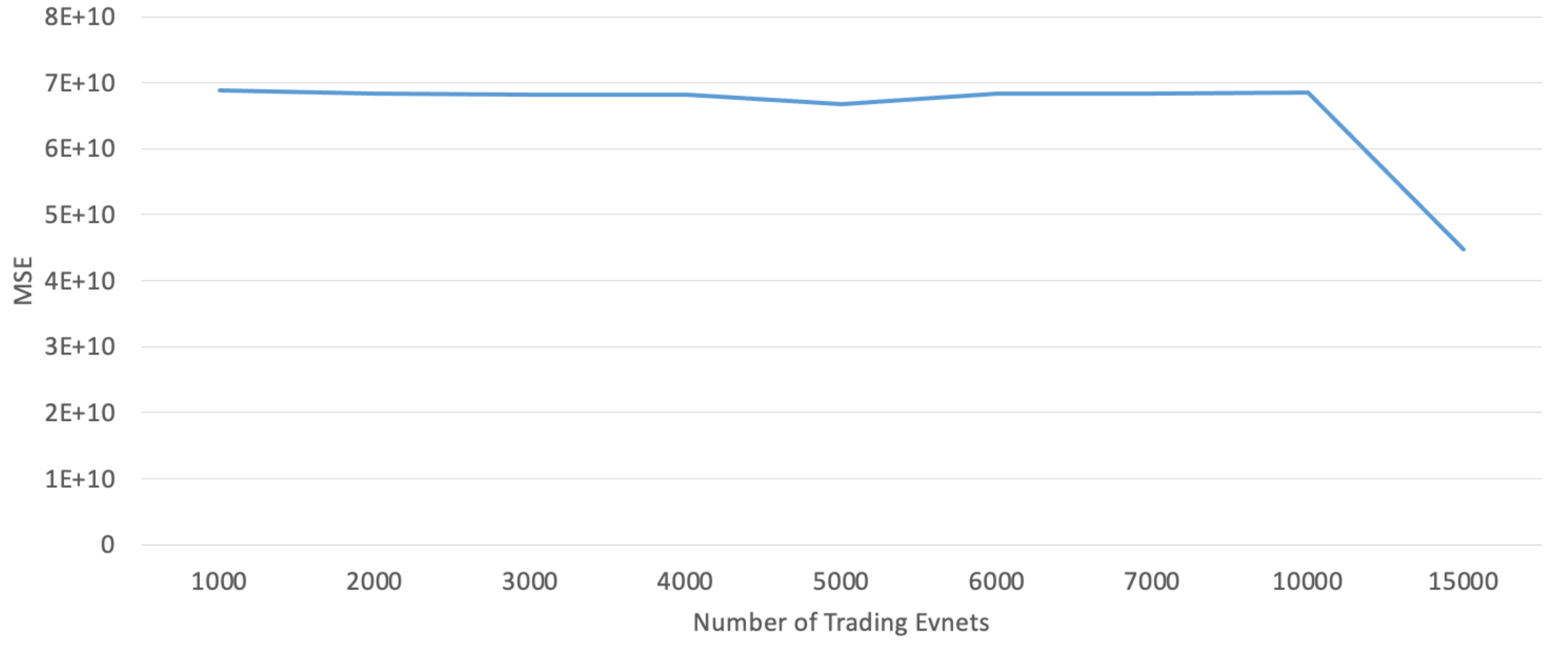}}}
  \subfloat[GRU testing MSE scores \label{1b}]{%
        \scalebox{0.53}{\includegraphics[width=0.65\linewidth]{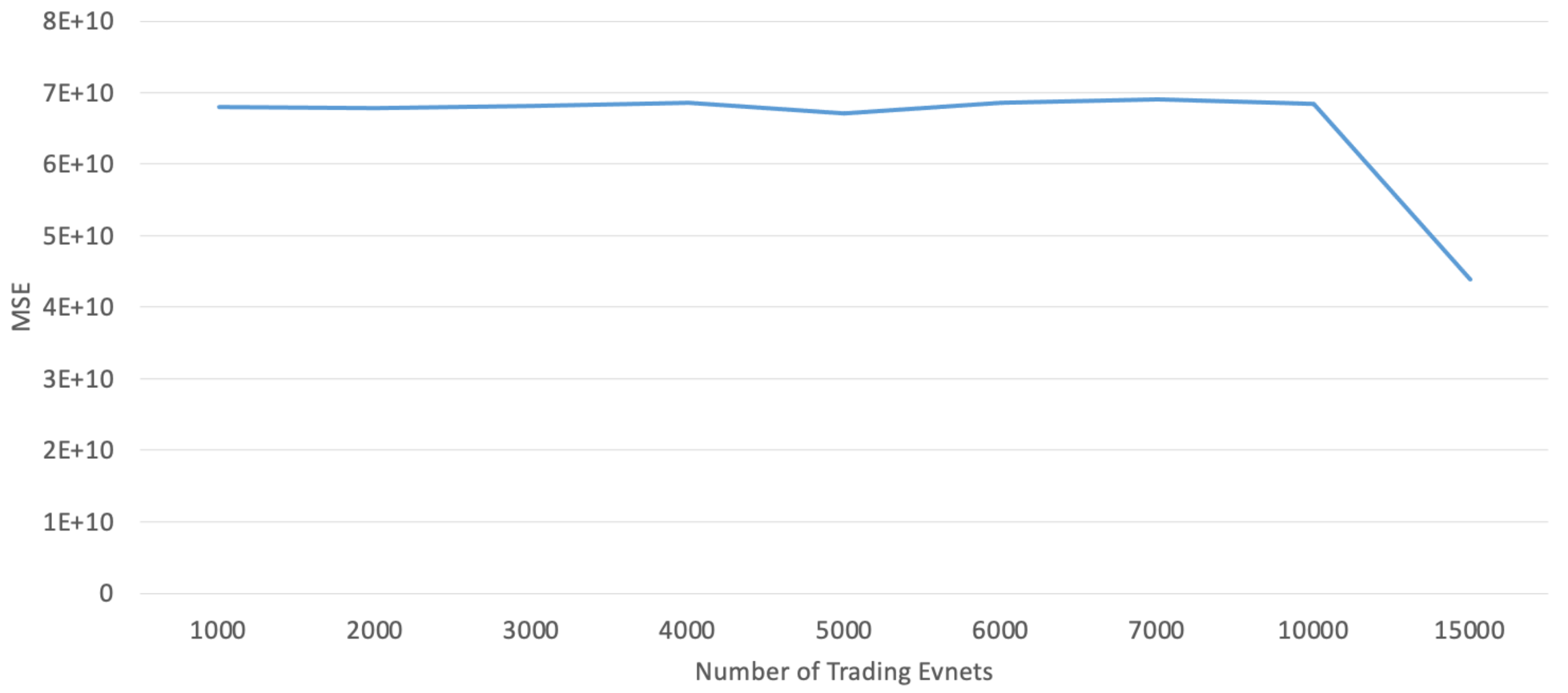}}}
    \\
  \subfloat[Hybrid training MSE scores \label{1a}]{%
       \scalebox{0.53}{\includegraphics[width=0.65\linewidth]{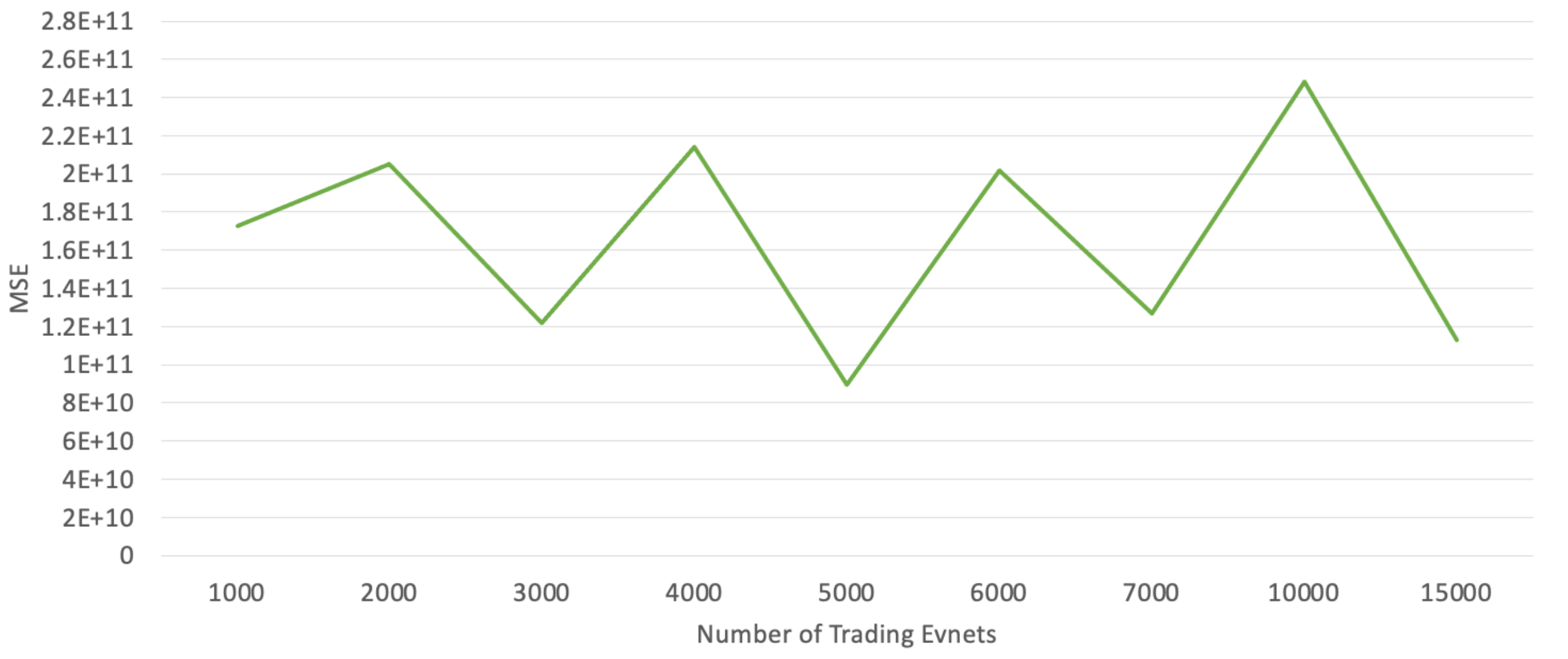}}}
  \subfloat[Hybrid testing MSE scores \label{1b}]{%
        \scalebox{0.53}{\includegraphics[width=0.65\linewidth]{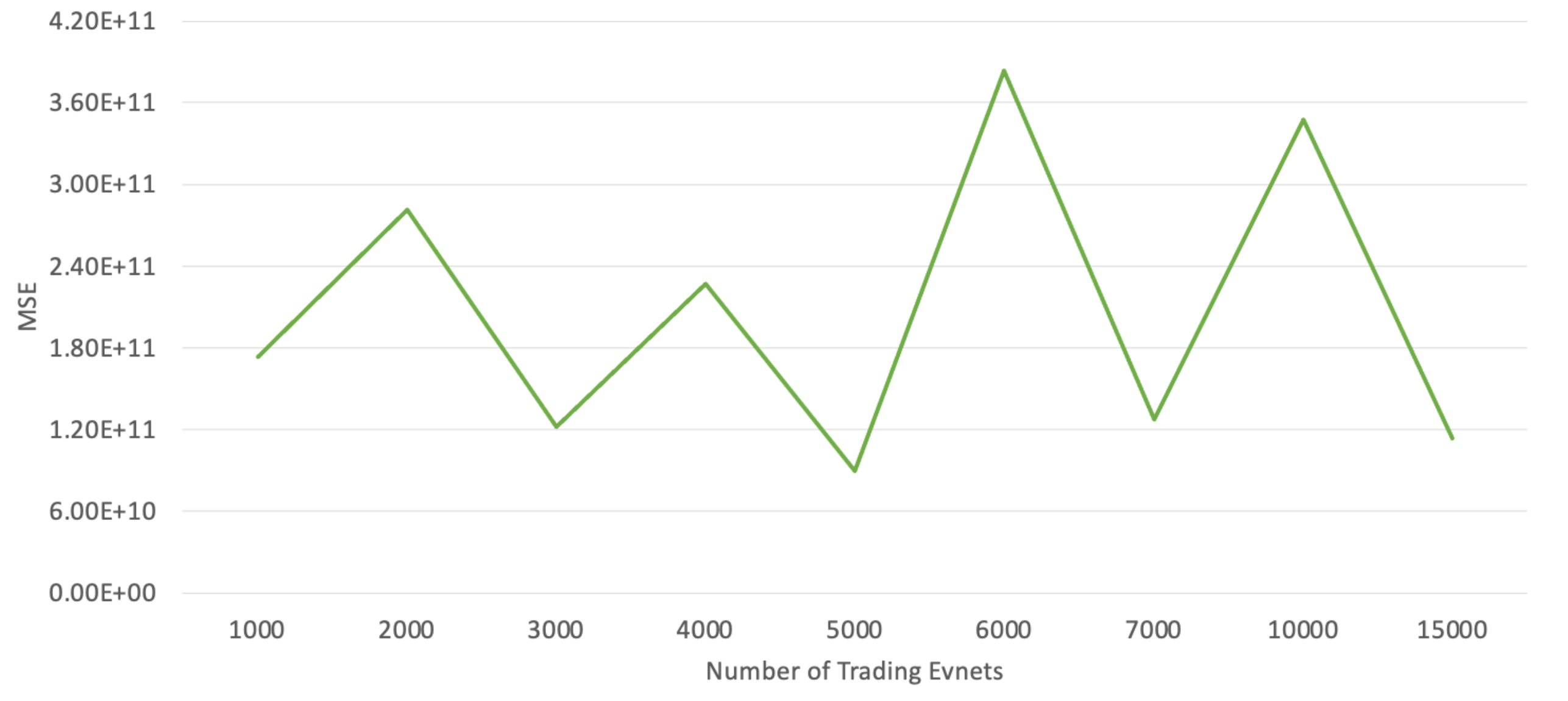}}}  
  \caption{Kesko Long MSE scores based on \hyperref[tab:KeskoLong]{Table \ref{tab:KeskoLong}}.}
  \label{fig:KeskoLong} 
\end{figure*}

\begin{table*}[hbt!]
\centering
\captionsetup{width=.70\textwidth}
\caption{Kesko Benchmark Training (left) and Benchmark Testing (right). Data sample is 50,000 trading events.}
\scalebox{0.60}{
\begin{tabular}{ccrlcccrl}
\cmidrule[2pt]{1-4}\cmidrule[2pt]{6-9}
\textbf{Input} & \textbf{Normalization} & \textbf{Model} & \textbf{MSE - Train} & \qquad & \textbf{Input} & \textbf{Normalization} & \textbf{Model} & \textbf{MSE - Test} \\
\cmidrule{1-4}\cmidrule{6-9}
 LOB Data & Raw & \textbf{OPTM-LSTM} & \textbf{1.28150E+04} & \qquad & LOB Data &  Raw & \textbf{OPTM-LSTM}& \textbf{9.99000E+02}\\   
 &               & LSTM          & 7.00089E+10                & \qquad &       &        & LSTM         & 7.20523E+10           \\ 
 &               & Attention     & 7.00074E+10               & \qquad &       &        & Attention    & 7.20518E+10          \\   
 &               & Bidirectional & 6.97331E+10              & \qquad &       &        & Bidirectional&  7.16165E+10         \\   
 &               & GRU           & 7.00073E+10              & \qquad &       &        & GRU          & 7.20519E+10         \\   
 &               & Hybrid        & 1.62985E+11             & \qquad &       &        & Hybrid       &  1.67990E+11        \\
 &               & Baseline      & 9.41815E+10              & \qquad &       &        & Baseline     &9.44458E+10  \\

\cmidrule{2-4}\cmidrule{7-9}
 & MinMax & \textbf{OPTM-LSTM} & \textbf{2.85920E-05}        & \qquad &       &  MinMax & \textbf{OPTM-LSTM}  & \textbf{9.98282E-05} \\   
 &             & LSTM           &  1.04000E-04               & \qquad &       &         & LSTM                &  1.14617E-04  \\ 
 &             & Attention      &  1.24770E-04               & \qquad &       &         & Attention           &  2.50912E-04 \\   
 &             & Bidirectional  & 1.30000E-04                & \qquad &       &         & Bidirectional       &  1.22839E-04 \\   
 &             & GRU            &  1.36740E-04               & \qquad &       &         & GRU                 &  1.46407E-04 \\   
 &             & Hybrid         &  3.60000E-04               & \qquad &       &         & Hybrid              &  2.65749E-04\\
 &             & Baseline       &   4.35225E-04              & \qquad &       &         & Baseline            &  4.12883E-04\\

\cmidrule{2-4}\cmidrule{7-9}
 & Zscore       & \textbf{OPTM-LSTM}& \textbf{2.13500E-01} & \qquad &         & Zscore & \textbf{OPTM-LSTM}  & \textbf{3.99179E-01} \\   
 &              & LSTM          &1.27620E+00  & \qquad &            &         & LSTM         & 1.00865E+00  \\ 
 &              & Attention     & 1.98440E+00 & \qquad &            &         & Attention    & 1.85445E+00 \\ 
 &              & Bidirectional & 1.96900E+00 & \qquad &            &         & Bidirectional& 1.13340E+00\\   
 &              & GRU           & 2.09650E+00 & \qquad &            &         & GRU          & 1.45136E+00\\   
 &              & Hybrid        & 2.20460E+00 & \qquad &            &         & Hybrid       & 1.32600E+00 \\
 &              & Baseline      & 3.24333E+00 & \qquad &            &         & Baseline     & 3.14158E+00 \\
 \cmidrule{1-4}\cmidrule{6-9}
Mid-price & Raw & \textbf{OPTM-LSTM} & \textbf{1.15897E+05} & \qquad & Mid-price &  Raw & \textbf{OPTM-LSTM}& \textbf{9.80490E+04}\\   
 &               & LSTM          & 6.80624E+07              & \qquad &       &        & LSTM         &  6.80760E+07          \\ 
 &               & Attention     & 6.82317E+07              & \qquad &       &        & Attention    &  6.82315E+07        \\   
 &               & Bidirectional &  6.39776E+07             & \qquad &       &        & Bidirectional&  6.32714E+07        \\   
 &               & GRU           & 6.38784E+07              & \qquad &       &        & GRU          &  6.31658E+07       \\   
 &               & Hybrid        & 1.10363E+08              & \qquad &       &        & Hybrid       &   1.10018E+08           \\
 &               & Persistence   & 7.99125E+07              & \qquad &       &        & Persistence  &  7.90350E+07           \\

\cmidrule{2-4}\cmidrule{7-9}
 & MinMax & \textbf{OPTM-LSTM} & \textbf{1.15280E-05}        & \qquad &       &  MinMax & \textbf{OPTM-LSTM}  & \textbf{1.26656E-05} \\   
 &             & LSTM           &  9.49250E-05               & \qquad &       &         & LSTM                & 9.28870E-05 \\ 
 &             & Attention      &  8.32390E-05               & \qquad &       &         & Attention           &  7.98481E-05\\   
 &             & Bidirectional  &  7.62860E-04               & \qquad &       &         & Bidirectional       & 8.46251E-04 \\   
 &             & GRU            &  7.70360E-04               & \qquad &       &         & GRU                 & 8.89751E-04 \\   
 &             & Hybrid         &  1.43340E-04               & \qquad &       &         & Hybrid              & 2.48957E-04\\
 &             & Persistence    &  7.05425E-04               & \qquad &       &         & Persistence         & 7.74876E-04 \\

\cmidrule{2-4}\cmidrule{7-9}
 & Zscore       & \textbf{OPTM-LSTM}& \textbf{1.05100E-01} & \qquad &         &  Zscore & \textbf{OPTM-LSTM}  & \textbf{1.32550E-01} \\   
 &              & LSTM          & 2.33020E+00 & \qquad &            &         & LSTM         &  1.87150E+00 \\ 
 &              & Attention     & 1.96070E+00  & \qquad &           &         & Attention    & 1.96074E+00 \\ 
 &              & Bidirectional & 1.98860E+00 & \qquad &            &         & Bidirectional& 1.98863E+00\\   
 &              & GRU           & 2.57540E+00 & \qquad &            &         & GRU          & 1.72275E+00 \\   
 &              & Hybrid        & 2.56590E+00 & \qquad &            &         & Hybrid       & 2.68000E+00\\
 &              & Persistence   & 2.68474E+00 & \qquad &            &         & Persistence  & 2.54835E+00\\
\cmidrule[2pt]{1-4}\cmidrule[2pt]{6-9}
\end{tabular}}
\medskip
\label{tab:KeskoBenchmark}
\end{table*}

\begin{table*}[hbt!]
\centering
\captionsetup{width=.70\textwidth}
\caption{Wartsila MSE scores under the Short experimental protocol.}
\scalebox{0.60}{
\begin{tabular}{rcrlcrcrlcrcrl}
\cmidrule[2pt]{1-6}\cmidrule[2pt]{6-9}\cmidrule[2pt]{9-14}
\textbf{Stock} & \textbf{Size} & \textbf{Model} & \textbf{MSE - Train} & \qquad & \textbf{Stock} & \textbf{Size} & \textbf{Model} & \textbf{MSE - Train} & \qquad & \textbf{Stock} & \textbf{Size} & \textbf{Model} & \textbf{MSE - Train}\\
\cmidrule{1-4}\cmidrule{6-9}\cmidrule{11-14}
Wartsila & 1,000  & \textbf{OPTM-LSTM}& \textbf{1.09804E+11} & \qquad &  Wartsila  &  2,000 & \textbf{OPTM-LSTM}    & \textbf{8.39142E+10} & \qquad & Wartsila & 3,000 & \textbf{OPTM-LSTM}       & \textbf{9.66711E+10}\\   
 &             & LSTM          & 1.21563E+11 & \qquad &            &         & LSTM                         & 1.21051E+11     & \qquad &       &        & LSTM          & 1.20788E+11\\ 
 &             & Attention     & 1.21562E+11 & \qquad &            &         & Attention                    & 1.21045E+11      & \qquad &       &        & Attention     & 1.20792E+11\\   
 &             & Bidirectional & 1.21559E+11 & \qquad &            &         & Bidirectional                & 1.21051E+11     & \qquad &       &        & Bidirectional & 1.20793E+11\\   
 &             & GRU           & 1.21562E+11 & \qquad &            &         & GRU                          & 1.21044E+11     & \qquad &       &        & GRU           & 1.20779E+11\\   
 &             & Hybrid        & 1.21395E+12 & \qquad &            &         & Hybrid                       & 1.10232E+11      & \qquad &       &        & Hybrid        & 5.69594E+11\\
\cmidrule{2-4}\cmidrule{7-9}\cmidrule{12-14}
 & 4,000 & \textbf{OPTM-LSTM}& \textbf{5.06770E+10} & \qquad &  &  5,000 & \textbf{OPTM-LSTM} & \textbf{5.37683E+10} & \qquad &  & 6,000 & \textbf{OPTM-LSTM} & \textbf{1.60922E+10}\\   
 &             & LSTM          & 1.20691E+11& \qquad &             &         & LSTM                    & 1.20673E+11& \qquad &       &        & LSTM          & 1.20643E+11\\ 
 &             & Attention     & 1.20690E+11& \qquad &             &         & Attention               & 1.20673E+11& \qquad &       &        & Attention     & 1.20642E+11\\   
 &             & Bidirectional & 1.20676E+11& \qquad &             &         & Bidirectional           & 1.20673E+11& \qquad &       &        & Bidirectional & 1.20642E+11\\   
 &             & GRU           & 1.20674E+11& \qquad &             &         & GRU                     & 1.20654E+11& \qquad &       &        & GRU           & 1.20642E+11\\   
 &             & Hybrid        & 2.28451E+11& \qquad &             &         & Hybrid                  & 1.95861E+11&\qquad &       &        & Hybrid        & 3.97941E+11\\
\cmidrule{2-4}\cmidrule{7-9}\cmidrule{12-14}
  & 7,000 & \textbf{OPTM-LSTM}& \textbf{3.48147E+10} & \qquad &     &  10,000 & \textbf{OPTM-LSTM}  & \textbf{3.92906E+09} & \qquad &  & 15,000 & \textbf{OPTM-LSTM} & \textbf{5.63909E+07}\\   
 &              & LSTM          & 1.19545E+11 & \qquad &            &         & LSTM         & 1.20258E+11 & \qquad &        &        & LSTM          & 1.20148E+11\\ 
 &              & Attention     & 1.19504E+11 & \qquad &            &         & Attention    & 1.20258E+11 & \qquad &        &        & Attention     & 1.20096E+11\\   
 &              & Bidirectional & 1.19511E+11 & \qquad &            &         & Bidirectional& 1.20257E+11 & \qquad &        &        & Bidirectional & 1.20151E+11\\   
 &              & GRU           & 1.19507E+11 & \qquad &            &         & GRU          & 1.20209E+11 & \qquad &        &        & GRU           & 1.20151E+11\\   
 &              & Hybrid        & 1.79659E+11 & \qquad &             &         & Hybrid       & 1.35232E+11  & \qquad &        &        & Hybrid & 4.92045E+11\\
\cmidrule{2-4}\cmidrule{7-9}\cmidrule{12-14}
 & 20,000 & \textbf{OPTM-LSTM} & \textbf{4.79600E+03} & \qquad &      &  35,000 & \textbf{OPTM-LSTM} & \textbf{3.12200E+03} & \qquad &  & 50,000 & \textbf{OPTM-LSTM} & \textbf{2.68100E+03}\\   
 &              & LSTM          & 1.19779E+11 & \qquad &            &         & LSTM         & 1.18918E+11 & \qquad &        &        & LSTM          & 1.18318E+11\\ 
 &              & Attention     & 1.19783E+11 & \qquad &             &         & Attention    & 1.18639E+11 & \qquad &        &        & Attention     & 1.18008E+11\\   
 &              & Bidirectional & 1.19663E+11 & \qquad &             &         & Bidirectional& 1.18918E+11 & \qquad &        &        & Bidirectional & 1.18325E+11\\   
 &              & GRU           & 1.19783E+11 & \qquad &            &         & GRU          & 1.18707E+11 & \qquad &        &        & GRU           & 1.18325E+11\\   
 &              & Hybrid        & 1.95015E+11  & \qquad &           &         & Hybrid       & 1.31175E+12 & \qquad & &               & Hybrid        & 4.00368E+11\\
\cmidrule{2-4}\cmidrule{7-9}\cmidrule{12-14}
  & 100,000 & \textbf{OPTM-LSTM}& \textbf{1.75200E+03} & \qquad &     &  400,000 & \textbf{OPTM-LSTM} & \textbf{3.02000E+03} & \qquad &  & 800,000 & \textbf{OPTM-LSTM} & \textbf{1.92900E+04}\\   
 &                & LSTM          & 1.17305E+11& \qquad &            &         & LSTM         & 1.21255E+11 & \qquad &         &        & LSTM          & 1.22637E+11\\ 
 &                & Attention     & 1.15883E+11& \qquad &            &         & Attention    & 1.21319E+11 & \qquad &         &        & Attention     & 1.22882E+11\\   
 &                & Bidirectional & 1.17321E+11  & \qquad &            &         & Bidirectional  & 8.40267E+10 & \qquad &        &        & Bidirectional & 1.22881E+11\\   
 &                & GRU           & 1.16074E+11& \qquad &            &         & GRU          & 8.43822E+10 & \qquad &         &        & GRU           & 1.22881E+11\\   
 &                & Hybrid        & 8.94838E+11& \qquad &            &         & Hybrid       & 4.94263E+11 & \qquad &         &        & Hybrid        & 2.17551E+11\\
\cmidrule{2-4}\cmidrule{7-9}\cmidrule{12-14}
 & 1,000,000 & \textbf{OPTM-LSTM}       & \textbf{2.01330E+04} & \qquad &   &  2,000,000 & \textbf{OPTM-LSTM}  & \textbf{2.07880E+04} & \qquad &  &  &  & \\   
 &                  & LSTM         & 1.29683E+11 & \qquad &             &         & LSTM    & 1.48478E+11     & \qquad &        &        &          & \\ 
 &                  & Attention    & 1.25959E+11 & \qquad &             &         & Attention& 1.27544E+11    & \qquad &        &        &      & \\   
 &                  & Bidirectional& 1.22556E+11 & \qquad &             &         & Bidirectional& 1.21888E+11& \qquad &        &        &  & \\   
 &                  & GRU          & 1.22669E+11 & \qquad &              &         & GRU          & 1.21567E+11& \qquad &        &        &            & \\   
 &                  & Hybrid       & 4.88503E+11 & \qquad &             &         & Hybrid       & 2.50929E+11& \qquad &        &        &         & \\
\cmidrule[2pt]{1-6}\cmidrule[2pt]{6-9}\cmidrule[2pt]{9-14}
\textbf{Stock} & \textbf{Size} & \textbf{Model} & \textbf{MSE - Test} & \qquad & \textbf{Stock} & \textbf{Size} & \textbf{Model} & \textbf{MSE - Test} & \qquad & \textbf{Stock} & \textbf{Size} & \textbf{Model} & \textbf{MSE - Test}\\
\cmidrule{1-4}\cmidrule{6-9}\cmidrule{11-14}
 Wartsila & 1,000 & \textbf{OPTM-LSTM} & \textbf{1.04203E+11} & \qquad & Wartsila & 2,000& \textbf{OPTM-LSTM}& \textbf{6.88044E+10}  & \qquad & Wartsila & 3,000 & \textbf{OPTM-LSTM}& \textbf{8.27156E+10}\\   
 &             & LSTM          & 1.20541E+11           & \qquad &       &        & LSTM         & 1.20273E+11            & \qquad &       &        & LSTM          & 1.20388E+11\\ 
 &             & Attention     & 1.20540E+11           & \qquad &       &        & Attention    & 1.20265E+11            & \qquad &       &        & Attention     & 1.20389E+11\\   
 &             & Bidirectional & 1.20536E+11           & \qquad &       &        & Bidirectional& 1.20272E+11            & \qquad &       &        & Bidirectional & 1.20374E+11\\   
 &             & GRU           & 1.20541E+11           & \qquad &       &        & GRU          & 1.20263E+11            & \qquad &       &        & GRU           & 1.20374E+11\\   
 &             & Hybrid        & 1.50009E+12              & \qquad &       &        & Hybrid     & 6.71980E+11               & \qquad &       &        & Hybrid     & 5.68991E+11\\
\cmidrule{2-4}\cmidrule{7-9}\cmidrule{12-14}
 & 4,000 & \textbf{OPTM-LSTM} & \textbf{2.17475E+10} & \qquad &        &  5,000  & \textbf{OPTM-LSTM}  & \textbf{2.19227E+10} & \qquad &  & 6,000 & \textbf{OPTM-LSTM} & \textbf{1.64676E+08}\\   
 &             & LSTM           & 1.20608E+11& \qquad &                       &         & LSTM         & 1.20489E+11 & \qquad &          &        & LSTM          & 1.20262E+11\\ 
 &             & Attention      & 1.20606E+11& \qquad &                       &         & Attention    & 1.20491E+11 & \qquad &          &        & Attention     & 1.20262E+11\\   
 &             & Bidirectional  & 1.20588E+11& \qquad &                       &         & Bidirectional& 1.20489E+11 & \qquad &          &        & Bidirectional & 1.20262E+11\\   
 &             & GRU            & 1.20586E+11& \qquad &                       &         & GRU          & 1.20464E+11 & \qquad &          &        & GRU           & 1.20262E+11\\   
 &             & Hybrid         & 2.28250E+11 & \qquad &                       &         & Hybrid      & 1.95808E+11 & \qquad &          &        & Hybrid        & 3.96597E+11\\
\cmidrule{2-4}\cmidrule{7-9}\cmidrule{12-14}
 & 7,000 & \textbf{OPTM-LSTM}& \textbf{1.69590E+09} & \qquad &      &  10,000 & \textbf{OPTM-LSTM}  & \textbf{2.17000E+02} & \qquad &  & 15,000 & \textbf{OPTM-LSTM} & \textbf{2.15600E+03}\\   
 &              & LSTM          & 1.19545E+11 & \qquad &            &         & LSTM         & 1.19716E+11 & \qquad &        &                           & LSTM   & 1.19533E+11\\ 
 &              & Attention     & 1.19504E+11 & \qquad &            &         & Attention    & 1.19716E+11 & \qquad &        &                           & Attention & 1.19454E+11\\ 
 &              & Bidirectional & 1.19511E+11 & \qquad &            &         & Bidirectional& 1.19716E+11 & \qquad &        &                           & Bidirectional & 1.19532E+11\\   
 &              & GRU           & 1.19507E+11 & \qquad &            &         & GRU          & 1.19648E+11 & \qquad &        &                           & GRU           & 1.19533E+11\\   
 &              & Hybrid        & 1.78198E+11 & \qquad &            &         & Hybrid       & 1.34849E+11 & \qquad &        &                           & Hybrid & 4.89492E+11\\
\cmidrule{2-4}\cmidrule{7-9}\cmidrule{12-14}
 & 20,000 & \textbf{OPTM-LSTM}& \textbf{1.72960E+04} & \qquad &     &  35,000 & \textbf{OPTM-LSTM}  & \textbf{3.03000E+02} & \qquad &  & 50,000 & \textbf{OPTM-LSTM} & \textbf{2.08000E+02}\\   
 &              & LSTM          & 1.17931E+11 & \qquad &            &         & LSTM         & 1.16948E+11 & \qquad &        &        & LSTM          & 1.17274E+11\\ 
 &              & Attention     & 1.17933E+11 & \qquad &            &         & Attention    & 1.16533E+11 & \qquad &        &        & Attention     & 1.16784E+11\\   
 &              & Bidirectional & 1.17762E+11 & \qquad &            &         & Bidirectional& 1.16951E+11 & \qquad &        &        & Bidirectional & 1.17281E+11\\   
 &              & GRU           & 1.17935E+11 & \qquad &            &         & GRU          & 1.16633E+11 & \qquad &        &        & GRU           & 1.17281E+11\\   
 &              & Hybrid        & 1.92199E+11 & \qquad &            &         & Hybrid       & 1.29029E+12 & \qquad &        &        & Hybrid        & 3.97123E+11\\
\cmidrule{2-4}\cmidrule{7-9}\cmidrule{12-14}
 & 100,000 & \textbf{OPTM-LSTM}& \textbf{2.44200E+03} & \qquad &  &  400,000 & \textbf{OPTM-LSTM}  & \textbf{1.13400E+03} & \qquad &  & 800,000 & \textbf{OPTM-LSTM} & \textbf{1.26000E+02}\\   
 &                & LSTM          & 1.16278E+11& \qquad &           &         & LSTM         & 1.30016E+11  & \qquad &         &        & LSTM          & 1.17921E+11\\ 
 &                & Attention     & 1.13813E+11& \qquad &           &         & Attention    & 1.30117E+11 & \qquad &         &        & Attention     & 1.18251E+11\\   
 &                & Bidirectional & 1.16301E+11& \qquad &           &         & Bidirectional& 5.85243E+10 & \qquad &         &        & Bidirectional & 1.18250E+11\\   
 &                & GRU           & 1.16301E+11& \qquad &           &         & GRU          & 5.93584E+10 & \qquad &         &        & GRU           & 1.18251E+11\\   
 &                & Hybrid        & 8.87207E+11 & \qquad &           &         & Hybrid      & 5.30482E+11 & \qquad &         &        & Hybrid        & 2.10445E+11\\
\cmidrule{2-4}\cmidrule{7-9}\cmidrule{12-14}
 & 1,000,000 & \textbf{OPTM-LSTM} & \textbf{1.15000E+02} & \qquad &  &  2,000,000 & \textbf{OPTM-LSTM} & \textbf{1.01000E+02}  & \qquad &  &  &  & \\   
 &                  & LSTM         & 1.16884E+11 & \qquad &            &         & LSTM            & 1.17877E+11       & \qquad &        &            &     & \\ 
 &                  & Attention    & 1.18254E+11 & \qquad &            &         & Attention       & 1.20885E+11       & \qquad &        &            &     & \\   
 &                  & Bidirectional& 1.17566E+11 & \qquad &            &         & Bidirectional   & 1.20701E+11       & \qquad &        &            &     & \\   
 &                  & GRU          & 1.15771E+11 & \qquad &            &         & GRU             & 1.17881E+11       & \qquad &        &            &     & \\   
 &                  & Hybrid       & 3.89878E+11 & \qquad &            &         & Hybrid          & 1.33446E+11       & \qquad &        &            &     & \\
\cmidrule[2pt]{1-6}\cmidrule[2pt]{6-9}\cmidrule[2pt]{9-14}
\end{tabular}}
\medskip
\label{tab:WartsilaShort}
\end{table*}

\begin{figure*}[hbt!]
    \centering
  \subfloat[OPTM-LSTM training MSE scores \label{1a}]{%
       \scalebox{0.53}{\includegraphics[width=0.65\linewidth]{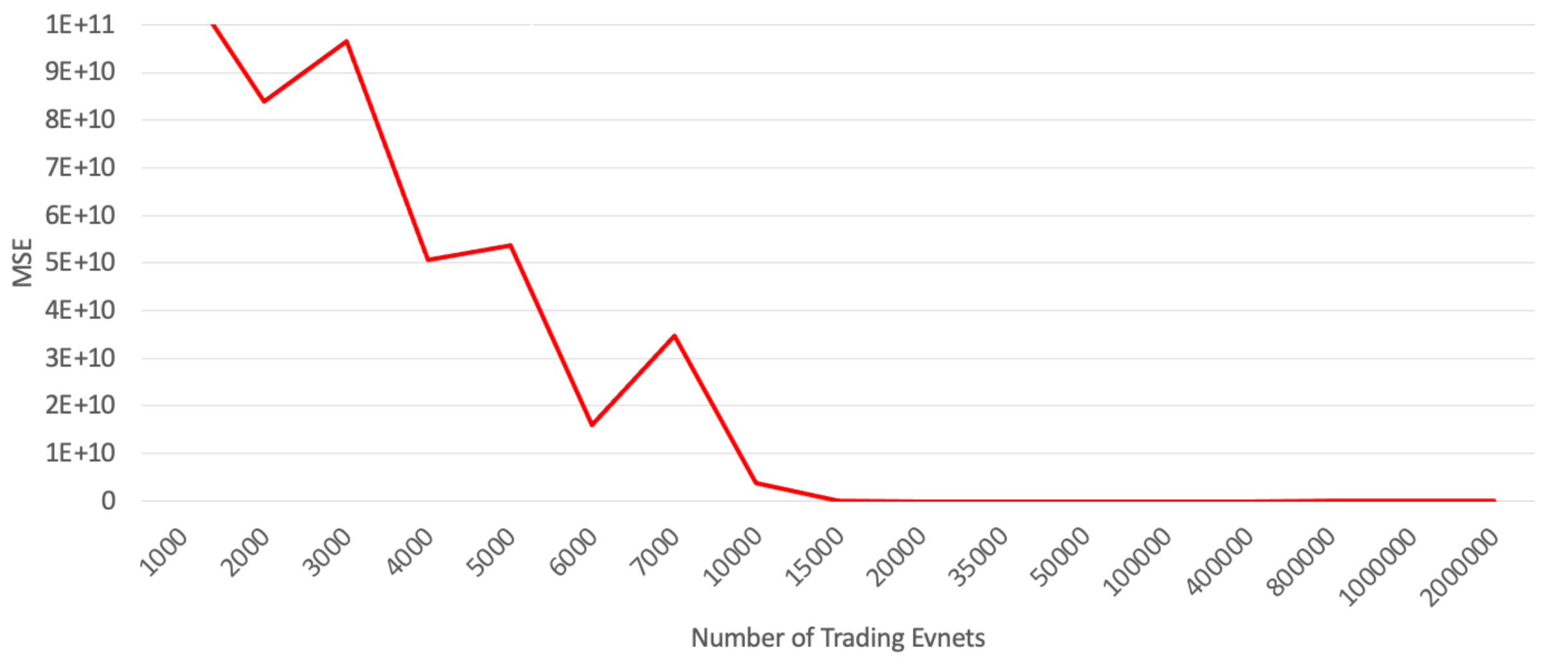}}}
  \subfloat[OPTM-LSTM testing MSE scores \label{1b}]{%
        \scalebox{0.53}{\includegraphics[width=0.65\linewidth]{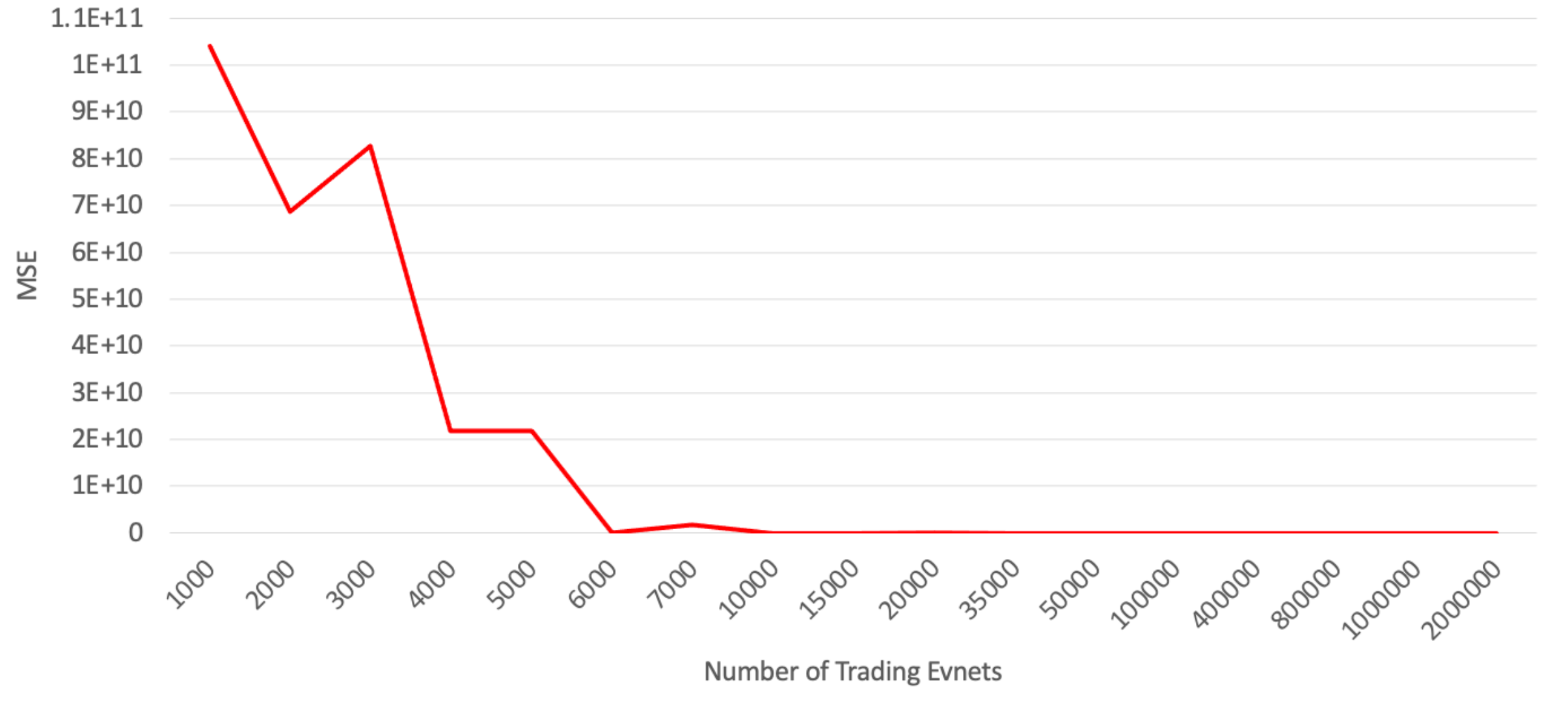}}}
    \\
  \subfloat[LSTM training MSE scores \label{1c}]{%
        \scalebox{0.53}{\includegraphics[width=0.65\linewidth]{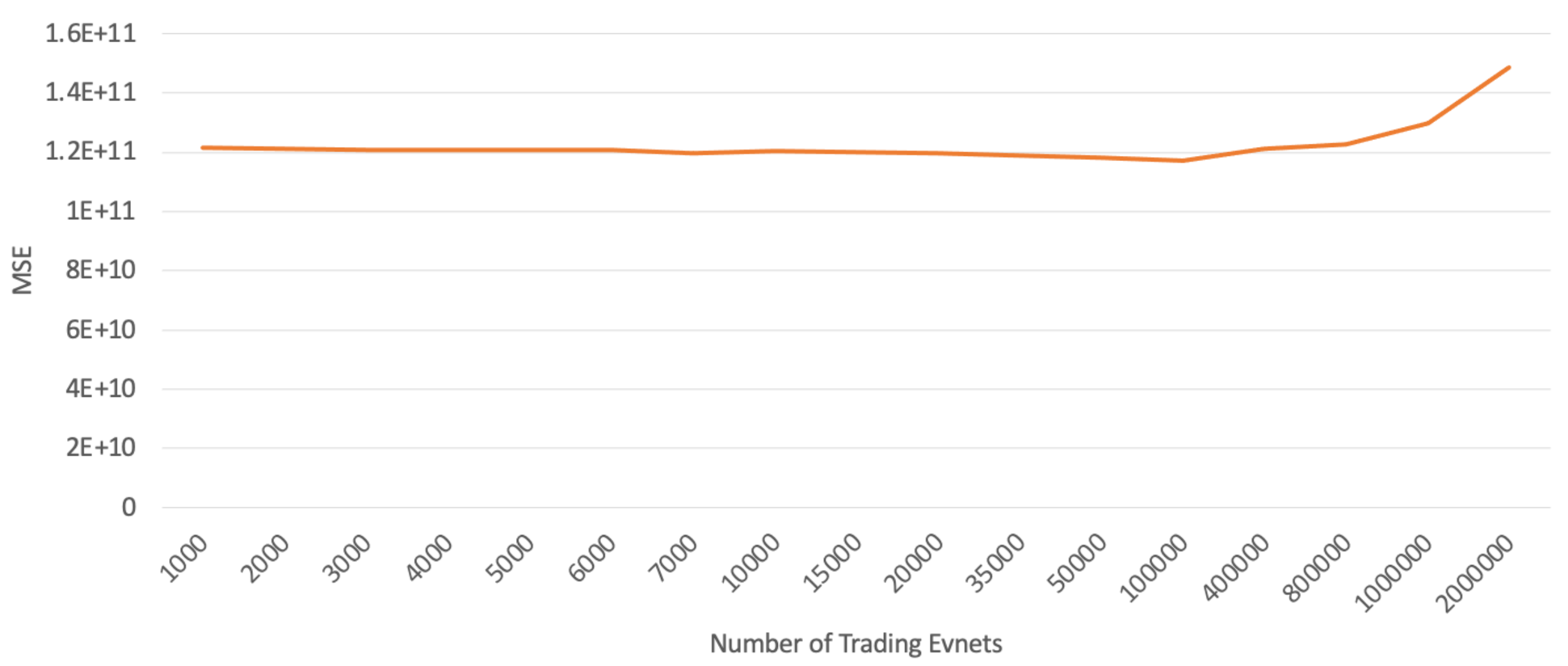}}}
  \subfloat[LSTM testing MSE scores \label{1d}]{%
        \scalebox{0.53}{\includegraphics[width=0.65\linewidth]{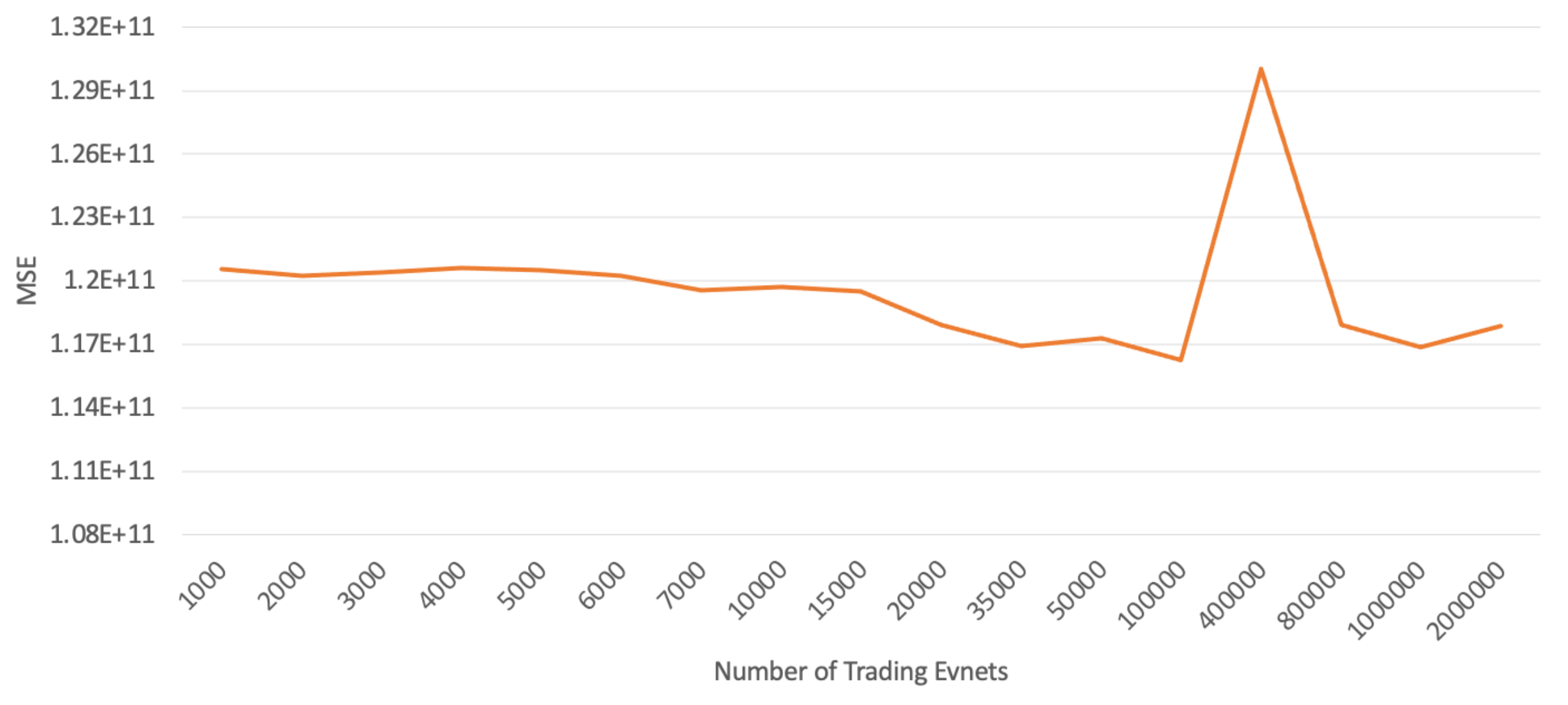}}}
    \\
  \subfloat[Attention LSTM training MSE scores\label{1c}]{%
        \scalebox{0.53}{\includegraphics[width=0.65\linewidth]{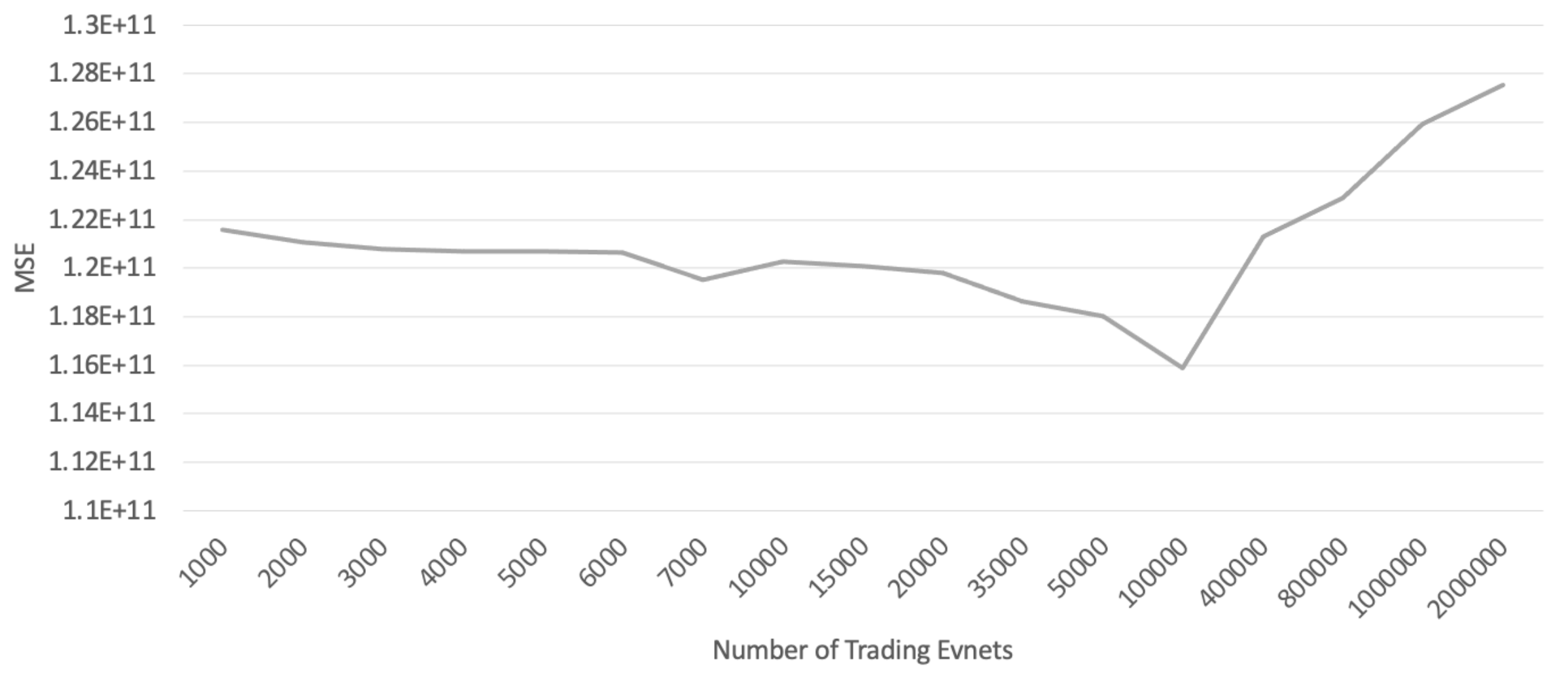}}}
  \subfloat[Attention LSTM testing MSE scores \label{1d}]{%
        \scalebox{0.53}{\includegraphics[width=0.65\linewidth]{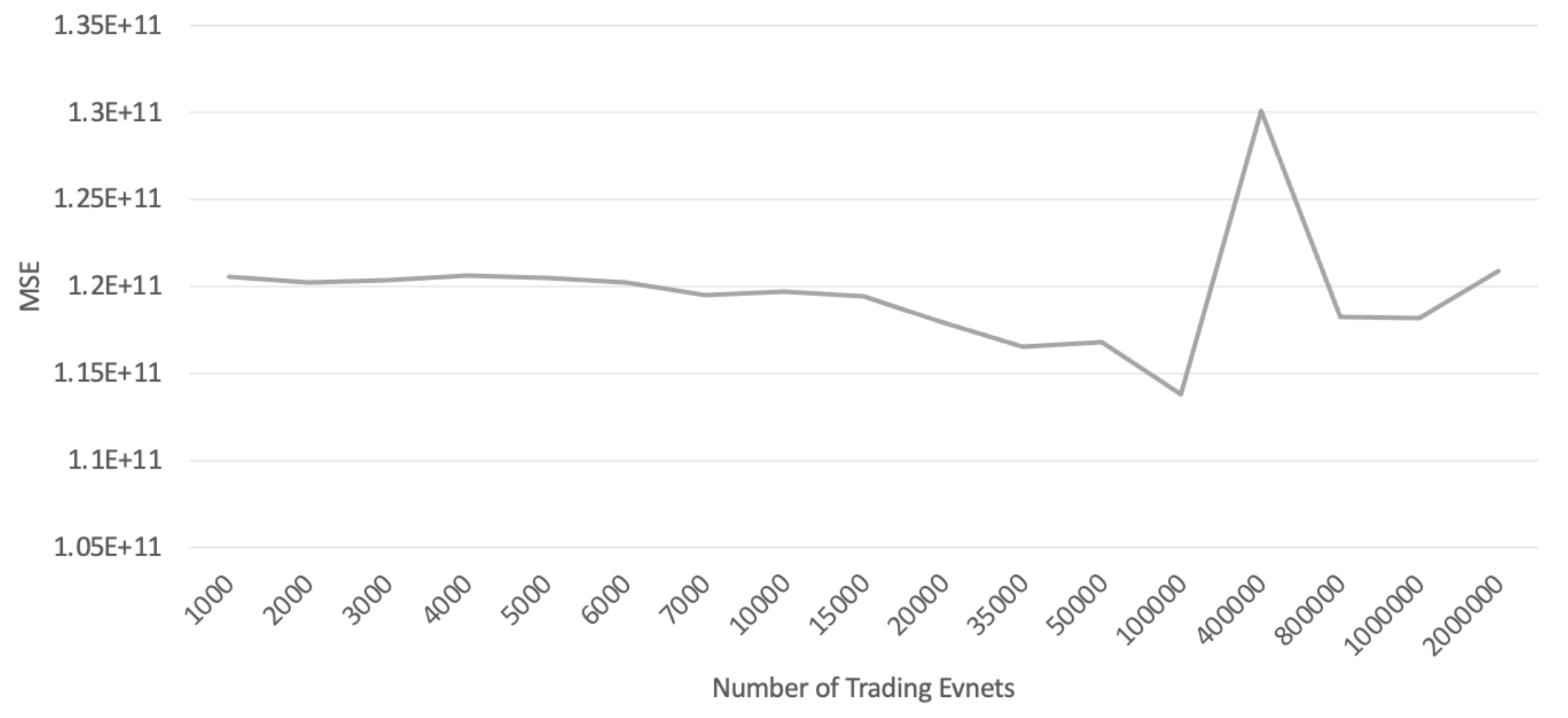}}}  
    \\
  \subfloat[Bidirectional training MSE scores \label{1a}]{%
       \scalebox{0.53}{\includegraphics[width=0.65\linewidth]{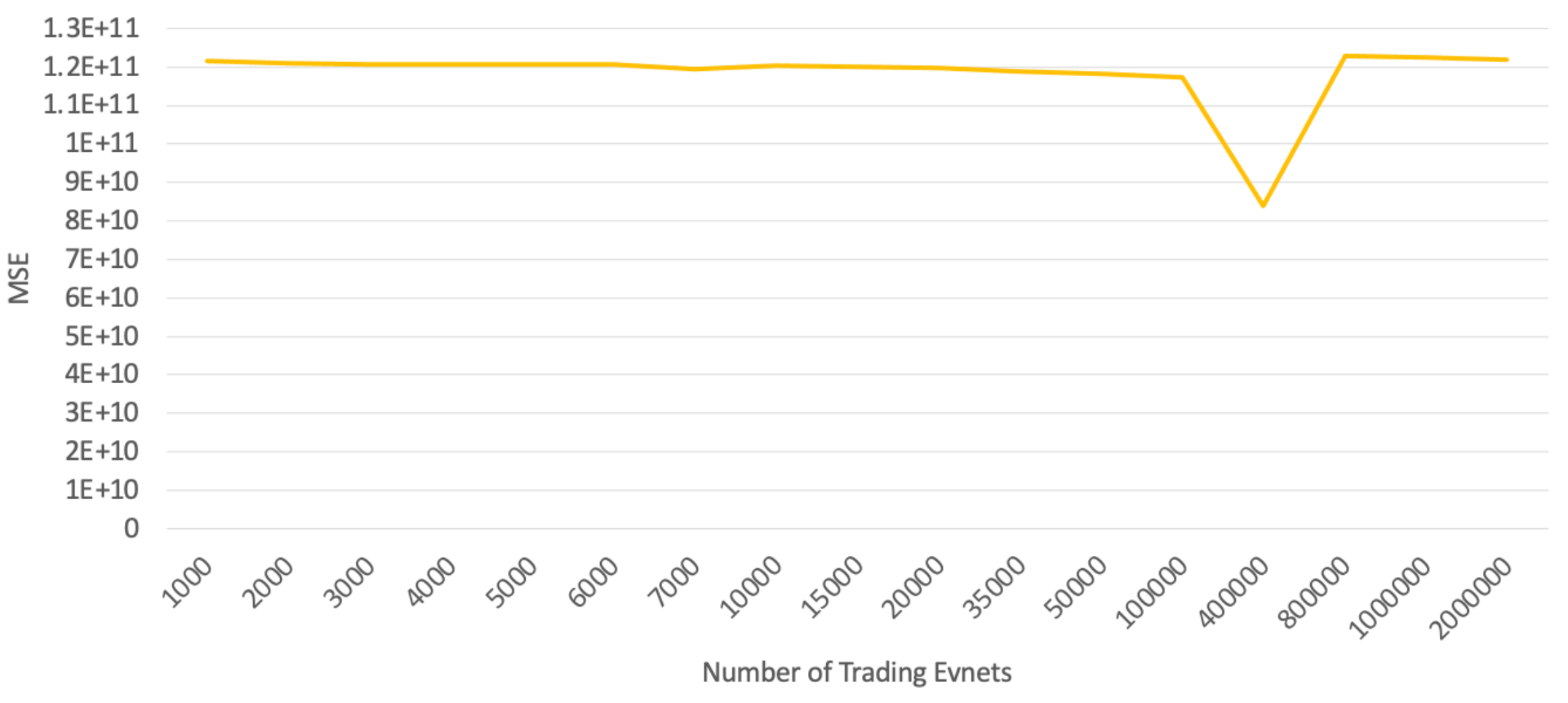}}}
  \subfloat[Bidirectional testing MSE scores \label{1b}]{%
        \scalebox{0.53}{\includegraphics[width=0.65\linewidth]{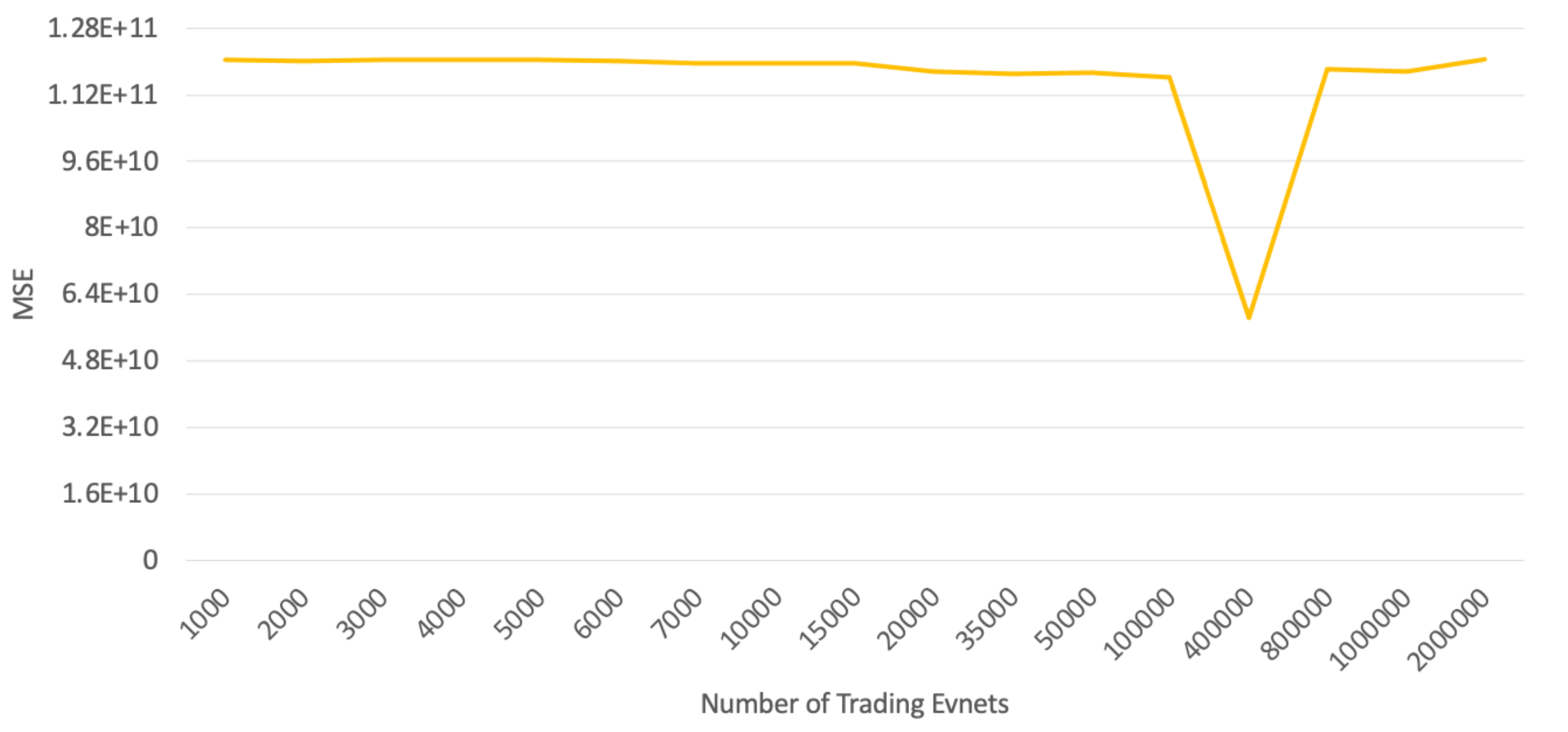}}}
    \\
  \subfloat[GRU training MSE scores \label{1a}]{%
       \scalebox{0.53}{\includegraphics[width=0.65\linewidth]{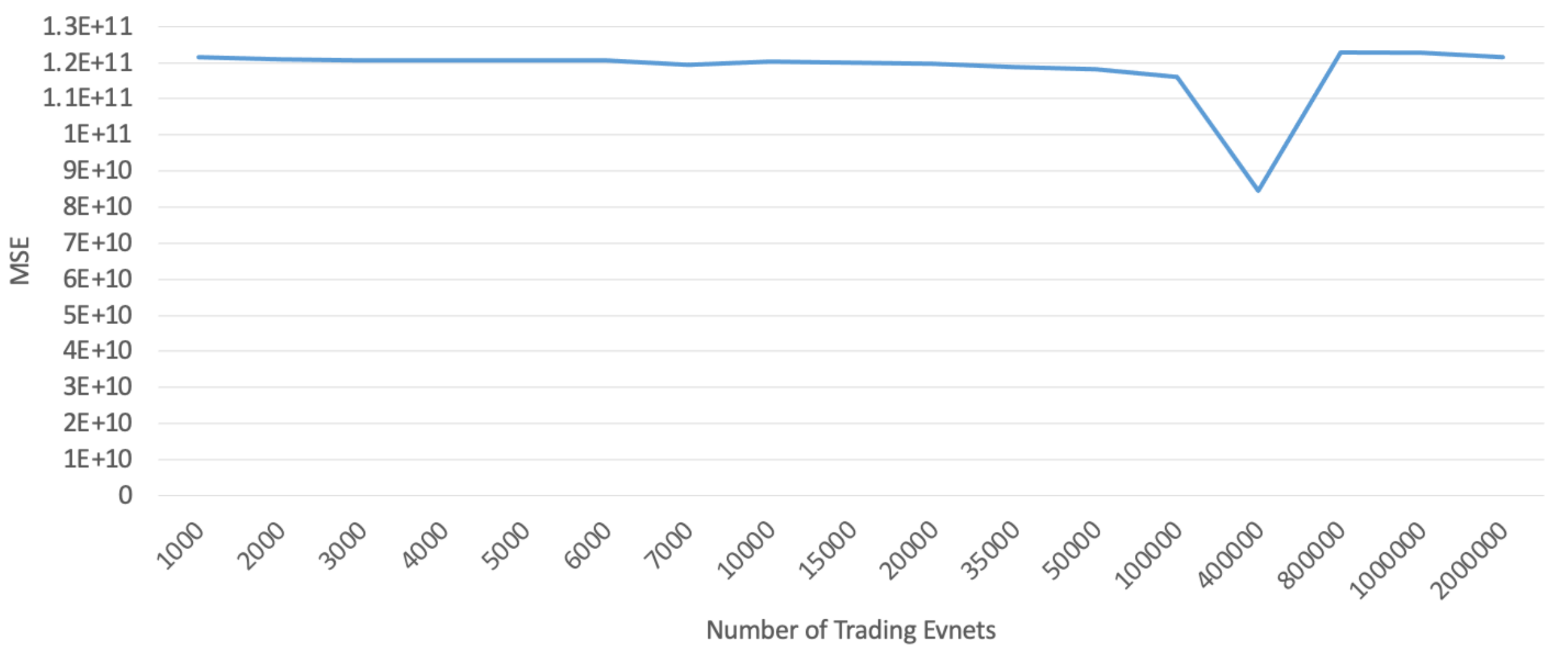}}}
  \subfloat[GRU testing MSE scores \label{1b}]{%
        \scalebox{0.53}{\includegraphics[width=0.65\linewidth]{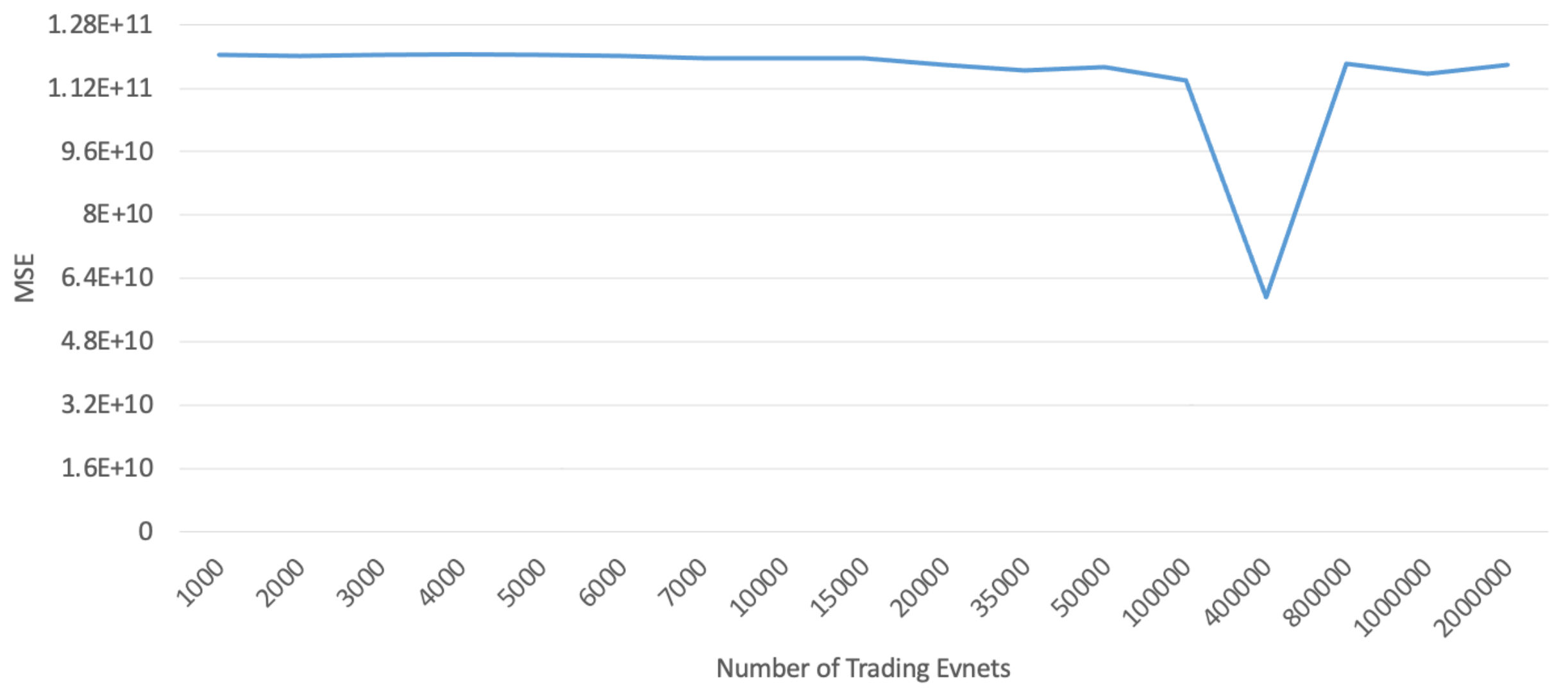}}}
    \\
  \subfloat[Hybrid training MSE scores \label{1a}]{%
       \scalebox{0.53}{\includegraphics[width=0.65\linewidth]{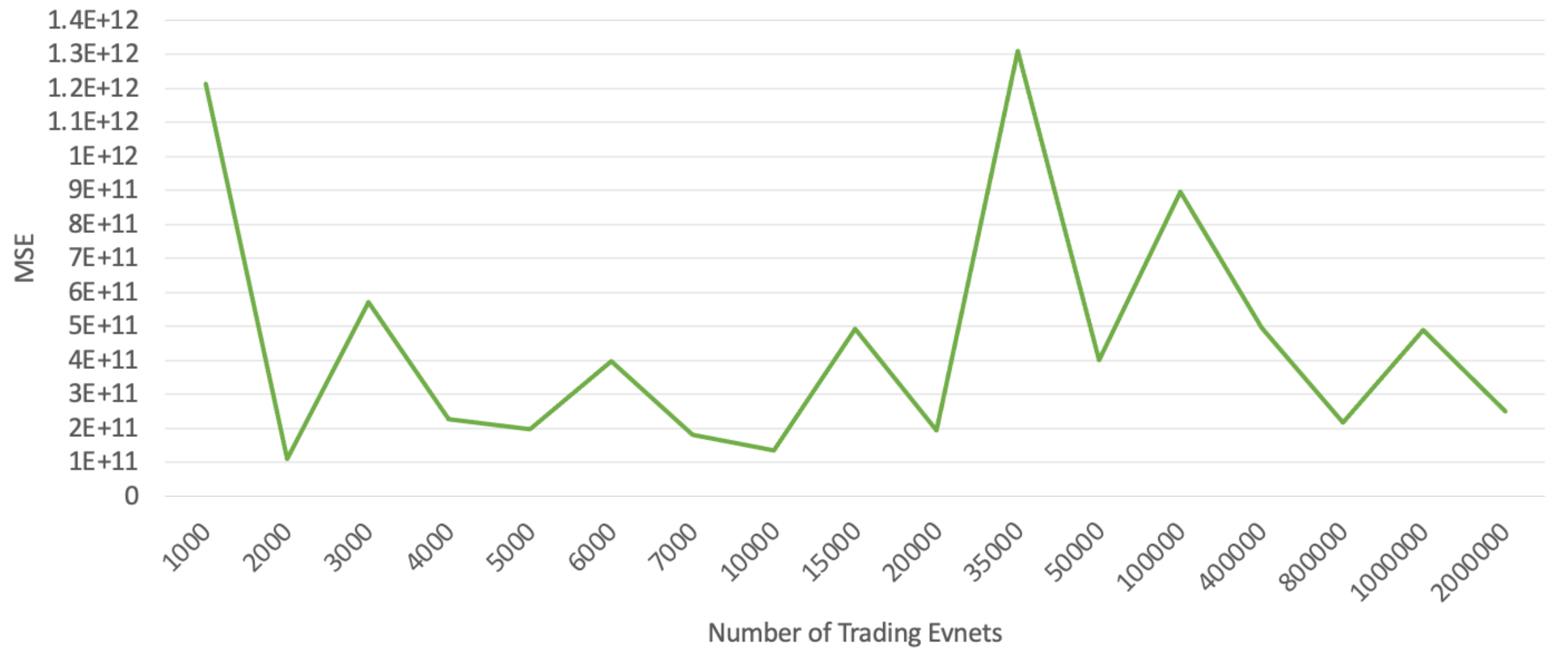}}}
  \subfloat[Hybrid testing MSE scores \label{1b}]{%
        \scalebox{0.53}{\includegraphics[width=0.65\linewidth]{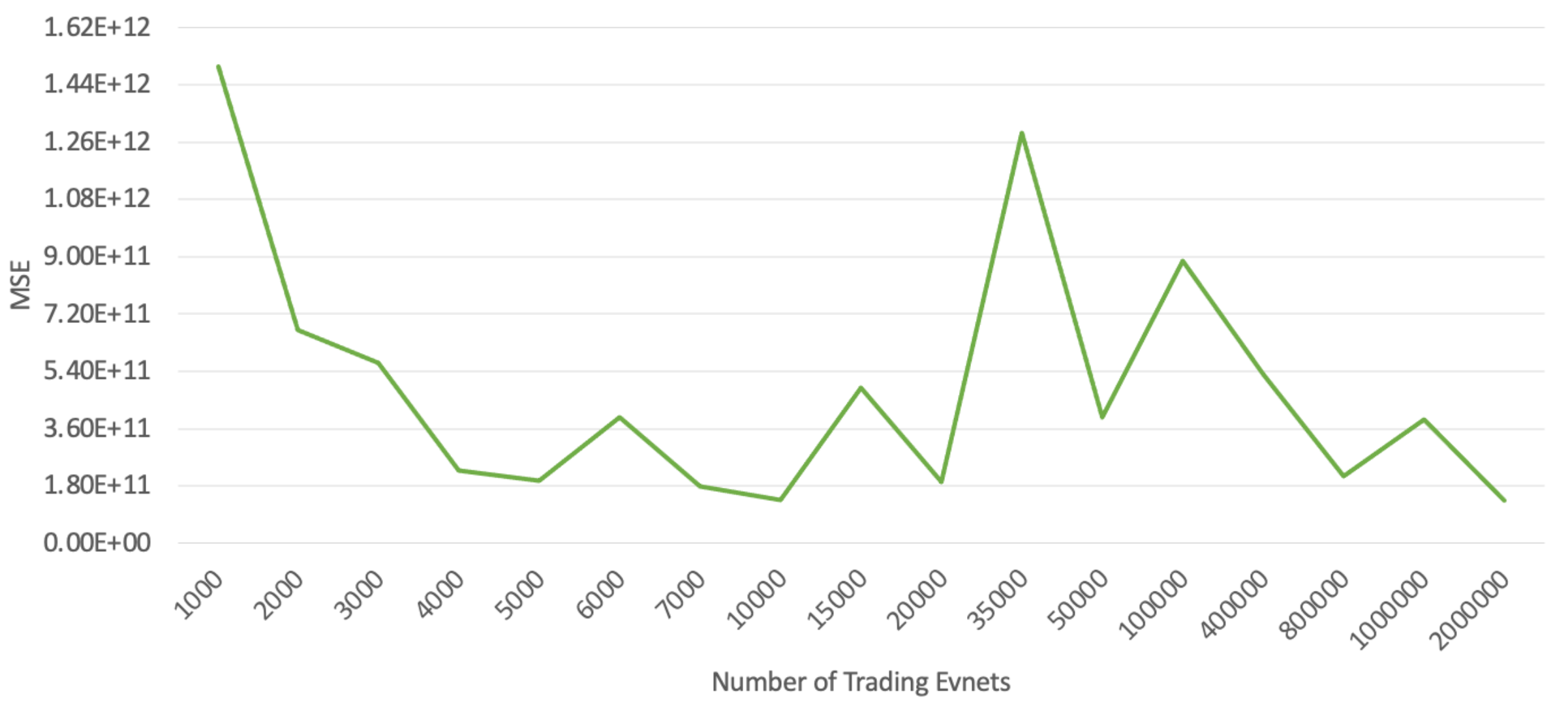}}}  
  \caption{Wartsila Short MSE scores based on \hyperref[tab:WartsilaShort]{Table \ref{tab:WartsilaShort}}.}
  \label{fig:WartsilaShort} 
\end{figure*}

\begin{table*}[hbt!]
\centering
\captionsetup{width=.70\textwidth}
\caption{Wartsila MSE scores under the Long experimental protocol.}
\scalebox{0.60}{
\begin{tabular}{rcrlcrcrlcrcrl}
\cmidrule[2pt]{1-6}\cmidrule[2pt]{6-9}\cmidrule[2pt]{9-14}
\textbf{Stock} & \textbf{Size} & \textbf{Model} & \textbf{MSE - Train} & \qquad & \textbf{Stock} & \textbf{Size} & \textbf{Model} & \textbf{MSE - Train} & \qquad & \textbf{Stock} & \textbf{Size} & \textbf{Model} & \textbf{MSE - Train}\\
\cmidrule{1-4}\cmidrule{6-9}\cmidrule{11-14}
 Wartsila & 1,000 & \textbf{OPTM-LSTM} & \textbf{5.67870E+04} & \qquad & Wartsila &  2,000 & \textbf{OPTM-LSTM}& \textbf{3.08010E+04}  & \qquad & Wartsila & 3,000 & \textbf{OPTM-LSTM}& \textbf{2.41600E+04}\\   
 &             & LSTM          & 1.21562E+11           & \qquad &       &        & LSTM         & 1.20695E+11           & \qquad &       &        & LSTM          & 1.20276E+11\\ 
 &             & Attention     & 1.21443E+11           & \qquad &       &        & Attention    & 1.21050E+11           & \qquad &       &        & Attention     & 1.19978E+11\\   
 &             & Bidirectional & 1.21562E+11           & \qquad &       &        & Bidirectional& 1.21055E+11           & \qquad &       &        & Bidirectional & 1.19986E+11\\   
 &             & GRU           & 1.21562E+11           & \qquad &       &        & GRU          & 1.21051E+11           & \qquad &       &        & GRU           & 1.20798E+11\\   
 &             & Hybrid        & 1.46884E+11           & \qquad &       &        & Hybrid       & 4.05923E+11           & \qquad &       &        & Hybrid        & 1.57073E+11\\
\cmidrule{2-4}\cmidrule{7-9}\cmidrule{12-14}
 & 4,000 & \textbf{OPTM-LSTM} & \textbf{1.90670E+04} & \qquad &        &  5,000  & \textbf{OPTM-LSTM}  & \textbf{1.66770E+04} & \qquad &  & 6,000 & \textbf{OPTM-LSTM} & \textbf{1.15560E+04}\\   
 &             & LSTM           & 1.19263E+11& \qquad &                       &         & LSTM         & 1.19089E+11 & \qquad &          &        & LSTM          & 1.20608E+11\\ 
 &             & Attention      & 1.20693E+11& \qquad &                       &         & Attention    & 1.20672E+11 & \qquad &          &        & Attention     & 1.17067E+11\\   
 &             & Bidirectional  & 1.20695E+11& \qquad &                       &         & Bidirectional& 1.18731E+11 & \qquad &          &        & Bidirectional & 1.17356E+11\\   
 &             & GRU            & 1.19233E+11& \qquad &                       &         & GRU          & 1.20673E+11 & \qquad &          &        & GRU           & 1.17483E+11\\   
 &             & Hybrid         & 1.61632E+11 & \qquad &                       &         & Hybrid      & 4.01191E+11 & \qquad &          &        & Hybrid        & 7.68799E+11\\
\cmidrule{2-4}\cmidrule{7-9}\cmidrule{12-14}
 & 7,000 & \textbf{OPTM-LSTM}& \textbf{1.08690E+04} & \qquad &      &  10,000 & \textbf{OPTM-LSTM}  & \textbf{8.25900E+03} & \qquad &  & 15,000 & \textbf{OPTM-LSTM} & \textbf{9.63900E+03}\\   
 &              & LSTM          & 1.20589E+11 & \qquad &            &         & LSTM         & 1.20257E+11 & \qquad &        &                           & LSTM   & 1.20087E+11\\ 
 &              & Attention     & 1.16692E+11 & \qquad &            &         & Attention    & 1.20258E+11 & \qquad &        &                           & Attention & 1.20151E+11\\ 
 &              & Bidirectional & 1.15956E+11 & \qquad &            &         & Bidirectional& 1.08053E+11 & \qquad &        &                           & Bidirectional & 9.66286E+10\\   
 &              & GRU           & 1.16061E+11 & \qquad &            &         & GRU          & 1.20258E+11 & \qquad &        &                           & GRU           & 1.20151E+11\\   
 &              & Hybrid        & 4.69414E+11 & \qquad &            &         & Hybrid       & 2.20184E+11 & \qquad &        &                           & Hybrid & 2.94234E+11\\
\cmidrule[2pt]{1-6}\cmidrule[2pt]{6-9}\cmidrule[2pt]{9-14}
\textbf{Stock} & \textbf{Size} & \textbf{Model} & \textbf{MSE - Test} & \qquad & \textbf{Stock} & \textbf{Size} & \textbf{Model} & \textbf{MSE - Test} & \qquad & \textbf{Stock} & \textbf{Size} & \textbf{Model} & \textbf{MSE - Test}\\
\cmidrule{1-4}\cmidrule{6-9}\cmidrule{11-14}
 Wartsila & 1,000 & \textbf{OPTM-LSTM} & \textbf{3.01000E+02} & \qquad & Wartsila &  2,000 & \textbf{OPTM-LSTM}& \textbf{7.29000E+02}  & \qquad & Wartsila & 3,000 & \textbf{OPTM-LSTM}& \textbf{1.98800E+03}\\   
 &             & LSTM          & 1.20539E+11           & \qquad &       &        & LSTM         & 1.19912E+11            & \qquad &       &        & LSTM          & 1.19869E+11\\ 
 &             & Attention     & 1.20419E+11           & \qquad &       &        & Attention    & 1.20272E+11            & \qquad &       &        & Attention     & 1.19560E+11\\   
 &             & Bidirectional & 1.20541E+11           & \qquad &       &        & Bidirectional& 1.20272E+11            & \qquad &       &        & Bidirectional & 1.19569E+11\\   
 &             & GRU           & 1.20541E+11           & \qquad &       &        & GRU          & 1.20272E+11            & \qquad &       &        & GRU           & 1.20389E+11\\   
 &             & Hybrid        & 1.46654E+11           & \qquad &       &        & Hybrid       & 4.04110E+11            & \qquad &       &        & Hybrid        & 1.56225E+11\\
\cmidrule{2-4}\cmidrule{7-9}\cmidrule{12-14}
 & 4,000 & \textbf{OPTM-LSTM} & \textbf{5.51000E+02} & \qquad &        &  5,000  & \textbf{OPTM-LSTM}  & \textbf{5.68000E+02} & \qquad & & 6,000 & \textbf{OPTM-LSTM} & \textbf{3.11260E+04}\\   
 &             & LSTM           & 1.19163E+11& \qquad &                       &         & LSTM         & 1.18888E+11 & \qquad &          &        & LSTM          & 1.20227E+11\\ 
 &             & Attention      & 1.20606E+11& \qquad &                       &         & Attention    & 1.20489E+11 & \qquad &          &        & Attention     & 1.16579E+11\\   
 &             & Bidirectional  & 1.20606E+11& \qquad &                       &         & Bidirectional& 1.18492E+11 & \qquad &          &        & Bidirectional & 1.16879E+11\\   
 &             & GRU            & 1.19112E+11& \qquad &                       &         & GRU          & 1.20491E+11 & \qquad &          &        & GRU           & 1.17007E+11\\   
 &             & Hybrid         & 1.61739E+11 & \qquad &                       &         & Hybrid      & 4.00412E+11 & \qquad &          &        & Hybrid        & 7.66015E+11\\
\cmidrule{2-4}\cmidrule{7-9}\cmidrule{12-14}
 & 7,000  & \textbf{OPTM-LSTM}& \textbf{2.17700E+03} & \qquad &     &  10,000 & \textbf{OPTM-LSTM}  & \textbf{5.62000E+02} & \qquad &  & 15,000 & \textbf{OPTM-LSTM} & \textbf{4.02000E+02}\\   
 &              & LSTM          & 1.19545E+11 & \qquad &            &         & LSTM         & 1.19716E+11 & \qquad &        &                           & LSTM   & 1.19469E+11\\ 
 &              & Attention     & 1.15535E+11 & \qquad &            &         & Attention    & 1.19716E+11 & \qquad &        &                           & Attention & 1.19532E+11\\ 
 &              & Bidirectional & 1.14774E+11 & \qquad &            &         & Bidirectional& 1.07105E+11 & \qquad &        &                           & Bidirectional & 9.52940E+10\\   
 &              & GRU           & 1.14889E+11 & \qquad &            &         & GRU          & 1.19716E+11 & \qquad &        &                           & GRU           & 1.19533E+11\\   
 &              & Hybrid        & 4.65431E+11 & \qquad &            &         & Hybrid       & 2.19444E+11 & \qquad &        &                           & Hybrid        & 2.92892E+11\\
\cmidrule[2pt]{1-6}\cmidrule[2pt]{6-9}\cmidrule[2pt]{9-14}
\end{tabular}}
\medskip
\label{tab:WartsilaLong}
\end{table*}

\begin{figure*}[hbt!]
    \centering
  \subfloat[OPTM-LSTM training MSE scores \label{1a}]{%
       \scalebox{0.53}{\includegraphics[width=0.65\linewidth]{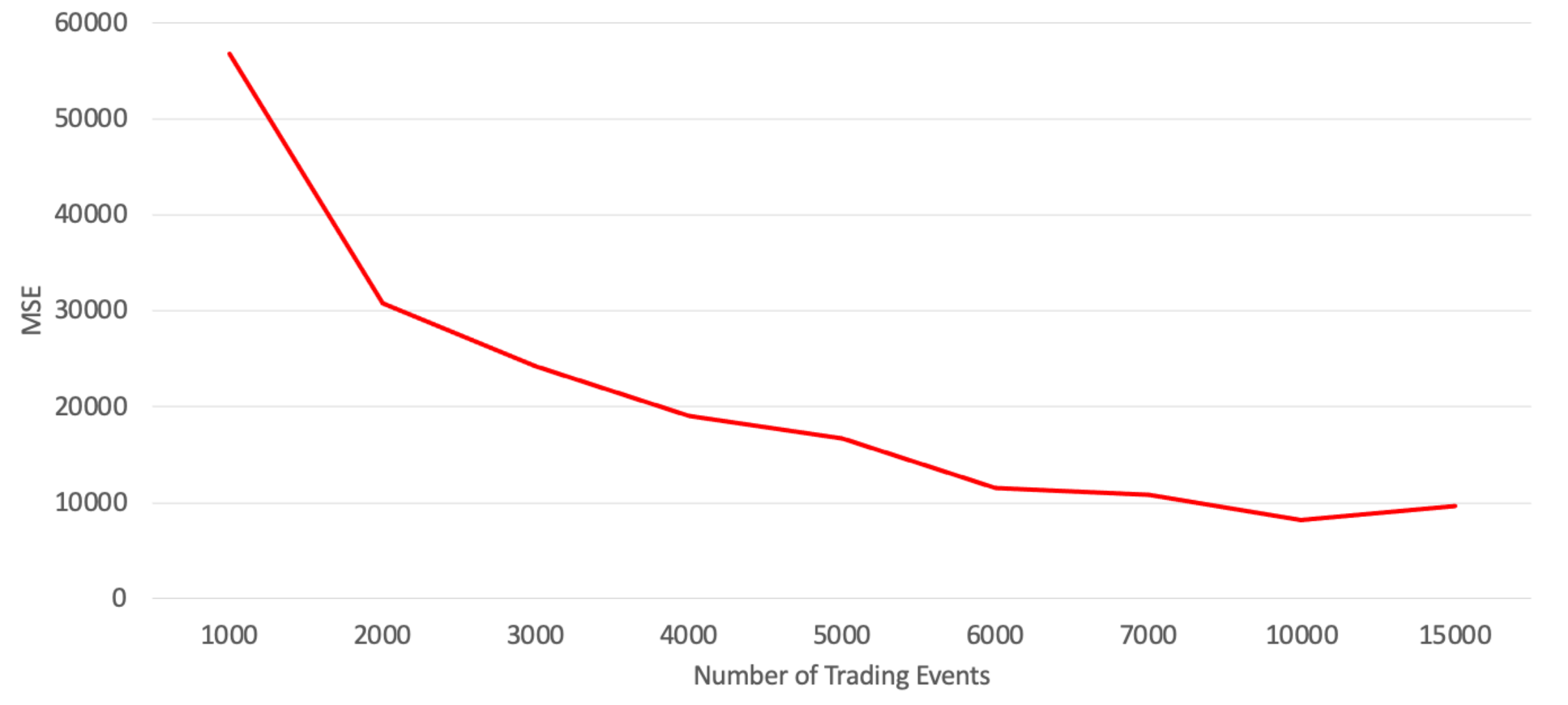}}}
  \subfloat[OPTM-LSTM testing MSE scores \label{1b}]{%
        \scalebox{0.53}{\includegraphics[width=0.65\linewidth]{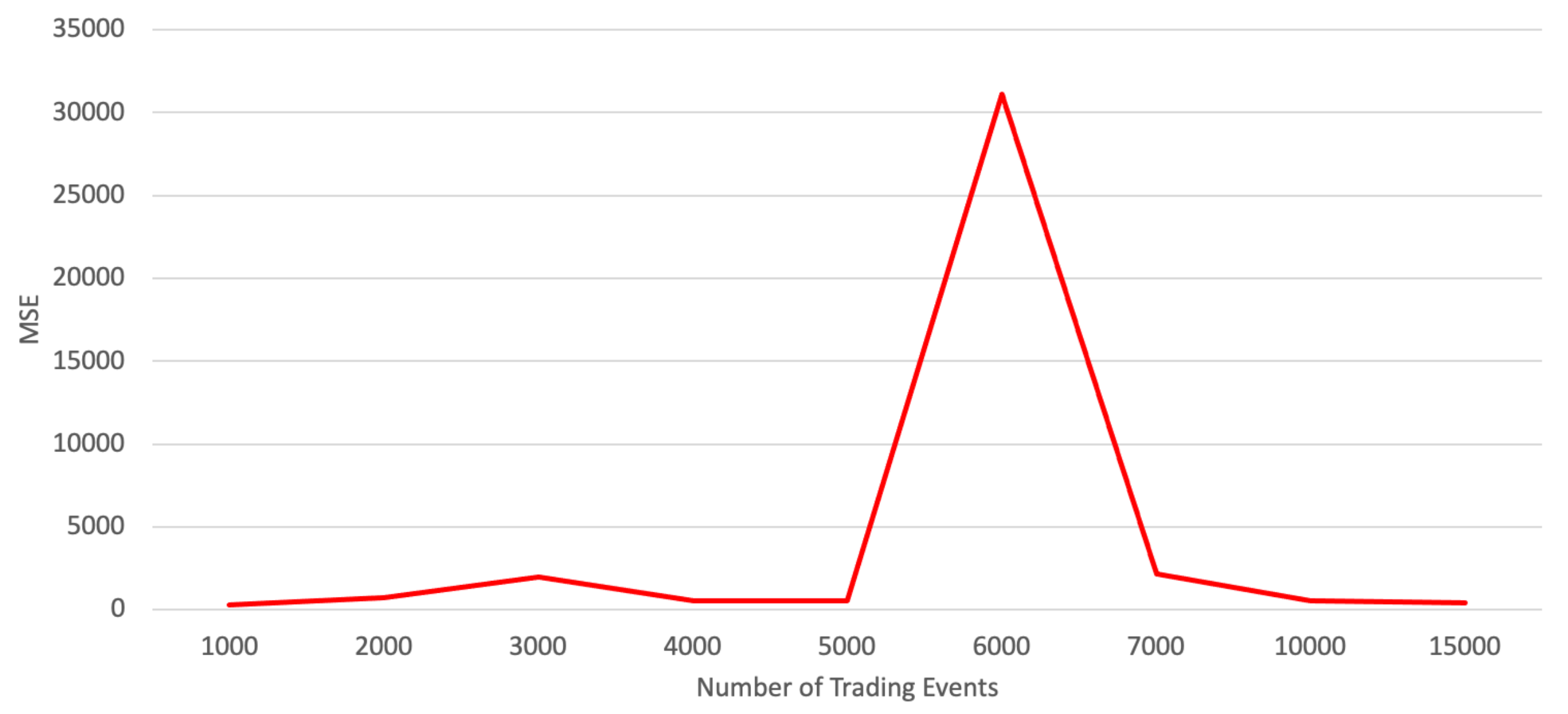}}}
    \\
  \subfloat[LSTM training MSE scores \label{1c}]{%
        \scalebox{0.53}{\includegraphics[width=0.65\linewidth]{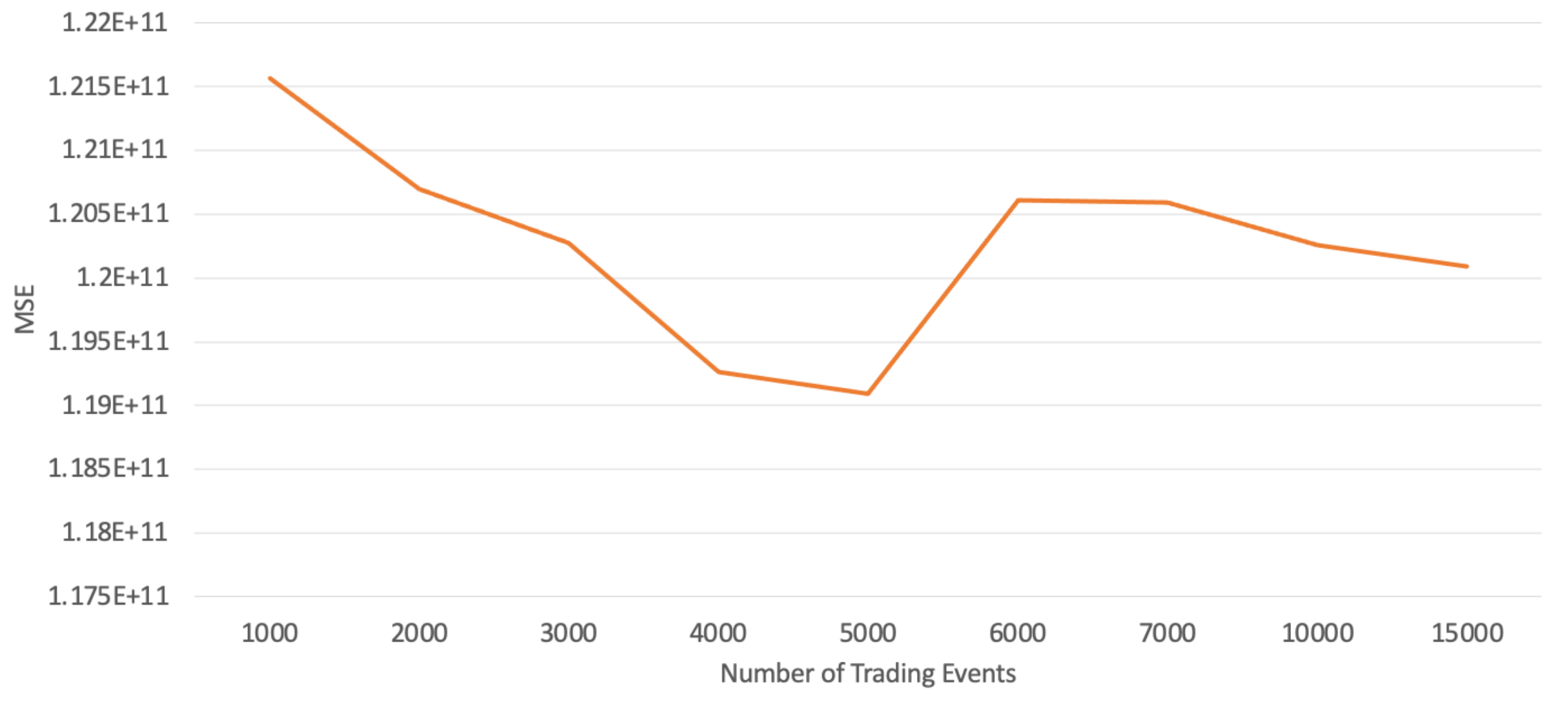}}}
  \subfloat[LSTM testing MSE scores \label{1d}]{%
        \scalebox{0.53}{\includegraphics[width=0.65\linewidth]{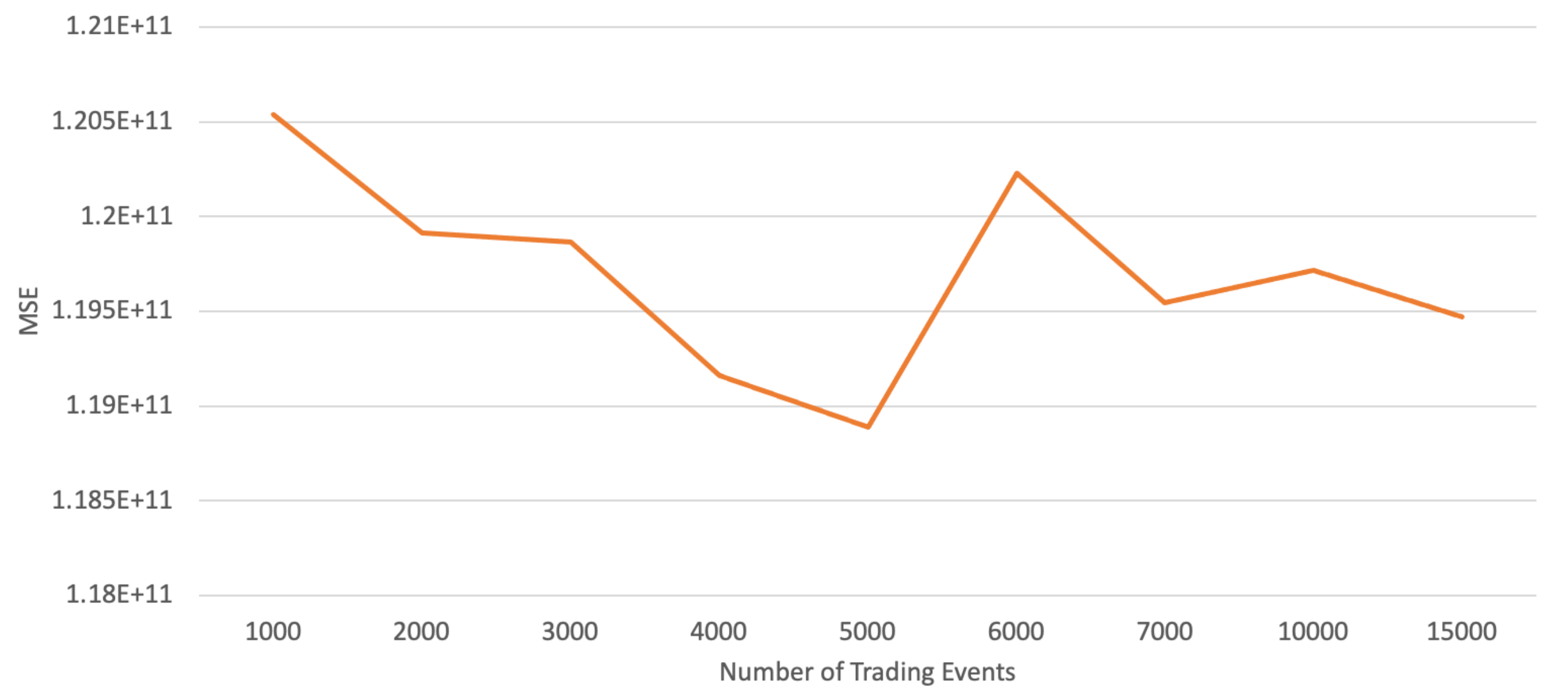}}}
    \\
  \subfloat[Attention LSTM training MSE scores\label{1c}]{%
        \scalebox{0.53}{\includegraphics[width=0.65\linewidth]{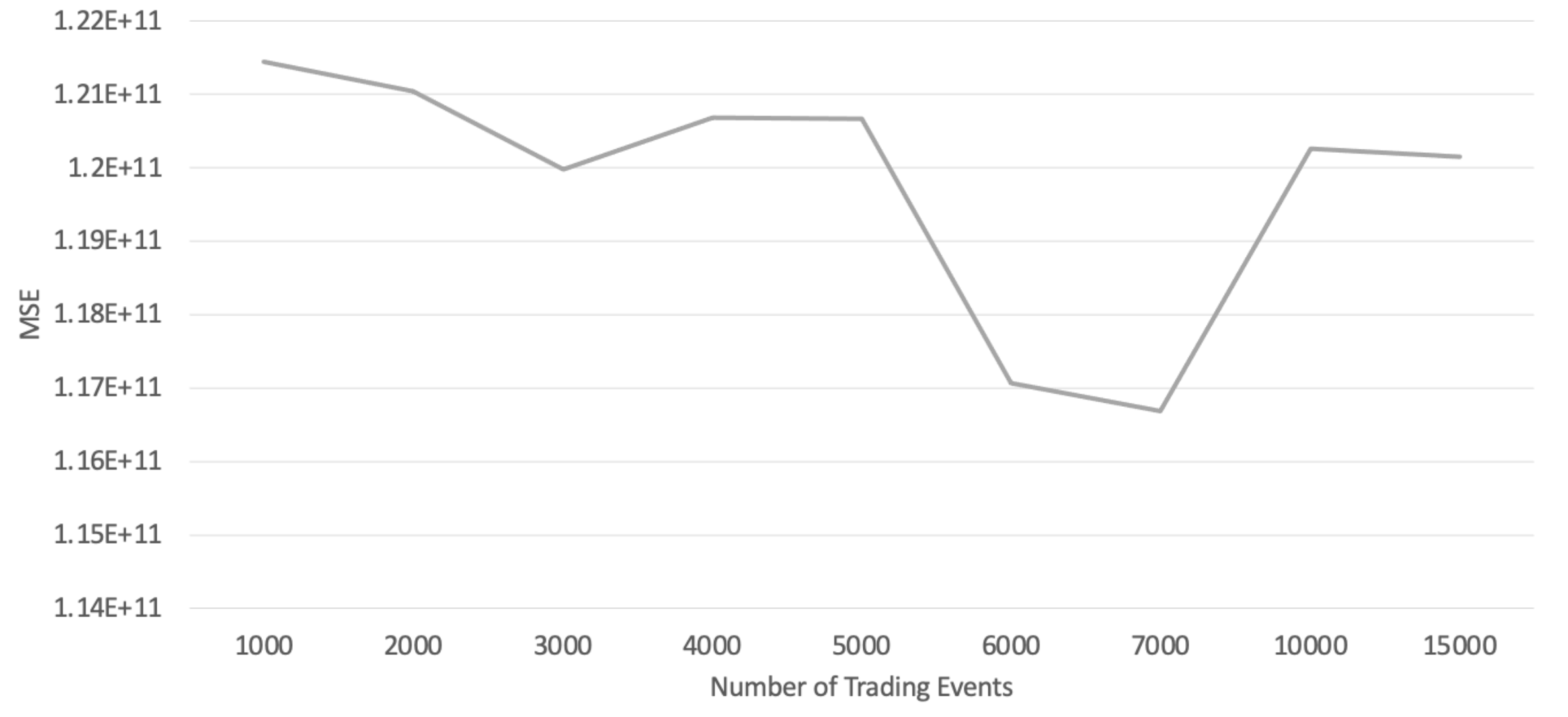}}}
  \subfloat[Attention LSTM testing MSE scores \label{1d}]{%
        \scalebox{0.53}{\includegraphics[width=0.65\linewidth]{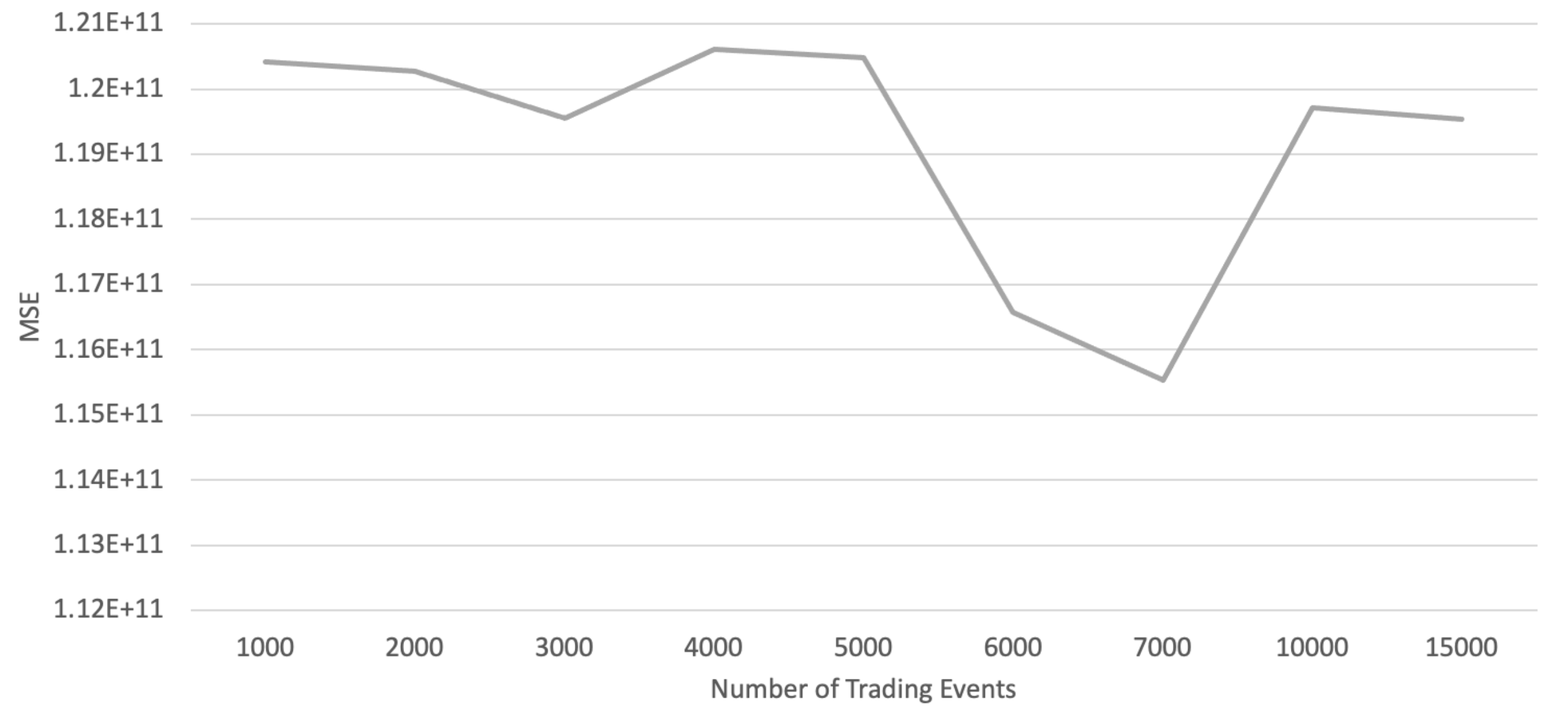}}}  
    \\
  \subfloat[Bidirectional training MSE scores \label{1a}]{%
       \scalebox{0.53}{\includegraphics[width=0.65\linewidth]{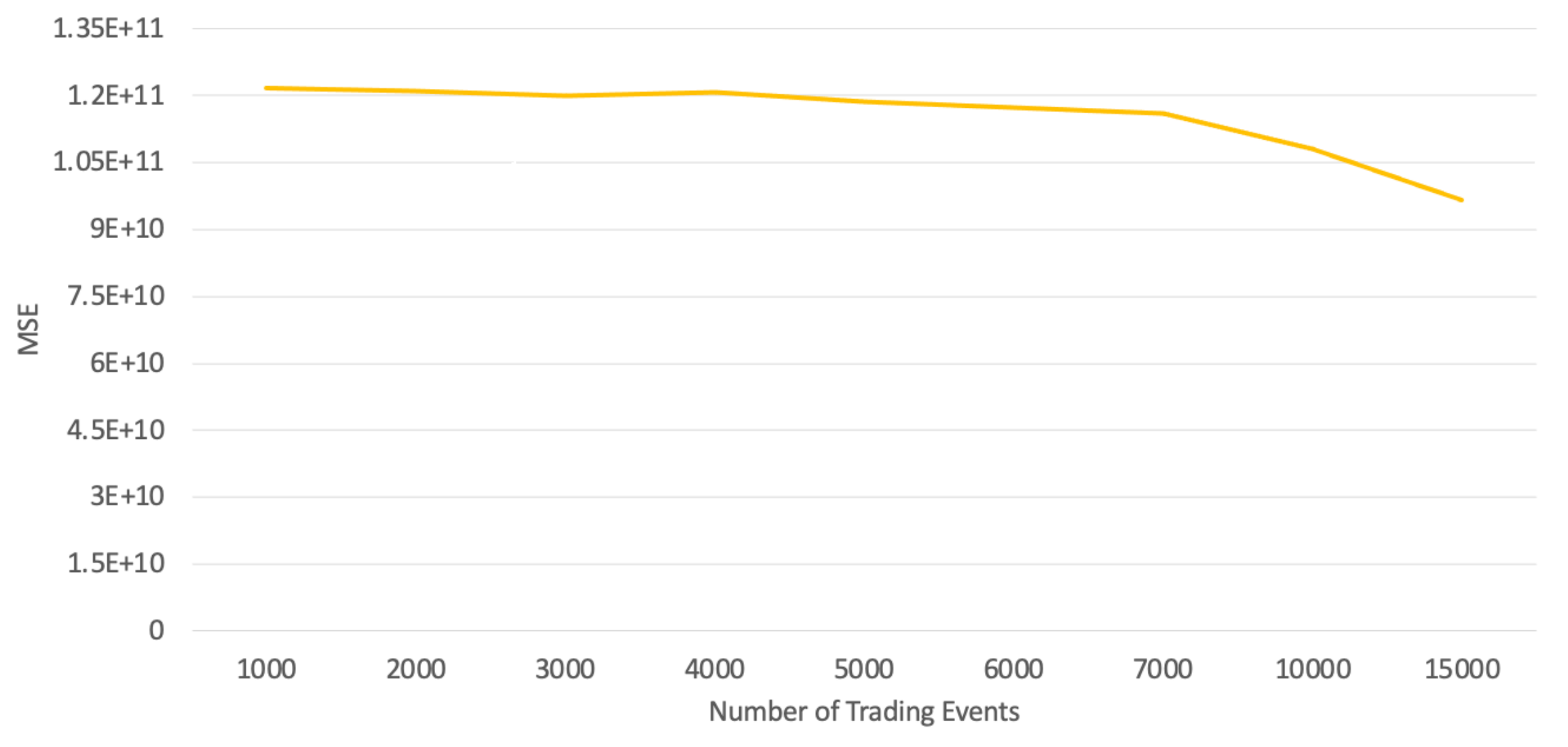}}}
  \subfloat[Bidirectional testing MSE scores \label{1b}]{%
        \scalebox{0.53}{\includegraphics[width=0.65\linewidth]{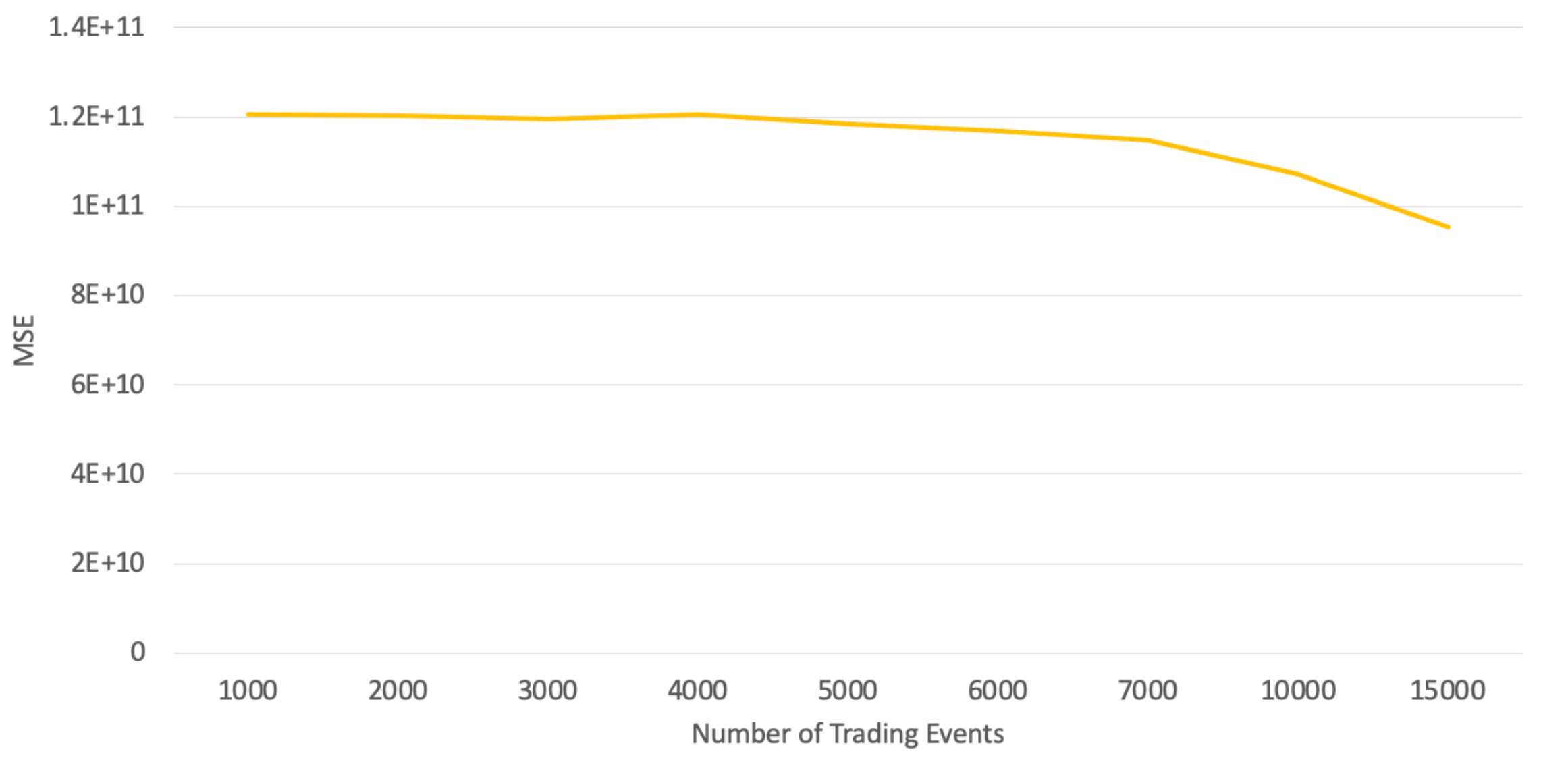}}}
    \\
  \subfloat[GRU training MSE scores \label{1a}]{%
       \scalebox{0.53}{\includegraphics[width=0.65\linewidth]{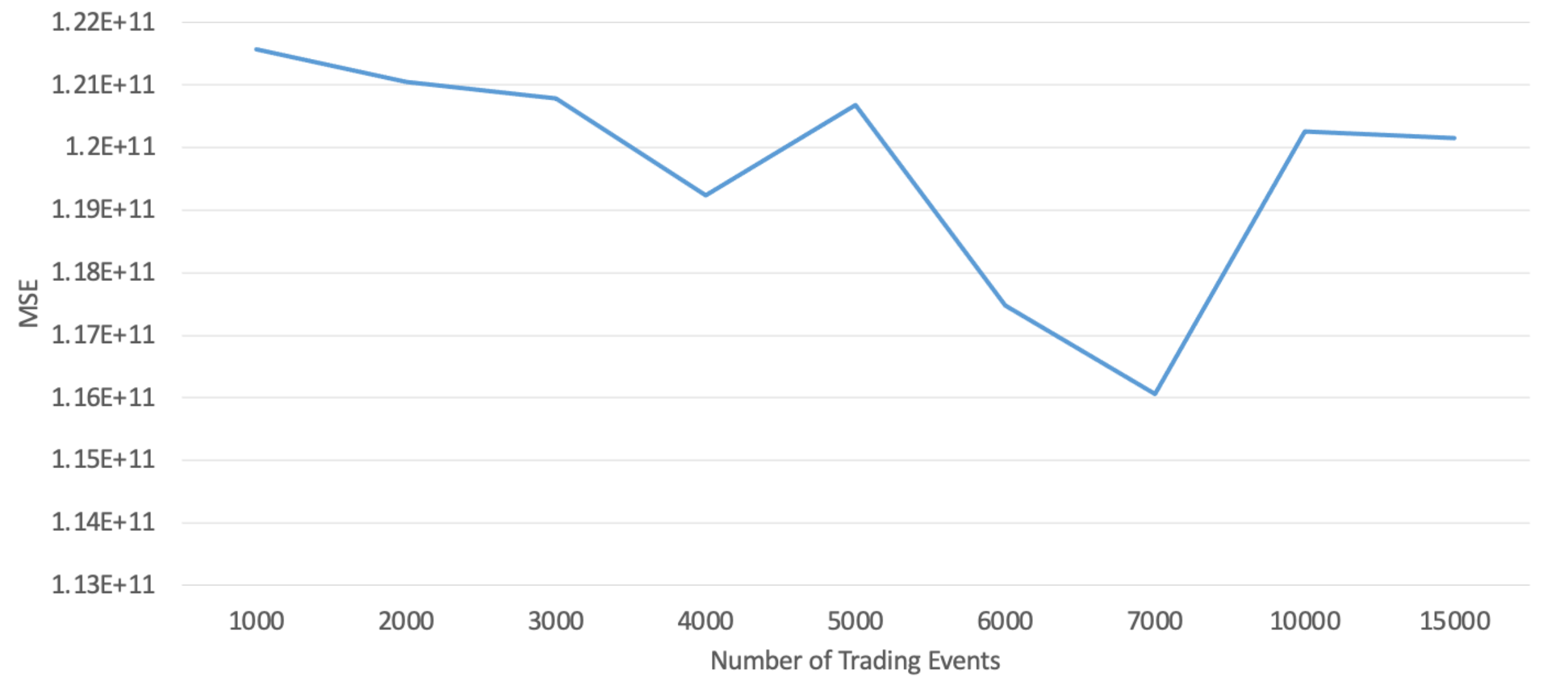}}}
  \subfloat[GRU testing MSE scores \label{1b}]{%
        \scalebox{0.53}{\includegraphics[width=0.65\linewidth]{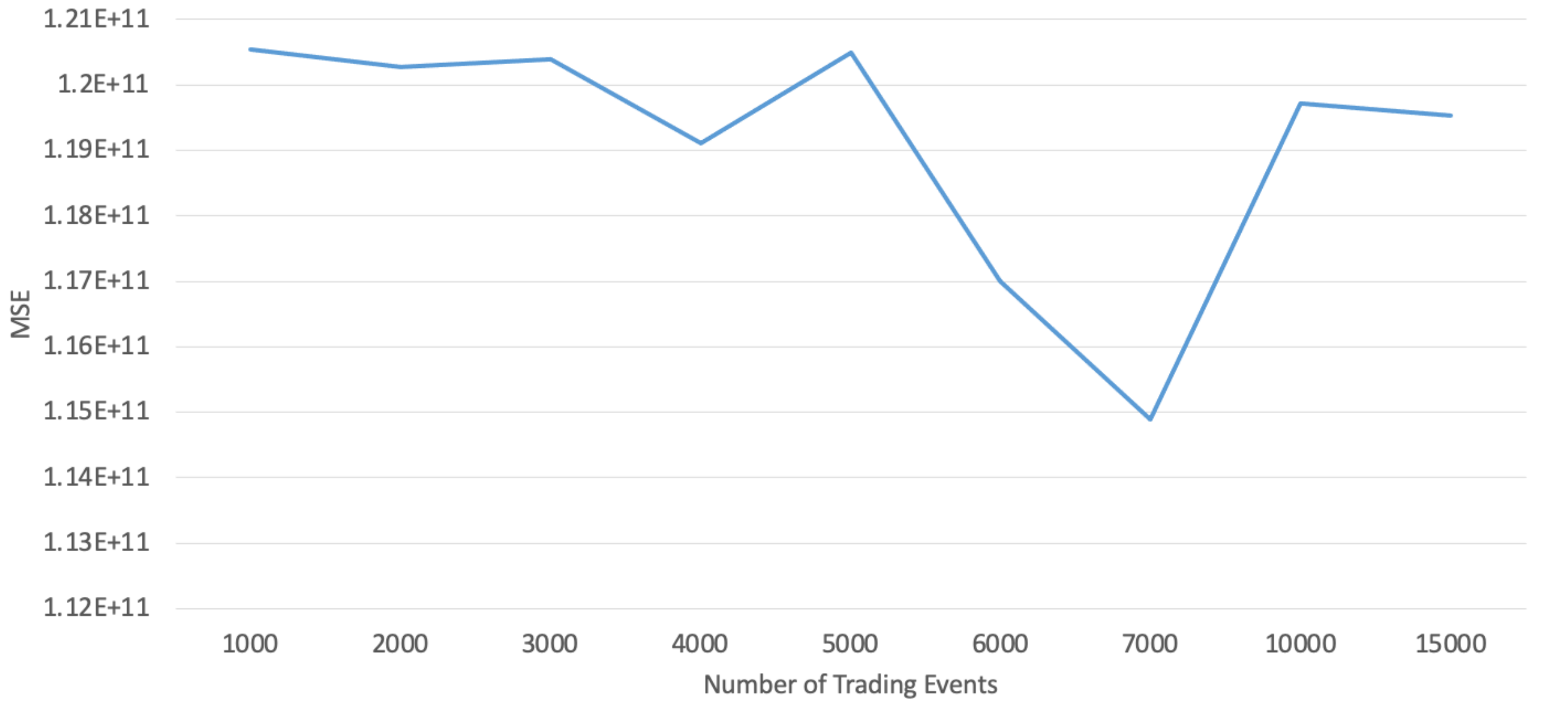}}}
    \\
  \subfloat[Hybrid training MSE scores \label{1a}]{%
       \scalebox{0.53}{\includegraphics[width=0.65\linewidth]{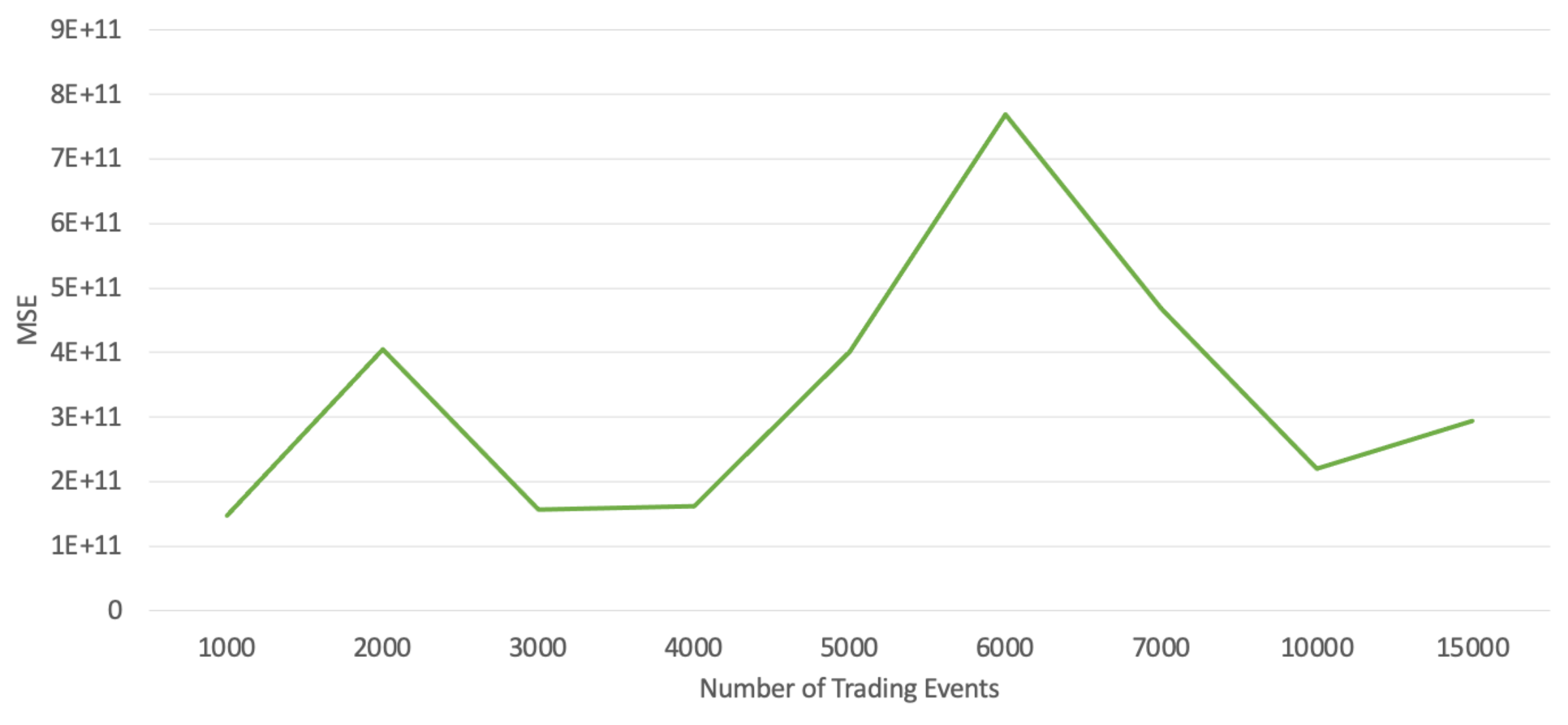}}}
  \subfloat[Hybrid testing MSE scores \label{1b}]{%
        \scalebox{0.53}{\includegraphics[width=0.65\linewidth]{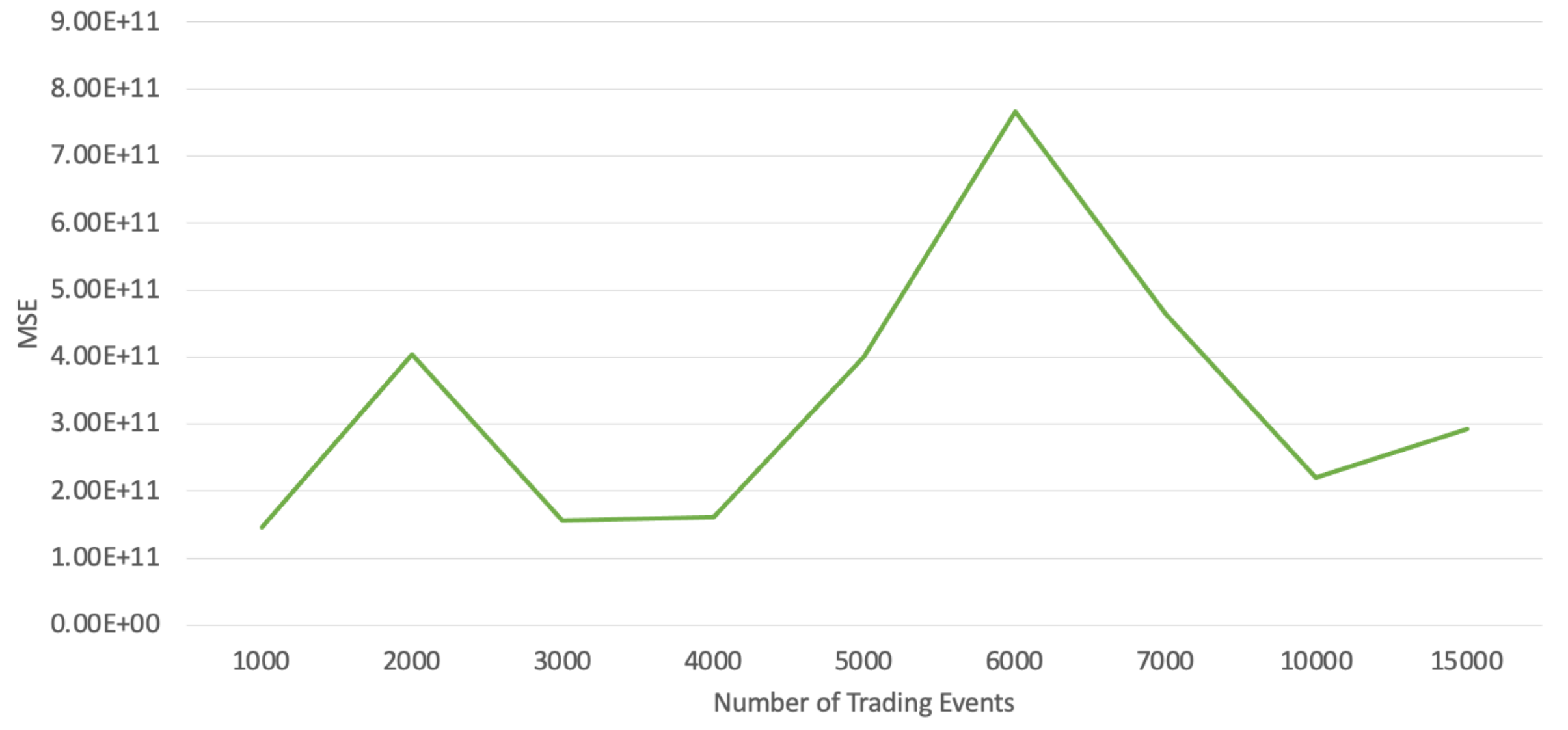}}}  
  \caption{Wartsila Long MSE scores based on \hyperref[tab:WartsilaLong]{Table \ref{tab:WartsilaLong}}.}
  \label{fig:WartsilaLong} 
\end{figure*}

\begin{table*}[hbt!]
\centering
\captionsetup{width=.70\textwidth}
\caption{Wartsila Benchmark Training (left) and Benchmark Testing (right). Data sample is 100,000 trading events.}
\scalebox{0.60}{
\begin{tabular}{ccrlcccrl}
\cmidrule[2pt]{1-4}\cmidrule[2pt]{6-9}
\textbf{Input} & \textbf{Normalization} & \textbf{Model} & \textbf{MSE - Train} & \qquad & \textbf{Input} & \textbf{Normalization} & \textbf{Model} & \textbf{MSE - Test} \\
\cmidrule{1-4}\cmidrule{6-9}
 LOB Data & Raw & \textbf{OPTM-LSTM} & \textbf{1.75200E+03} & \qquad & LOB Data &  Raw & \textbf{OPTM-LSTM}& \textbf{2.44200E+03}\\   
 &               & LSTM          & 1.17305E+11               & \qquad &       &        & LSTM         &  1.16278E+11         \\ 
 &               & Attention     &  1.15883E+11              & \qquad &       &        & Attention    &  1.13813E+11       \\   
 &               & Bidirectional &  1.17321E+11              & \qquad &       &        & Bidirectional&  1.16301E+11         \\   
 &               & GRU           &  1.16074E+11              & \qquad &       &        & GRU          &  1.16301E+11        \\   
 &               & Hybrid        &  8.94838E+11              & \qquad &       &        & Hybrid       &  8.87207E+11       \\
 &               & Baseline      &  6.81022E+11              & \qquad &       &        & Baseline     &  6.76325E+11\\
\cmidrule{2-4}\cmidrule{7-9}
 & MinMax & \textbf{OPTM-LSTM} & \textbf{1.10690E-04}       & \qquad &       &  MinMax & \textbf{OPTM-LSTM}  & \textbf{1.09039E-04} \\   
 &             & LSTM           & 1.92230E-04               & \qquad &       &         & LSTM                & 1.87033E-04   \\ 
 &             & Attention      & 2.17910E-04               & \qquad &       &         & Attention           & 1.95100E-04  \\   
 &             & Bidirectional  & 3.06550E-04               & \qquad &       &         & Bidirectional       & 4.05580E-04\\   
 &             & GRU            & 2.85060E-04               & \qquad &       &         & GRU                 & 3.79129E-04\\   
 &             & Hybrid         & 8.86610E-04               & \qquad &       &         & Hybrid              & 8.74760E-04\\
 &             & Baseline       & 7.14814E-04               & \qquad &       &         & Baseline            & 5.60481E-04 \\

\cmidrule{2-4}\cmidrule{7-9}
 & Zscore       & \textbf{OPTM-LSTM}& \textbf{1.76000E-02} & \qquad &         & Zscore & \textbf{OPTM-LSTM}  & \textbf{3.22796E-02} \\   
 &              & LSTM          & 1.32100E-01  & \qquad &            &         & LSTM        & 2.80338E-01 \\ 
 &              & Attention     & 3.74500E-01 & \qquad &            &         & Attention    & 7.54337E-01 \\ 
 &              & Bidirectional & 3.69200E-01 & \qquad &            &         & Bidirectional& 7.03396E-01 \\   
 &              & GRU           & 3.32800E-01 & \qquad &            &         & GRU          & 7.49759E-01 \\   
 &              & Hybrid        & 1.56070E+00 & \qquad &            &         & Hybrid       & 1.86654E+00 \\
 &              & Baseline      & 7.06162E-01 & \qquad &            &         & Baseline     & 6.26625E-01 \\
 \cmidrule{1-4}\cmidrule{6-9}
Mid-price & Raw & \textbf{OPTM-LSTM} & \textbf{2.15000E+03} & \qquad & Mid-price &  Raw & \textbf{OPTM-LSTM}& \textbf{2.05800E+03}\\   
 &               & LSTM          & 1.17975E+11              & \qquad &       &        & LSTM         & 1.16916E+11           \\ 
 &               & Attention     & 1.18014E+11              & \qquad &       &        & Attention    & 1.16952E+11         \\   
 &               & Bidirectional & 1.02525E+11              & \qquad &       &        & Bidirectional& 1.19952E+11        \\   
 &               & GRU           & 1.02197E+11              & \qquad &       &        & GRU          & 1.21952E+11       \\   
 &               & Hybrid        & 2.24065E+11              & \qquad &       &        & Hybrid       & 2.22092E+11            \\
 &               & Persistence   & 2.23065E+11              & \qquad &       &        & Persistence  & 2.23065E+11           \\

\cmidrule{2-4}\cmidrule{7-9}
 & MinMax & \textbf{OPTM-LSTM} & \textbf{1.28090E-04}        & \qquad &       &  MinMax & \textbf{OPTM-LSTM}  & \textbf{2.45917E-04} \\   
 &             & LSTM           & 3.36300E-04                & \qquad &       &         & LSTM                & 4.08478E-04  \\ 
 &             & Attention      & 5.03490E-04                & \qquad &       &         & Attention           & 6.09052E-04\\   
 &             & Bidirectional  & 7.65230E-04                & \qquad &       &         & Bidirectional       & 8.37438E-04 \\   
 &             & GRU            & 9.88666E-04                & \qquad &       &         & GRU                 & 9.72481E-04 \\   
 &             & Hybrid         & 5.27550E-04                & \qquad &       &         & Hybrid              & 5.23988E-04\\
 &             & Persistence    & 6.06519E-04                & \qquad &       &         & Persistence         & 6.04269E-04\\

\cmidrule{2-4}\cmidrule{7-9}
 & Zscore       & \textbf{OPTM-LSTM}& \textbf{2.44100E-01} & \qquad &         &  Zscore & \textbf{OPTM-LSTM}  & \textbf{2.55842E-01} \\   
 &              & LSTM          & 5.65200E-01  & \qquad &            &         & LSTM        & 6.04132E-01  \\ 
 &              & Attention     & 5.82000E-01 & \qquad &            &         & Attention    & 8.90469E-01 \\ 
 &              & Bidirectional & 6.79000E-01 & \qquad &            &         & Bidirectional& 7.98851E-01 \\   
 &              & GRU           & 6.62600E-01 & \qquad &            &         & GRU          & 9.50139E-01\\   
 &              & Hybrid        & 8.81800E-01 & \qquad &            &         & Hybrid       & 1.15688E+00 \\
 &              & Persistence   & 1.08168E+00 & \qquad &            &         & Persistence  & 1.09168E+00 \\
\cmidrule[2pt]{1-4}\cmidrule[2pt]{6-9}
\end{tabular}}
\medskip
\label{tab:WartsilaBenchmark}
\end{table*}

\begin{thebibliography}{53}
\bibliographystyle{IEEEtran}
\balance
\bibitem{lstmoriginal}
S. Hochreiter, and J. Schmidhuber, ``Long short-term memory", {\it{Neural Computation}}, vol. 9, issue 8, pp. 1735-1780, Nov 1997.

\bibitem{lstmacc}
T. Zia, and U. Zahid (2019), "Long short-term memory recurrent neural network architectures for Urdu acoustic modeling", {\it{International Journal of Speech Technology}}, vol. 22, issue 1, pp. 21-30, Mar 2019

\bibitem{lstmacc2}
A. Zeyer, P. Doetsch, P. Voigtlaender, R. Schlüter, and H. Ney,`` Comprehensive study of deep bidirectional LSTM RNNs for acoustic modeling in speech recognition", {\it{In 2017 IEEE international conference on acoustics, speech and signal processing (ICASSP)}}, pp. 2462-2466, Mar 2017.

\bibitem{lstmacc3}
Q. Wang, C. Downey, L. Wan, P. A. Mansfield, and I. L. Moreno, ``Speaker diarization with LSTM", {\it{In IEEE International Conference on Acoustics, Speech and Signal Processing (ICASSP)}}, Calgary, 2018, pp. 5239-5243.

\bibitem{lstmnlp}
P. Zhou, W. Shi, J. Tian, Z. Qi, B. Li, H. Hao, and B. Xu, ``Attention-based bidirectional long short-term memory networks for relation classification" {\it{In Proceedings of the 54th annual meeting of the association for computational linguistics}}, vol. 2 Short papers, pp. 207-212, Aug 2016. 

\bibitem{nlp2}
O. Melamud, J.. Goldberger, and I. Dagan, (2016, August). ``context2vec: Learning generic context embedding with bidirectional lstm", {\it{In Proceedings of the 20th SIGNLL conference on computational natural language learning}}, Berlin, 2016, pp. 51-61.

\bibitem{nlp3}
G. Liu, and J. Guo, (2019), ``Bidirectional LSTM with attention mechanism and convolutional layer for text classification" {\it{Neurocomputing}}, vol. 337, pp. 325-338, Apr 2019.

\bibitem{lstmmed}
K. M. Tsiouris, V. C. Pezoulas, M. Zervakis, S. Konitsiotis, D. D. Koutsouris, and D. I. Fotiadis, ``A long short-term memory deep learning network for the prediction of epileptic seizures using EEG signals", {\it{ Computers in biology and medicine}}, vol. 99, pp. 24-37,  Aug 2018.

\bibitem{bio2}
\"{O}. Yildirim, ``A novel wavelet sequence based on deep bidirectional LSTM network model for ECG signal classification", {\it{Computers in biology and medicine}}, vol 96, pp. 189-202, May 2018.

\bibitem{bio3}
S. L. Oh, E. Y. Ng, R. San Tan, and U. R. Acharya, ``Automated diagnosis of arrhythmia using combination of CNN and LSTM techniques with variable length heart beats'', {\it{Computers in biology and medicine}}, vol. 102, pp. 278-287, Nov 2018.

\bibitem{lstmvis}
R. R. Varior, B. Shuai, J. Lu, D. Xu, and G. Wang, ``A siamese long short-term memory architecture for human re-identification"", {\it{In European conference on computer vision}}, Cham, 2016, pp. 135-153. 

\bibitem{lstmvis2}
X. Liang, X. Shen, J. Feng, L. Lin, and S. Yan,(2016, October), ``Semantic object parsing with graph lstm", {\it{ In European Conference on Computer Vision, Springer}}, Cham, 2016, pp. 125-143. 

\bibitem{vis3}
H. Xue, D. Q. Huynh, and M. Reynolds, ``SS-LSTM: A hierarchical LSTM model for pedestrian trajectory prediction", {\it{In 2018 IEEE Winter Conference on Applications of Computer Vision (WACV), IEEE}}, Nevada, 2018, pp. 1186-1194.

\bibitem{fin1}
T. Fischer, and C. Krauss, ``Deep learning with long short-term memory networks for financial market predictions", {\it{European Journal of Operational Research}}, vol. 270, issue 2, pp. 654-669, Oct 2018.

\bibitem{fin2}
W. Bao, J. Yue, and Y. Rao, ``A deep learning framework for financial time series using stacked autoencoders and long-short term memory'', {\it{PloS One}}, vol. 12, issue 7, Jul 2017.

\bibitem{fin3}
R. Akita, A. Yoshihara, T. Matsubara, and K. Uehara, ``Deep learning for stock prediction using numerical and textual information", {\it{In 2016 IEEE/ACIS 15th International Conference on Computer and Information Science (ICIS), IEEE}}, Okayama, 2016, pp. 1-6.

\bibitem{fin4}
D. M. Nelson, A. C. Pereira, and R. A. De Oliveira, ``Stock market's price movement prediction with LSTM neural networks", {\it{In 2017 International joint conference on neural networks (IJCNN), IEEE}}, Anchorage, 2017, pp. 1419-1426.

\bibitem{fin5}
M. Fabbr, and G. Moro, ``Dow Jones Trading with Deep Learning: The Unreasonable Effectiveness of Recurrent Neural Networks", {\it{Data}}, pp. 142-153, Jul 2018.

\bibitem{siri1}
J. A. Sirignano, ``Deep learning for limit order books", {\it{Quantitative Finance}}, vol. 19, issue 4, pp. 549-570, Apr 2019.

\bibitem{siri2}
J. Sirignano, and R. Cont, ``Universal features of price formation in financial markets: perspectives from deep learning", {\it{ Quantitative Finance}}, vol 19, issue 9, pp. 1449-1459, Sep 2019.

\bibitem{deeplob}
Z. Zhang, S. Zohren, and S. Roberts, ``Deeplob: Deep convolutional neural networks for limit order books'', {\it{IEEE Transactions on Signal Processing}}, vol 67, issue 11, 3001-3012, Mar 2019.

\bibitem{ntak1}
A. Ntakaris, G. Mirone, J. Kanniainen, M. Gabbouj and A. Iosifidis, ``Feature Engineering for Mid-Price Prediction With Deep Learning", {\it{IEEE Access}}, vol. 7, pp. 82390-82412, Jun 2019 

\bibitem{ntak2}
A. Ntakaris, M. Magris, J. Kanniainen, M. Gabbouj, A. and Iosifidis. ``Benchmark dataset for mid‐price forecasting of limit order book data with machine learning methods", \it{Journal of Forecasting}, vol. 37, issue 8, pp.852-866, Dec 2018.

\bibitem{pass1}
A. Tsantekidis, N. Passalis, A. Tefas, J. Kanniainen, M. Gabbouj and A. Iosifidis,``Using deep learning to detect price change indications in financial markets," {\it{2017 25th European Signal Processing Conference (EUSIPCO)}}, 2017, pp. 2511-2515.

\bibitem{dix1}
M. Dixon, ``Sequence classification of the limit order book using recurrent neural networks", {\it{Journal of computational science}}, vol. 24, pp. 277-286, Jan 2018.

\bibitem{pass2}
A. Tsantekidis, N. Passalis, A. Tefas, J. Kanniainen, M. Gabbouj, and A. Iosifidis, ``Using deep learning for price prediction by exploiting stationary limit order book features"", {\it{Applied Soft Computing}}, vol. 93, Aug 2020.

\bibitem{Ch1}
Y. Li, L. Li, X. Zhao, T. Ma, Y. Zou, and M. Chen, ``An Attention-Based LSTM Model for Stock Price Trend Prediction Using Limit Order Books'', {\it{In Journal of Physics: Conference Series}}, vol. 1575, issue. 1, pp. 012124, Jun 2020.

\bibitem{Maki1}
Y. M\"{a}kinen, J. Kanniainen, M. Gabbouj, and A. Iosifidis, ``Forecasting jump arrivals in stock prices: new attention-based network architecture using limit order book data'', {\it{Quantitative Finance}}, vol. 19, issue 12, pp. 2033-2050, Dec 2019.

\bibitem{markmak}
T. Sun, D. Huang, and J. Yu, ``Market Making Strategy Optimization via Deep Reinforcement Learning", {\it{IEEE Access}}, vol. 10, pp. 9085-9093, Jan 2022. 

\bibitem{Goo}
R. Jozefowicz, W. Zaremba, and I. Sutskever, ``An empirical exploration of recurrent network architectures", {\it{In International conference on machine learning}, PMLR}, 2015, pp. 2342-2350.

\bibitem{simple}
Y. Lu, and F. M. Salem, ``Simplified gating in long short-term memory (lstm) recurrent neural networks", {\it{In 2017 IEEE 60th International Midwest Symposium on Circuits and Systems} (MWSCAS)} Medford, 2017, pp. 1601-1604.

\bibitem{chrono}
C. Tallec, and Y. Ollivier, ``Can recurrent neural networks warp time?", arXiv preprint arXiv:1804.11188, Mar 2013.

\bibitem{improve}
A. Gu, C. Gulcehre, T. Paine, M. Hoffman, and R. Pascanu, ``Improving the gating mechanism of recurrent neural networks", {\it{In International Conference on Machine Learning}, PMLR}, 2020, pp. 3800-3809).

\bibitem{zhang}
S. Zhang, Y. Wu, T. Che, Z. Lin, R. Memisevic, R. R. Salakhutdinov, and Y. Bengio, ``Architectural complexity measures of recurrent neural networks", {\it{Advances in neural information processing systems}}, vol. 29, 2016

\bibitem{Unitary}
M. Arjovsky, A. Shah, and Y. Bengio, ``Unitary evolution recurrent neural networks", {\it{In International Conference on Machine Learning}, PMLR}, New York, 2016  pp. 1120-1128.

\bibitem{ororbia}
A. G. Ororbia II, T. Mikolov and D. Reitter, ``Learning Simpler Language Models with the Differential State Framework", {\it{in Neural Computation}}, vol. 29, no. 12, pp. 3327-3352, Dec. 2017.

\bibitem{lee}
Q. V. Le, N. Jaitly, and G. E. Hinton, ``A simple way to initialize recurrent networks of rectified linear units", arXiv preprint arXiv:1504.00941, Apr 2015

\bibitem{greff}
K. Greff, R. K. Srivastava, J. Koutník, B. R. Steunebrink, and J. Schmidhuber, J., ``LSTM: A search space odyssey", {\it{Transactions on neural networks and learning systems}, IEEE}, vol. 28, issue 10, pp. 2222-2232, Jul 2016.

\bibitem{gers}
F. A. Gers, J. Schmidhuber, and F. Cummins, ``Learning to forget: Continual prediction with LSTM", {\it{Neural computation}}, vol 12, issue 10, pp. 2451-2471, Oct 2000.

\bibitem{dat}
D. T. Tran, A. Iosifidis, J. Kanniainen and M. Gabbouj,``Temporal Attention-Augmented Bilinear Network for Financial Time-Series Data Analysis", {\it{Transactions on Neural Networks and Learning Systems}}, vol. 30, no. 5, pp. 1407-1418, May 2019. 

\bibitem{he}
Z. He, S. Gao, L. Xiao, D. Liu, H. He, and D. Barber, ``Wider and deeper, cheaper and faster: Tensorized lstms for sequence learning", {\it{in Advances in neural information processing systems}, NIPS}, California, 2017.

\bibitem{zoneout}
D. Krueger, T. Maharaj, J. Kramár, M. Pezeshki, N. Ballas, N. R. Ke, N.R., A. Goyal, Y. Bengio, A. Courville, A. and C. Pal, ``Zoneout: Regularizing rnns by randomly preserving hidden activations", arXiv preprint arXiv:1606.01305, June 2016.

\bibitem{phased}
D. Neil, M. Pfeiffer, and S. C. Liu, ``Phased lstm: Accelerating recurrent network training for long or event-based sequences", {\it{Advances in neural information processing systems},NIPS}, Barcelona, 2016. 

\bibitem{stoch}
M. Fraccaro, S. K. Sønderby, U. Paquet, and O. Winther, ``Sequential neural models with stochastic layers", {\it{Advances in neural information processing systems}, NIPS}, Barcelona, 2016.

\bibitem{yao}
K. Yao, T. Cohn, K. Vylomova, K. Duh, and C. Dyer, ``Depth-gated recurrent neural networks", arXiv preprint arXiv:1508.03790, Aug. 2015.

\bibitem{rot}
R. Dangovski, L. Jing, P. Nakov, M. Tatalović, and M. Solja\v{c}ić, ``Rotational unit of memory: a novel representation unit for RNNs with scalable applications", {\it{Transactions of the Association for Computational Linguistics}}, vol. 7, pp. 121-138, Aug 2019.

\bibitem{cnn_1}
Y. LeCun, L. Bottou, Y. Bengio, and P. Haffner 1998. ``Gradient-based learning applied to document recognition". \it{In of Proceedings of the IEEE}, 86(11), pp.2278-2324, Nov 1998.
Vancouver	

\bibitem{keras}
F. Chollet, 2015, ``keras'', \it{Github}, GitHub repository, \url{https://github.com/fchollet/keras}, commit: 5bcac37

\bibitem{tensorflow}
 M. Abadi, A. Agarwal, P. Barham, E. Brevdo, Z. Chen, C. Citro, G.S. Corrado, A. Davis, J. Dean, M. Devin, and S. Ghemawat, 2016. Tensorflow: ``Large-scale machine learning on heterogeneous distributed systems''. arXiv preprint arXiv:1603.04467.

\bibitem{bidi}
S. Mootha, S. Sridhar, R. Seetharaman,  and S. Chitrakala, 2020. ``Stock price prediction using bi-directional LSTM based sequence to sequence modeling and multitask learning''. In {\it{11th IEEE Annual Ubiquitous Computing, Electronics \& Mobile Communication Conference}, UEMCON}, New York, 2020.

\bibitem{bidi_2}
M.A.I. Sunny, M.M.S.Maswood, and A.G., Alharbi, 2020, October. ``Deep learning-based stock price prediction using LSTM and bi-directional LSTM model''. In \it{Novel Intelligent and Leading Emerging Sciences Conference''}, (NILES), Egypt, 2020. 

\bibitem{bidi_3}
K.A. Althelaya, E.S.M. El-Alfy, and S. Mohammed. ``Evaluation of bidirectional LSTM for short-and long-term stock market prediction''. In {\it{9th international conference on information and communication systems}, (ICICS)}, France, 2018.

\bibitem{cnn_1}
W. Lu,, J. Li, Y. Li, A. Sun, and J. Wang, {\it{''A CNN-LSTM-based model to forecast stock prices.}} Complexity, 2020.

\bibitem{cnn_2}
S., Mehtab, and J. Sen. {\it{``Stock price prediction using CNN and LSTM-based deep learning models.}} In {\it{International Conference on Decision Aid Sciences and Application (DASA)}, IEEE}, Bahrain, 2020.

\bibitem{cnn_3}
J.M.T. Wu, Z. Li, N. Herencsar, B. Vo, and J.C.W. Lin. \it{``A graph-based CNN-LSTM stock price prediction algorithm with leading indicators}. Multimedia Systems, pp.1-20, Feb. 2021.

\bibitem{cnn_4}
Z. Nourbakhsh, and N. Habibi. \it{``Combining LSTM and CNN methods and fundamental analysis for stock price trend prediction}. Multimedia Tools and Applications, 2022. 

\bibitem{phd_1}
T. Mikolov, ``Statistical Language Models Based on Neural Networks'', PhD Thesis, Department of Computer Graphics and Multimedia, BRNO University of Technology, Brno, Czechia, 2012 

\bibitem{init}
M. Mehdipour Ghazi, M. Nielsen, A. Pai, M. Modat, M.J. Cardoso, S. Ourselin, S. and L. Sorensen. ``On the initialization of long short-term memory networks". \textit{In International Conference on Neural Information Processing} (pp. 275-286). Springer, Cham, Dec. 2019

\bibitem{vani}
H.Y.S. Chien, J.S. Turek, N. Beckage, V.A. Vo, C.J. Honey, and T.L. Willke. \textit{``Slower is Better: Revisiting the Forgetting Mechanism in LSTM for Slower Information Decay".} arXiv preprint arXiv:2105.05944, May 2021.

\bibitem{dynamic}
B. Laperre, J. Amaya, and G. Lapenta. \textit{``Dynamic time warping as a new evaluation for dst forecast with machine learning''}. Frontiers in Astronomy and Space Sciences, , p.39. 2020

\bibitem{mlops}
 G. Gürses-Tran, and A. Monti. \textit{``Advances in Time Series Forecasting Development for Power Systems''} Operation with MLOps. Forecasting, 4(2), pp.501-524. May 2022.
\end{thebibliography}
\end{document}